\documentclass[11pt]{article}

\usepackage[preprint]{acl}

\usepackage{times}
\usepackage{latexsym}

\usepackage[T1]{fontenc}

\usepackage[utf8]{inputenc}

\usepackage{microtype}

\usepackage{inconsolata}

\usepackage{graphicx}
\usepackage{amsmath}        
\usepackage{cleveref}            
\usepackage{tcolorbox}
\usepackage{booktabs}       
\usepackage{multirow}
\usepackage{subcaption}

\usepackage{placeins}
\usepackage{amsfonts}       
\tcbset{
    left*=10pt, right*=10pt,
    top=0pt, bottom=0pt,
    colback=white!10!white,
    colframe=black!75!black,
    fonttitle=\bfseries\large,
    subtitle style={boxrule=0pt,colback=gray!50!white},
}

\newcommand{\OlmoSevenB}{OLMo-2-7B IT}
\newcommand{\OlmoThreeSevenB}{OLMo-3-7B IT}
\newcommand{\OlmoThreeThirtyTwoB}{OLMo-3.1-32B IT}
\newcommand{\OlmoThirteenB}{OLMo-2-13B IT}
\newcommand{\Gemma}{Gemma-2-9B IT}
\newcommand{\GemmaTwentySeven}{Gemma-2-27B IT}

\title{LMs as Task-Specific Knowledge Bases: An Interpretability Analysis}

\author{
 \textbf{Amit Elhelo\textsuperscript{1}}\quad
 \textbf{Amir Globerson\textsuperscript{1,2,$*$}}\quad
 \textbf{Mor Geva\textsuperscript{1,$*$}}
\\
 \textsuperscript{1}Blavatnik School of Computer Science and AI, Tel Aviv University
 \\
 \textsuperscript{2}Google Research
\\
 \small{
   {\{amitelhelw@mail,gamir@tauex,morgeva@tauex\}.tau.ac.il}
 }
}

\begin{document}
\maketitle
\def\thefootnote{*}\footnotetext{Equal senior authorship.}\def\thefootnote{\arabic{footnote}}

\begin{abstract}
Language models (LMs) capture large amounts of factual knowledge applicable to a wide range of tasks, motivating the view of their parameters as a knowledge base. 
An important property of knowledge bases is that different queries for the same fact return consistent results, drawing on a single source of truth. 
We investigate whether LMs satisfy this property through behavioral and mechanistic analyses. Our results suggest that they encode knowledge in a task-specific manner. 
Behaviorally, facts acquired on one task frequently fail to co-emerge on others during training. Parameter localization experiments suggest a mechanistic explanation, revealing distinct parameter subsets underlying different tasks for the same fact. 
Finally, we show that chain-of-thought reasoning draws part of its effectiveness from engaging task-specific parameters beyond those tied to the evaluation task.
Our findings suggest that what the model knows and how it is asked are intertwined in parameter space, undermining the ``knowledge base'' analogy and carrying implications for the reliability and controllability of factual knowledge in LMs.
\end{abstract}

\begin{figure}[t]
    \centering
    \includegraphics[width=\linewidth]{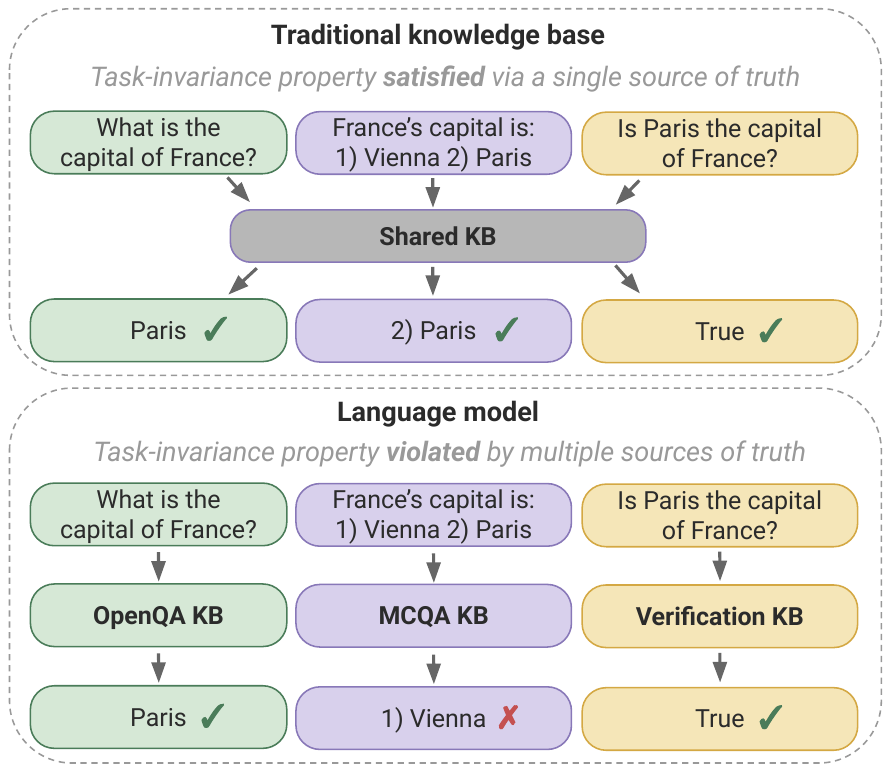}
    \caption{Language models and task-invariance. A traditional knowledge base (top) draws on a single source of truth regardless of query format. Here we show that LMs are better described by a scheme where each task has its own KB (bottom), so the same question can be answered differently, as in the multiple-choice case.}
    \label{fig:figure1}
\end{figure}

\section{Introduction}

Language models (LMs) encode vast amounts of knowledge in their parameters which is utilized in various contexts, such as dialogue, summarization, and reasoning \citep{Hendrycks2020MeasuringMM}.
As such, LMs are often viewed as information systems whose parameters act as a knowledge base \citep{petroni2019language, roberts-etal-2020-much}.

In a well-designed knowledge base, different queries for the same fact draw on a single source of truth, guaranteeing consistent results.
For example, a knowledge base should retrieve \texttt{Paris} for both \textit{``What is the capital of France?''} and \textit{``The capital of France is \_\_\_''}.
Violating this introduces risks to system reliability, consistency, and updateability \citep{10.1145/362384.362685, DBLP:books/aw/AbiteboulHV95}. In this work, we ask whether LMs satisfy this property across knowledge-querying formats (which we refer to as \textit{tasks}) that require different operations on the same underlying fact, such as open-ended QA and statement verification.

We investigate this through two experiments. 
First, a behavioral analysis where we track across training checkpoints of \OlmoThreeSevenB{} and \OlmoThreeThirtyTwoB{} \citep{olmo2025olmo} how knowledge of individual facts, drawn from datasets of (subject, relation, object) triplets, co-emerges across tasks. 
If facts were stored in a task-invariant manner, a model acquiring a fact for one task should simultaneously acquire it for other tasks it is already competent at. 
We find that co-emergence is limited, varying substantially across tasks, suggesting that knowledge acquisition in LMs is task dependent.

Next, we analyze how task-specific knowledge encodings are manifested in model parameters.
If knowledge is stored independently of task format, it should not be possible to isolate parameters that are specific to individual (fact, task) pairs.
We study this through a mechanistic analysis, adapting the localization framework of \citet{bayazit2024discovering}. 
For each (fact, task) pair we identify a sparse subset of parameters whose removal degrades model performance on that pair with little effect on other facts on the same task or the same fact on other tasks.
Across four models and five relational datasets, we consistently find such subsets.
Together with the behavioral results, this suggests that LMs maintain task-dependent parametric encodings of individual facts, instead of drawing from a shared, task-invariant store. \Cref{fig:figure1} illustrates this. 

We find that the degree of this separation is not uniform, as some (fact, task) encodings can be well isolated from other pairs while others show partial overlap. 
This raises the question of which tasks tend to have separate versus shared encodings \citep{zamir2018taskonomy}. 
To quantify this overlap, we develop metrics that measure how separable each (fact, task) encoding is from other pairs, and find that discrimination tasks (e.g., \textsc{Multiple Choice QA}) are consistently more entangled than generation tasks (e.g., \textsc{Fill-in-the-Blank}).
Moreover, facts acquired through generation tasks generally co-emerge on other tasks, but not vice versa.

Finally, we hypothesize that part of the effectiveness of chain-of-thought (CoT) reasoning in recovering knowledge inaccessible to direct answering (without intermediate reasoning;  \citealt{gekhman2026thinking}) comes from engaging parametric encodings beyond those tied to the evaluation task.
We confirm this by removing the localized (fact, task) encodings. CoT largely recovers performance lost when a task's own encoding is ablated, yet drops more than direct answering when \emph{other} tasks' encodings are removed, suggesting it relies on them more than direct answering does.

Together, these findings show that knowledge in LMs is not cleanly separated from task structure, as what the model knows and how it is asked are intertwined in parameter space. 
This undermines the ``knowledge base'' analogy, whose guarantees of reliability and controllability rest on knowledge being task-invariant. For instance, knowledge editing or unlearning interventions targeting a single task format may leave other formats intact, and single-task evaluation may provide only a partial view of what the model encodes. We release our code and data at 
\url{https://github.com/amitelhelo/TaskInvariance}.

\section{Task-specific knowledge encodings}
 \label{sec:behavioral}

In a well-designed knowledge base, querying a given fact in different ways should return the same result, drawing on the same internal source of truth.
We call this property \emph{task-invariance}, and investigate it in LMs through a \textit{behavioral} experiment (detailed in this section), tracking how acquisition of individual facts co-emerges across tasks during training, and a \textit{mechanistic} experiment (\S\ref{sec:mechanistic_analysis}), asking whether the parameters that support a fact differ across tasks.
Our analysis shows that knowledge is fragmented across task-specific encodings; facts acquired on one task often fail to transfer to other tasks, and it is possible to localize distinct parameters that encode the same fact for different tasks.

\subsection{Experimental setup}
We track factual knowledge in LMs using relational datasets \citep{10.1145/2629489, hernandez2024linearity}, where facts are formulated as (subject, relation, object) triplets. For example, the fact that Paris is the capital of France can be represented as the triplet \texttt{(France, capital-of, Paris)}. We use datasets of five relations:
\texttt{(country, capital-of, city)}, \texttt{(country, official language, language)}, \texttt{(landmark, in-country, country)}, \texttt{(company, HQ-in-city, city)}, and \texttt{(person, plays-instrument, instrument)}. 
Each fact is probed via six task formats, differing in the operation applied to it: next-token completion (\textsc{Completion}), fill-in-the-blank (\textsc{FiTB}), open-ended QA (\textsc{OpenQA}), multiple-choice QA (\textsc{MCQA}), negative MCQA (\textsc{Neg~MCQA}; select the \emph{incorrect} answer), and binary statement verification (\textsc{Verification}). 

For each dataset-task pair we composed 10 prompt paraphrases, which we use to evaluate the model's knowledge of the facts for the task. For discrimination tasks, each paraphrase is further expanded by rotating the correct answer through all positions (4 for \textsc{MCQA}, 2 for \textsc{Neg~MCQA}) or by pairing it with both a true and a false statement (\textsc{Verification}).
For additional dataset details see \S\ref{appendix:datasets}. Representative prompts for the different tasks are provided in Appendix \Cref{fig:prompt-examples}.
All datasets are down-sampled to 46 facts each (matching the size of the smallest dataset), yielding 230 facts. 

\subsection{Co-emergence hypothesis}
The task-invariance property entails predictions about training. Specifically, it implies that facts should co-emerge across tasks, a hypothesis which we formalize as follows: 
if different tasks retrieve a given fact from the same task-invariant parametric store, then once the model can retrieve a fact for some task (e.g., correctly answer an open question about the capital of France), then it should retrieve that fact for other tasks it is competent on (e.g., correctly answer a multiple choice question about the capital of France).

\begin{figure}[t]
    \centering
    \includegraphics[width=\linewidth]{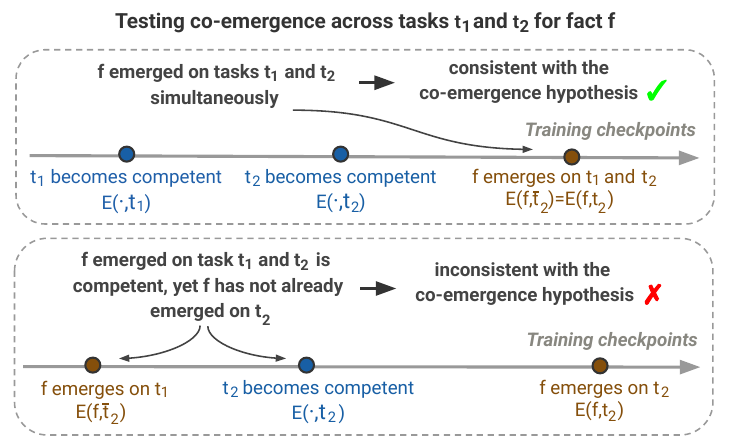}
    \caption{Example observations consistent (top) and inconsistent (bottom) with the co-emergence hypothesis.}    
    \label{fig:behavioral_explained}
\end{figure}

Formally, let $\mathcal{T}$ denote a set of tasks. We write $E(f,t)$ for the \emph{emergence step} of fact $f$ on task $t$, defined as the first checkpoint at which the model reliably retrieves $f$ on $t$ (operationalized below), or $\infty$ if this never occurs. We write $E(\cdot, t)$ for the emergence step of task $t$, defined as the first checkpoint in which a substantial fraction of facts are reliably retrieved on $t$. Finally, we write $E(f, \bar{t})$ for the earliest checkpoint at which $f$ emerges on any task other than $t$, defined as $E(f, \bar{t}) = \min_{t' \in \mathcal{T} \setminus \{t\}} E(f, t')$.
See \Cref{fig:behavioral_explained} for illustration.
Under the co-emergence hypothesis, once both prerequisites are met (the fact is retrieved on some task and the target task is competent), the fact should be retrieved on the target task as well. That is, for all facts $f$ and tasks $t$:
\begin{equation}
E(f, t) \;\leq\; \max\!\bigl(E(f, \bar{t}),\; E(\cdot, t)\bigr)
\label{eq:co_emergence}
\end{equation}

\paragraph{Testing the co-emergence hypothesis}
We use \OlmoThreeSevenB{} and \OlmoThreeThirtyTwoB{} \citep{olmo2025olmo}, since their intermediate training checkpoints are publicly available. We describe the setup for the 7B model; the 32B model follows the same protocol, with minor differences detailed in \S\ref{appendix:behavioral}.
We track the model's performance for each (fact, task) pair across training.  
Concretely, we examine 105 checkpoints covering the pretraining stage (100 checkpoints), midtraining and long context (2 checkpoints), and post-training (3 checkpoints).

To determine emergence of a (fact,task) pair, we take the model's probability of the first token of the correct answer per paraphrase, normalize it by the task's chance level, and consider the fact reliably retrieved when the mean probability over paraphrases exceeds $\theta = 0.6$.
This ensures meaningful preference for the correct answer while allowing imperfect performance mid-training.\footnote{We repeated the analysis with $\theta = 0.4$ and $\theta = 0.8$ and observed similar trends.}
The emergence step $E(f, t)$ is the first checkpoint at which this criterion is met.
Similarly, task $t$'s emergence step $E(\cdot, t)$ is the first checkpoint where at least 25\% of facts are reliably retrieved on $t$.
We retain only facts that the final Instruct model retrieves correctly on at least one task, and exclude (fact, task) pairs that cannot meaningfully test the prediction. 
These include pairs where the fact first emerged on $t$ (no prior source to ``co-emerge'' with) and pairs where, at the expected step (RHS of Eq.~\ref{eq:co_emergence}), the fact is no longer retrieved elsewhere or $t$ is no longer competent. 
We apply this second condition only to pairs that fail to co-emerge, so a lapsed prerequisite is not counted against the hypothesis.
This yields 1,031 pairs.
See \S\ref{appendix:behavioral} for additional details.

When co-emergence occurs by the predicted step (Eq.~\ref{eq:co_emergence}), we say the observation is \emph{consistent} with the hypothesis, and otherwise \emph{inconsistent}. \Cref{fig:behavioral_explained} illustrates both cases. 
Notably, consistent observations on their own do not support the hypothesis, since a fact may emerge for reasons unrelated to shared storage; but inconsistent observations provide direct evidence against it.

\paragraph{Results}
We find that the co-emergence hypothesis is frequently violated. In 47.9\% of (fact, task) pairs ($50.3\%$ for the 32B model), the fact does not emerge on the target task by the expected step, suggesting that factual knowledge does not transfer reliably across tasks during training. Violations remain frequent across thresholds: $50.9\%$ at $\theta{=}0.4$, $49.2\%$ at $\theta{=}0.8$ ($44.7\%$ and $33.5\%$, respectively, for the 32B model). For both models, $\theta{=}0.8$ leaves only about two-thirds as many testable pairs, so rates at that threshold are estimated on a smaller, more filtered subset.
In \S\ref{sec:entanglement} we analyze these results at the task-pair level, asking which pairings show reliable co-emergence and which do not.

\subsection{Testing for (fact,task) interaction}
The above results suggest that the data does not agree with a single, task-invariant store of factual knowledge.
To test the task-invariant store hypothesis statistically, we formalize it as an additive decomposition of the log-probability of a correct answer into a fact term and a task term, which corresponds to a two-way ANOVA with no interaction term. 
This is the expected behavior in a model where retrieving a particular fact does not depend on the task for which it is retrieved.
We test whether the interaction is zero.
We run the test on the chance-normalized log-probabilities of the 7B model, using prompt paraphrases as replications within each (fact, task) cell. 
The null hypothesis is rejected at every checkpoint ($p \approx 0$).
The interaction also grows across training, explaining $23\%$ of the variance in the final model. Thus we conclude that the data does not support a task-invariant model (see \S\ref{appendix:interaction} for more details and full results).

\section{Mechanistic analysis}
\label{sec:mechanistic_analysis}
Having established that knowledge acquisition is task-dependent at the behavioral level, we turn to investigate how this manifests in the model weights. 
Specifically, we ask whether the same fact relies on different parameters for different tasks.
We search for small subsets of model components (attention heads and MLP neurons) that are \emph{necessary}, \emph{sufficient}, and \emph{specific} for individual (fact, task) pairs.
Existence of subsets satisfying all three criteria would support the hypothesis that LMs maintain task-dependent parametric encodings of individual facts. We show that such subsets can be found.
 
 \subsection{Experimental setup}
We use the datasets and tasks from \S\ref{sec:behavioral} (without the downsampling to 46 facts), dropping \textsc{Completion}, which is incompatible with Instruct models, and adding two multi-hop reasoning tasks where the fact's relation is part of a two-step chain.\footnote{The downsampling in \S\ref{sec:behavioral} enabled cross-dataset aggregation; here we analyze each dataset separately. \textsc{Completion} evaluates next-token prediction on plain sentences (e.g., \textit{The capital city of France is}), incompatible with Instruct models' chat-template. Multi-hop tasks complicate co-emergence tracking through their bridging entities.}
In first-hop (\textsc{Multi-Hop-1}) the target relation is the first step, and in second-hop (\textsc{Multi-Hop-2}) it is the second.
For example, the prompt \textit{``What is the capital city of the country containing the landmark called The Kharga Oasis?''} follows the reasoning path \texttt{landmark $\to$ country $\to$ capital}. It can serve as a \textsc{Multi-Hop-1} prompt for the \texttt{(landmark, in-country, country)} dataset, and as a \textsc{Multi-Hop-2} prompt for the \texttt{(country, capital-of, city)} dataset.
The exact task set varies by dataset, depending on the availability of intermediate relations for multi-hop tasks, and facts below a baseline performance threshold in any task are filtered out (see \S\ref{appendix:params_exp}).
We examine four models: \OlmoSevenB, \OlmoThirteenB{} \citep{OLMo20242O2}, \Gemma{}, and \GemmaTwentySeven{} \citep{Riviere2024Gemma2I}, the last on two datasets.

\paragraph{Localization via learned binary masks}
We adapt the framework of \citet{bayazit2024discovering}, who trained binary masks over model parameters to find \emph{knowledge-critical subnetworks}, and extend it to localize subsets of parameters that are necessary, sufficient, and specific for (fact, task) pairs. Concretely, for a target pair $(f^*, t^*)$, we learn a binary mask $\mathbf{m} \in \{0,1\}^{|N|+|H|}$ over the sets of MLP neurons $N$ and attention heads $H$ in the model. 
We parameterize $\mathbf{m}$ as continuous logits passed through a sigmoid, binarized at threshold $0.5$ via a straight-through estimator \citep{bengio2013estimating}. 
Each mask is optimized to minimize:
\begin{equation}
\resizebox{\columnwidth}{!}{$\displaystyle
\mathcal{L}(\mathbf{m}) = \mathcal{L}_{\text{nec}}(\mathbf{m}) + \mathcal{L}_{\text{suff}}(\mathbf{m}) + \mathcal{L}_{\text{spec}}(\mathbf{m}) + \beta \mathcal{L}_{\text{spar}}(\mathbf{m})
$}
\label{eq:mask_objective}
\end{equation}
where $\mathcal{L}_{\text{nec}}$, $\mathcal{L}_{\text{suff}}$, and $\mathcal{L}_{\text{spec}}$ encourage the identified parameters to be necessary, sufficient, and specific for $(f^*, t^*)$, respectively, and $\mathcal{L}_{\text{spar}}$ encourages sparsity. 
We define each term below.
 
\textit{Necessity.}\;
The necessity loss ensures that removing the localized parameters hurts performance on the target pair, establishing that they are necessary for it. 
Let $p(f, t, \theta \circ \mathbf{m})$ denote the probability of the first token of the correct answer for task $t$ on fact $f$ when the model parameters $\theta$ are masked by $\mathbf{m}$. Let $p(f, t, \theta)$ denote the unmasked model's probability. Masking zeros out the activations of the selected MLP neurons. For attention heads, it zeros the output vectors before the output projection. Both are equivalent to zeroing the parameters themselves.
The loss drives this probability toward chance level $\tau$ ($\tau{=}0$ for generation tasks, $0.25$ for \textsc{MCQA}, $0.5$ for binary tasks):
\begin{equation}
\mathcal{L}_{\text{nec}}(\mathbf{m}) = \text{MSE}\!\bigl(p(f^*, t^*, \theta \circ \mathbf{m}),\; \tau\bigr)
\label{eq:nec_loss}
\end{equation}
For discrimination tasks, an additional MSE term encourages the aggregate probability of the distractors to rise to $1-\tau$, so that ablating the identified parameters changes the model's answer rather than disrupting its ability to perform the task (see \S\ref{appendix:params_exp} for details). 
For evaluation, we measure the relative change in accuracy under masking, where predictions that differ from the target only in formatting are not penalized (see \S\ref{appendix:variant_tolerance}).

\textit{Sufficiency.}\; 
The sufficiency loss ensures that the localized components carry enough information to retrieve the fact even when the prompt is corrupted. 
Following \citet{yona2026friends}, we corrupt the prompt by replacing the subject entity with placeholder tokens (e.g., \texttt{France} $\to$ \texttt{xx}), removing the part that identifies the fact. 
We then run two forward passes: (i)~a pass on the original prompt, caching the localized components' activations; (ii)~a pass on the corrupted prompt, in which the cached activations replace the corrupted ones. 
As in necessity, this targets the MLP neuron activations and attention head outputs before the output projection.
The loss encourages the patched model's probability of the correct answer on the corrupted prompt to match the unintervened model's probability on the original prompt:
\begin{equation}
\mathcal{L}_{\text{suff}}(\mathbf{m}) = \text{MSE}\!\bigl(\tilde{p}(f^*, t^*, \theta \circ \mathbf{m}),\; p(f^*, t^*, \theta)\bigr)
\label{eq:suff_loss}
\end{equation}
where $\tilde{p}(f^*, t^*, \theta \circ \mathbf{m})$ denotes the patched model's probability on the corrupted prompt.
For evaluation, we report the \emph{reconstruction rate}: the fraction of the accuracy lost to corruption that is recovered by patching the localized components' activations (see \S\ref{appendix:params_exp} for the formal definition).

\begin{figure}[t]
    \centering
    \includegraphics[width=\linewidth]{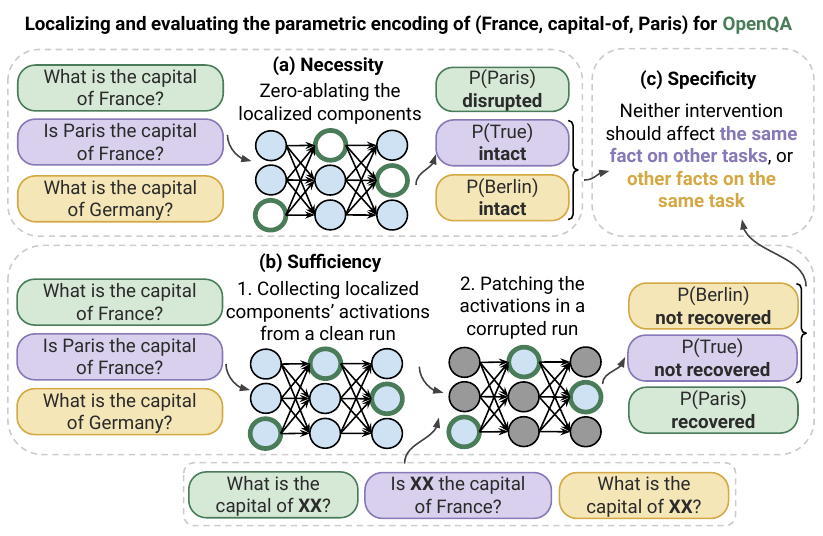}
    \caption{Criteria for localizing and evaluating (fact, task) parametric encodings, illustrated for the encoding of (\texttt{France, capital-of, Paris}) on \textsc{OpenQA}.}
    \label{fig:figure2}
\end{figure}

\textit{Specificity.}\; Intervening on the localized parameters should not affect the model's performance on the same fact under other tasks, or on other facts under the same task. To this end, we add a specificity term to the necessity loss, penalizing interference with non-target pairs:
\begin{equation}
\resizebox{\columnwidth}{!}{$\displaystyle
\begin{aligned}
\mathcal{L}_{\text{spec}}^{\text{nec}}(\mathbf{m})
&= \underbrace{\mathbb{E}_{f' \neq f^*}\!\Big[\text{MSE}\!\bigl(p(f', t^*, \theta \circ \mathbf{m}),\; p(f', t^*, \theta)\bigr)\Big]}_{\text{other facts, same task}} \\
&\quad + \underbrace{\sum_{t' \neq t^*} \text{MSE}\!\bigl(p(f^*, t', \theta \circ \mathbf{m}),\; p(f^*, t', \theta)\bigr)}_{\text{same fact, other tasks}}\end{aligned}
$}
\label{eq:spec_nec_loss}
\end{equation}
The same specificity constraint also applies to sufficiency, requiring that patching the identified components' activations into corrupted prompts for non-target pairs will \emph{not} recover performance. See \S\ref{appendix:params_exp} for the full loss term.
Multi-hop tasks extend the target relation with an additional hop (e.g., \emph{landmark $\to$ country} becomes \emph{landmark $\to$ country $\to$ language}). To ensure the mask does not target the added hop, we add a control chain sharing it (e.g., \emph{capital $\to$ country $\to$ language}) to the retention pool.
\Cref{fig:figure2} illustrates the necessity, sufficiency, and specificity criteria for an example fact.

\textit{Sparsity.}\; The mask should be as sparse as possible.  We apply an L1 penalty to $\mathbf{1}-\mathbf{m}$ (the indicator of selected components), normalized by the total number of components, weighted by $\beta=10.0$. See \S\ref{appendix:params_exp} for implementation details.

\paragraph{Mask training and evaluation}
For each fact, we train one mask per task, sequentially in random order. Components selected by earlier masks are excluded from later ones, resulting in fully disjoint masks.\footnote{Without the disjointness constraint, results are similar and the masks still end up almost disjoint (see \S\ref{appendix:params_results}).}
Since each (fact,task) pair has multiple prompt paraphrases (see \S\ref{sec:behavioral}), all loss terms average over them.
We evaluate masks on necessity, sufficiency and specificity using held-out prompt paraphrases (5 training, 2 evaluation per task). 
The pool of other facts used to evaluate same-task specificity is split into 75\%/25\% train/evaluation, resampled for each target fact.
We average over prompt paraphrases to obtain a per-fact accuracy score, then report the mean and std across facts.

\begin{figure}[t]
    \centering
    \includegraphics[width=\linewidth]{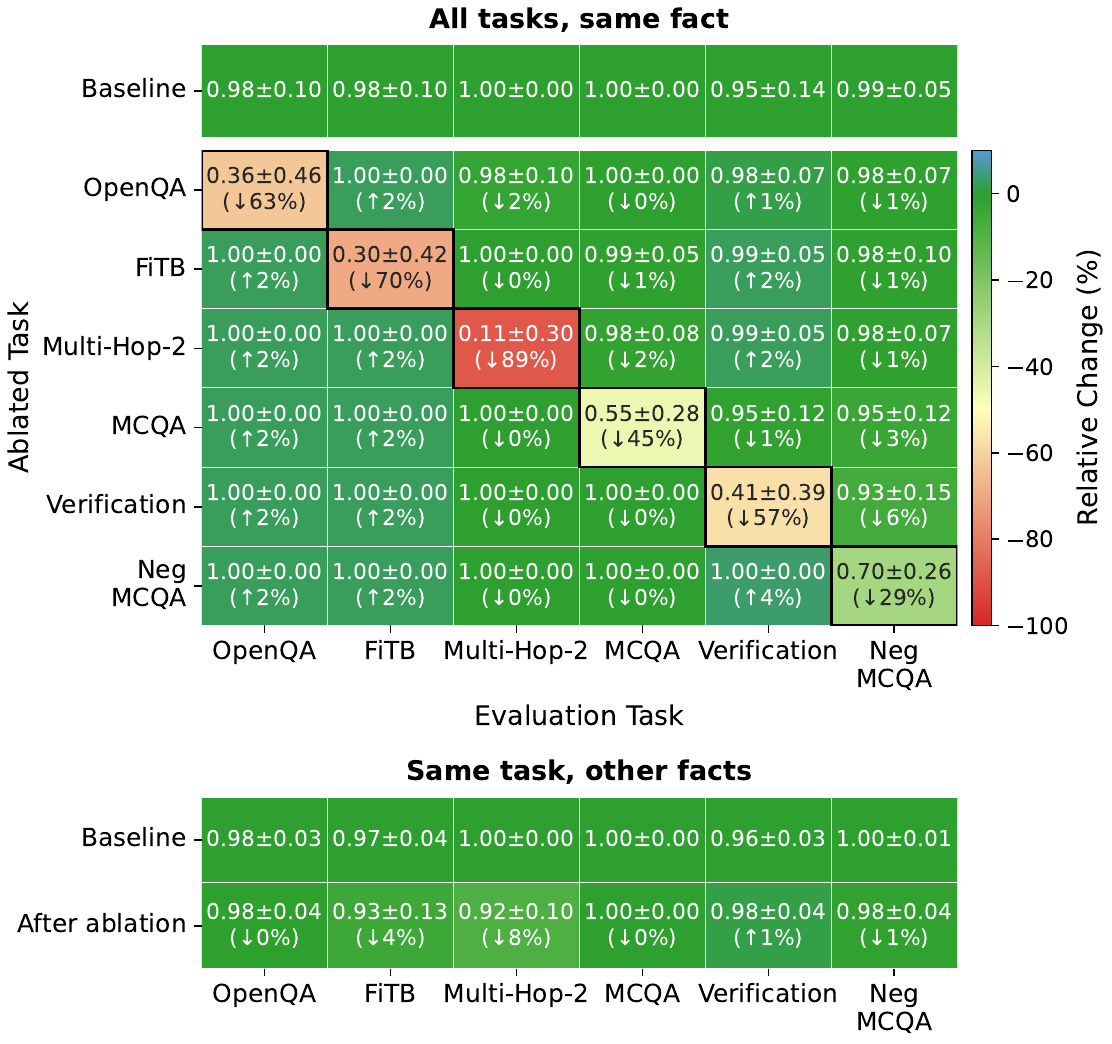}
    \caption{Necessity results for \texttt{(country, official language, language)} on \OlmoSevenB{}. 
    For each fact, we ablate the heads and neurons localized per task. Rows correspond to the ablated task, columns to the evaluation task.
    Values average over facts. 
    Color reflects relative change from baseline (baseline rows pinned green).
    Large diagonal drops confirm necessity; near baseline off-diagonal and bottom-row values specificity.}
    \label{fig:necessity}
\end{figure}

\paragraph{Results}
Across all datasets and models, we observe that individual (fact, task) pairs are supported by distinct, task-specific parameter subsets.
\Cref{fig:necessity} presents representative results for the \texttt{(country, official language, language)} dataset on \OlmoSevenB{} (full results in \S\ref{appendix:params_results}).
Zero-ablating the identified components causes a large performance drop on the targeted (fact, task) pair (diagonal cells: 29\%--89\% relative decrease), while the same fact on other tasks (off-diagonal columns) and other facts on the same task (bottom row) remain largely unaffected (all cells $\leq$8\%). This confirms that the identified subsets are both \emph{necessary} for and \emph{specific} to individual (fact, task) pairs. 
For \emph{sufficiency}, patching the identified components' activations into a corrupted prompt achieves high recovery on the targeted pair (69\%--102\% reconstruction rate); recovery on non-target pairs remains small. The pattern is consistent across other model-dataset combinations.
The 27B model shows similar task-specific pattern, with diagonal necessity drops of 10\%--93\% (see \S\ref{appendix:params_results}).

\textit{Together with the behavioral results, these findings suggest that LMs maintain task-specific parametric encodings of individual facts, instead of drawing from a shared, task-invariant store.}
 
\section{Quantifying cross-task entanglement}
\label{sec:entanglement}
Our previous results show that factual knowledge is often encoded in a task-specific manner. Yet, certain tasks show some degree of overlap; ablating the encoding for one task causes collateral damage on some tasks but not on others, and pairwise co-emergence rates differ between task pairs.
For example, facts acquired via \textsc{FiTB} reliably co-emerge in \textsc{OpenQA}, but facts acquired via \textsc{MCQA} show late or absent acquisition on \textsc{Completion}.
We refer to this overlap as \textit{cross-task entanglement}, and introduce metrics that quantify it from both the behavioral and parametric perspectives. Our results show that task format is a dominant predictor of such entanglement, with discrimination tasks being consistently more entangled than generation tasks.

\subsection{Methodology}
 
\paragraph{Behavioral analysis}
We break down the co-emergence rates from the behavioral experiment (\S\ref{sec:behavioral}) into \emph{directional task pairwise co-emergence rates}.
For each ordered pair of tasks (source~$s$, target~$t$), we measure what fraction of facts that emerged on $s$ co-emerge on~$t$ by the expected step. 
This reveals asymmetries in co-emergence between task pairs.
Implementation details are in \S\ref{appendix:behavioral}. 

\paragraph{Parameter-level entanglement metric}
Our parametric experiment (\S\ref{sec:mechanistic_analysis}) localizes for each (fact, task) pair a sparse parametric encoding necessary and sufficient for the model's performance on that pair.
As \Cref{fig:necessity} shows,
ablating one (fact, task) encoding can degrade other pairs as well. The specificity penalty (Eq.~\ref{eq:spec_nec_loss}) limits this collateral damage, but for some (fact, task) pairs, it constrains how fully the mask can suppress its target (Eq.~\ref{eq:nec_loss}).

We define an entanglement score $\operatorname{Ent}(f,t)$ that measures, for a single (fact, task) encoding, how cleanly it can be ablated \emph{without} affecting other (fact, task) pairs. 
This metric collapses each row of the necessity heatmap into a single number. 
Concretely, for each pair $(f, t)$, we ablate its identified parameters and measure three quantities:
(a) \emph{target drop} $\Delta_{\text{target}}(f, t)$: how much the ablation degrades performance on the targeted pair, (b) \emph{collateral change} $\Delta_{\text{coll}}(f', t)$: the effect on other facts $f' \neq f$ on the same task, and (c) \emph{collateral change} $\Delta_{\text{coll}}(f, t')$: the effect on the same fact under other tasks $t' \neq t$.
The score $\operatorname{Ent}(f,t)$ averages these:
\begin{equation}
\begin{array}{r@{}l}
\resizebox{0.9\columnwidth}{!}{$\displaystyle
\frac{1}{3}\Bigg[\;\bigl(1 - \Delta_{\text{target}}(f,t)\bigr) + \frac{1}{|\mathcal{F}|-1}\!\sum_{f' \neq f}\!\Delta_{\text{coll}}(f',t)
$} \\[6pt]
\resizebox{0.6\columnwidth}{!}{$\displaystyle
\quad + \frac{1}{|\mathcal{T}|-1}\!\sum_{t' \neq t}\!\Delta_{\text{coll}}(f,t')\;\Bigg]
$}
\end{array}
\label{eq:E_score}
\end{equation}
$\operatorname{Ent}(f,t) = 0$ is achieved when the target drop is maximal and collateral damage is zero on both axes. 
This means that the encoding is fully necessary and specific for the targeted pair. Higher values indicate greater entanglement. 
Averaging this score over facts yields a per-task score $\operatorname{Ent}_{\text{task}}(t) = \frac{1}{|\mathcal{F}|} \sum_{f} \operatorname{Ent}(f,t)$, which we report for three models and five datasets. 
Formal definitions are provided in \S\ref{appendix:entanglement}.

\paragraph{Pairwise task entanglement}
To test whether certain task pairs are more entangled than others, we train a separate mask for each directed pair of tasks $(t_A, t_B)$. 
We use the same objective as in \S\ref{sec:mechanistic_analysis}, except that the specificity loss penalizes interference only with $t_B$ (rather than with all other tasks).

\subsection{Findings}
We report the key results for the two analyses below, with full per-task tables and heatmaps provided in \S\ref{appendix:behavioral} (behavioral) and \S\ref{appendix:entanglement} (parametric). 
Pairwise entanglement results are  consistent with the aggregative $\operatorname{Ent}_{\text{task}}$ scores, thus we report them in \S\ref{appendix:entanglement}.
 
\paragraph{Generation tasks are less entangled, discrimination tasks are more entangled}
The parametric results show that discrimination tasks (\textsc{MCQA}, \textsc{Verification}, \textsc{Neg~MCQA}) are markedly more entangled than generation tasks (\textsc{OpenQA}, \textsc{FiTB}, \textsc{Multi-Hop}). 
Aggregated over 15 model-dataset combinations, the mean $\operatorname{Ent}_{\text{task}}$ is $0.21$ for discrimination versus $0.11$ for generation. 
\textsc{Verification} and \textsc{Neg~MCQA} are the most entangled tasks ($0.25$ and $0.24$, respectively), while \textsc{Multi-Hop-2} is the least ($0.08$).

\paragraph{Discrimination tasks are weak sources of cross-task co-emergence}
Among facts that have emerged on a discrimination task before or alongside a given target task, co-emergence rates are 3\%-42\% on non-\textsc{Verification} targets, compared to 40\%-90\% for facts that have emerged on a generation task.
\textsc{Verification} shows higher co-emergence rates overall, which we hypothesize is due to its late emergence in training, but even there discrimination tasks are the weakest sources of co-emergence (63\%-70\%, versus 76\%-94\% for generation sources). This is consistent with discrimination tasks requiring less knowledge of a fact than generation tasks do.

\section{The role of task-specific encodings in chain-of-thought reasoning}
\label{sec:cot_ablation}
In this section, we expand our analysis to examine how task-specific encodings are utilized during generation, focusing on chain-of-thought (CoT) reasoning \citep{wei2022chain}. The mechanistic experiment (\S\ref{sec:mechanistic_analysis}) established that under \emph{direct answering}, where the model produces an answer without intermediate reasoning, different tasks rely on distinct parameter subsets to retrieve the same fact. 
Given that reasoning has been shown to unlock factual knowledge inaccessible to direct answering \citep{gekhman2026thinking, ma2026improving, calderon2026empty}, a natural hypothesis is that reasoning draws part of its power from engaging parametric encodings beyond those tied to the evaluation format.
If this holds, 
then CoT should help the model recover performance lost when a task's localized parameters are ablated, by rerouting through alternative encodings. 
Moreover, if CoT draws on encodings of other tasks, cross-task collateral damages should be larger than under direct answering.
The ablation framework from \S\ref{sec:mechanistic_analysis} lets us test both predictions.
 
\paragraph{Experiment}
We apply the zero-ablations from \S\ref{sec:mechanistic_analysis}, but now compare model accuracy under both direct answering and CoT.
We exclude the multi-hop tasks, whose two-step chains contain bridging knowledge that may confound cross-task attribution, and use facts that meet a baseline CoT performance threshold on all tasks.
We report accuracy averaged across facts before and after ablation, under each setting. To measure cross-task collateral damage, we average the worst-case cross-task drops across facts resulting from encoding ablations.
Additional details are in \S\ref{appendix:cot_ablation}.

\begin{figure}[t]
    \centering
    \begin{subfigure}{\linewidth}
        \centering
        \includegraphics[width=\linewidth]{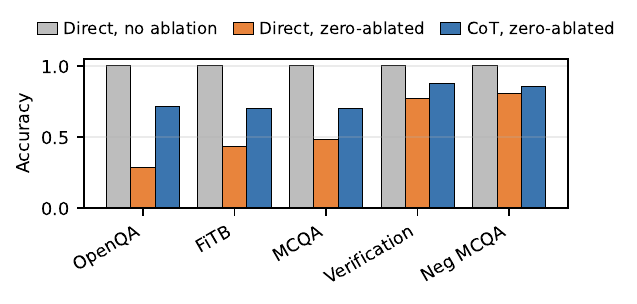}
        \caption{Same-task ablation}
        \label{fig:cot_recovery_own}
    \end{subfigure}
    \par\vspace{6pt}
    \begin{subfigure}{\linewidth}
        \centering
        \includegraphics[width=\linewidth]{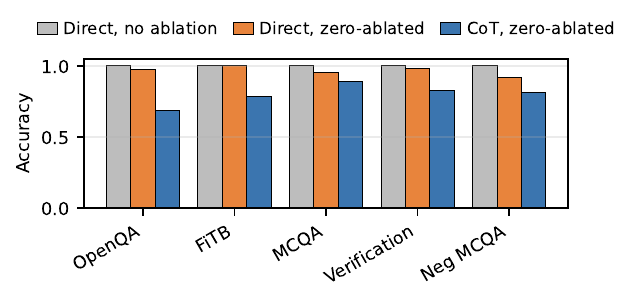}
        \caption{Cross-task ablation}
        \label{fig:cot_recovery_cross}
    \end{subfigure}
    \caption{CoT versus direct answering under zero-ablation on \texttt{(landmark, in-country)} for \Gemma{}, reported as accuracy. (a)~Ablating each (fact, task) pair's own encoding. 
    (b)~For each pair, ablating the other task's encoding causing the largest drop.}
    \label{fig:cot_recovery}
\end{figure}

\paragraph{Results}
\Cref{fig:cot_recovery} presents results for \texttt{(landmark, in-country, country)} on \Gemma{}; other models and datasets show similar patterns (see \S\ref{appendix:cot_ablation}). 
Zero-ablating the localized parameters reduces direct accuracy by 20\%--72\% (varying across tasks), whereas CoT loses only 12\%--30\%, staying closer to the unablated baseline (\Cref{fig:cot_recovery_own}), confirming the predicted recovery under CoT.
When we ablate, per fact, the other task whose encoding most damages each condition, direct accuracy falls by at most 8\% while CoT drops by 11\%--31\% (\Cref{fig:cot_recovery_cross}), confirming the predicted increase in cross-task collateral damage under CoT.
Together, these results support the hypothesis that CoT routes through multiple task-specific encodings.

\section{Related work}
\paragraph{LMs as knowledge bases}
LM parameters encode vast relational knowledge \citep{petroni2019language, roberts-etal-2020-much}, motivating their view as knowledge bases. Several works showed that factual recall is sensitive to query form:
paraphrased prompts yield inconsistent predictions for the same facts \citep{elazar-etal-2021-measuring}, models trained on ``A is B'' fail to infer the reverse \citep{berglund2024reversal}, and knowledge edits often fail to generalize to related queries \citep{cohen2024evaluating}. 
We study this sensitivity along another axis, the task, which changes the operation applied to the fact. We further ask not only what the model retrieves but how it is stored, showing that the same fact is supported by different parameters under different tasks.

\paragraph{Redundancy in factual encodings}
Recent work has suggested that factual knowledge in LMs is not stored in a single location. \citet{bayazit2024discovering,chen2024journey} suggested that different subsets of parameters can encode the same knowledge, and \citet{chen-etal-2025-cracking} showed that such redundancies contribute to robustness under input perturbations. 
\citet{pham2026knowledge} localized \emph{conflicting} parametric encodings of the same facts arising from inconsistent pretraining data. 
\citet{feng2025extractive} demonstrated that facts learned during finetuning are stored redundantly across layers, 
supporting different multi-hop reasoning tasks. 
We find that task-specific storage is not limited to finetuned multi-hop knowledge but applies broadly to pretrained facts across diverse task formats.

\paragraph{Cross-lingual transfer} 
Language is another dimension along which the same fact can be queried in different surface forms. \citet{blum2025beyond} showed that models can develop either unified or separate representations of the same facts across languages; \citet{liu-etal-2025-tracing} traced cross-lingual factual recall across OLMo-7B checkpoints, finding it largely predicted by fact frequency rather than transfer from other languages;  \citet{bandarkar2026knowledge} leveraged cross-lingual inconsistency to identify knowledge-critical experts in MoE models. 
Overall, whether facts are stored in shared or language-specific parameters remains largely unanswered. 
Our work addresses the analogous question along the task dimension, showing that storage is organized, at least in part, by task format.

\paragraph{Cognitive parallels}
Findings in cognitive science show that access to memory depends on the relationship between how information is encoded at study and how it is later probed \citep{tulving1973encoding, MORRIS1977519}. 
Our work echoes this principle in LMs, showing that what the model knows and how it is asked are intertwined in the parameters. Whether this arises from the format in which facts are encountered during training is an interesting direction for future work.

\FloatBarrier

\section{Conclusion and discussion}
We investigate the task-invariance property expected of knowledge bases in LMs, and find it is largely violated. Behaviorally, facts acquired on one task frequently fail to co-emerge on others during training. Our mechanistic analysis offers an explanation, revealing distinct parameters underlying different tasks for the same fact. 
This separation varies by task, with discrimination-task encodings consistently more entangled with other (fact, task) pairs than those for generation tasks.
Our experiments further suggest that CoT reasoning draws on encodings beyond those tied to the evaluation task, offering a mechanistic account for how reasoning unlocks otherwise inaccessible knowledge.

\textbf{Practical implications} Our findings have implications for model development and evaluation.
Knowledge editing and unlearning methods that target a single task format may leave the targeted fact intact on others \citep{Ye2025LLMUS}, or recoverable through chain-of-thought prompting (\S\ref{sec:cot_ablation}), and evaluations that probe only one format provide an incomplete picture of what the model encodes.
More fundamentally, training schemes that encourage task-invariant encodings may yield more trustworthy and controllable models; our finding that CoT reasoning bridges across task-specific encodings suggests incorporating intermediate reasoning during training as a candidate scheme.
Task-specific encodings may also carry a capacity cost that task-invariant storage would avoid.


\section*{Limitations}
In the behavioral experiment (\S\ref{sec:behavioral}), we observe training at periodic checkpoints, so the exact step at which a fact emerges on a given task is only approximate. 
That said, our analysis relies on the relative ordering of when facts emerge across tasks rather than on exact step counts, so this approximation is unlikely to affect our conclusions.

All of our experiments focus on relational knowledge expressible as (subject, relation, object) triplets, and the mechanistic (\S\ref{sec:mechanistic_analysis}) and CoT (\S\ref{sec:cot_ablation}) experiments further restrict to facts that meet a high baseline performance threshold on all tasks. Our findings therefore primarily describe well-known relational knowledge. 
Notably, a great deal of knowledge, and specifically the kind invoked by the knowledge base analogy, is relational factual knowledge. We therefore view this as a fairly general setting.

It is important to distinguish our localized task-specific encodings from simple redundant encodings, where the same fact is stored in multiple interchangeable locations. Our discovery of encodings that are necessary, sufficient, and specific to (fact, task) pairs reveals that factual storage is organized, at least in part, by task format, but does not characterize \emph{how many} parameter subsets encode a given fact within or across tasks, which we leave for future work. 
One could, for example, apply our framework iteratively within a task, excluding previously localized components and asking whether additional necessary and sufficient subsets can be found for the same pair.
Relatedly, our sufficiency results (\S\ref{sec:mechanistic_analysis}) show that patching a localized subset recovers performance on a (fact,task) pair, but this does not mean the subset ``fully'' encodes the fact, as other parameters may contribute in ways our masks do not capture.

\section*{Acknowledgments}
We thank Yoav Gur-Arieh for providing valuable feedback. This research was supported
in part by grants 1083/24 and 2247/23 from The Israel Science Foundation.
\bibliography{custom}

\appendix

\section{Datasets and prompt construction}
\label{appendix:datasets}
Our experiments rely on five relational datasets (\S\ref{sec:behavioral}). Here we provide additional details on dataset sources, fact filtering criteria, prompt templates, and distractor selection.

\paragraph{Relational datasets}
We use five relational datasets, each consisting of (subject, relation, object) triplets.
\texttt{(landmark, in-country, country)}, \texttt{(person, plays-instrument, instrument)} and \texttt{(company, HQ-in-city, city)} are sourced from \citet{hernandez2024linearity};
\texttt{(country, capital-of, city)} and \texttt{(country, official language, language)} were obtained using Wikidata SPARQL queries \citep{10.1145/2629489}. 
Countries with more than one official language in the \texttt{(country, official language, language)} dataset were filtered out. All the datasets are in English.

Three of the datasets include multi-hop reasoning tasks, each paired with a control task that shares one hop of the multi-hop chain with the target relation. The control is used in the specificity loss to ensure the mask targets the relation rather than the shared hop.
\Cref{tab:multihop_relations} presents the multi-hop relations and their controls.
For \OlmoSevenB{}, we use a variant of the \texttt{(person, plays-instrument, instrument)} dataset with uncapitalized object names (e.g., \texttt{guitar} rather than \texttt{Guitar}), since the baseline unintervened performance of the model is substantially better on the uncapitalized version.

\begin{table*}[t]
\caption{Multi-hop chains and their corresponding control chains.}
\label{tab:multihop_relations}
\centering
\small
\resizebox{\textwidth}{!}{%
\begin{tabular}{lll}
\toprule
\textbf{Dataset} & \textbf{Multi-hop relation} & \textbf{Control relation} \\
\midrule
\texttt{(landmark, in-country, country)}           & landmark $\to$ country $\to$ language & capital $\to$ country $\to$ language \\
\texttt{(country, capital-of, city)}       & landmark $\to$ country $\to$ capital  & landmark $\to$ country $\to$ language \\
\texttt{(country, official language, language)}  & landmark $\to$ country $\to$ language & landmark $\to$ country $\to$ capital  \\
\bottomrule
\end{tabular}%
}
\end{table*}
 
\paragraph{Multi-hop intermediate entities selection}
The multi-hop chains we use have \texttt{(landmark, in-country, country)} as the first hop. Since multiple landmarks exist per country in the dataset, we selected one for each country using a model-based procedure with Gemma-3-1B IT \citep{Kamath2025Gemma3T}. 
For the \texttt{(landmark, in-country, country)} dataset, we evaluated each candidate landmark on the \textsc{Multi-Hop-1} prompt paraphrases and selected the landmark with highest mean probability assigned by the model to the correct answer. 
For datasets with \textsc{Multi-Hop-2} tasks (\texttt{(country, capital-of, city)}, \texttt{(country, official language, language)}), we evaluated each candidate landmark on both the main and control \textsc{Multi-Hop-2} paraphrases and selected the landmark with the highest average probability across the two. 

\paragraph{Prompt templates}
For each (dataset, task) pair, we curated 10 prompt paraphrases, using LLM suggestions as a starting point. The same base paraphrases are shared between \textsc{MCQA} and \textsc{OpenQA}, and between \textsc{FiTB} and \textsc{Completion}.
The prompts of all tasks aside from \textsc{Completion} begin with a task instruction (e.g., \textit{``Answer the following question.''}), include a formatting guideline (\textit{``Your response should be formatted as: `Answer: \{your answer\}'\,''}), and end with the task-specific query. The \textsc{Completion} task does not include instructions or guidelines, since it evaluates next-token prediction on plain sentences. \Cref{fig:prompt-examples} shows representative prompts for different tasks, all for the fact \texttt{(France, capital-of, Paris)} from the \texttt{(country, capital-of, city)} dataset.

\paragraph{Distractors selection}
Distractor answers for \textsc{MCQA} (3 incorrect choices), \textsc{Neg~MCQA} (1 incorrect choice), and \textsc{Verification} (the object in the false statement) are sampled uniformly from the full set of objects in the dataset before any filtering.

\begin{figure}[t]
\begin{tcolorbox}[title=Example Prompts, fonttitle=\bfseries,
    left*=10pt, right*=10pt, top=2pt, bottom=2pt,
    colback=white!10!white, colframe=black!75!black]
\small
All examples are for the fact \texttt{(France, capital-of, Paris)}. Target completion is evaluated at the position following the answer prefix (or at the sentence end for \textsc{Completion}).

\tcbsubtitle{\textsc{OpenQA} \hfill Target: \texttt{Paris}}
\textit{Answer the following question. Your response should be formatted as: `Answer: \{your answer\}'. Question: What is the capital city of France?}

\tcbsubtitle{\textsc{FiTB} \hfill Target: \texttt{Paris}}
\textit{Complete the missing word. Your response should be formatted as: `Answer: \{your answer\}'. Sentence: The capital city of France is \_\_.}

\tcbsubtitle{\textsc{Completion} \hfill Target: \texttt{Paris}}
\textit{The capital city of France is} \hfill(no answer prefix)

\tcbsubtitle{\textsc{MCQA} \hfill Target: \texttt{1)}}
\textit{Answer the following question. \ldots\ Question: What is the capital city of France? Choices: 1) Paris~~2) Hanoi~~3) Bras\'ilia~~4) Sofia}

\tcbsubtitle{\textsc{Neg MCQA} \hfill Target: \texttt{1)}}
\textit{Answer the following question. \ldots\ Question: What is NOT the capital city of France? Choices: 1) Dublin~~2) Paris}

\tcbsubtitle{\textsc{Verification} \hfill Target: \texttt{True}}
\textit{Determine whether the following statement is True or False. \ldots\ Statement: The capital city of France is Paris.}

\tcbsubtitle{\textsc{Multi-Hop-2} \hfill Target: \texttt{Paris}}
\textit{Answer the following question. \ldots\ Question: What is the name of the capital city of the country containing the landmark called The Bourg-la-Reine?}
\end{tcolorbox}
\caption{Example prompts for each task, shown for the fact (France, capital-of, Paris).}
\label{fig:prompt-examples}
\end{figure}

\section{Behavioral experiment: additional details}
\label{appendix:behavioral}

In \S\ref{sec:behavioral} we tested the co-emergence hypothesis by tracking facts acquisition across tasks over training checkpoints. Here, we provide additional implementation details and the full co-emergence rates.

\subsection{Additional implementation details}
 
\paragraph{Checkpoints}
The 105 checkpoints span three pretraining stages and three post-training models: 100 stage~1 (pretraining) checkpoints, selected at approximately uniform intervals (every ${\sim}14$k training steps); the final checkpoints of stage~2 (midtraining) and stage~3 (long context); the ``main'' checkpoints of three post-training models: SFT, DPO and Instruct.
 
\paragraph{Prompt formatting}
For all tasks except \textsc{Completion} we append an answer prefix. The prefix is \textit{``Answer:~''} (with trailing space) for the multiple-choice tasks (\textsc{MCQA}, \textsc{Neg~MCQA}) and \textit{``Answer:''} (without trailing space) for all others. On pretraining checkpoints the prefix is appended to the prompt after a newline. On post-training checkpoints the prompt is first wrapped in the model's default chat template, and the prefix is appended directly afterward.

\paragraph{Directional co-emergence rates: implementation details}
The analysis in \S\ref{sec:behavioral} attributes each fact to its earliest source across all tasks other than $t$, through $E(f,\bar{t})$. To study task pairwise co-emergence rates, we fix an ordered pair of tasks (source $s$, target $t$) and replace $E(f,\bar{t})$ with $E(f,s)$. We observe every fact that emerges on $s$ no later than on $t$ ($E(f,s) \leq E(f,t)$), and count the observation as \emph{consistent} when $E(f, t) \leq \max\!\bigl(E(f, s),\, E(\cdot, t)\bigr)$. The directional co-emergence rate of $(s,t)$ is the fraction of these that are consistent, with the same liveness exclusions as in \S\ref{sec:behavioral}. 

\paragraph{Larger model}
We repeat the analysis on \OlmoThreeThirtyTwoB{} over 102 checkpoints. These include 97 stage~1 checkpoints at ${\sim}7$k-step intervals, the final checkpoints of stage~2 (midtraining) and stage~3 (long-context), and the ``main'' checkpoints of the SFT, DPO, and Instruct post-training models. 
As in \S\ref{sec:behavioral}, all datasets are down-sampled to 47 facts each (matching the size of the smallest dataset), yielding 235 facts. Of the 1{,}410 (fact, task) pairs, 336 are skipped, leaving 1{,}074 evaluated. 
144 pairs were skipped because $t$ was the task on which the fact first emerged, 133 because the target task was no longer competent at the expected step (RHS of Eq.~\ref{eq:co_emergence}), and 59 because the fact was no longer retrieved on any other task. 

\subsection{Results}

\begin{figure*}[t]
    \centering
    \includegraphics[width=\linewidth]{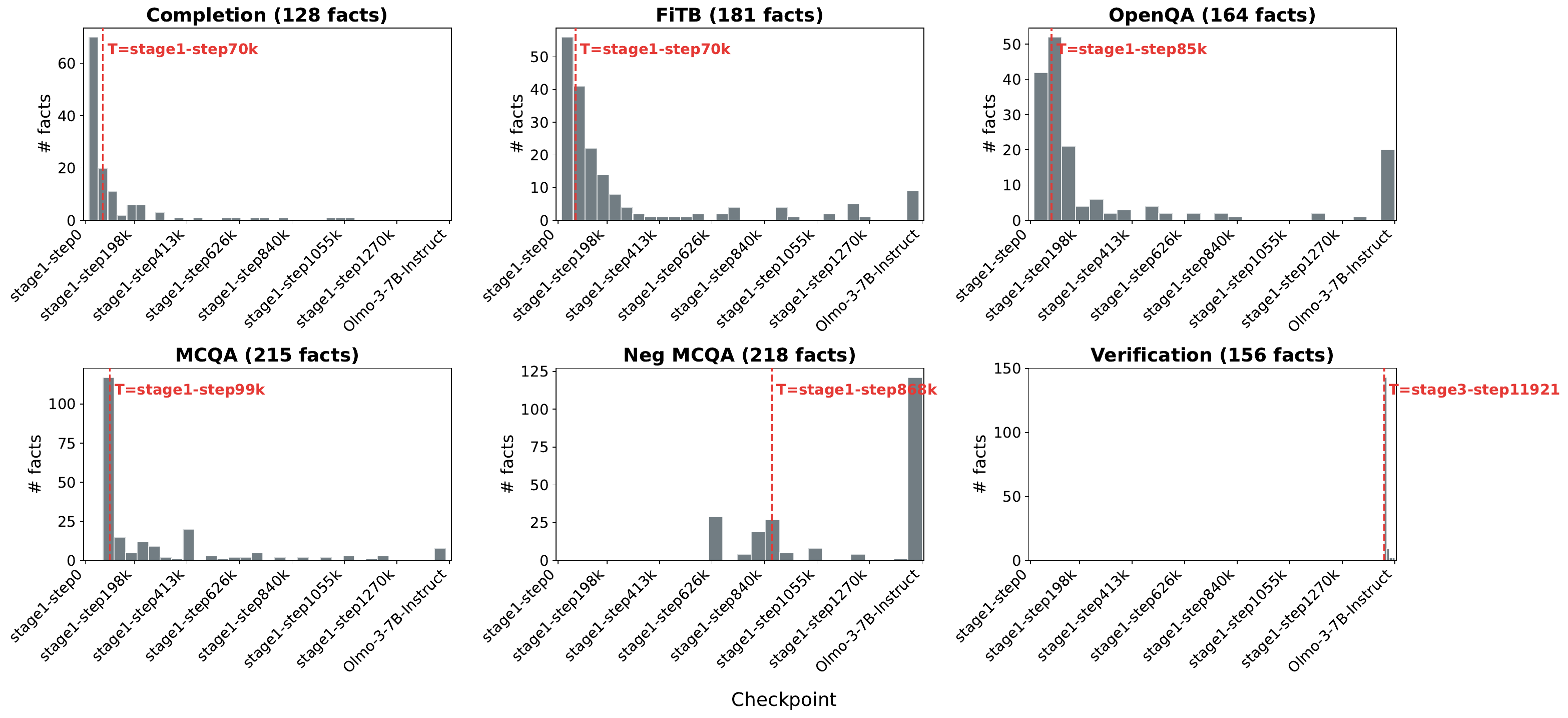}
    \caption{Distribution of fact emergence steps per task. Red dashed lines mark task emergence ($\geq$25\% of facts known).}
    \label{fig:emergence-per-task-appendix}
\end{figure*}

\paragraph{Task emergence}
\Cref{fig:emergence-per-task-appendix} shows per-task histograms of fact emergence steps across training. Generation tasks become competent early in training, with \textsc{Completion} and \textsc{FiTB} emerging at 70k steps, and \textsc{OpenQA} at 85k steps. Discrimination tasks emerge later. \textsc{MCQA} emerges at 99k steps, while \textsc{Neg MCQA} and \textsc{Verification} emerge much later, at 868k steps and at the long-context checkpoint (pretraining stage 3), respectively.

\paragraph{Skipped pairs} Of the 1{,}380 (fact, task) pairs, 349 are excluded, leaving 1{,}031 evaluated.
Of the excluded pairs, 174 were dropped because $t$ was the task on which the fact first emerged, 113 because the target task was no longer competent at the expected step (RHS of Eq.~\ref{eq:co_emergence}), and 62 because the fact was no longer retrieved on any other task.

\paragraph{Per-task co-emergence rates}
\Cref{tab:hit-rate-updated} breaks the co-emergence rates down by task (for each target task $t$, the fraction of evaluated facts that co-emerge on $t$ by the expected step). Generation tasks generally have lower rates, with \textsc{Completion} at $31.9\%$, \textsc{FiTB} at $47.4\%$, and \textsc{OpenQA} at $49.7\%$. Discrimination tasks generally have higher rates, with \textsc{MCQA} at $72.3\%$ and \textsc{Verification} at $65.3\%$. \textsc{Neg MCQA} is the exception at $41.9\%$.

\begin{table}[t]
\caption{Per-task co-emergence rates. $N$ is the number of facts tested per task.}
\label{tab:hit-rate-updated}
\centering
\small
\begin{tabular}{lrc}
\toprule
\textbf{Task} & \textbf{N} & \textbf{co-emergence rate (\%)} \\
\midrule
\textsc{MCQA} & 148 & 72.3 \\
\textsc{Verification} & 219 & 65.3 \\
\textsc{OpenQA} & 163 & 49.7 \\
\textsc{FiTB} & 175 & 47.4 \\
\textsc{Neg MCQA} & 191 & 41.9 \\
\textsc{Completion} & 135 & 31.9 \\
\midrule
\textbf{Overall} & \textbf{230} & \textbf{52.1} \\
\bottomrule
\end{tabular}
\end{table}

\begin{figure*}[t]
    \centering
    \includegraphics[width=\linewidth]{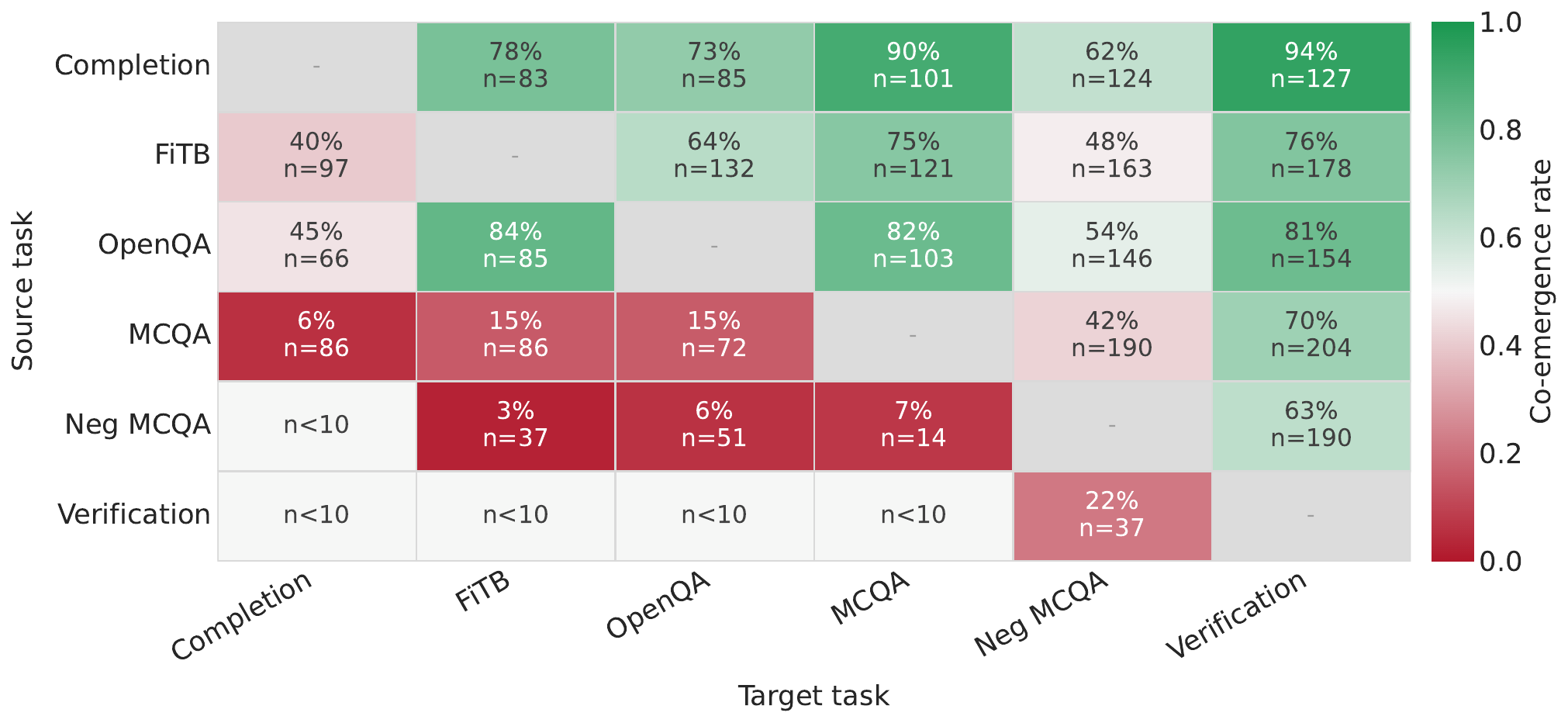}
    \caption{Directional co-emergence rates on \OlmoThreeSevenB{}. Each cell reports the co-emergence rate, with pair count $n$.}
\label{fig:pairwise-directional-heatmap}
\end{figure*}

\begin{figure*}[t]
    \centering
    \captionsetup[subfigure]{skip=2pt}
    \begin{subfigure}{0.95\linewidth}
        \centering
        \includegraphics[width=\linewidth]{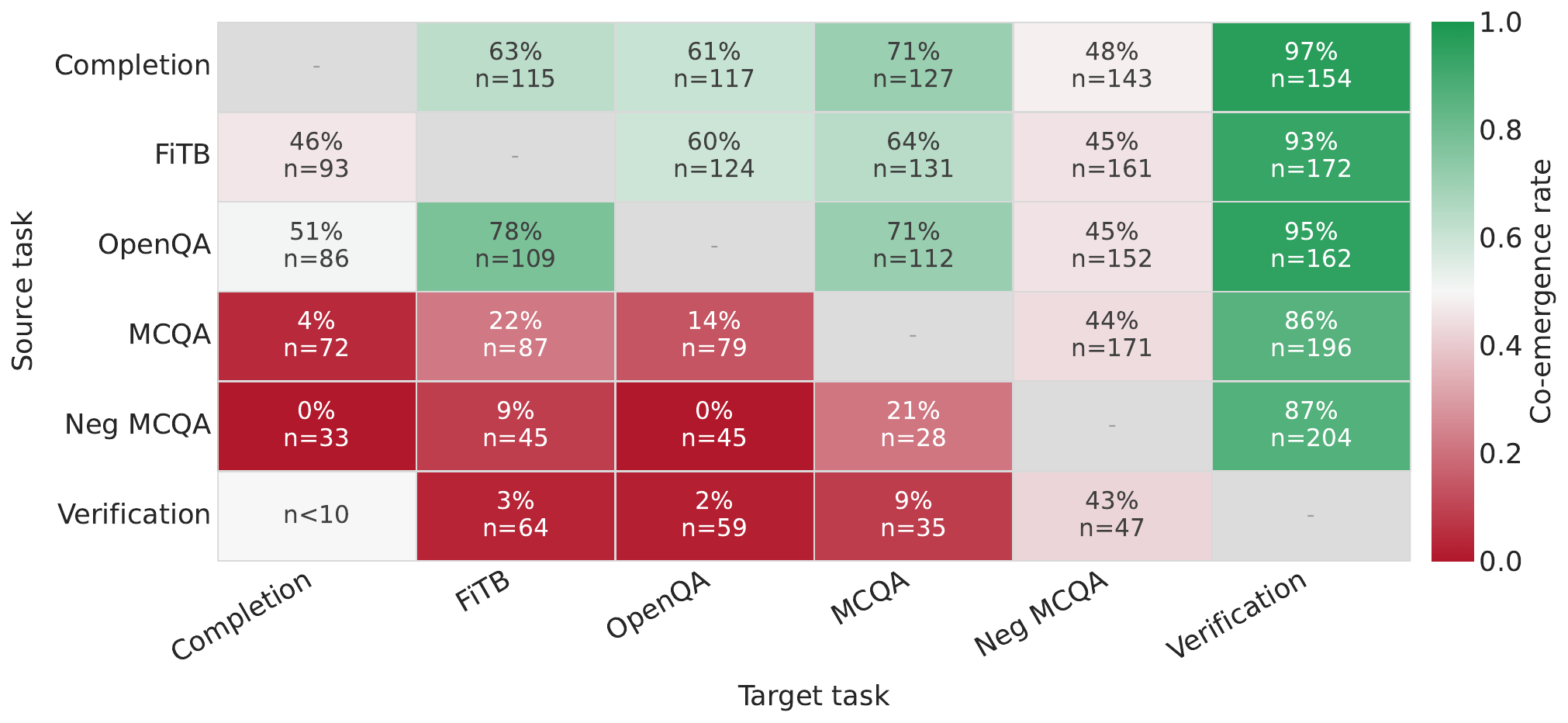}
        \caption{$\theta = 0.4$}
        \label{fig:pairwise-directional-t04}
    \end{subfigure}
    \\[2pt]
    \begin{subfigure}{0.95\linewidth}
        \centering
        \includegraphics[width=\linewidth]{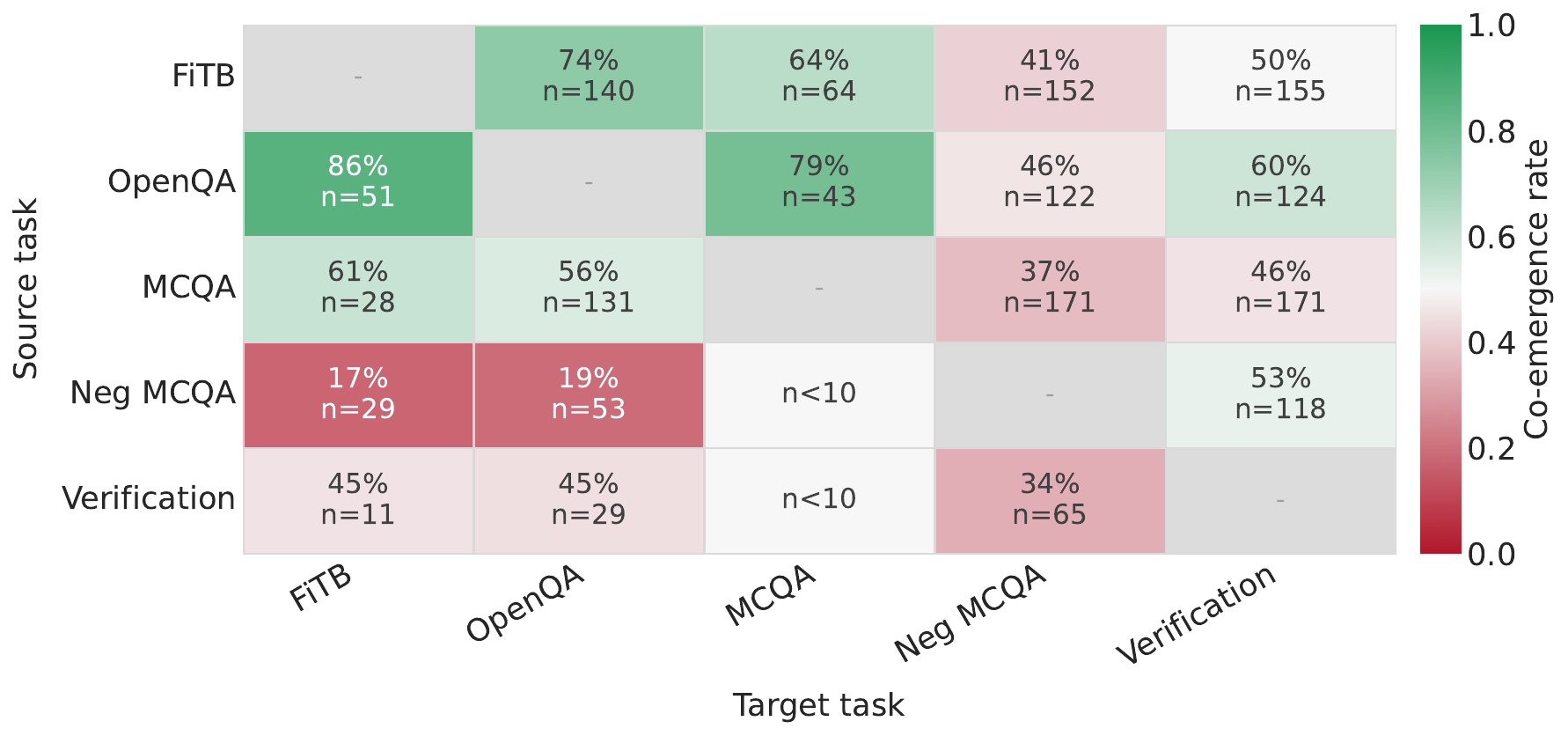}
        \caption{$\theta = 0.8$}
        \label{fig:pairwise-directional-t08}
    \end{subfigure}
    \\[2pt]
    \begin{subfigure}{0.95\linewidth}
        \centering
        \includegraphics[width=\linewidth]{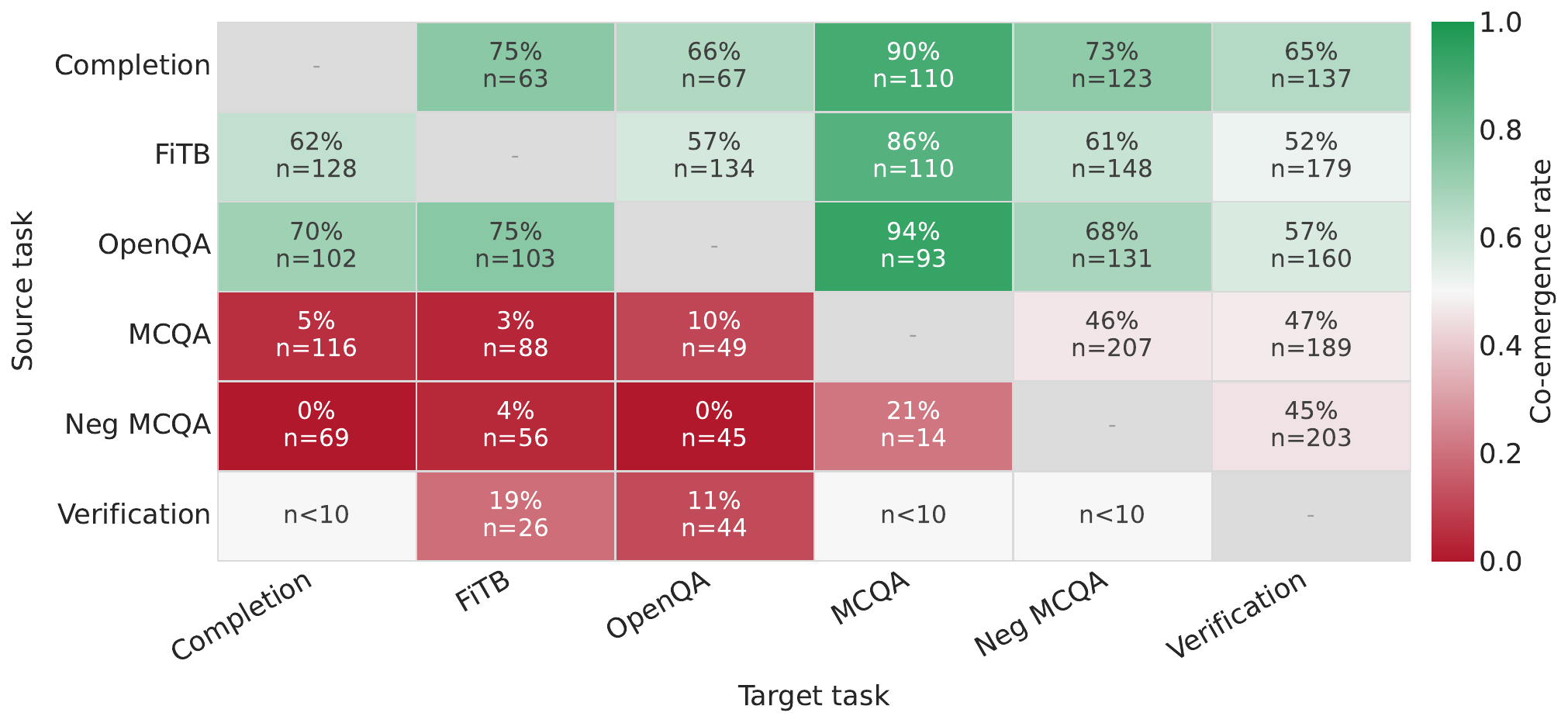}
        \caption{\OlmoThreeThirtyTwoB{}, $\theta = 0.6$}
        \label{fig:pairwise-directional-32b}
    \end{subfigure}
    \caption{Directional co-emergence rates. (a,~b)~\OlmoThreeSevenB{} under a looser ($\theta = 0.4$) and stricter ($\theta = 0.8$) threshold, complementing the $\theta = 0.6$ rates in \Cref{fig:pairwise-directional-heatmap}. (c)~\OlmoThreeThirtyTwoB{} at $\theta = 0.6$. Each cell reports the co-emergence rate, with pair count $n$. At $\theta = 0.8$, \textsc{Completion} does not reach task emergence and is omitted.}
    \label{fig:pairwise-directional-thresholds}
\end{figure*}

\paragraph{Directional co-emergence rates}
\Cref{fig:pairwise-directional-heatmap} presents the full directional pairwise co-emergence rates. 
Generation tasks are reliable sources of co-emergence, with facts emerging on \textsc{Completion} co-emerging on other tasks at rates of $62\%$-$94\%$, and \textsc{OpenQA} and \textsc{FiTB} showing similar patterns  ($45\%$-$84\%$ and $40\%$-$76\%$, respectively). 
Discrimination tasks are weaker sources, with facts acquired via \textsc{MCQA} or \textsc{Neg MCQA} co-emerging on tasks other than \textsc{Verification} only $3\%$-$42\%$ of the time. 
The exception is \textsc{Verification} as a target, which shows high co-emergence rates regardless of the source task. 
Generally similar directional structure holds at looser and stricter thresholds ($\theta = 0.4$ and $\theta = 0.8$) and for \OlmoThreeThirtyTwoB{} (\Cref{fig:pairwise-directional-thresholds}).
 
\section{Fact-task interaction test: additional details}
\label{appendix:interaction}

In \S\ref{sec:behavioral} we tested the task-invariant store hypothesis by formalizing it as an additive decomposition of the log-probability of a correct answer into a fact term and a task term, and found that the interaction is significant. Here we give additional implementation details and results.

\paragraph{Hypothesis and model}
Task-invariance predicts that the log-probability of a correct answer decomposes additively into a fact term and a task term. 
For each (fact, task) cell we take the log-probability of the correct answer's first token, $y_{f,t,p}$, and fit a two-way ANOVA
\begin{equation}
y_{f,t,p} = \mu + \alpha_f + \beta_t + \gamma_{f,t} + \varepsilon_{f,t,p},
\end{equation}
where $p$ denotes prompt paraphrases, that serve as replications within each cell, and $\mu$ is the grand mean (the average log-probability across all cells).
Task-invariance corresponds to $\gamma_{f,t}=0$ (the no-interaction model). We test $\gamma_{f,t}$ with an $F$-test against the within-cell (paraphrase) variance, and report each term's effect size as its share of the total variance, $\eta^2=\mathrm{SS}/\mathrm{SS}_{\text{total}}$.

\paragraph{Data and filtering}
We use the same facts, tasks, checkpoints, and filtering as the behavioral experiment (\S\ref{sec:behavioral}): 230 facts (46 per dataset), six tasks on the base-model checkpoints and five on the post-training checkpoints (\textsc{Completion} dropped, as it does not fit naturally within chat templates), 10 paraphrases per (fact, task) cell, across the 105 checkpoints. 
For \textsc{MCQA} and \textsc{Neg~MCQA} we average over the rotations of the correct answer's position, and for \textsc{Verification} we average over the true and false statements, so each paraphrase contributes one observation.

\paragraph{Experimental setting}
We run the same ANOVA and $F$-test in three settings.
\begin{enumerate}
\item \textbf{Per-checkpoint, unnormalized}, on the raw correct-answer probability, where different chance levels across tasks are absorbed by $\beta_t$.
\item \textbf{Per-checkpoint, normalized}, on the chance-normalized probability (subtract the task's chance level and rescale to $[0,1]$, as in \S\ref{sec:behavioral}).
\item \textbf{Global (stage 1)}, a single ANOVA on the chance-normalized probability, pooled over all 100 pretraining stage-1 checkpoints (before midtraining and long-context), with the noise estimated within each (checkpoint, fact, task) cell and $\gamma_{f,t}$ is the checkpoint-averaged interaction.
This asks whether a consistent interaction persists across pretraining.
\end{enumerate}
We add a small constant ($\epsilon=1e-7$) to every probability before taking the log.

\paragraph{Results}
The (fact,task) interaction is significant at every checkpoint ($p\approx0$); \Cref{tab:interaction} gives the normalized decomposition across training. 
Early in pretraining the variance is dominated by the task main effect ($0.84$ at step ${\sim}14$k), but as facts are learned and performance across tasks improves, this term drops (to $0.03$ in the final model) while the fact, interaction, and noise shares rise; the interaction climbs (unevenly) from ${\approx}0.04$ to $0.10$ across stage~1 and reaches $0.21$-$0.23$ in the post-training models.
On unnormalized probabilities the interaction also accounts for a large share of the variance ($0.41$ in the final model).
In the global (stage-1) test, the interaction is again significant ($p\approx0$) with $\eta^2=0.023$ of the pooled total. Overall, these results confirm that the data does not support a task-invariant model.

 \begin{table}[t]
\caption{Variance decomposition of the (fact, task) ANOVA test at different checkpoints. Columns are the share of total variance ($\eta^2$) attributed to the task main effect, fact main effect, (fact, task) interaction, and within-cell paraphrase noise. Every checkpoint rejects the no-interaction model ($p\approx0$).}
\label{tab:interaction}
\centering
\small
\begin{tabular}{lcccc}
\toprule
\textbf{Checkpoint} & \textbf{Task} & \textbf{Fact} & \textbf{Inter.} & \textbf{Noise} \\
\midrule
Stage 1 (early, ${\sim}$14k) & 0.84 & 0.03 & 0.04 & 0.09 \\
Stage 1 (final) & 0.60 & 0.10 & 0.10 & 0.19 \\
Stage 2 (midtraining) & 0.36 & 0.13 & 0.12 & 0.39 \\
Stage 3 (long context) & 0.12 & 0.22 & 0.18 & 0.48 \\
SFT & 0.01 & 0.32 & 0.21 & 0.46 \\
DPO & 0.04 & 0.35 & 0.22 & 0.40 \\
Instruct (final) & 0.03 & 0.35 & 0.23 & 0.39 \\
\bottomrule
\end{tabular}
\end{table}

\section{Mechanistic analysis: additional details}
\label{appendix:params_exp}

In \S\ref{sec:mechanistic_analysis} we presented the localization framework used to identify parametric encodings of (fact,task) pairs. Here we provide additional details on filtering, mask optimization, and evaluation metrics, as well as the complete necessity and sufficiency results for all models and datasets.

\begin{table*}[t]
\caption{Tasks available per dataset in the mechanistic analysis. All datasets include \textsc{OpenQA}, \textsc{FiTB}, \textsc{MCQA},
\textsc{Neg~MCQA}, and \textsc{Verification}. Multi-hop tasks depend
on the availability of intermediate entities. Cell entries are
formatted as \#trained~/~\#filter-passing facts.
$^{\dag}$Due to compute budget, masks are trained on a random subset
of the filter-passing pool; the subset is taken as a contiguous
prefix of the dataset, whose fact ordering is a deterministic
random permutation set at dataset construction.}
\label{tab:tasks_per_dataset}
\centering
\resizebox{\textwidth}{!}{%
\begin{tabular}{lc@{\hspace{6pt}}c@{\hspace{6pt}}ccc}
\toprule
& \multicolumn{3}{c}{\textbf{\# Target facts / \# Total facts}} & & \\
\cmidrule(lr){2-4}
\multirow{2}{*}{\textbf{Dataset}} 
  & \multirow{2}{*}{\Gemma} 
  & \multirow{2}{*}{\OlmoSevenB} 
  & \multirow{2}{*}{\OlmoThirteenB} 
  & \textbf{Multi-} & \textbf{Multi-} \\
& & & & \textbf{Hop-1} & \textbf{Hop-2} \\
\midrule
\texttt{(landmark, in-country, country)}            & 38/38   & 17/17                  & 23/23   & \checkmark & ---        \\
\texttt{(country, capital-of, city)}             & 32/32   & 23/23                  & 28/28   & ---        & \checkmark \\
\texttt{(country, official language, language)}   & 33/33   & 22/22                  & 29/29   & ---        & \checkmark \\
\texttt{(company, HQ-in-city, city)}$^{\dag}$  & 51/317  & 34/92                  & 32/172  & ---        & ---        \\
\texttt{(person, plays-instrument, instrument)}$^{\dag}$ & 25/133  & 25/50 & 25/70   & ---        & ---        \\
\bottomrule
\end{tabular}%
}
\end{table*}

\subsection{Fact filtering and paraphrase selection}
\label{appendix:dataset_construction}
  
\paragraph{Paraphrase filtering and splitting}
Before training masks, we evaluate the model's performance across all 10 paraphrases for every (dataset, task) pair and discard the 3 templates for which the model's performance is the lowest. 
The remaining 7 paraphrases are split into 5 training and 2 evaluation templates. This split is per (model, dataset) pair and is fixed across all facts for this pair. 
 
\paragraph{Fact filtering}
To ensure we target facts the model can retrieve for all tasks, we filtered out facts for which the model's performance is below a task-specific threshold in any task. The thresholds are: 0.85 for \textsc{MCQA}, \textsc{Neg~MCQA}, and \textsc{Verification}; 0.75 for \textsc{OpenQA}, \textsc{FiTB}, and \textsc{Multi-Hop} tasks. \Cref{tab:tasks_per_dataset} lists the tasks and the number of facts retained per dataset and model after filtering.
 
\subsection{Mask optimization details}
\label{appendix:mask_optimization}
 
\paragraph{Hook placements}
Our masks target attention heads and MLP neurons. Interventions are implemented via PyTorch forward hooks \citep{paszke2019pytorch} on HuggingFace Transformers models \citep{wolf2019transformers}.
To mask individual \textbf{attention heads}, we register a forward pre-hook on the output projection $W_O$ that multiplies each head's $d_{\text{head}}$-dimensional slice by the
corresponding scalar mask value ($0$ or $1$).
Recall that gated MLPs \citep{shazeer2020glu} use three weight matrices. Given input $\mathbf{x}$, the MLP computes an intermediate representation $\mathbf{h} = \text{act}(W_{\text{gate}}\,\mathbf{x}) \odot W_{\text{up}}\,\mathbf{x} \in \mathbb{R}^{d_{\text{mlp}}}$, then projects back via $W_{\text{down}} \in \mathbb{R}^{d_{\text{model}} \times d_{\text{mlp}}}$: $\text{MLP}(\mathbf{x}) = W_{\text{down}}\,\mathbf{h}$. 
To mask individual \textbf{MLP neurons}, we register a forward pre-hook on $W_{\text{down}}$ that multiplies each entry of $\mathbf{h}$ independently by its corresponding mask value. Since zeroing a factor zeros the product, this is equivalent to zeroing $\text{act}(W_{\text{gate}}\,\mathbf{x})$ at the corresponding entries, which is the formulation we use at evaluation.
 
\paragraph{Sparsity objective}
In practice, we optimize separate sub-masks for attention heads ($\mathbf{m}_H$) and MLP neurons ($\mathbf{m}_N$), each with its own normalized L1 penalty term. 
This is because the number of neurons $|N|$ is roughly two orders of magnitude larger than the number of attention heads $|H|$. 
The full sparsity term in Eq.~\ref{eq:mask_objective} is therefore:
\begin{equation}
\mathcal{L}_{\text{spar}}(\mathbf{m}) = \frac{1}{|H|}\|\mathbf{1}-\mathbf{m}_H\|_1 + \frac{1}{|N|}\|\mathbf{1}-\mathbf{m}_N\|_1
\end{equation}

\paragraph{Necessity loss for discrimination tasks}
For discrimination tasks (\textsc{MCQA}, \textsc{Neg~MCQA}, \textsc{Verification}), Eq.~\ref{eq:nec_loss} includes an additional MSE term that drives the aggregated probability of the distractors up to $1-\tau$, so that the ablation changes the model's answer rather than disrupting its ability to perform the task.

\paragraph{Sufficiency specificity}
The specificity loss contains a sufficiency term that mirrors the necessity specificity term of Eq.~\ref{eq:spec_nec_loss}. Let $\tilde{p}(f,t,\theta)$ denote the model's probability on the corrupted prompt without any intervention. Patching the activations of the localized components should \emph{not} restore performance on the corrupted prompt for non-target pairs:
\begin{equation}
\resizebox{\columnwidth}{!}{$\displaystyle
\begin{aligned}
\mathcal{L}_{\text{spec}}^{\text{suff}}(\mathbf{m})
&= \underbrace{\mathbb{E}_{f' \neq f^*}\!\Big[\text{MSE}\!\bigl(\tilde{p}(f', t^*, \theta \circ \mathbf{m}),\; \tilde{p}(f', t^*, \theta)\bigr)\Big]}_{\text{other facts, same task}} \\
&\quad + \underbrace{\sum_{t' \neq t^*} \text{MSE}\!\bigl(\tilde{p}(f^*, t', \theta \circ \mathbf{m}),\; \tilde{p}(f^*, t', \theta)\bigr)}_{\text{same fact, other tasks}}
\end{aligned}
$}
\label{eq:spec_suff_loss}
\end{equation}
  
\paragraph{Optimization hyperparameters}
All logits are initialized to $0$ (sigmoid value $0.5$). Masks are trained for 2{,}500 steps using Adam with a learning rate of $0.1$, a mini-batch size of $8$, and an L1 penalty weight of $\beta = 10.0$. 

\paragraph{Per-step sampling}
For the necessity specificity term on non-target facts on the target task, at every step we sample a mini-batch of those facts' paraphrases without replacement. For the sufficiency terms, we sample paraphrases with replacement, corrupting each sampled prompt with a randomly drawn placeholder.

 \subsection{Experimental Protocol}
\paragraph{Verification task handling}
From each verification template, we generated both a true-statement and a false-statement prompt. 
When \textsc{Verification} is the targeted task, we filter out false-statement prompts for the targeted fact, so that the mask is optimized and evaluated solely on true statements. 
When \textsc{Verification} serves as a retention task (i.e., not the targeted task), both true and false prompts are used. When the baseline performance of the two modes (true statements vs true and false statements) differs by at least $0.02$, we mention both numbers in the necessity or sufficiency heatmaps.
 
\paragraph{Sufficiency patching protocol}
The \textsc{Multi-Hop-2} task, and the {Multi-Hop-1 Control} task are excluded from the sufficiency evaluation because the subject entity of the targeted fact does not appear in the prompt, making the subject-replacement corruption procedure inapplicable.
 
\paragraph{Subject corruption}
We replace the subject string of every prompt with repeated copies of a placeholder token. The placeholders pool contains 16 strings: four base characters (\textit{x}, \textit{y}, \textit{z}, \textit{w}), their uppercase variants, and space-prefixed variants of all eight. 
The number of repetitions is adjusted so that the tokenized length of the replacement exactly matches that of the original subject. 
During training, a fresh placeholder is sampled per prompt before each mini-batch forward. At evaluation, we sample one placeholder per prompt.

\subsection{Evaluation}
 
\paragraph{Metric definitions}
For each prompt paraphrase, the model scores $1$ if its top-1 token matches the first token of the correct answer and $0$ otherwise, subject to formatting variants described below. We then average across paraphrases per fact and report the mean and std across facts.

\paragraph{Tolerance to formatting variants of the target}
\label{appendix:variant_tolerance}
A strict exact-match criterion would penalize correct answers produced in a slightly different form (e.g., a leading space, different capitalization, or a multi-token split). We therefore apply two post-hoc checks; if either passes, we count the model's prediction as correct.

\textit{Top-1 token variant.}\;
We compare the decoded top-1 token $p$ to the decoded first token of the target $t_0$, accepting the prediction under whitespace-only, case-only, or combined normalization, as well as when one is a prefix of the other (length $\geq 2$), which handles multi-token answers.

\textit{Short continuation.}\;
We further extract a 3-token continuation (via greedy decoding) and compare each cumulative generated prefix to the \emph{full} target completion $t$ using the same normalization procedures described above, plus substring containment (catching outputs like \textit{``Hmm, Paris''}). For \textsc{MCQA} and \textsc{Neg~MCQA}, where the target completion includes a closing parenthesis (e.g., \textit{``3)''}), we additionally accept the bare digit alone (e.g., \textit{``3''}).
These checks validate that measured drops in the model's performance on a (fact,task) pair reflect genuine failures rather than tokenization artifacts.

 \paragraph{Sufficiency metric} We measure the model's accuracy under three conditions: \emph{clean} (unmodified prompt), \emph{corrupted} (subject replaced with placeholder tokens), and \emph{patched} (corrupted prompt with the encoding's clean activations stitched in). 
The reconstruction rate is:
\begin{equation}
\frac{\text{acc}_{\text{patched}} - \text{acc}_{\text{corrupted}}}{\text{acc}_{\text{clean}} - \text{acc}_{\text{corrupted}}} \times 100\%.
\end{equation}
A value of $100\%$ indicates full recovery to clean accuracy, $0\%$ indicates no improvement over the corrupted baseline, and negative values indicate that patching further degrades accuracy.

\paragraph{Cross-experiment aggregation in the sufficiency heatmaps}
Reconstruction rates are computed against a clean and a corrupted baseline. The corrupted prompts are resampled in each patching experiment, so for a fixed evaluation task the corrupted accuracy varies slightly depending on which task was patched. To keep reconstruction rates comparable, we pool the clean and corrupted accuracies across all patched tasks that share an evaluation task and use that as the common baseline for that evaluation task.
\textsc{Verification} is handled separately: when \textsc{Verification} is itself the patched task it is scored on true statements only, and when it is the evaluation task for another patched task it is scored on the full true+false set, so we pool and normalize the two cases separately.

 \subsection{Additional results}
\label{appendix:params_results}

\begin{figure*}[t]
    \centering
    \begin{minipage}[t]{0.49\linewidth}
        \centering
        \begin{subfigure}{\linewidth}
            \centering
            \includegraphics[width=\linewidth]{figures/necessity_language_olmo_7b_top1.pdf}
            \caption{\texttt{(country, official language, language)}}
            \label{fig:necessity_language_olmo}
        \end{subfigure}\\
        \begin{subfigure}{\linewidth}
            \centering
            \includegraphics[width=\linewidth]{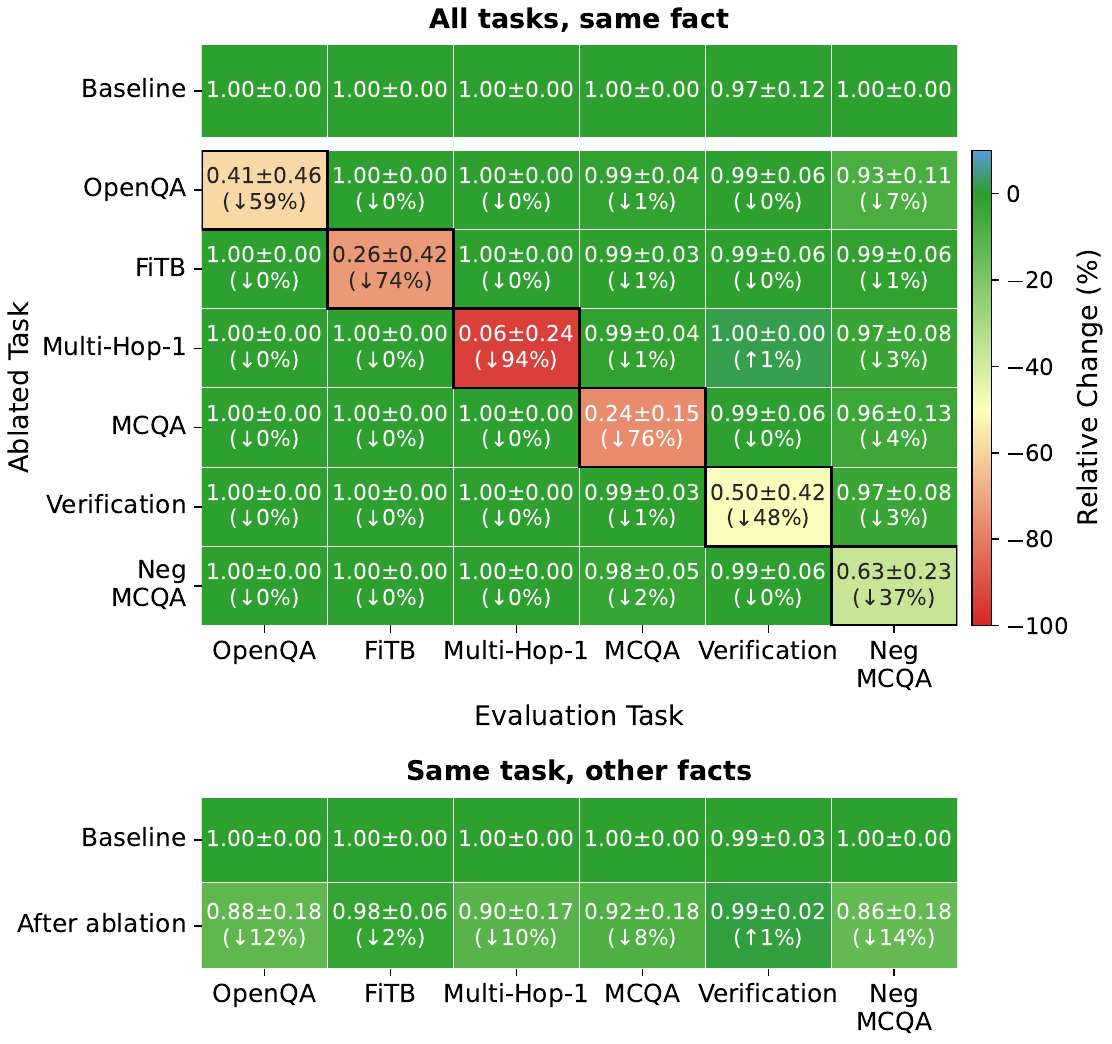}
            \caption{\texttt{(landmark, in-country, country)}}
            \label{fig:necessity_landmark_olmo}
        \end{subfigure}\\
        \begin{subfigure}{\linewidth}
            \centering
            \includegraphics[width=\linewidth]{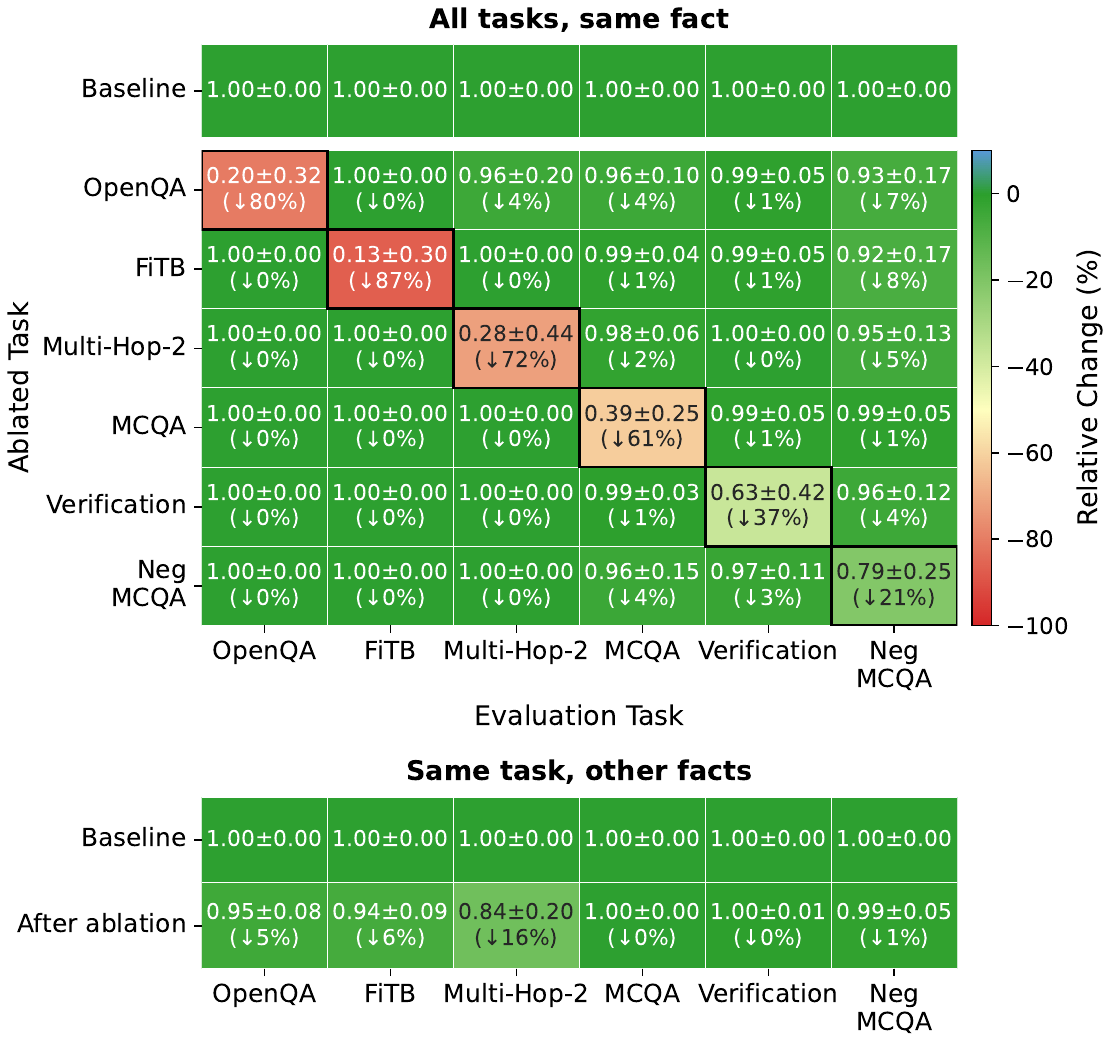}
            \caption{\texttt{(country, capital-of, city)}}
            \label{fig:necessity_capital_olmo}
        \end{subfigure}
    \end{minipage}\hfill
    \begin{minipage}[t]{0.49\linewidth}
        \centering
        \begin{subfigure}{\linewidth}
            \centering
            \includegraphics[width=\linewidth]{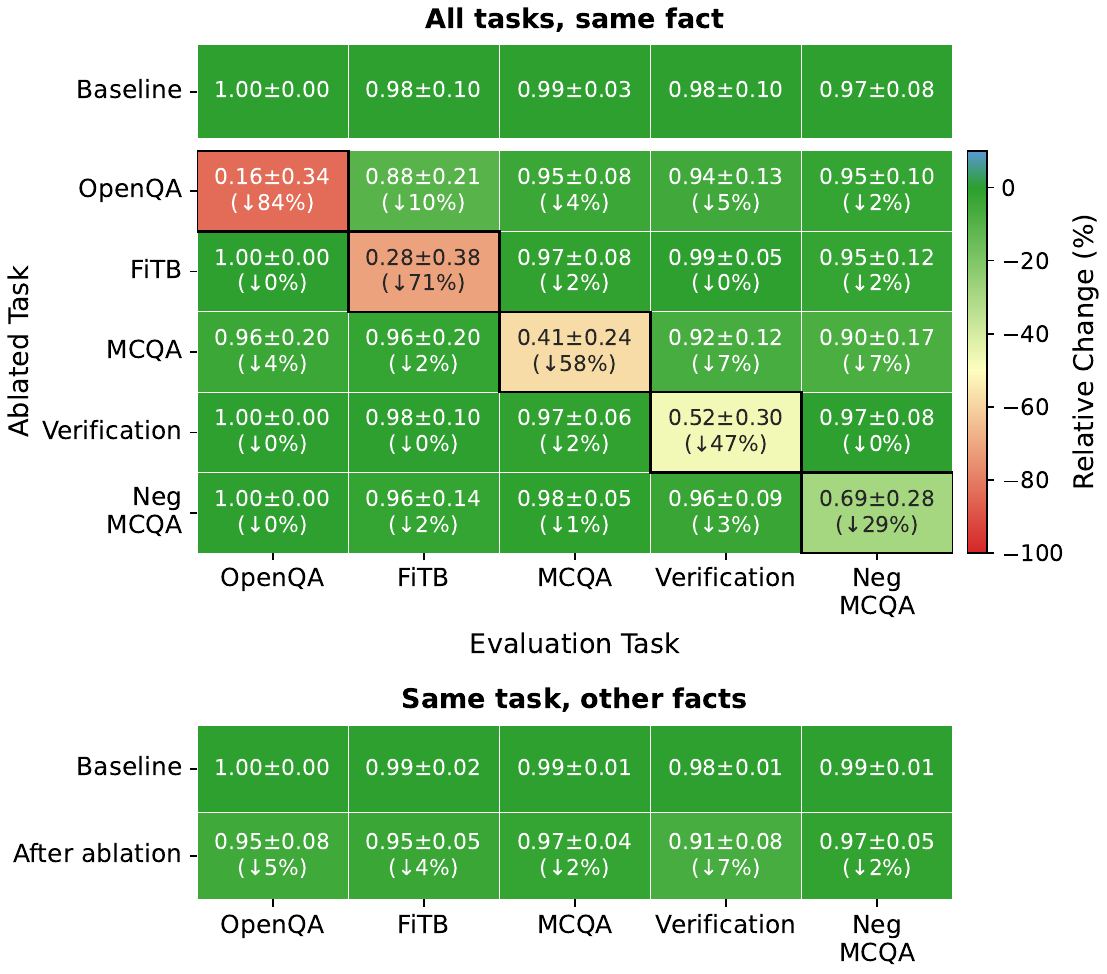}
            \caption{\texttt{(person, plays-instrument, instrument)}}
            \label{fig:necessity_instrument_olmo}
        \end{subfigure}\\
        \begin{subfigure}{\linewidth}
            \centering
            \includegraphics[width=\linewidth]{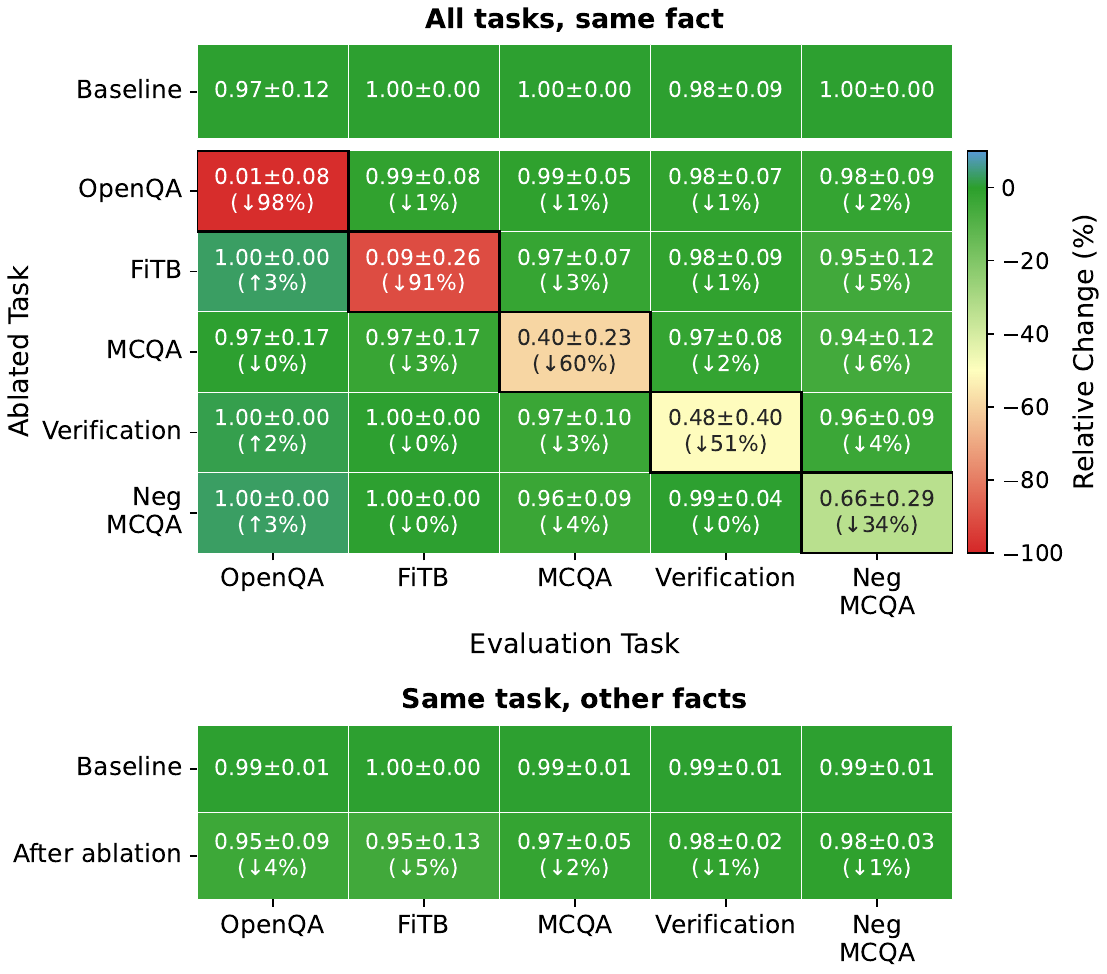}
            \caption{\texttt{(company, HQ-in-city, city)}}
            \label{fig:necessity_company_olmo}
        \end{subfigure}
    \end{minipage}
    \caption{Necessity results on \OlmoSevenB{}. Same layout as \Cref{fig:necessity}.}
    \label{fig:necessity_olmo_7b}
\end{figure*}

\begin{figure*}[t]
    \centering
    \begin{minipage}[t]{0.49\linewidth}
        \centering
        \begin{subfigure}{\linewidth}
            \centering
            \includegraphics[width=0.9\linewidth]{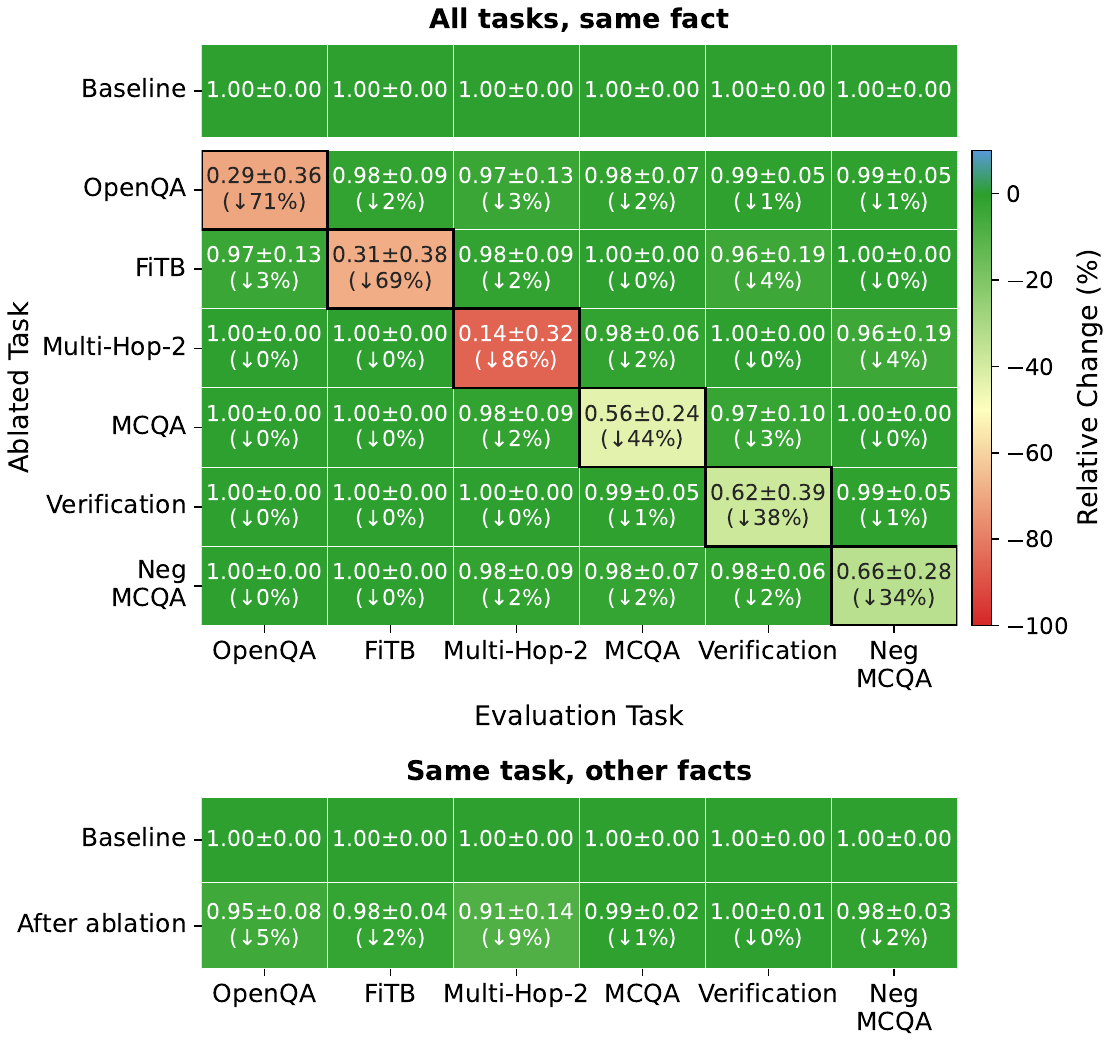}
            \caption{\texttt{(country, official language, language)}}
            \label{fig:necessity_language_olmo_13b}
        \end{subfigure}\\
        \begin{subfigure}{\linewidth}
            \centering
            \includegraphics[width=0.9\linewidth]{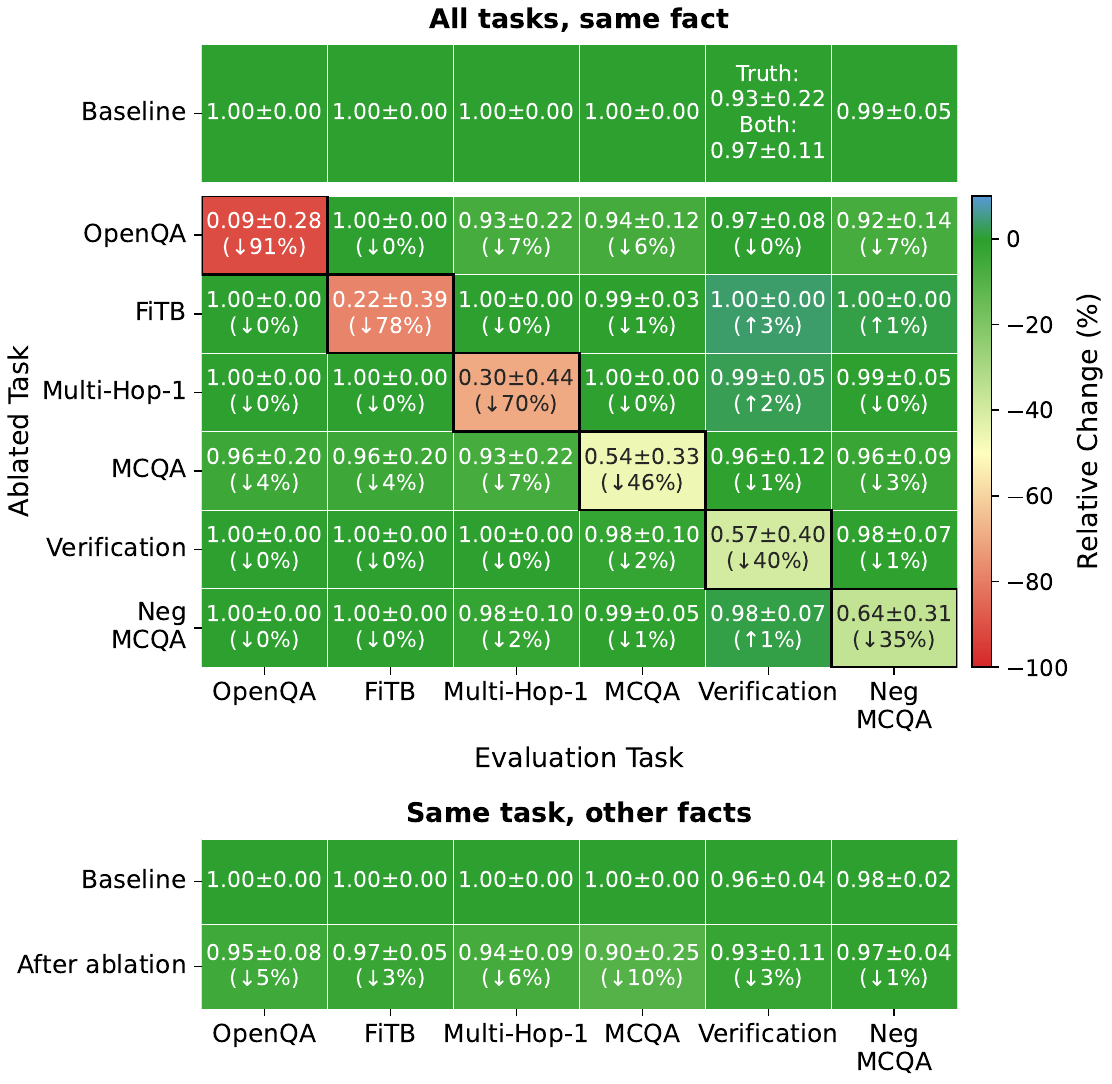}
            \caption{\texttt{(landmark, in-country, country)}}
            \label{fig:necessity_landmark_olmo_13b}
        \end{subfigure}\\
        \begin{subfigure}{\linewidth}
            \centering
            \includegraphics[width=0.9\linewidth]{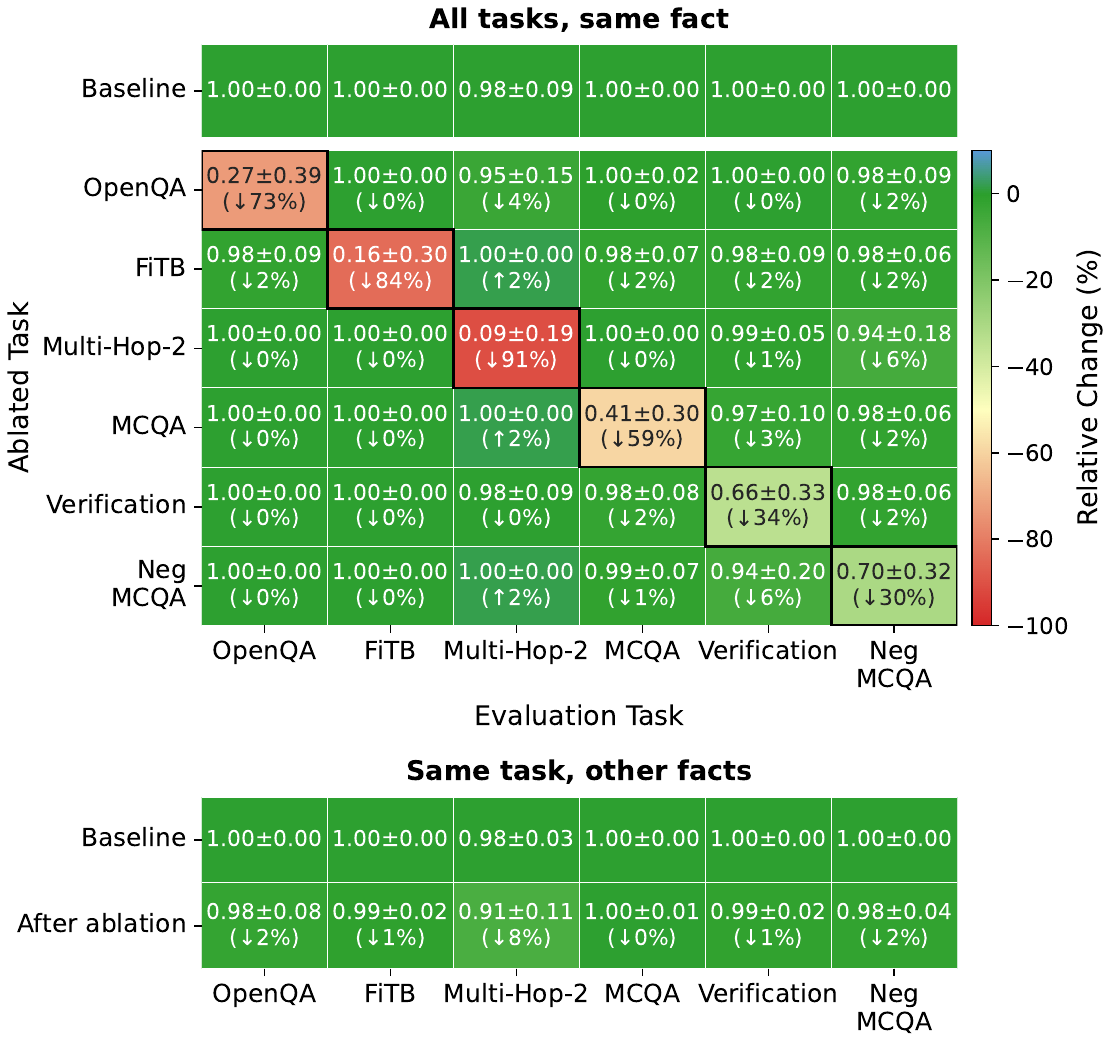}
            \caption{\texttt{(country, capital-of, city)}}
            \label{fig:necessity_capital_olmo_13b}
        \end{subfigure}
    \end{minipage}\hfill
    \begin{minipage}[t]{0.49\linewidth}
        \centering
        \begin{subfigure}{\linewidth}
            \centering
            \includegraphics[width=0.9\linewidth]{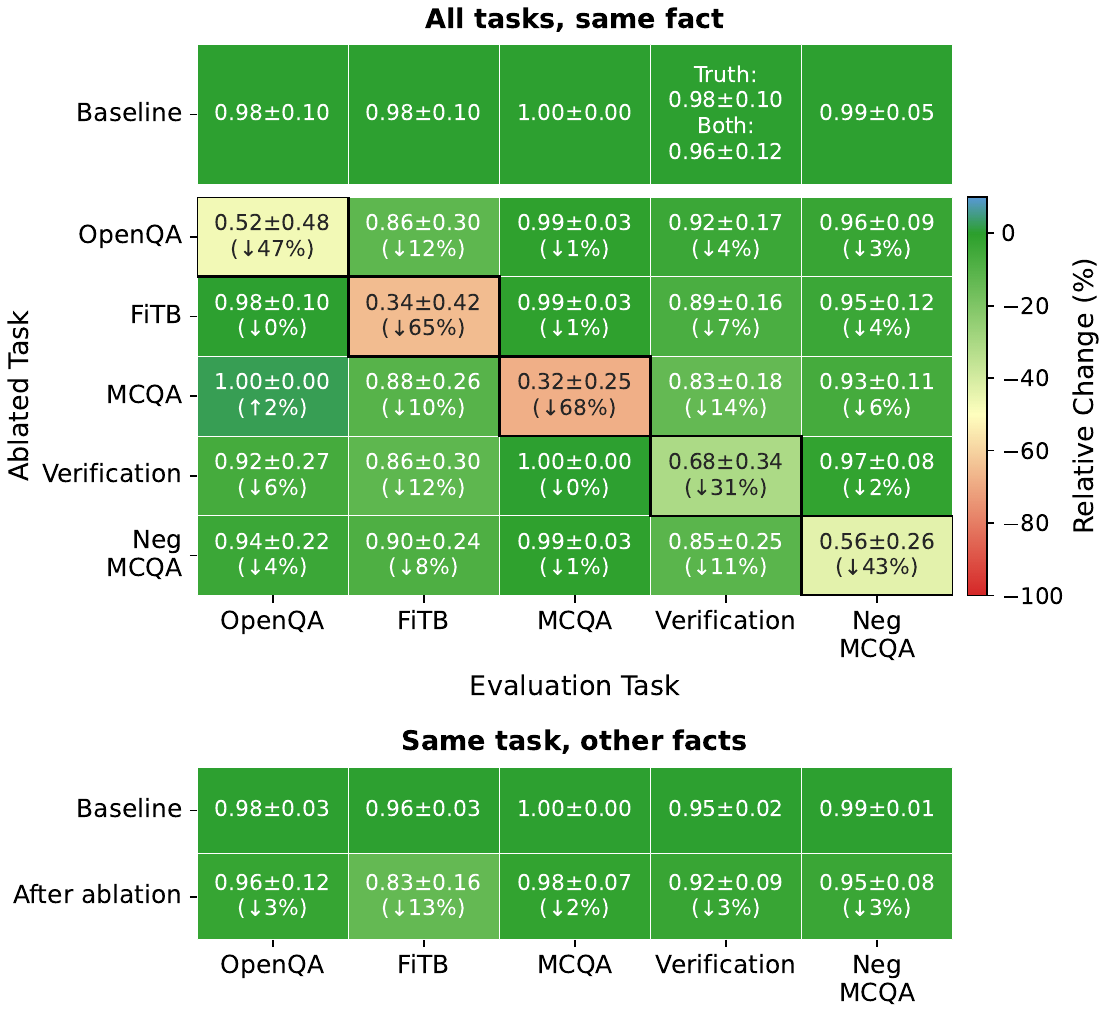}
            \caption{\texttt{(person, plays-instrument, instrument)}}
            \label{fig:necessity_instrument_olmo_13b}
        \end{subfigure}\\
        \begin{subfigure}{\linewidth}
            \centering
            \includegraphics[width=0.9\linewidth]{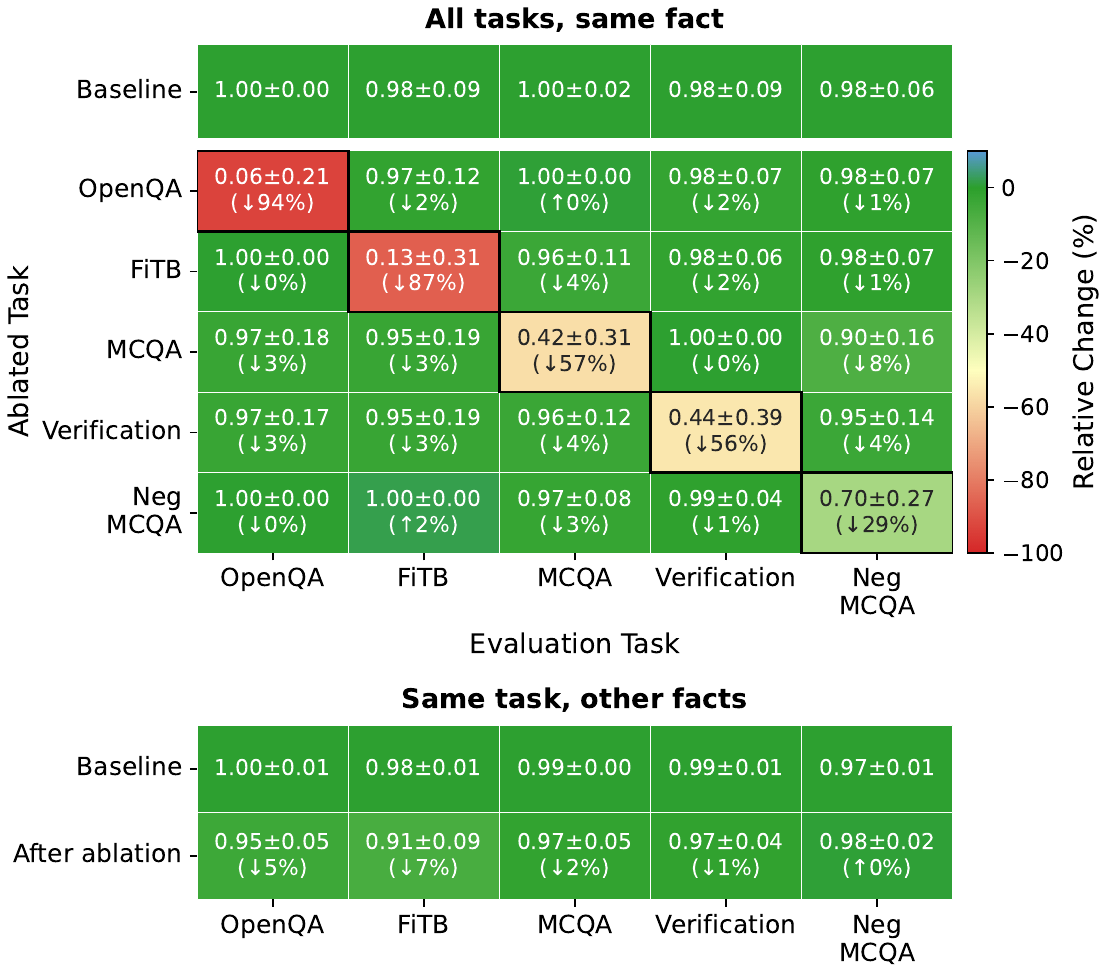}
            \caption{\texttt{(company, HQ-in-city, city)}}
            \label{fig:necessity_company_olmo_13b}
        \end{subfigure}
    \end{minipage}
    \caption{Necessity results on \OlmoThirteenB{}. Same layout as \Cref{fig:necessity}.}
    \label{fig:necessity_olmo_13b}
\end{figure*}

\begin{figure*}[t]
    \centering
    \begin{minipage}[t]{0.49\linewidth}
        \centering
        \begin{subfigure}{\linewidth}
        \centering
        \includegraphics[width=\linewidth]{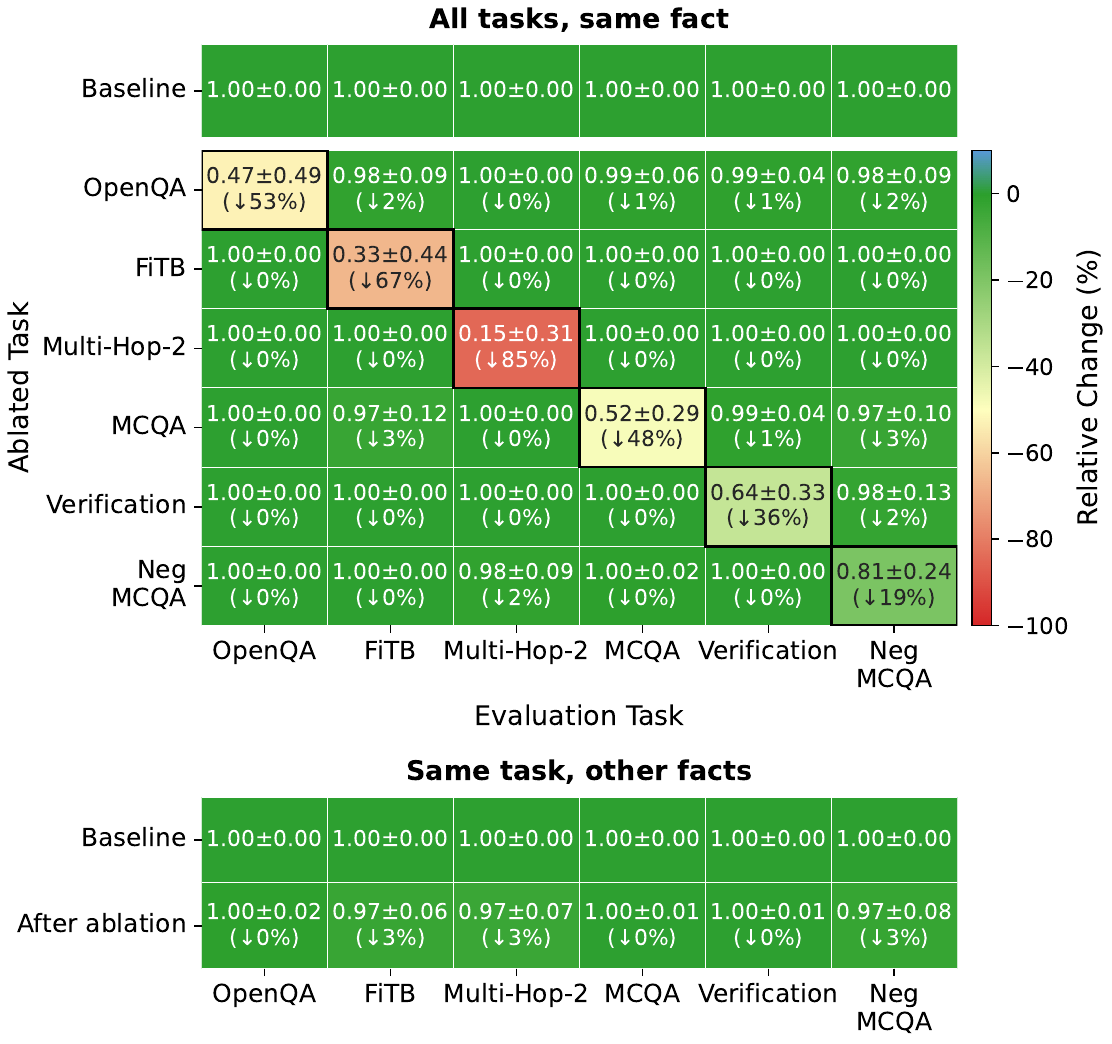}
        \caption{\texttt{(country, official language, language)}}
        \label{fig:necessity_language_gemma}
    \end{subfigure}
        \begin{subfigure}{\linewidth}
            \centering
            \includegraphics[width=\linewidth]{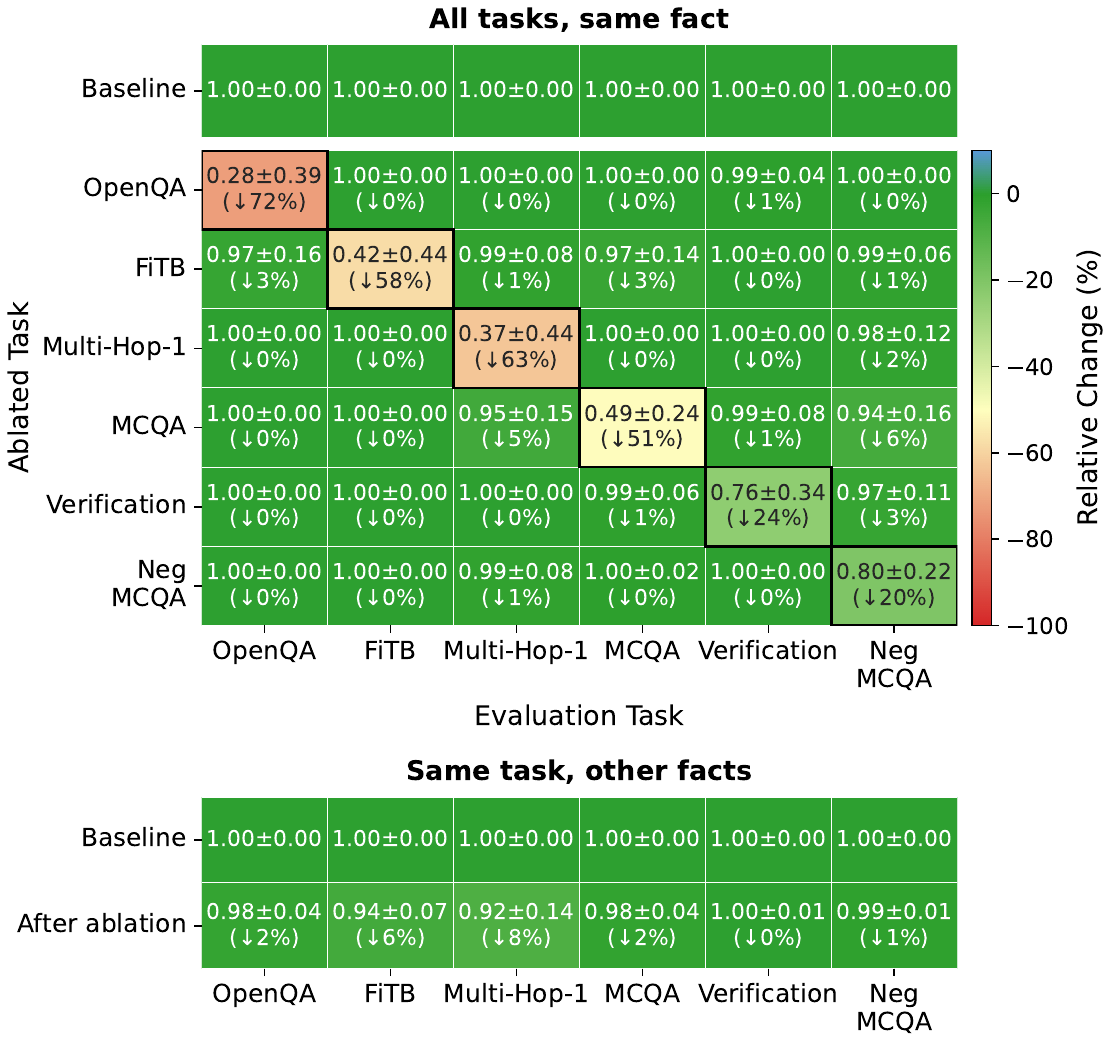}
            \caption{\texttt{(landmark, in-country, country)}}
            \label{fig:necessity_landmark_gemma}
        \end{subfigure}\\
        \begin{subfigure}{\linewidth}
            \centering
            \includegraphics[width=\linewidth]{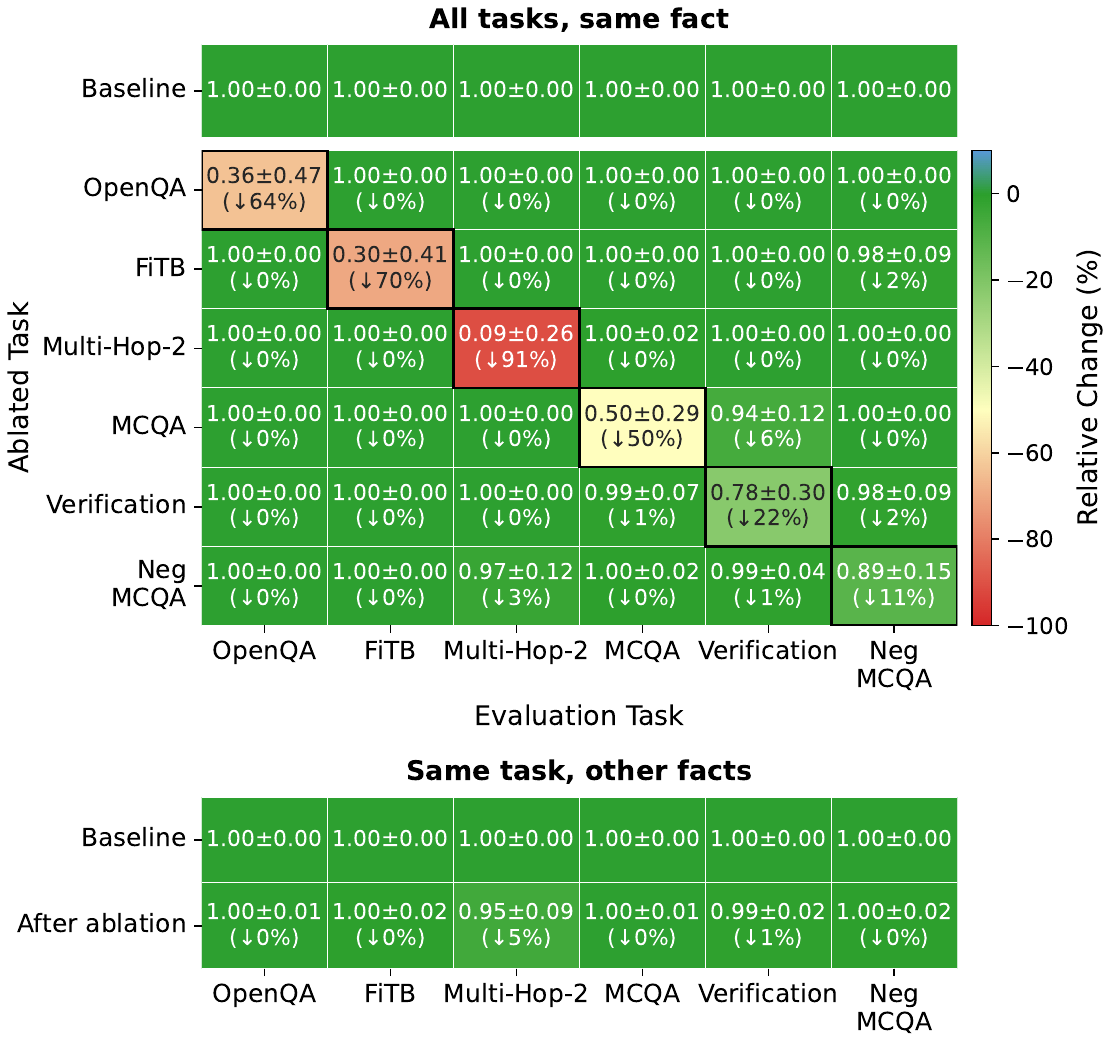}
            \caption{\texttt{(country, capital-of, city)}}
            \label{fig:necessity_capital_gemma}
        \end{subfigure}
    \end{minipage}\hfill
    \begin{minipage}[t]{0.49\linewidth}
        \centering
        \begin{subfigure}{\linewidth}
            \centering
            \includegraphics[width=\linewidth]{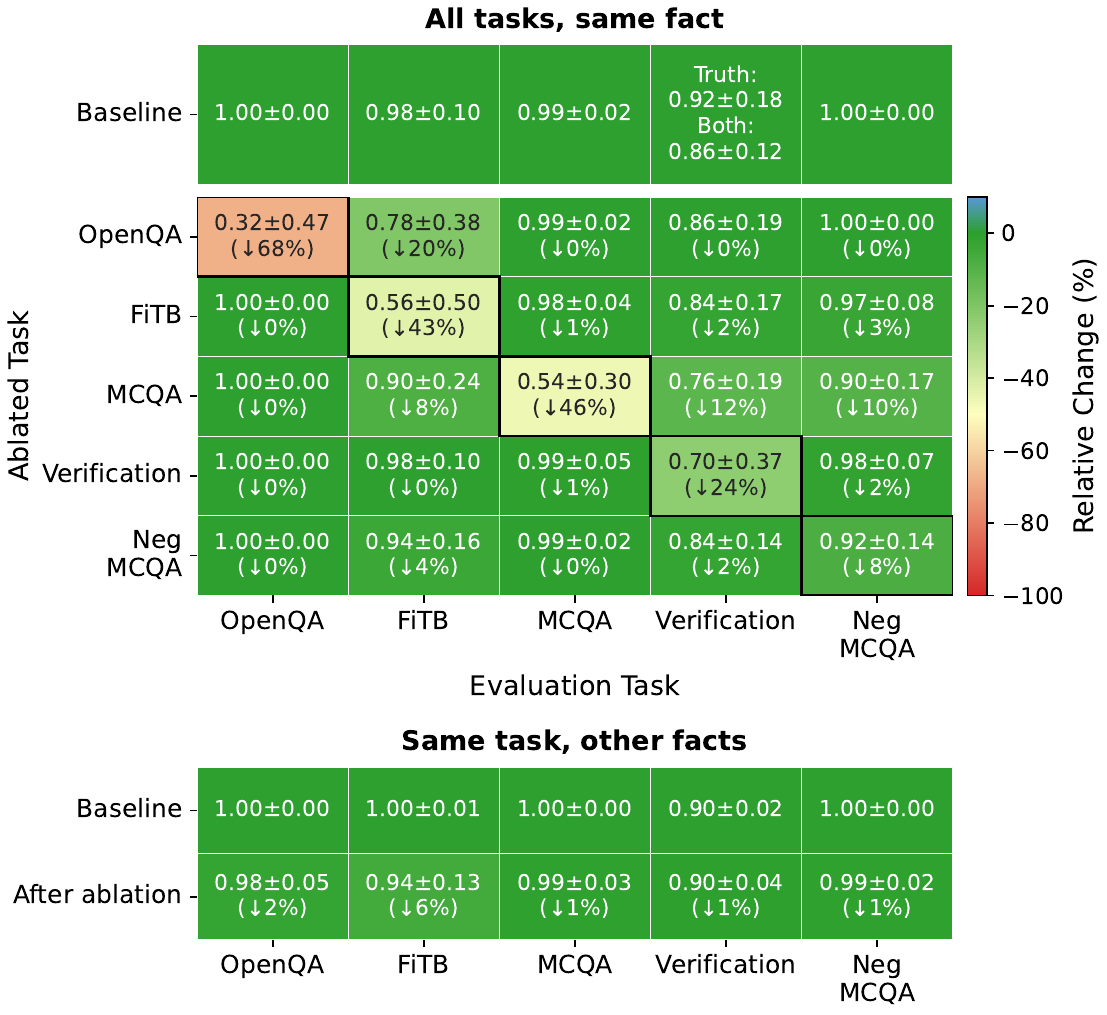}
            \caption{\texttt{(person, plays-instrument, instrument)}}
            \label{fig:necessity_instrument_gemma}
        \end{subfigure}\\
        \begin{subfigure}{\linewidth}
            \centering
            \includegraphics[width=\linewidth]{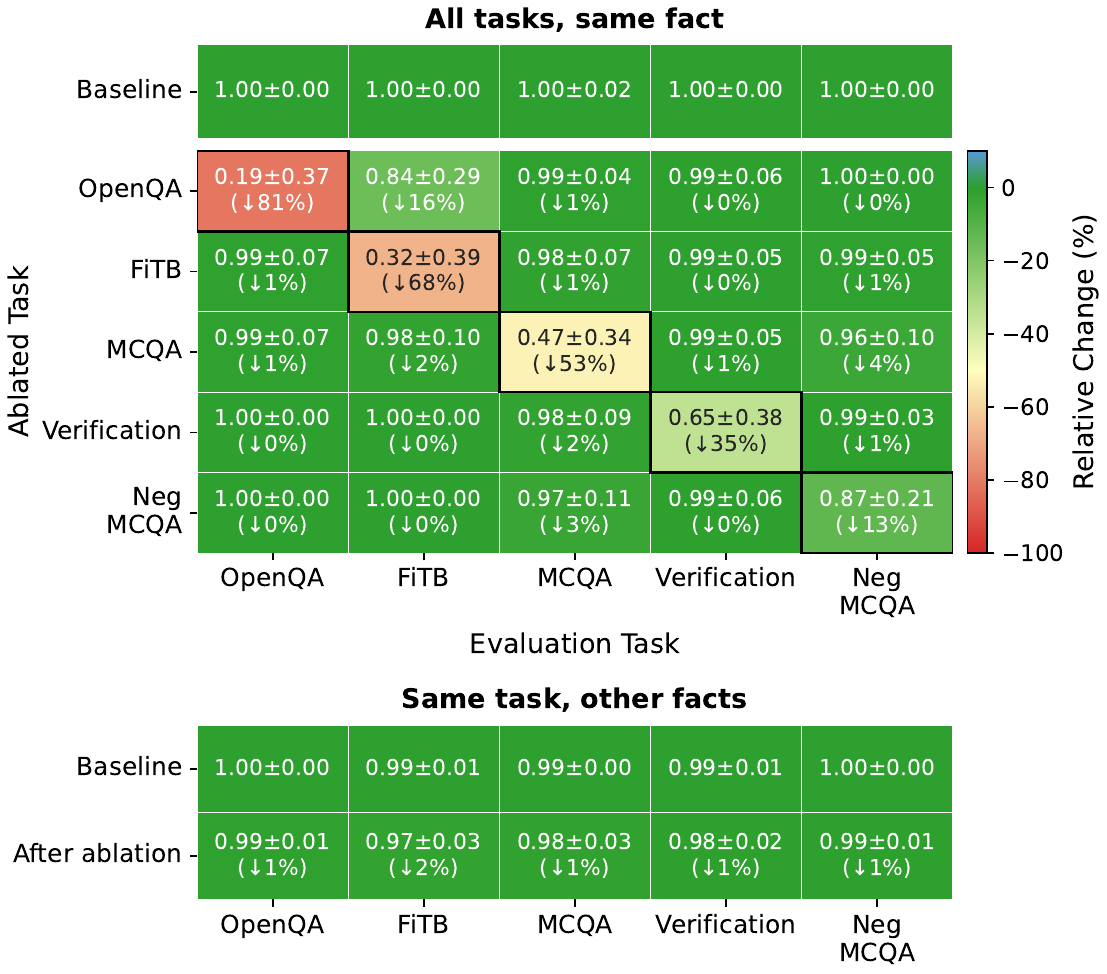}
            \caption{\texttt{(company, HQ-in-city, city)}}
            \label{fig:necessity_company_gemma}
        \end{subfigure}
    \end{minipage}
    \caption{Necessity results on \Gemma{}. Same layout as \Cref{fig:necessity}.}
    \label{fig:necessity_gemma}
\end{figure*}

\paragraph{Full necessity results}
The pattern from \Cref{fig:necessity} holds across all models and all datasets (\Cref{fig:necessity_olmo_7b,fig:necessity_olmo_13b,fig:necessity_gemma}). 
Ablating the (fact, task) parametric encoding reduces the model's performance on the targeted pair (diagonal), while off-diagonal cells (describing the model's performance on the same fact for other tasks) and the bottom row (describing the model's performance on other facts on the targeted task) stay near the baseline. 
The magnitude of the drops varies by task, with generative tasks (\textsc{OpenQA}, \textsc{FiTB}, \textsc{Multi-Hop}) generally showing the largest drops, while \textsc{Neg~MCQA} drops are typically the smallest yet still notable. In \S\ref{sec:entanglement} we analyze this further.

\begin{figure*}[t]
    \centering
    \begin{minipage}[t]{0.49\linewidth}
        \centering
        \begin{subfigure}{\linewidth}
            \centering
            \includegraphics[width=0.88\linewidth,trim=0 15pt 0 0pt,clip]{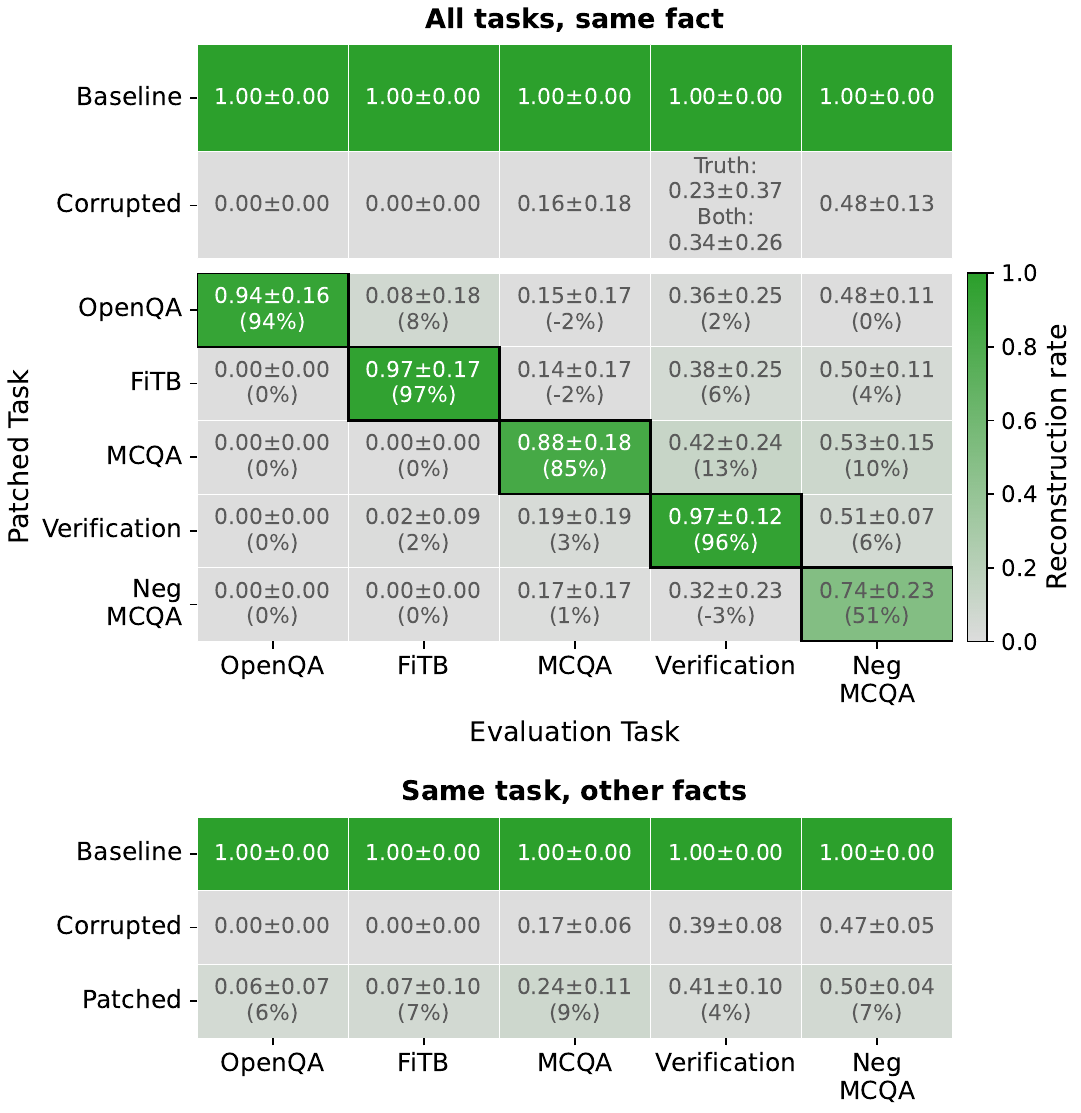}
            \caption{\texttt{(country, official language, language)}}
            \label{fig:sufficiency_language_gemma}
        \end{subfigure}\\
        \begin{subfigure}{\linewidth}
            \centering
            \includegraphics[width=0.88\linewidth,trim=0 15pt 0 0pt,clip]{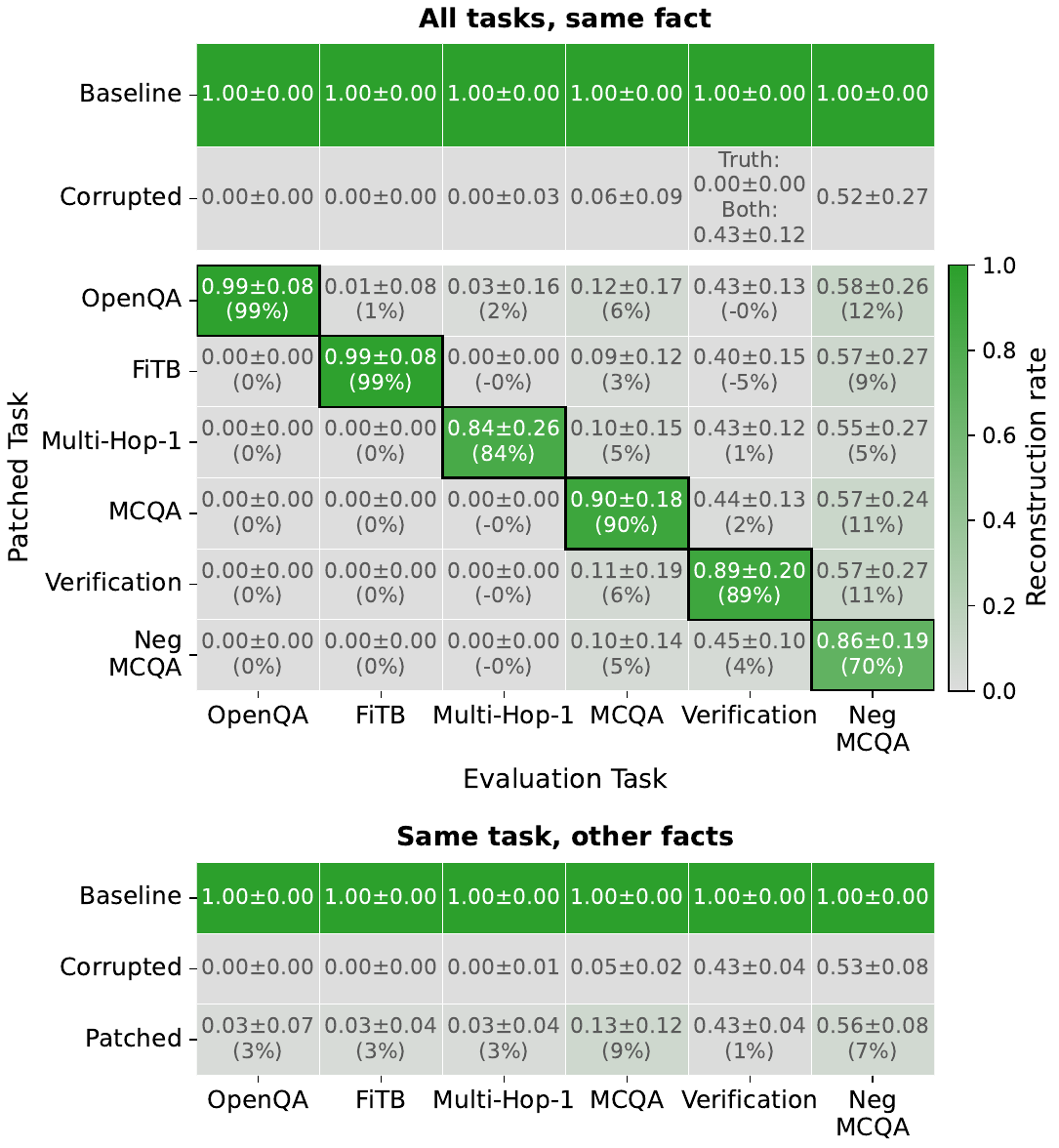}
            \caption{\texttt{(landmark, in-country, country)}}
            \label{fig:sufficiency_landmark_gemma}
        \end{subfigure}\\
        \begin{subfigure}{\linewidth}
            \centering
            \includegraphics[width=0.88\linewidth,trim=0 15pt 0 0pt,clip]{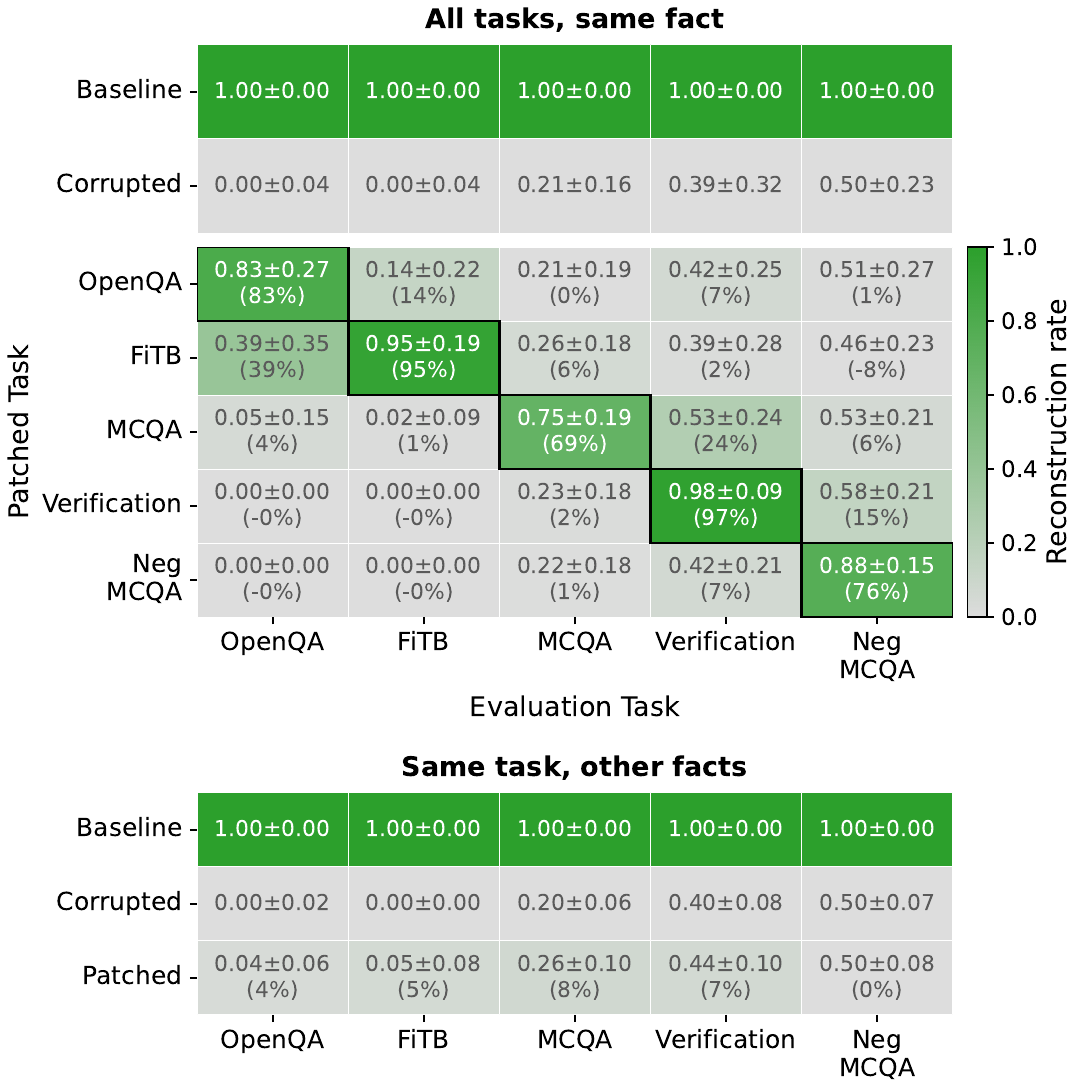}
            \caption{\texttt{(country, capital-of, city)}}
            \label{fig:sufficiency_capital_gemma}
        \end{subfigure}
    \end{minipage}\hfill
    \begin{minipage}[t]{0.49\linewidth}
        \centering
        \begin{subfigure}{\linewidth}
            \centering
            \includegraphics[width=0.88\linewidth,trim=0 15pt 0 0pt,clip]{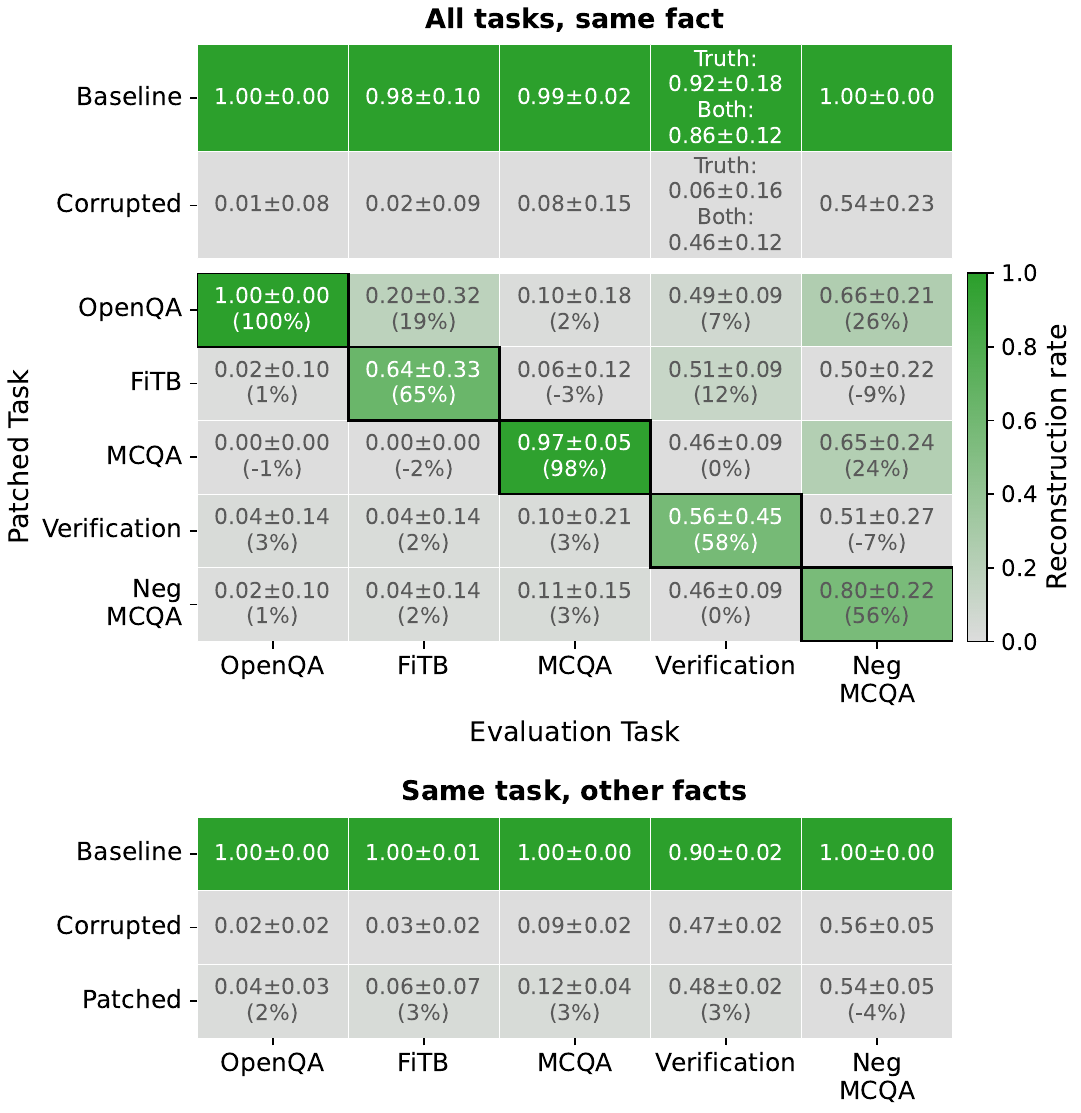}
            \caption{\texttt{(person, plays-instrument, instrument)}}
            \label{fig:sufficiency_instrument_gemma}
        \end{subfigure}\\
        \begin{subfigure}{\linewidth}
            \centering
            \includegraphics[width=0.88\linewidth,trim=0 15pt 0 0pt,clip]{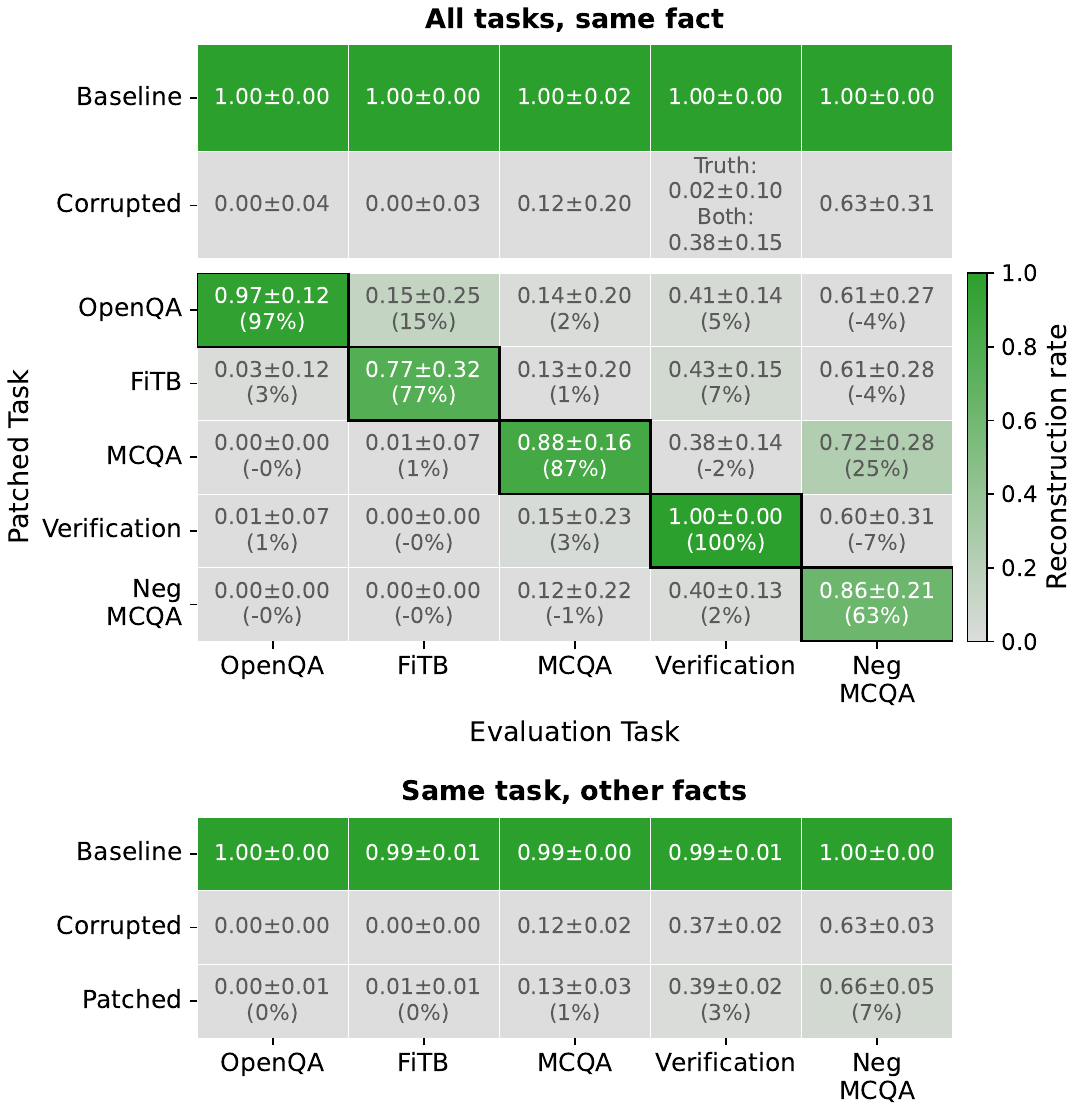}
            \caption{\texttt{(company, HQ-in-city, city)}}
            \label{fig:sufficiency_company_gemma}
        \end{subfigure}
    \end{minipage}
    \caption{Sufficiency results on \Gemma{}. 
    Each row shows the reconstruction rate after patching the parametric encoding optimized for one task. 
    Cell color reflects the reconstruction rate: dark green indicates full recovery, light gray indicates no recovery. 
    The top baseline row is pinned to green and the corrupted row to gray for reference.}
    \label{fig:sufficiency_gemma}
\end{figure*}

\begin{figure*}[t]
    \centering
    \begin{minipage}[t]{0.49\linewidth}
        \centering
        \begin{subfigure}{\linewidth}
            \centering
            \includegraphics[width=0.9\linewidth]{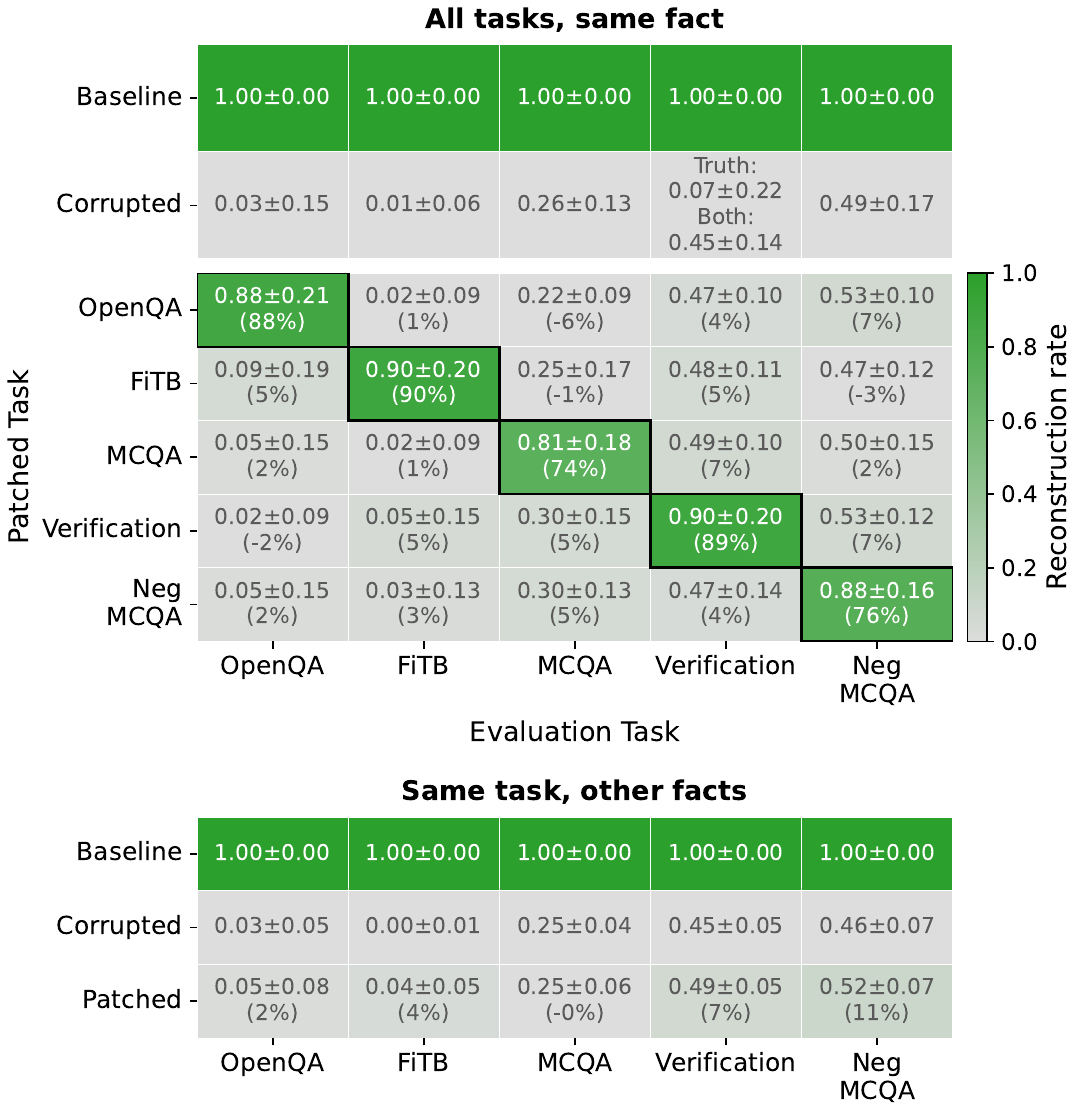}
            \caption{\texttt{(country, official language, language)}}
            \label{fig:sufficiency_language_olmo_13b}
        \end{subfigure}\\
        \begin{subfigure}{\linewidth}
            \centering
            \includegraphics[width=0.9\linewidth]{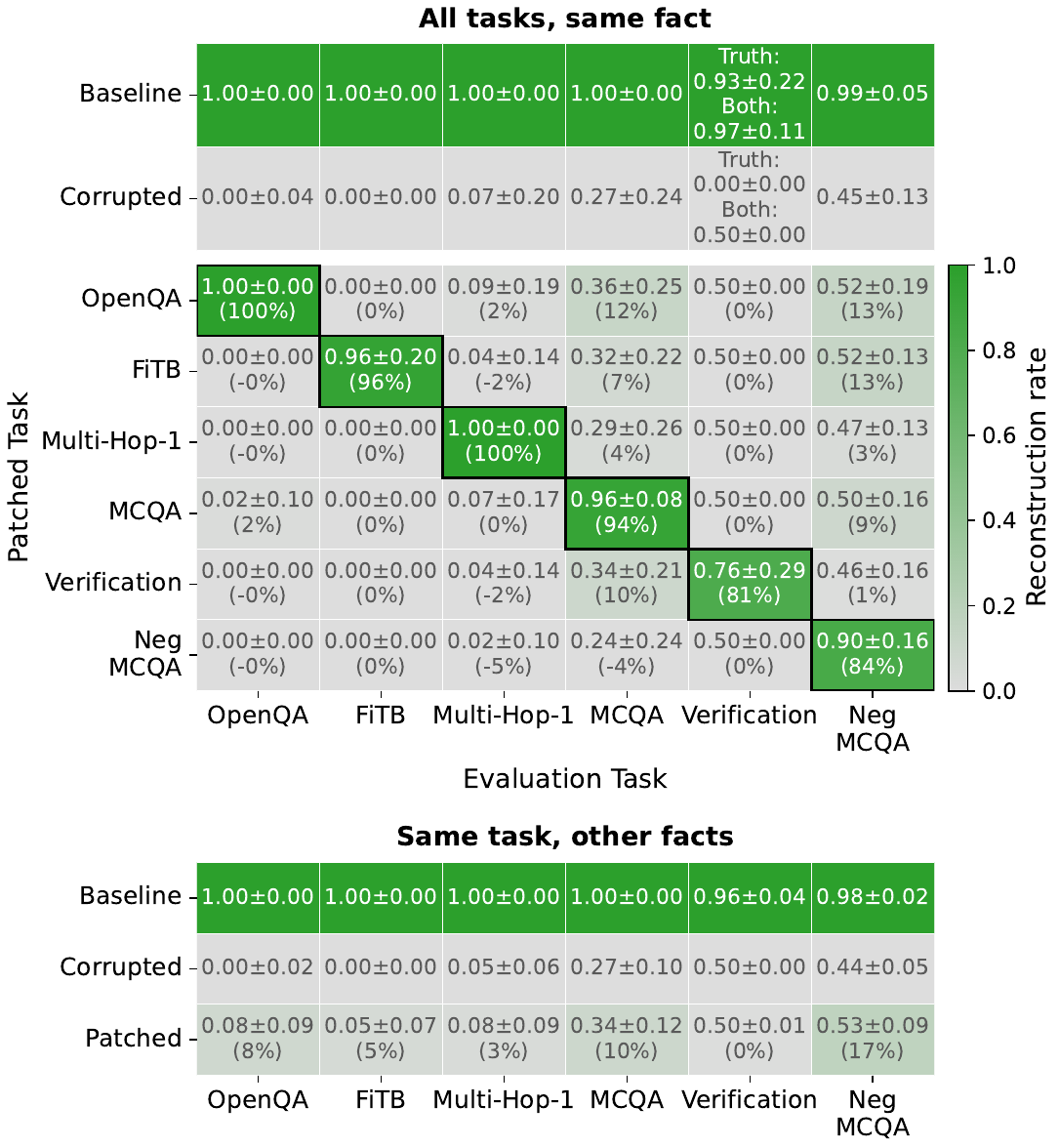}
            \caption{\texttt{(landmark, in-country, country)}}
            \label{fig:sufficiency_landmark_olmo_13b}
        \end{subfigure}\\
        \begin{subfigure}{\linewidth}
            \centering
            \includegraphics[width=0.9\linewidth]{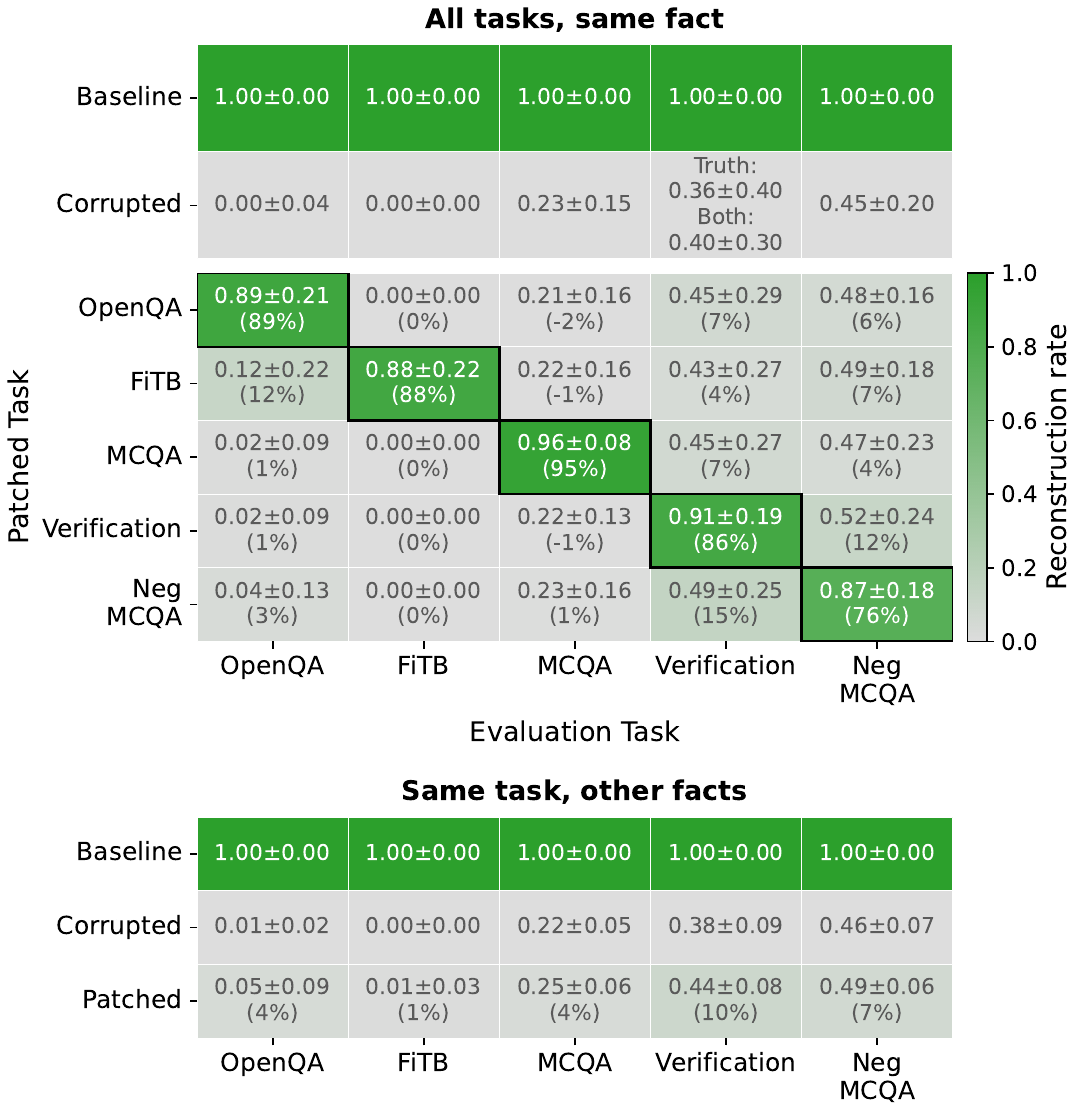}
            \caption{\texttt{(country, capital-of, city)}}
            \label{fig:sufficiency_capital_olmo_13b}
        \end{subfigure}
    \end{minipage}\hfill
    \begin{minipage}[t]{0.49\linewidth}
        \centering
        \begin{subfigure}{\linewidth}
            \centering
            \includegraphics[width=0.9\linewidth]{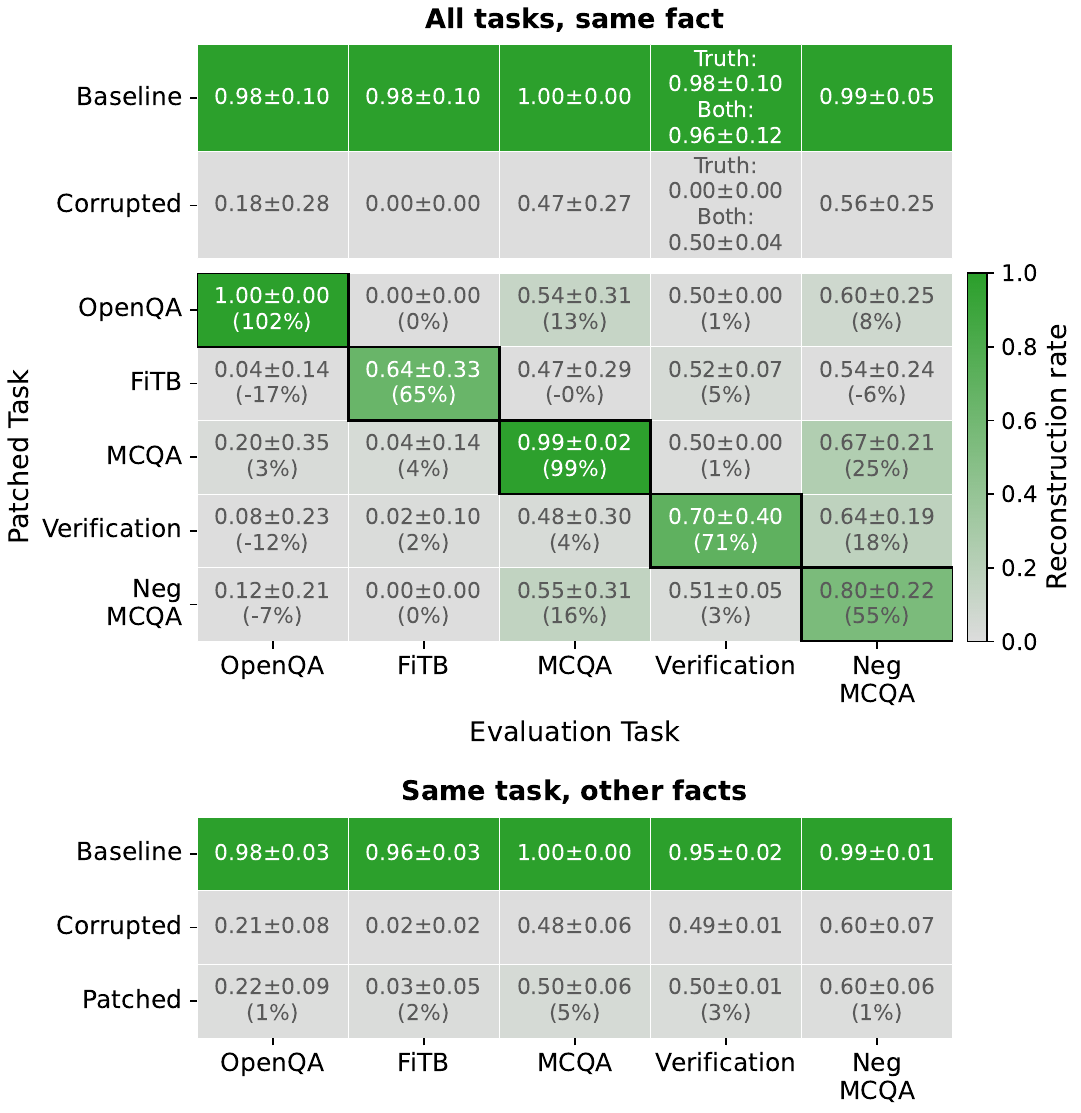}
            \caption{\texttt{(person, plays-instrument, instrument)}}
            \label{fig:sufficiency_instrument_olmo_13b}
        \end{subfigure}\\
        \begin{subfigure}{\linewidth}
            \centering
            \includegraphics[width=0.9\linewidth]{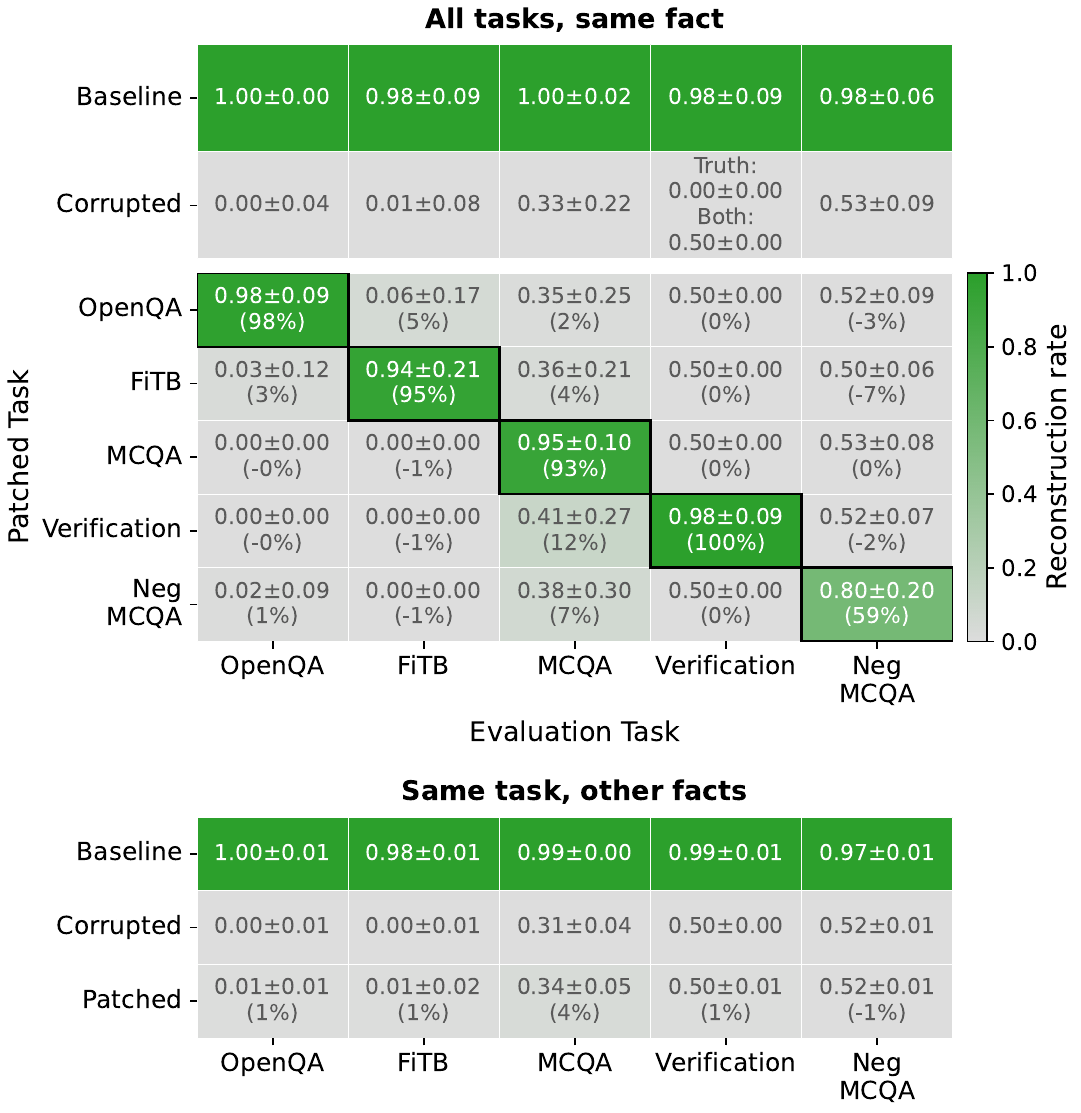}
            \caption{\texttt{(company, HQ-in-city, city)}}
            \label{fig:sufficiency_company_olmo_13b}
        \end{subfigure}
    \end{minipage}
    \caption{Sufficiency results on \OlmoThirteenB{}. Same layout as \Cref{fig:sufficiency_gemma}.}
    \label{fig:sufficiency_olmo_13b}
\end{figure*}

\begin{figure*}[t]
    \centering
    \begin{minipage}[t]{0.49\linewidth}
        \centering
        \begin{subfigure}{\linewidth}
            \centering
            \includegraphics[width=0.9\linewidth]{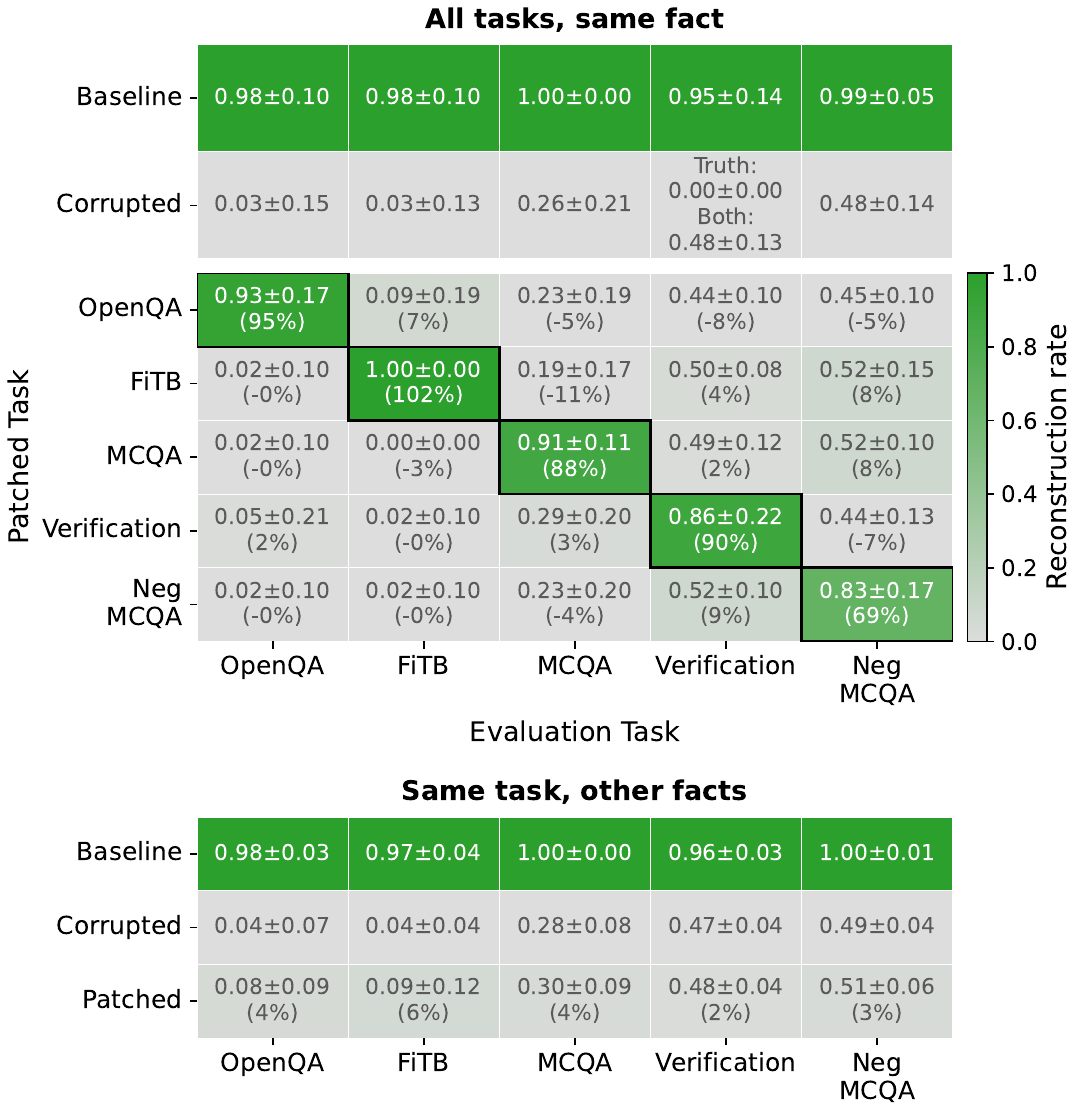}
            \caption{\texttt{(country, official language, language)}}
            \label{fig:sufficiency_language_olmo}
        \end{subfigure}\\
        \begin{subfigure}{\linewidth}
            \centering
            \includegraphics[width=0.9\linewidth]{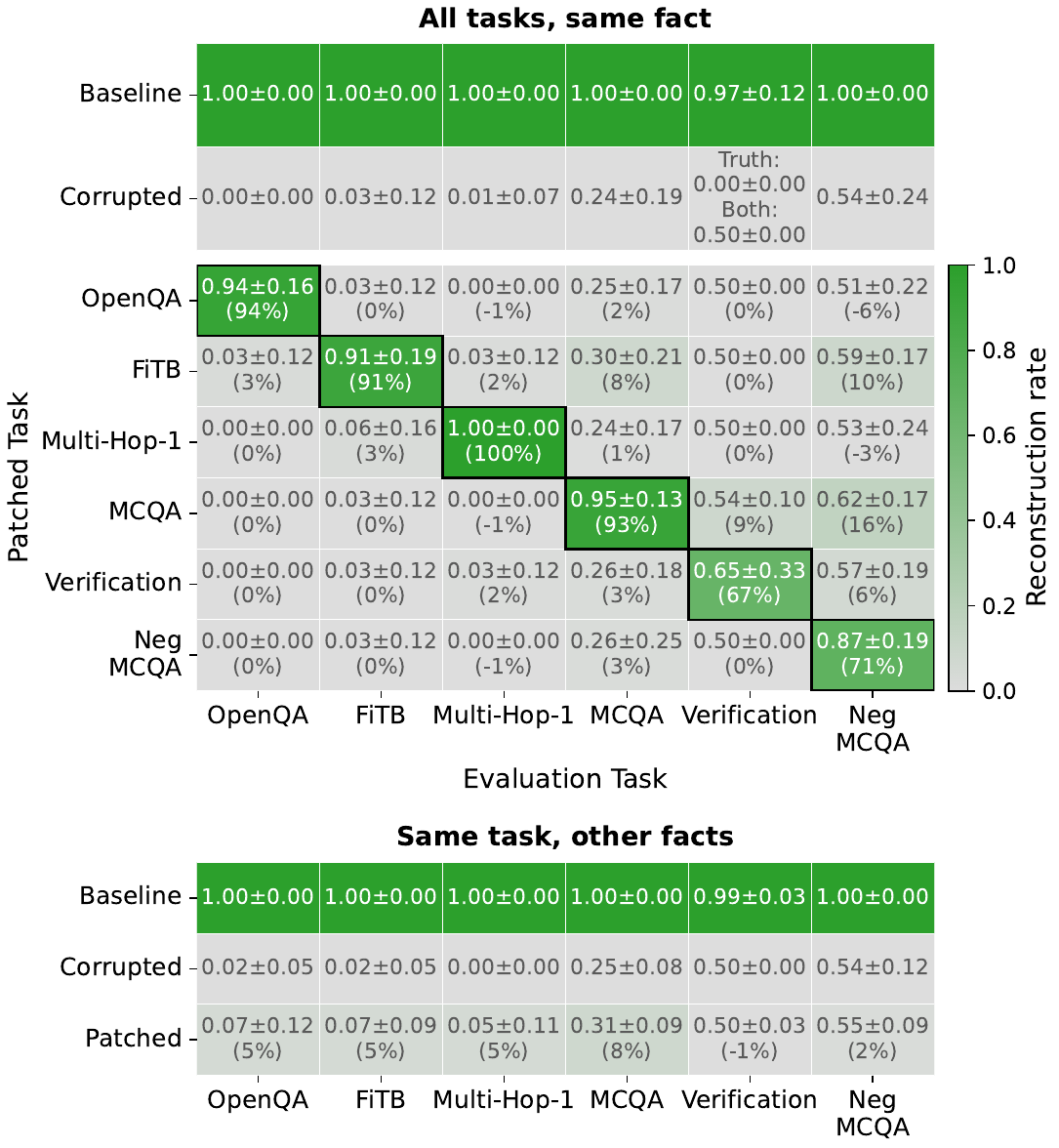}
            \caption{\texttt{(landmark, in-country, country)}}
            \label{fig:sufficiency_landmark_olmo}
        \end{subfigure}\\
        \begin{subfigure}{\linewidth}
            \centering
            \includegraphics[width=0.9\linewidth]{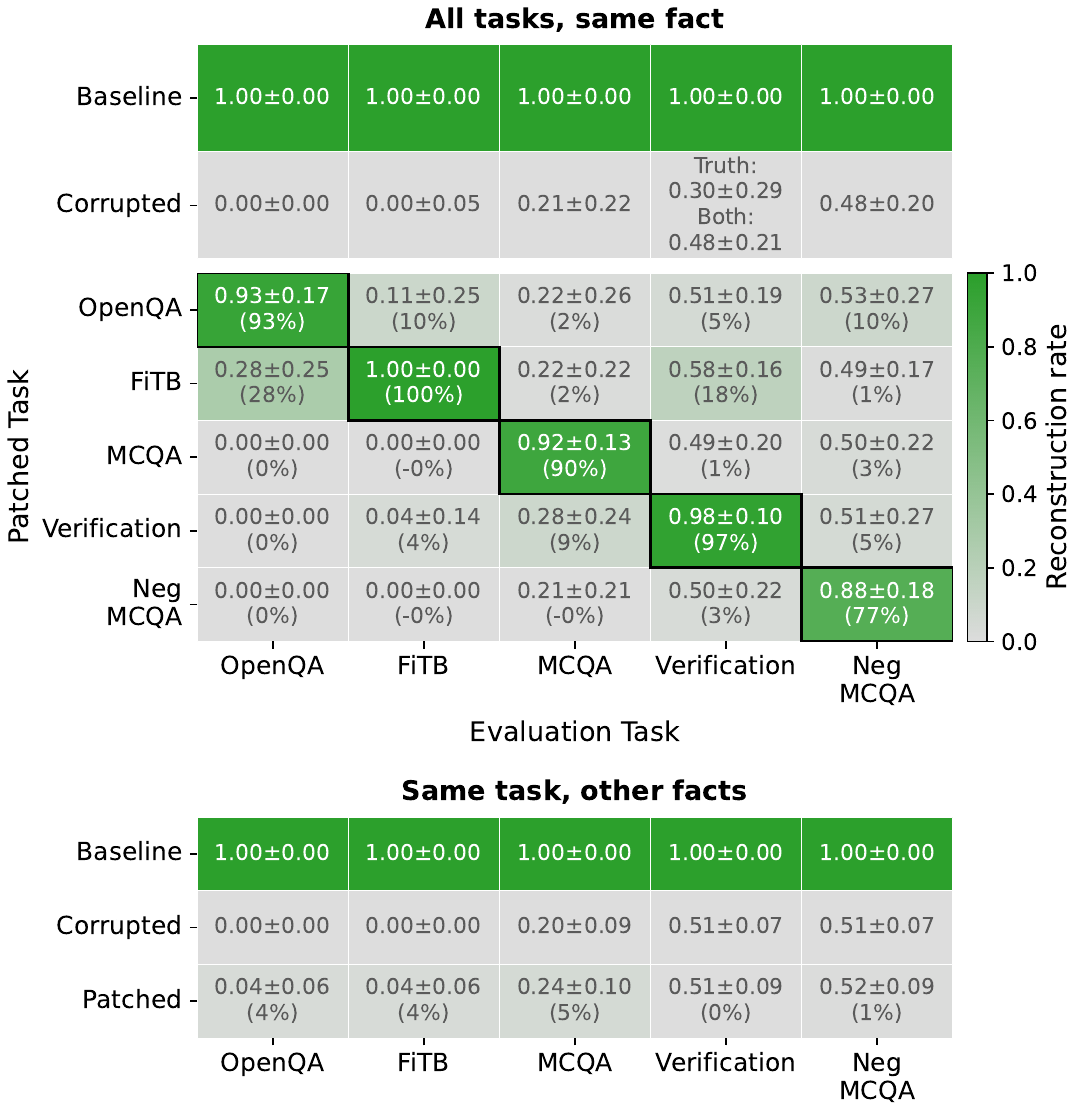}
            \caption{\texttt{(country, capital-of, city)}}
            \label{fig:sufficiency_capital_olmo}
        \end{subfigure}
    \end{minipage}\hfill
    \begin{minipage}[t]{0.49\linewidth}
        \centering
        \begin{subfigure}{\linewidth}
            \centering
            \includegraphics[width=0.9\linewidth]{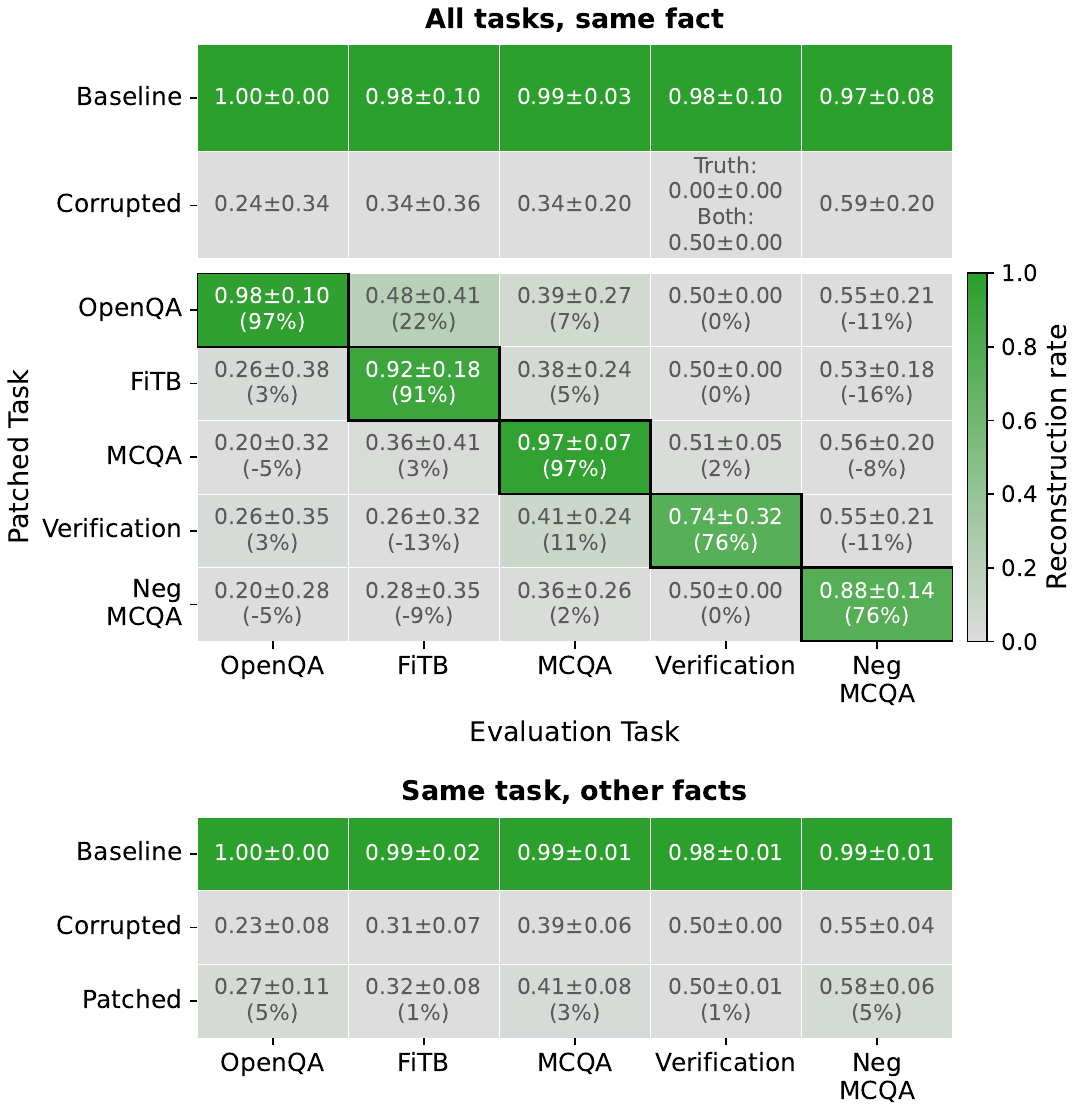}
            \caption{\texttt{(person, plays-instrument, instrument)}}
            \label{fig:sufficiency_instrument_olmo}
        \end{subfigure}\\
        \begin{subfigure}{\linewidth}
            \centering
            \includegraphics[width=0.9\linewidth]{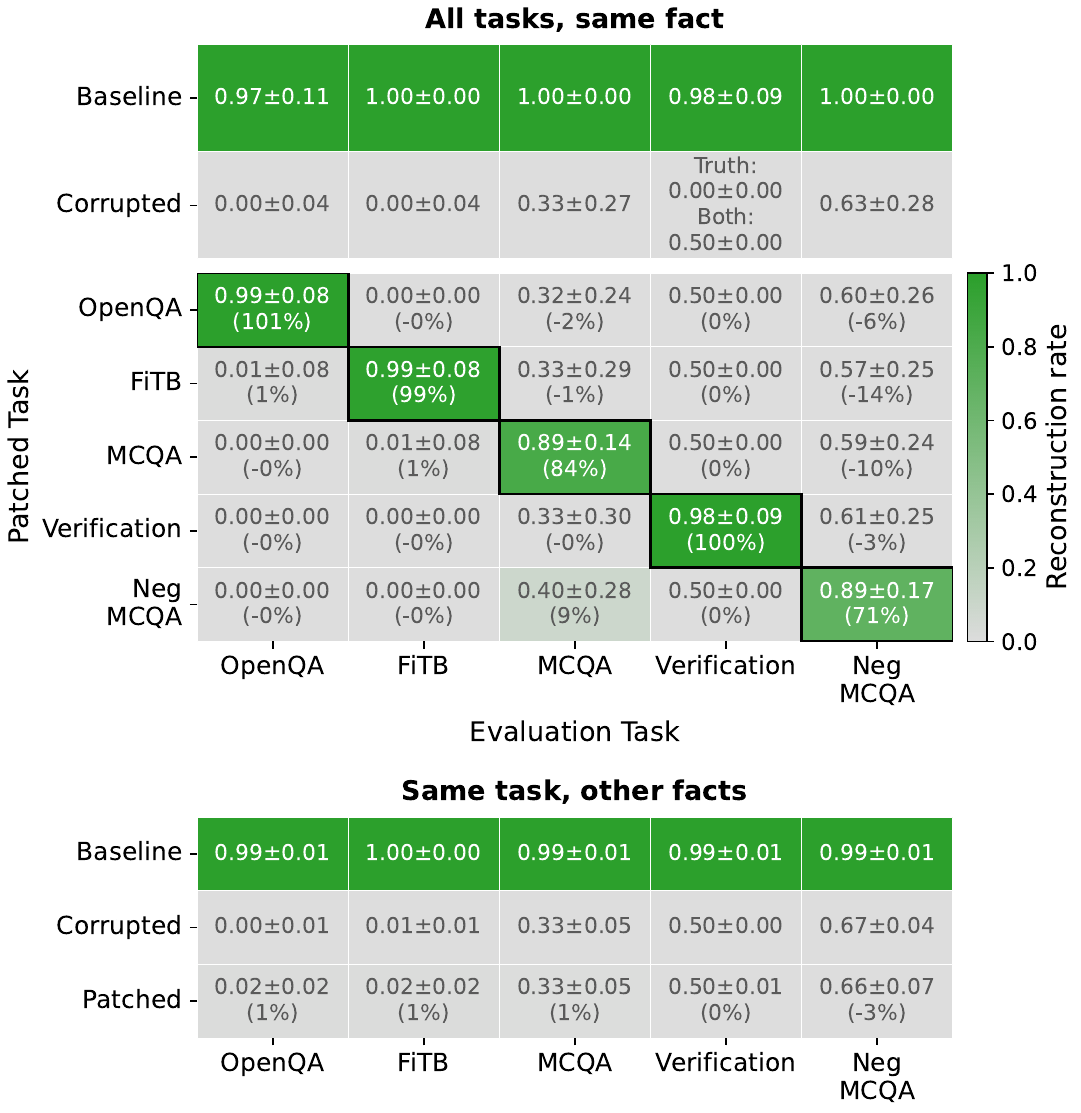}
            \caption{\texttt{(company, HQ-in-city, city)}}
            \label{fig:sufficiency_company_olmo}
        \end{subfigure}
    \end{minipage}
    \caption{Sufficiency results on \OlmoSevenB{}. Same layout as \Cref{fig:sufficiency_gemma}.}
    \label{fig:sufficiency_olmo_7b}
\end{figure*}

\paragraph{Full sufficiency results}
Sufficiency results demonstrate the same task-specific pattern across all three models and five datasets (\Cref{fig:sufficiency_gemma,fig:sufficiency_olmo_13b,fig:sufficiency_olmo_7b}). Patching the localized (fact,task) components' activations into the model's run on a corrupted prompt recovers performance primarily on the (fact, task) diagonal, with small recovery for the target fact on other tasks, or other facts on the target task.

\paragraph{Ruling out search artifacts}
A possible concern is that these subsets exist merely because the search is expressive enough to find them in a large network. However, the reported metrics are measured under a fixed protocol,
on held-out paraphrases and facts, 
independently of the search. 

\paragraph{Ruling out task-specific circuitry}
The two specificity axes (same task evaluated on other facts, and other tasks evaluated on the same fact) constrain the space of explanations for the mechanistic results. A shared, task-invariant store accessed through task-specific readout circuitry predicts that ablating the readout degrades \emph{all} facts on that task, and that ablating the store degrades that fact on \emph{all} tasks. The bottom-row and off-diagonal near-baseline values rule out both predictions. What remains is conditioned jointly on fact and task. Notably, any readout with that conditioning already encodes which fact it serves, so the storage/readout distinction does not affect our conclusion.

\begin{figure*}[t]
    \centering
    \begin{minipage}[t]{0.49\linewidth}
        \centering
        \begin{subfigure}{\linewidth}
            \centering
            \includegraphics[width=\linewidth]{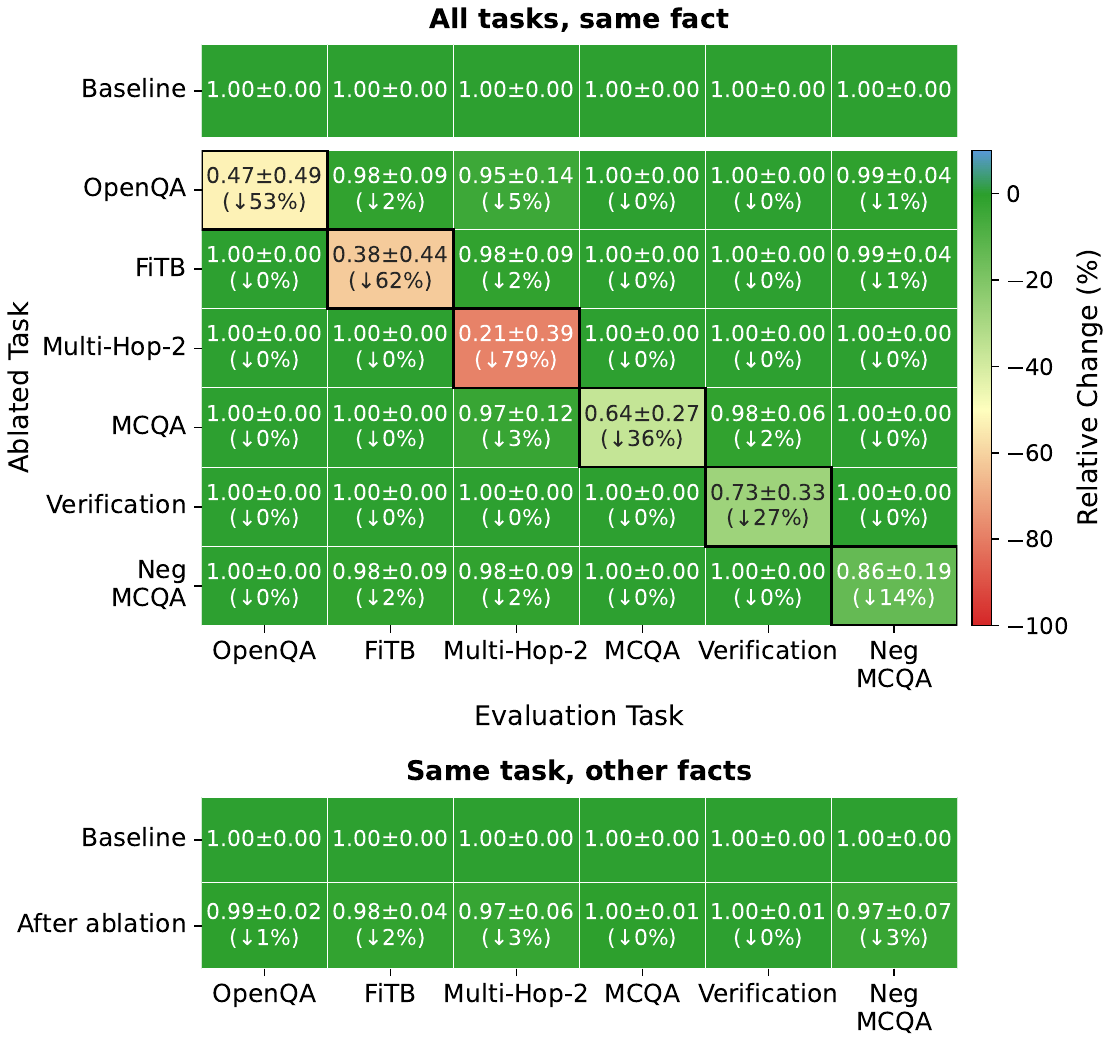}
            \caption{\texttt{(country, official language, language)}}
            \label{fig:necessity_nd_language_gemma_9b}
        \end{subfigure}\\
        \begin{subfigure}{\linewidth}
            \centering
            \includegraphics[width=\linewidth]{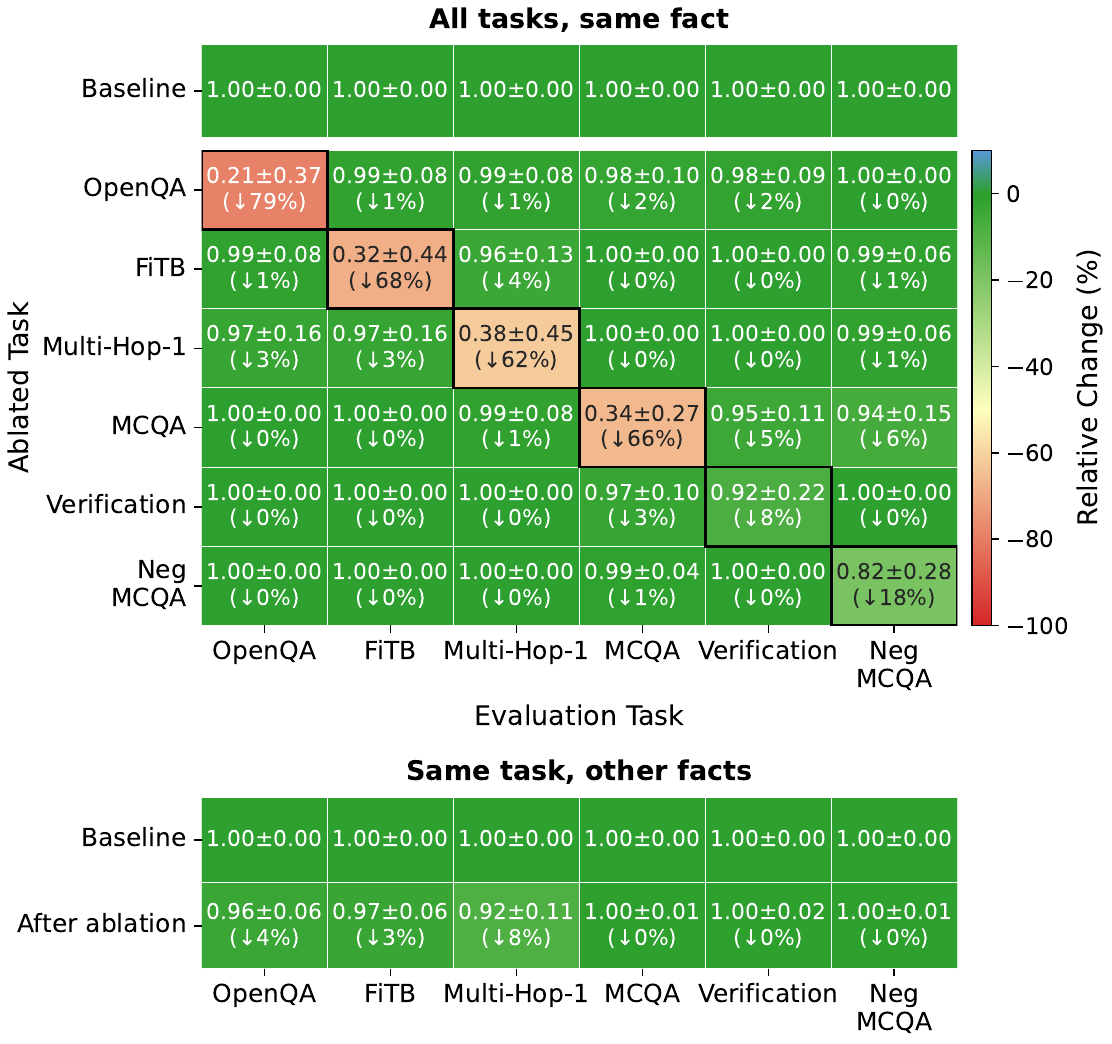}
            \caption{\texttt{(landmark, in-country, country)}}
            \label{fig:necessity_nd_landmark_gemma_9b}
        \end{subfigure}\\
        \begin{subfigure}{\linewidth}
            \centering
            \includegraphics[width=\linewidth]{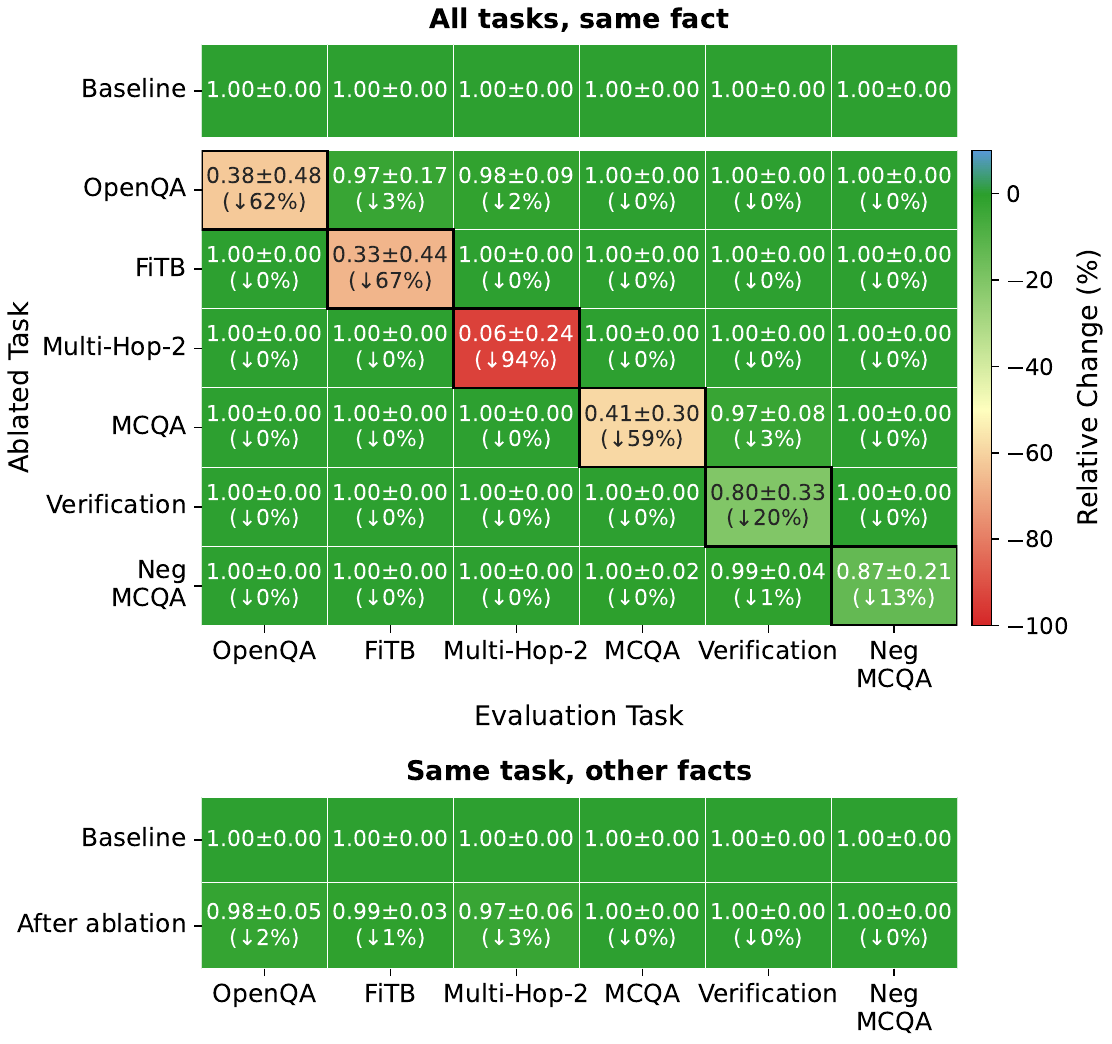}
            \caption{\texttt{(country, capital-of, city)}}
            \label{fig:necessity_nd_capital_gemma_9b}
        \end{subfigure}
    \end{minipage}\hfill
    \begin{minipage}[t]{0.49\linewidth}
        \centering
        \begin{subfigure}{\linewidth}
            \centering
            \includegraphics[width=\linewidth]{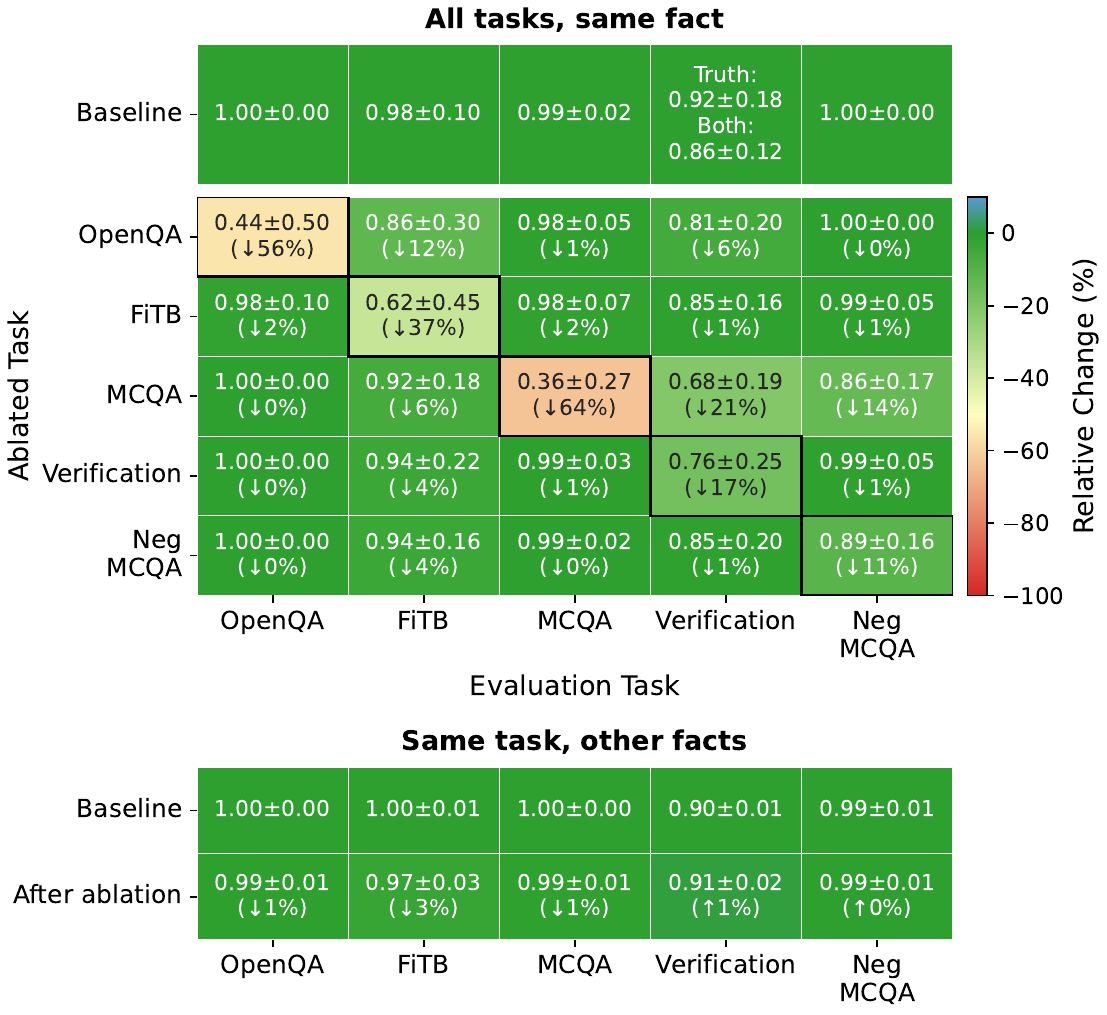}
            \caption{\texttt{(person, plays-instrument, instrument)}}
            \label{fig:necessity_nd_instrument_gemma_9b}
        \end{subfigure}\\
        \begin{subfigure}{\linewidth}
            \centering
            \includegraphics[width=\linewidth]{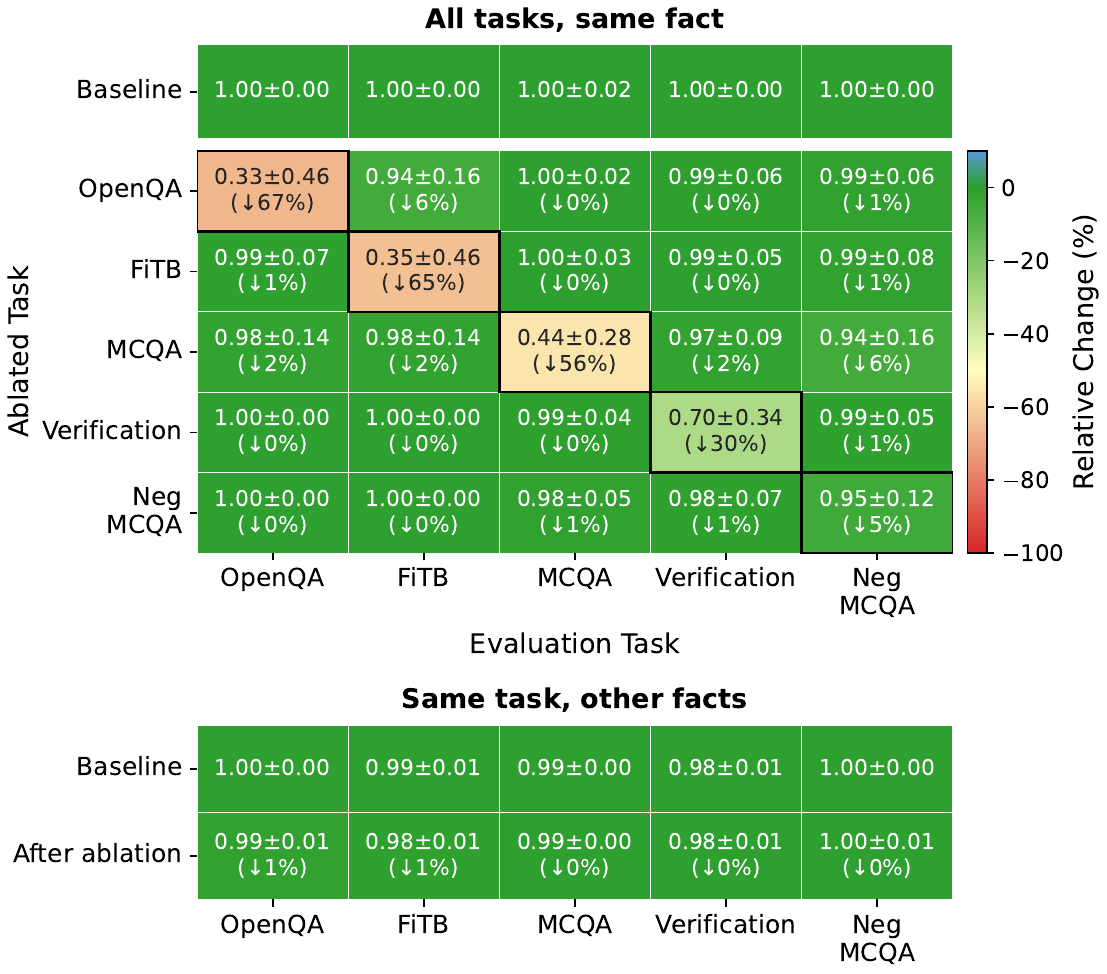}
            \caption{\texttt{(company, HQ-in-city, city)}}
            \label{fig:necessity_nd_company_gemma_9b}
        \end{subfigure}
    \end{minipage}
    \caption{Necessity results on \Gemma{} under \emph{non-disjoint} mask training. Same layout as \Cref{fig:necessity}.}
    \label{fig:necessity_nd_gemma_9b}
\end{figure*}

\begin{figure*}[t]
    \centering
    \begin{minipage}[t]{0.49\linewidth}
        \centering
        \begin{subfigure}{\linewidth}
            \centering
            \includegraphics[width=\linewidth]{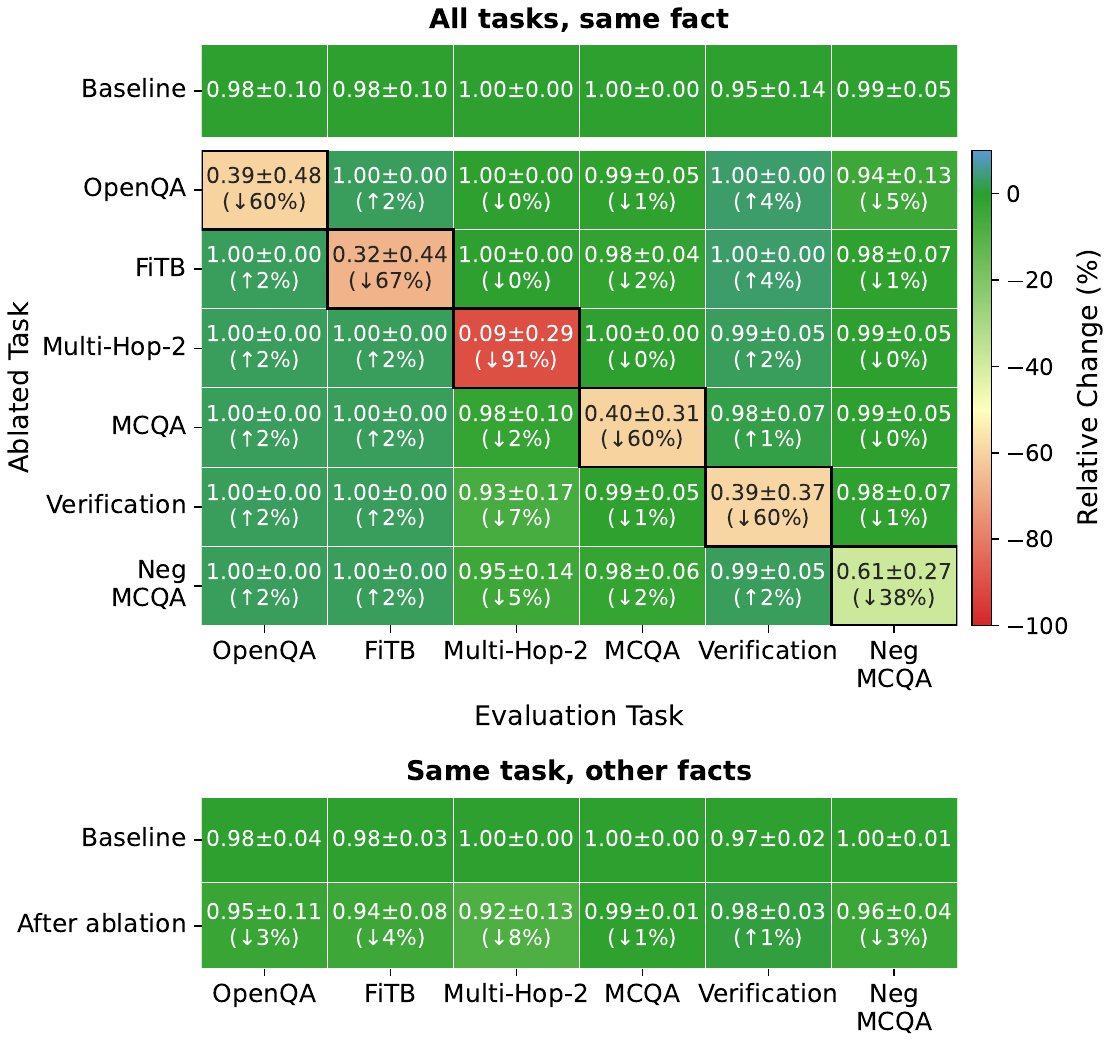}
            \caption{\texttt{(country, official language, language)}}
            \label{fig:necessity_nd_language_olmo_7b}
        \end{subfigure}\\
        \begin{subfigure}{\linewidth}
            \centering
            \includegraphics[width=\linewidth]{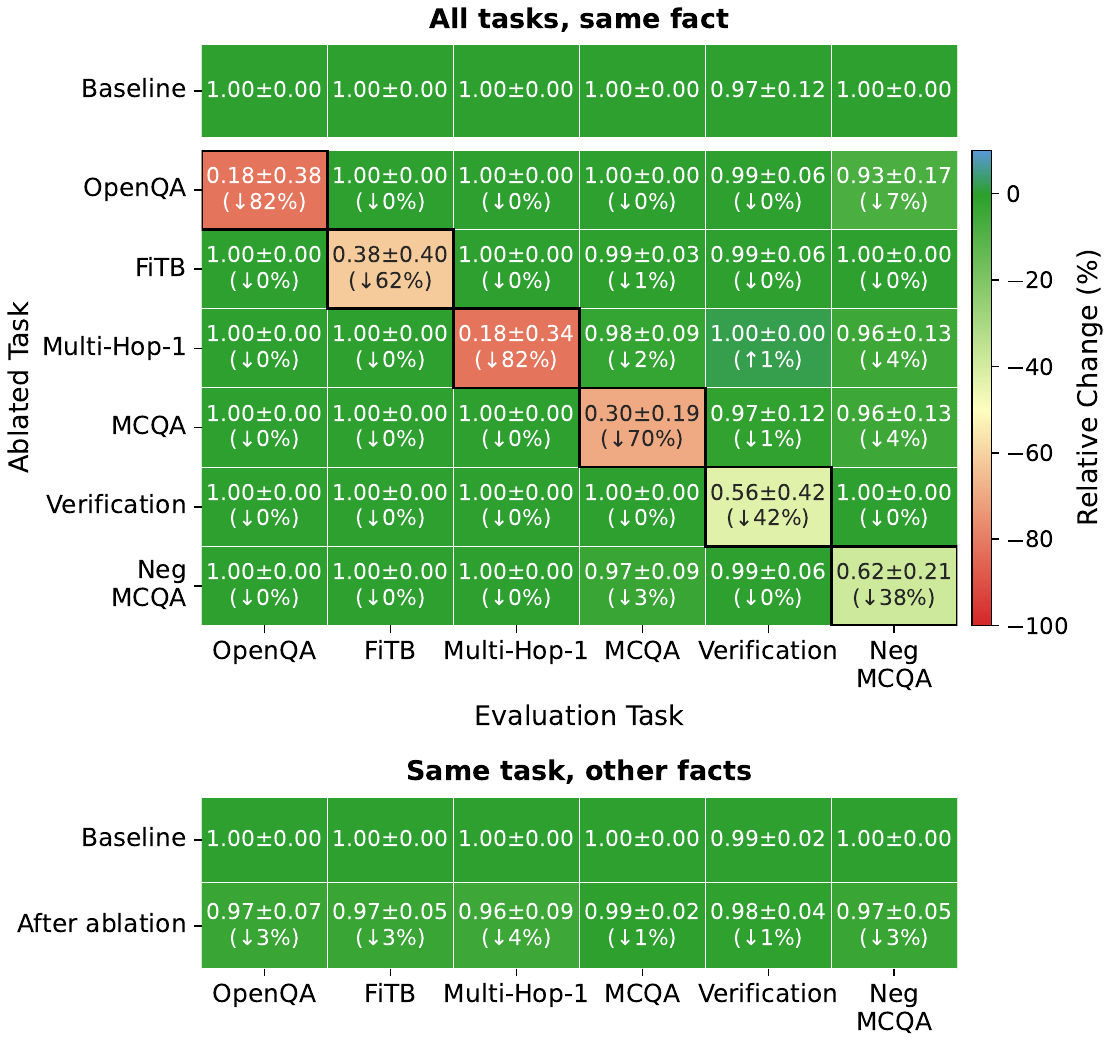}
            \caption{\texttt{(landmark, in-country, country)}}
            \label{fig:necessity_nd_landmark_olmo_7b}
        \end{subfigure}\\
        \begin{subfigure}{\linewidth}
            \centering
            \includegraphics[width=\linewidth]{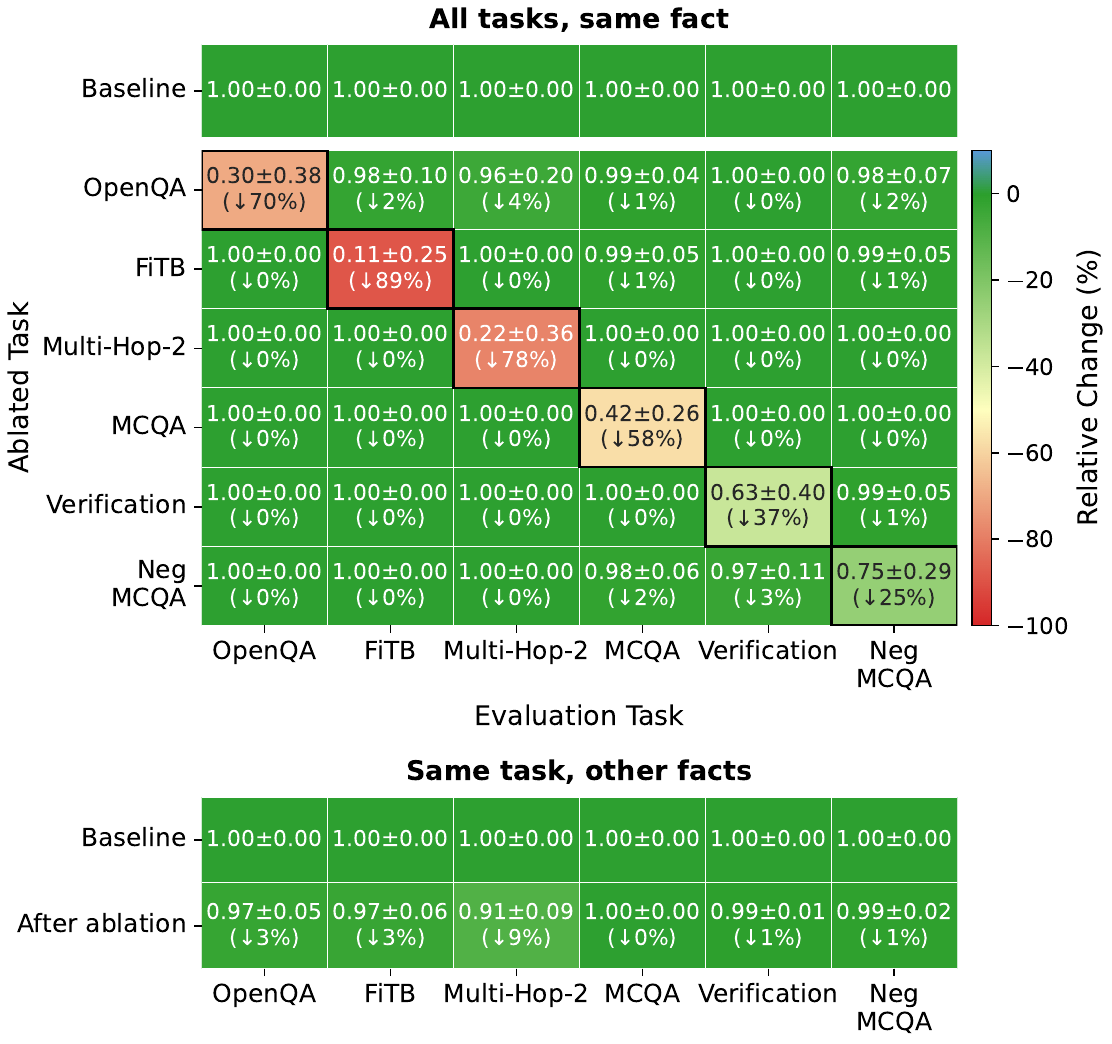}
            \caption{\texttt{(country, capital-of, city)}}
            \label{fig:necessity_nd_capital_olmo_7b}
        \end{subfigure}
    \end{minipage}\hfill
    \begin{minipage}[t]{0.49\linewidth}
        \centering
        \begin{subfigure}{\linewidth}
            \centering
            \includegraphics[width=\linewidth]{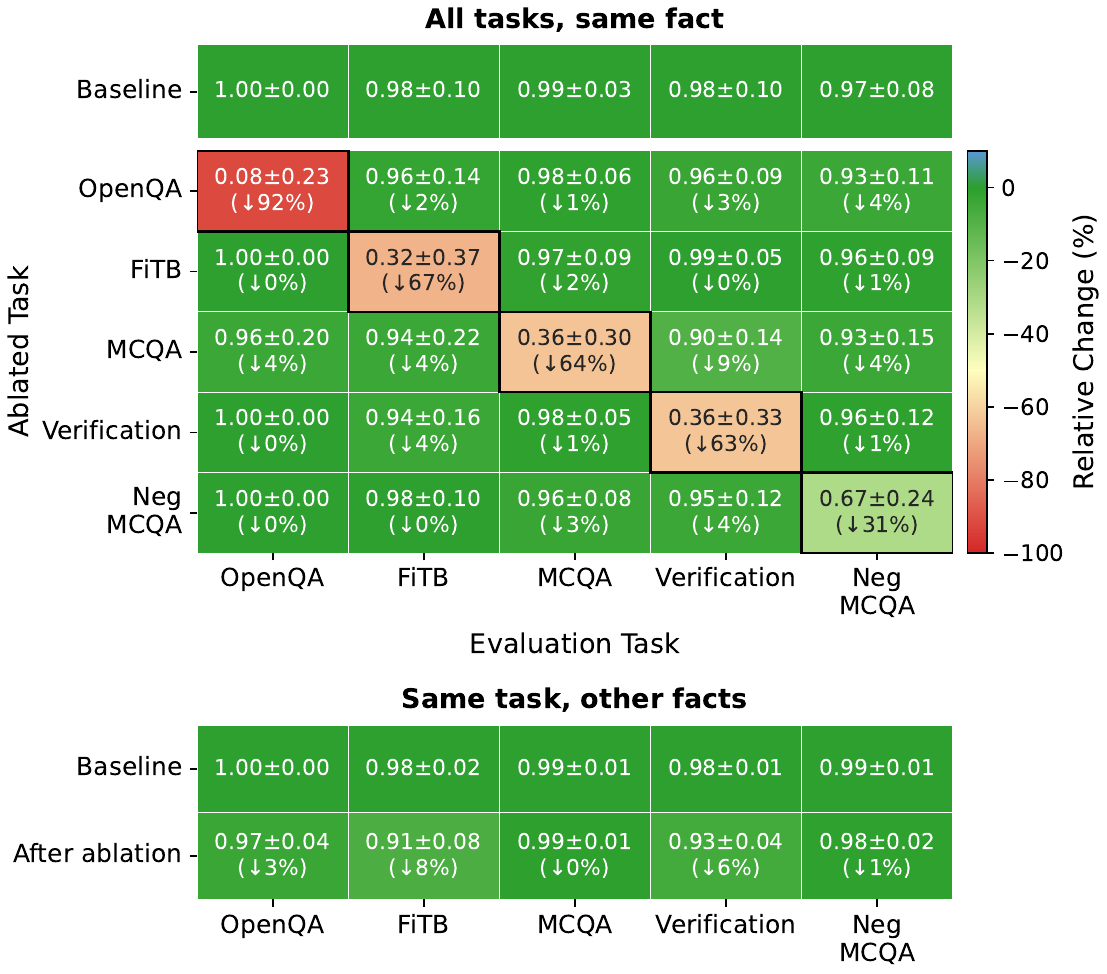}
            \caption{\texttt{(person, plays-instrument, instrument)}}
            \label{fig:necessity_nd_instrument_olmo_7b}
        \end{subfigure}\\
        \begin{subfigure}{\linewidth}
            \centering
            \includegraphics[width=\linewidth]{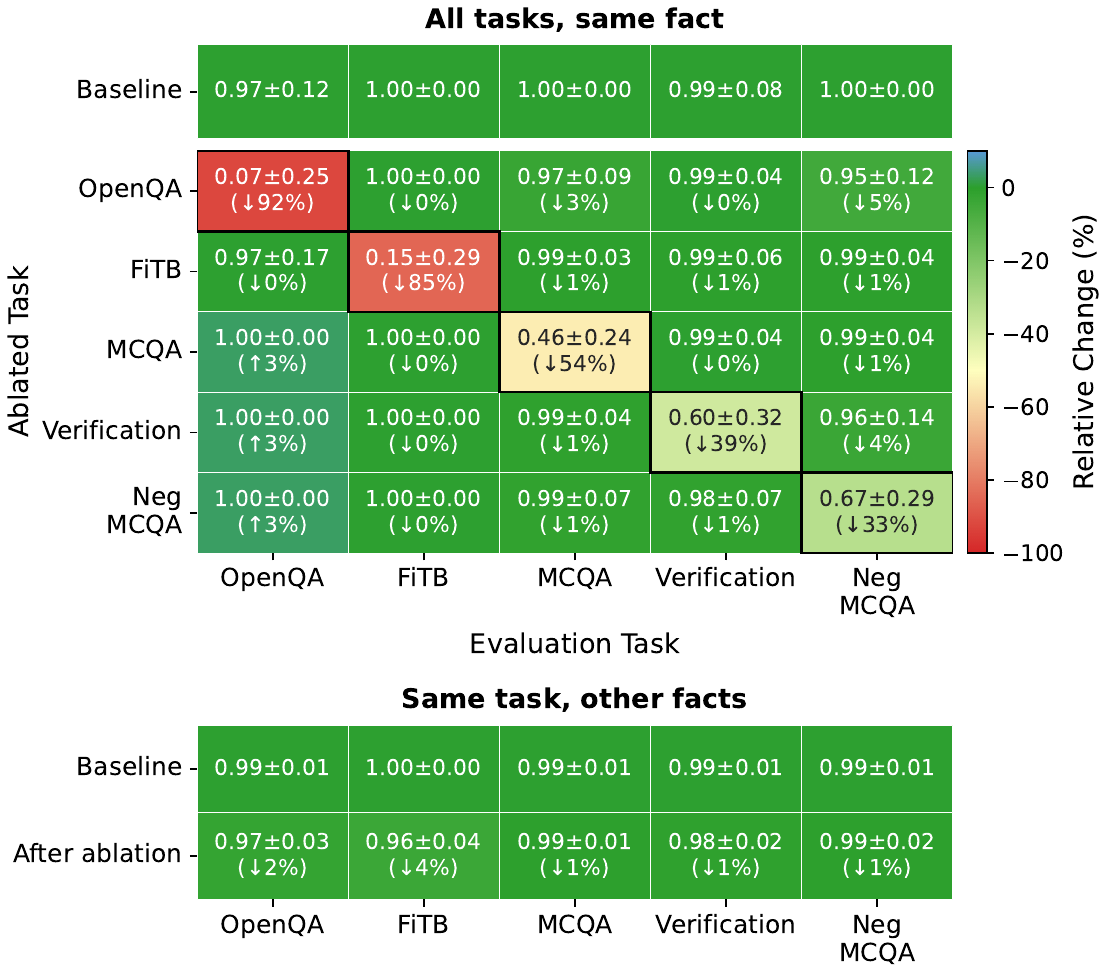}
            \caption{\texttt{(company, HQ-in-city, city)}}
            \label{fig:necessity_nd_company_olmo_7b}
        \end{subfigure}
    \end{minipage}
    \caption{Necessity results on \OlmoSevenB{} under \emph{non-disjoint} mask training. Same layout as \Cref{fig:necessity}.}
    \label{fig:necessity_nd_olmo_7b}
\end{figure*}

\begin{figure*}[t]
    \centering
    \begin{minipage}[t]{0.49\linewidth}
        \centering
        \begin{subfigure}{\linewidth}
            \centering
            \includegraphics[width=0.9\linewidth]{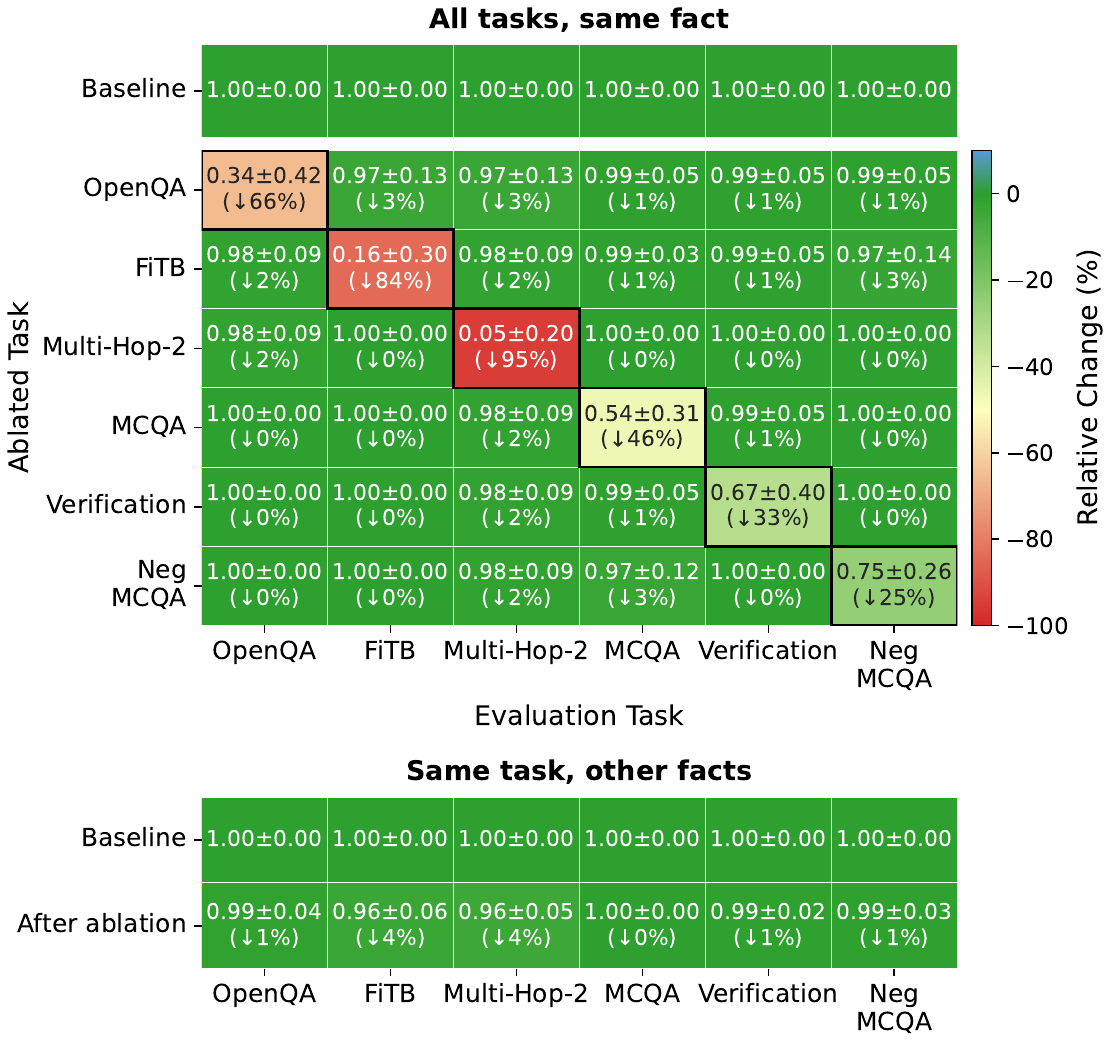}
            \caption{\texttt{(country, official language, language)}}
            \label{fig:necessity_nd_language_olmo_13b}
        \end{subfigure}\\
        \begin{subfigure}{\linewidth}
            \centering
            \includegraphics[width=0.9\linewidth]{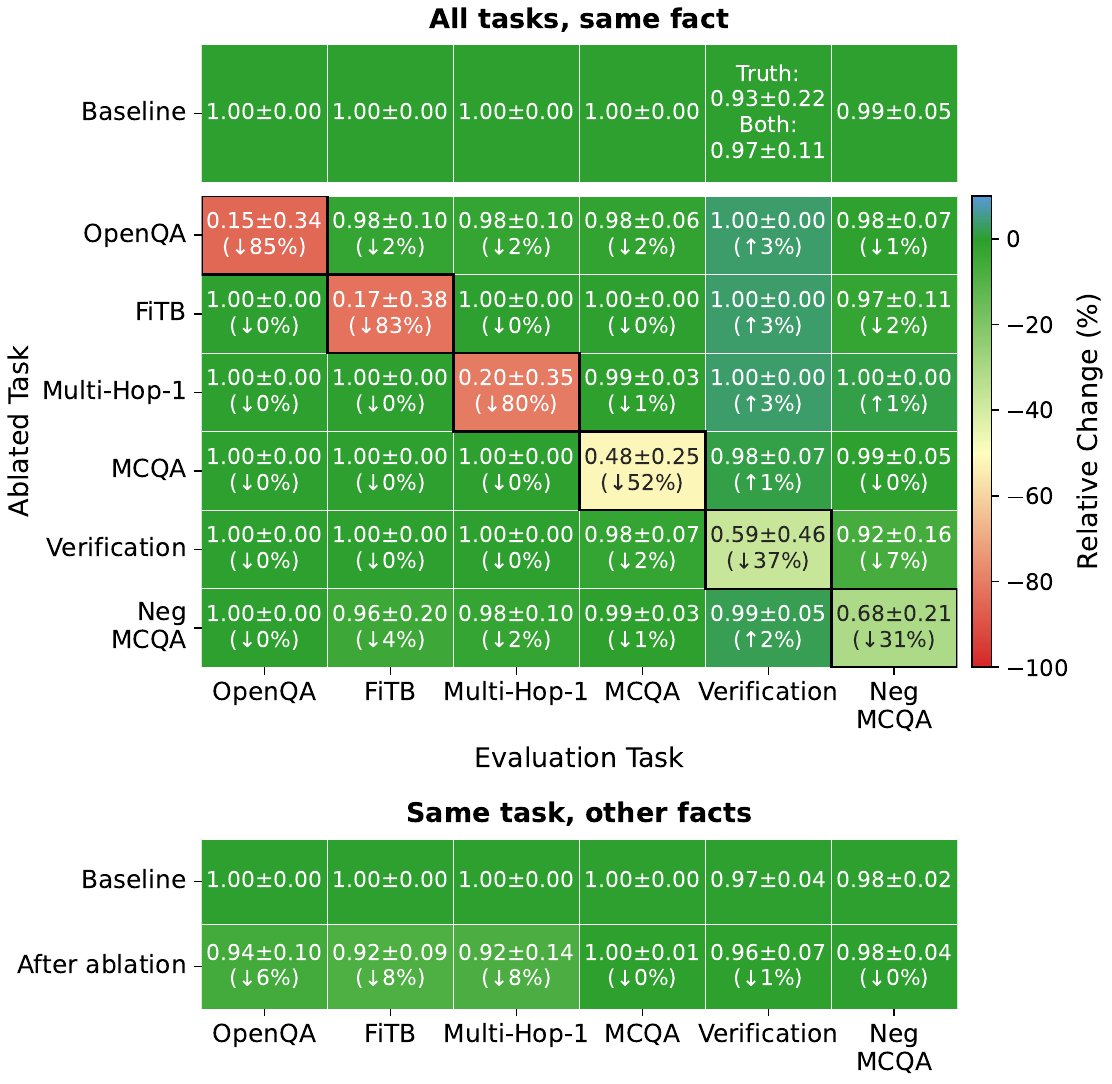}
            \caption{\texttt{(landmark, in-country, country)}}
            \label{fig:necessity_nd_landmark_olmo_13b}
        \end{subfigure}\\
        \begin{subfigure}{\linewidth}
            \centering
            \includegraphics[width=0.9\linewidth]{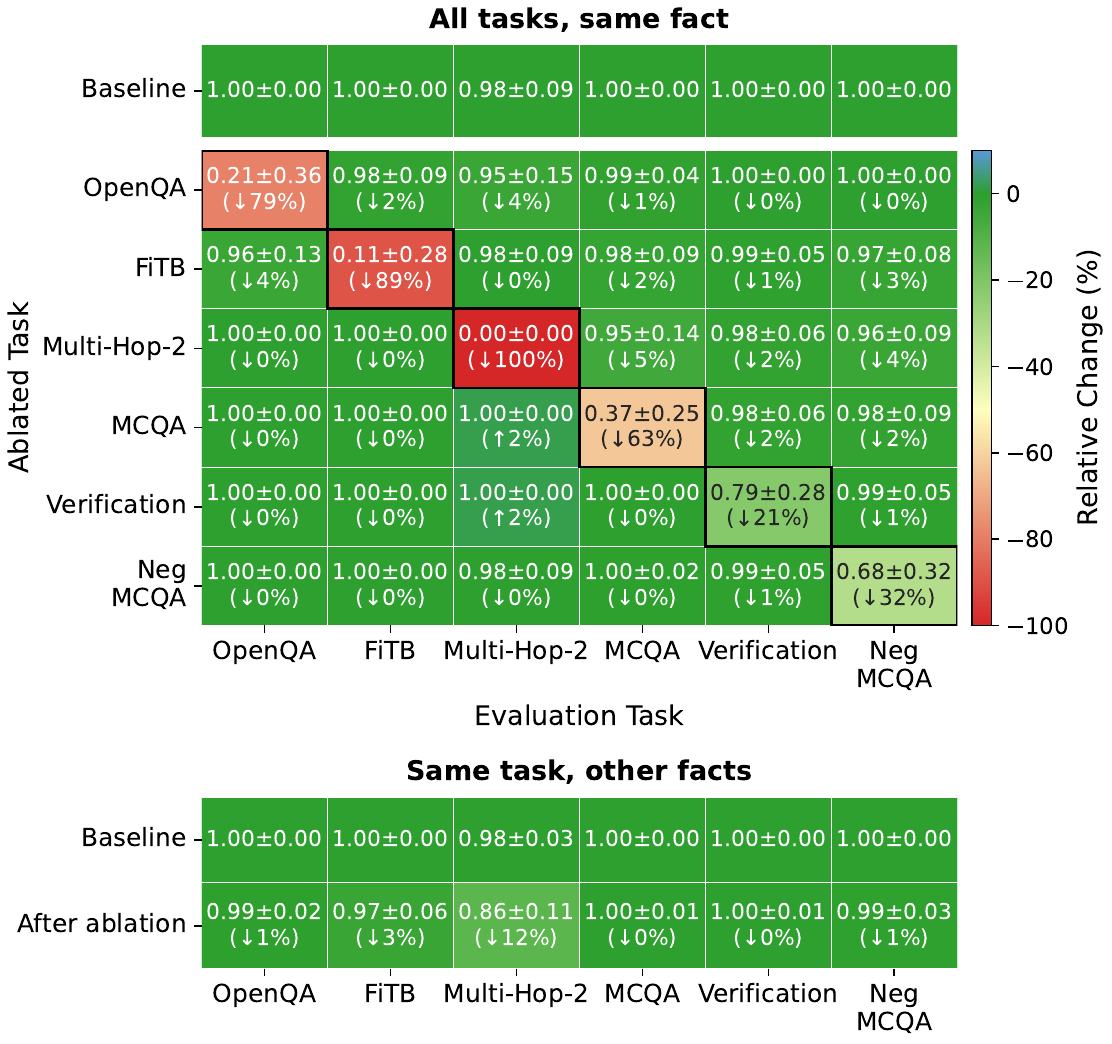}
            \caption{\texttt{(country, capital-of, city)}}
            \label{fig:necessity_nd_capital_olmo_13b}
        \end{subfigure}
    \end{minipage}\hfill
    \begin{minipage}[t]{0.49\linewidth}
        \centering
        \begin{subfigure}{\linewidth}
            \centering
            \includegraphics[width=0.9\linewidth]{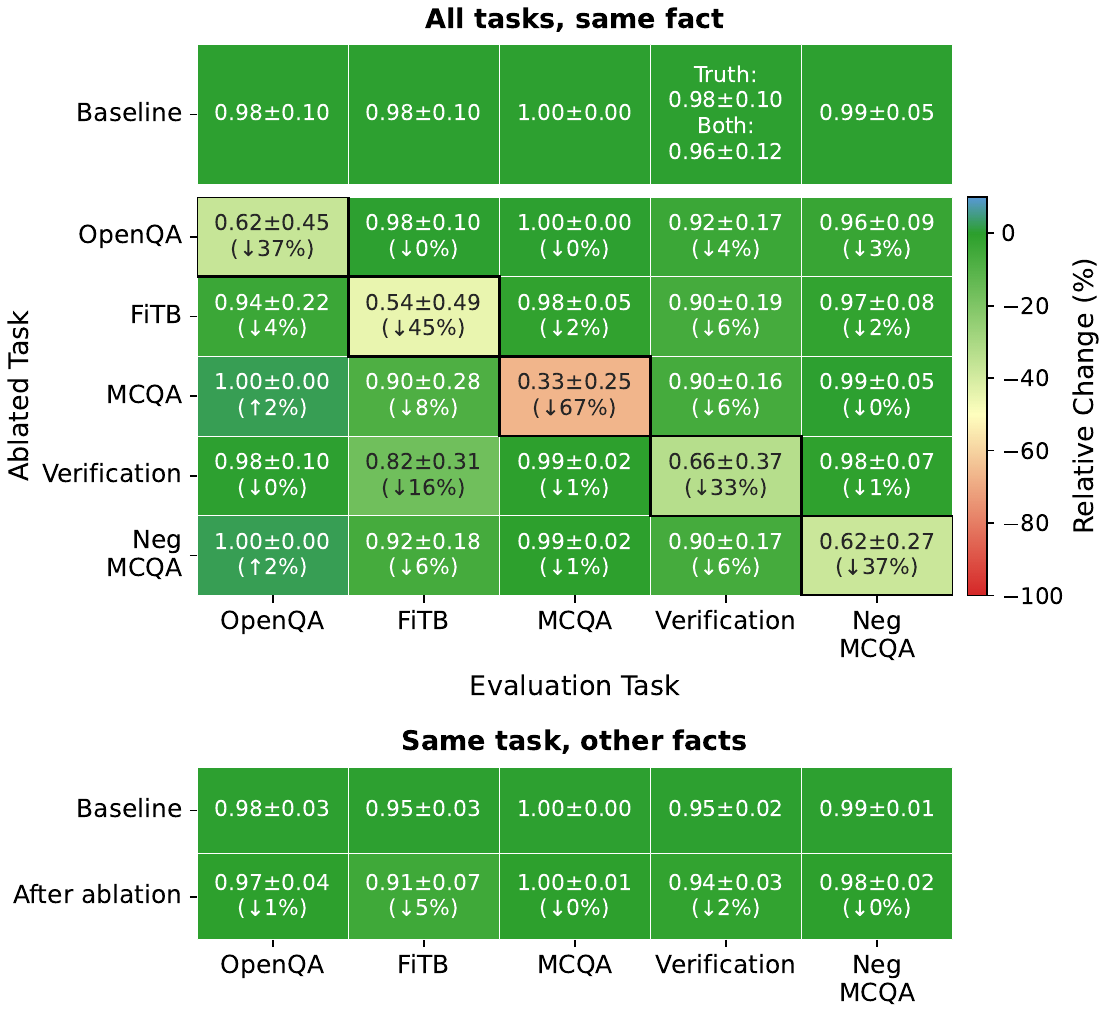}
            \caption{\texttt{(person, plays-instrument, instrument)}}
            \label{fig:necessity_nd_instrument_olmo_13b}
        \end{subfigure}\\
        \begin{subfigure}{\linewidth}
            \centering
            \includegraphics[width=0.9\linewidth]{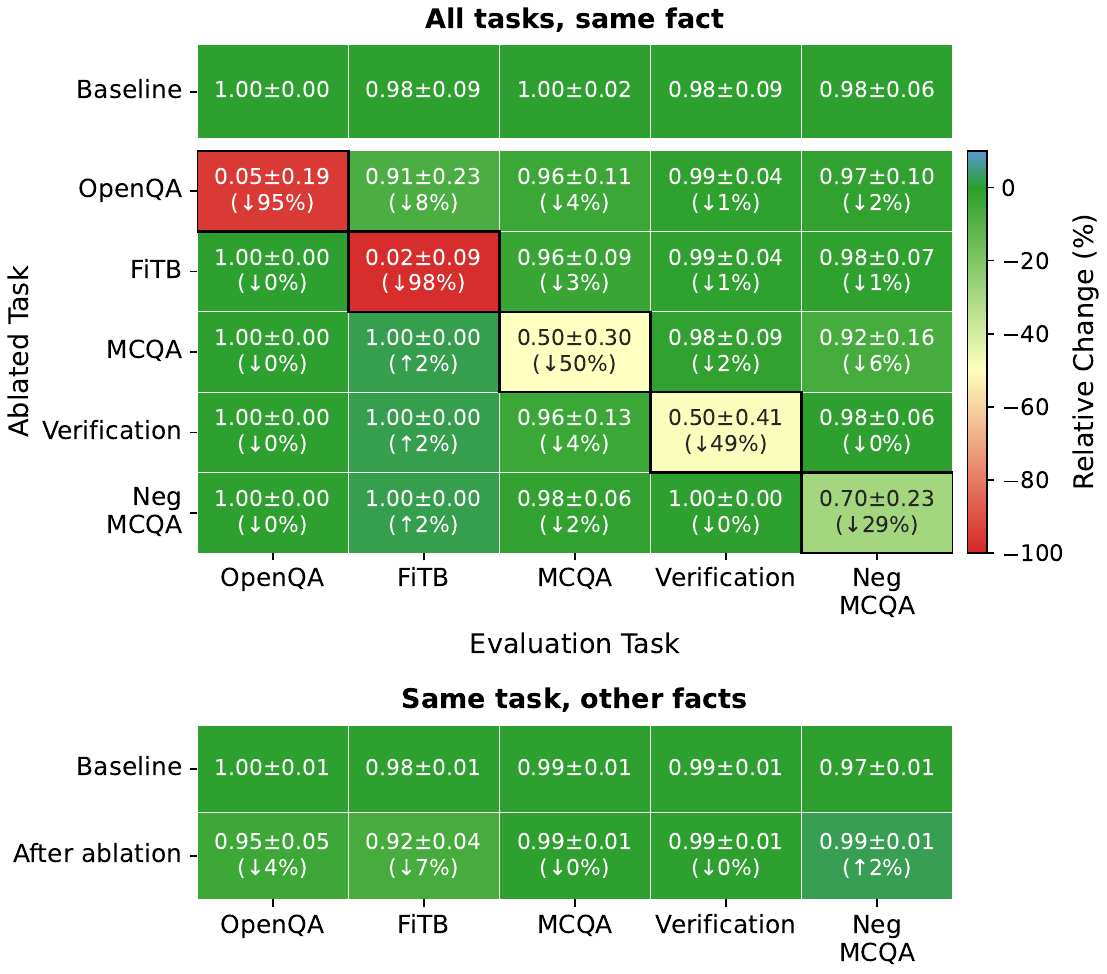}
            \caption{\texttt{(company, HQ-in-city, city)}}
            \label{fig:necessity_nd_company_olmo_13b}
        \end{subfigure}
    \end{minipage}
    \caption{Necessity results on \OlmoThirteenB{} under \emph{non-disjoint} mask training. Same layout as \Cref{fig:necessity}.}
    \label{fig:necessity_nd_olmo_13b}
\end{figure*}

\begin{figure*}[t]
    \centering
    \begin{minipage}[t]{0.49\linewidth}
        \centering
        \begin{subfigure}{\linewidth}
            \centering
            \includegraphics[width=0.88\linewidth,trim=0 15pt 0 0pt,clip]{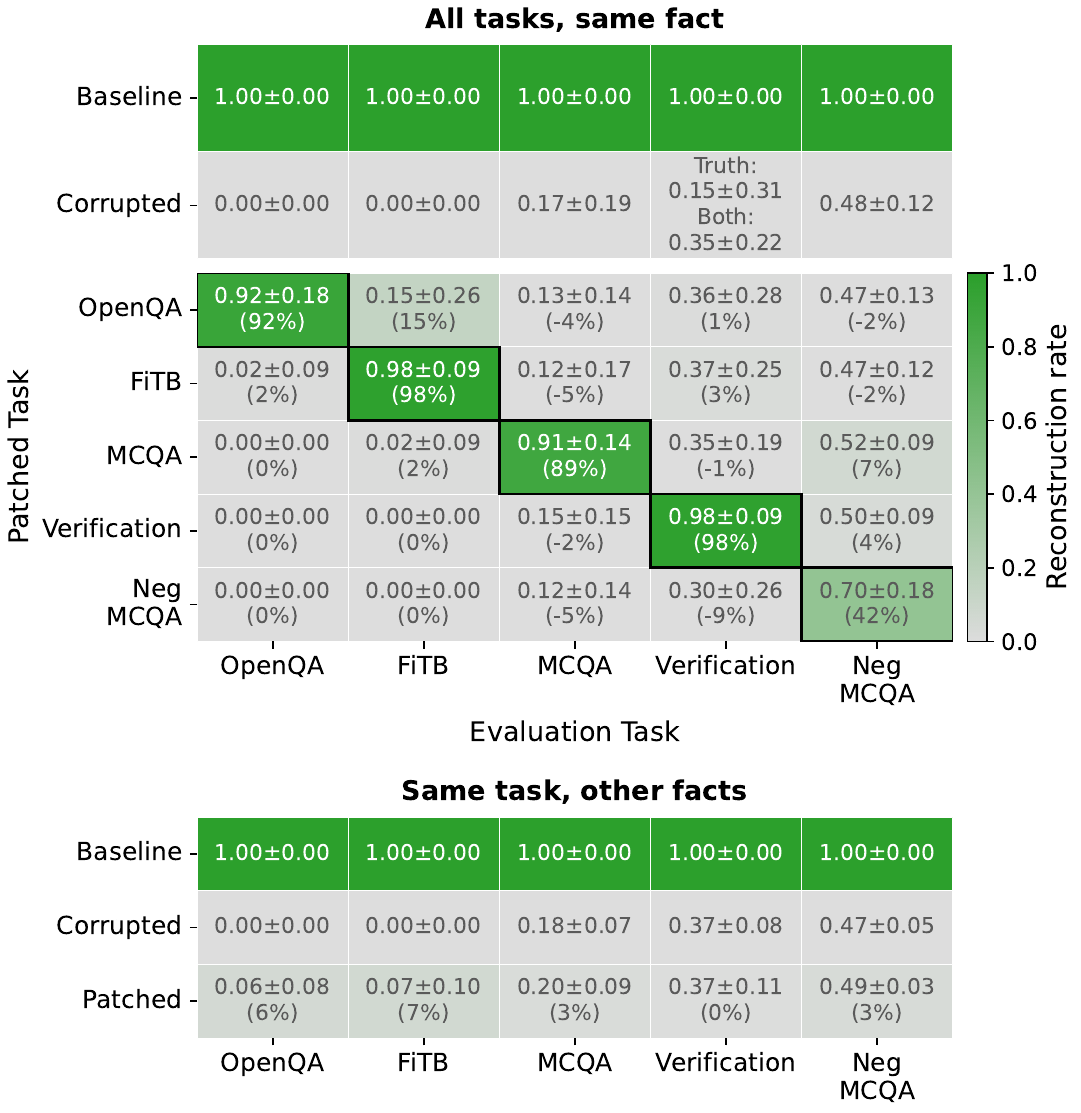}
            \caption{\texttt{(country, official language, language)}}
            \label{fig:sufficiency_nd_language_gemma_9b}
        \end{subfigure}\\
        \begin{subfigure}{\linewidth}
            \centering
            \includegraphics[width=0.88\linewidth,trim=0 15pt 0 0pt,clip]{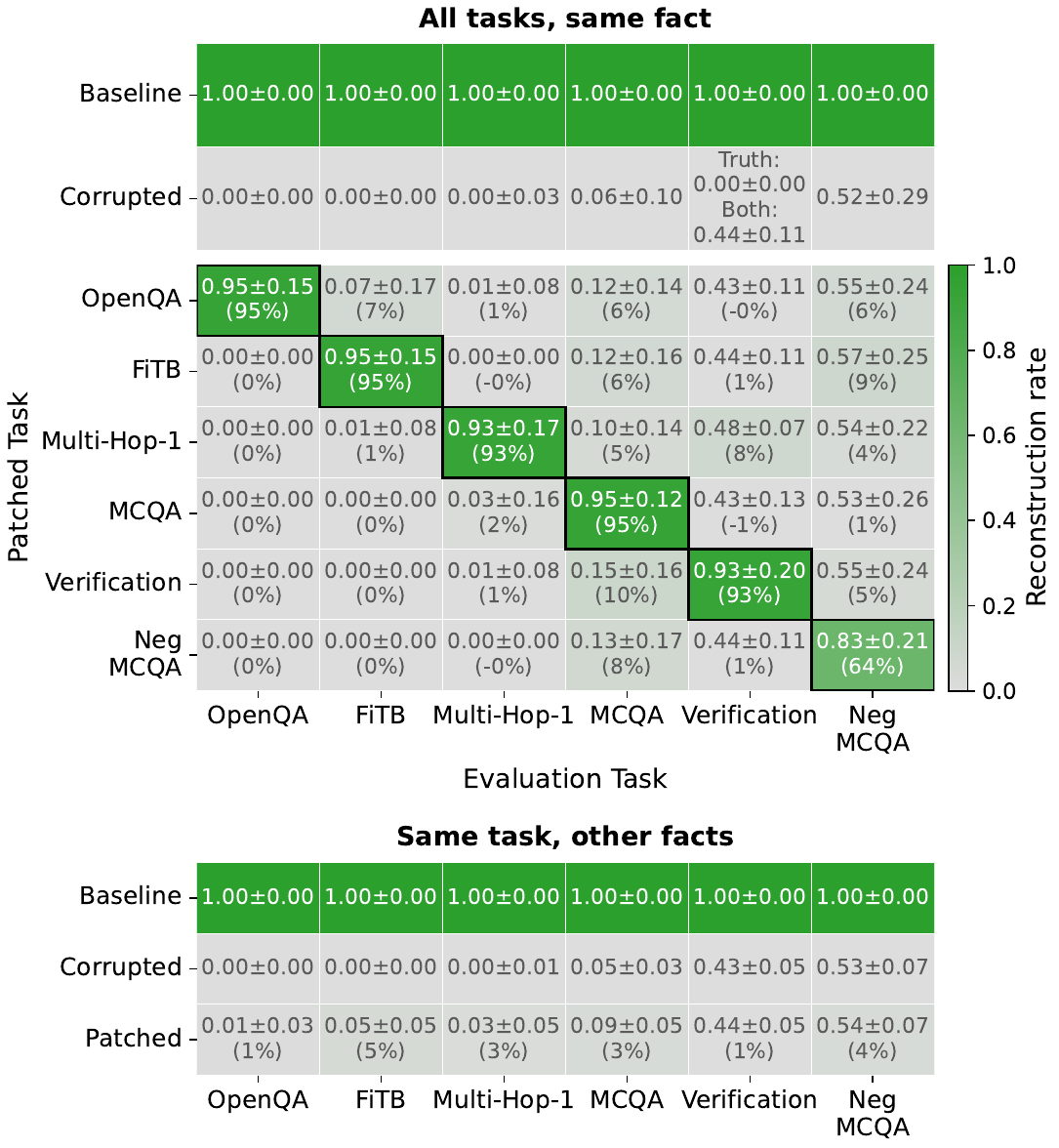}
            \caption{\texttt{(landmark, in-country, country)}}
            \label{fig:sufficiency_nd_landmark_gemma_9b}
        \end{subfigure}\\
        \begin{subfigure}{\linewidth}
            \centering
            \includegraphics[width=0.88\linewidth,trim=0 15pt 0 0pt,clip]{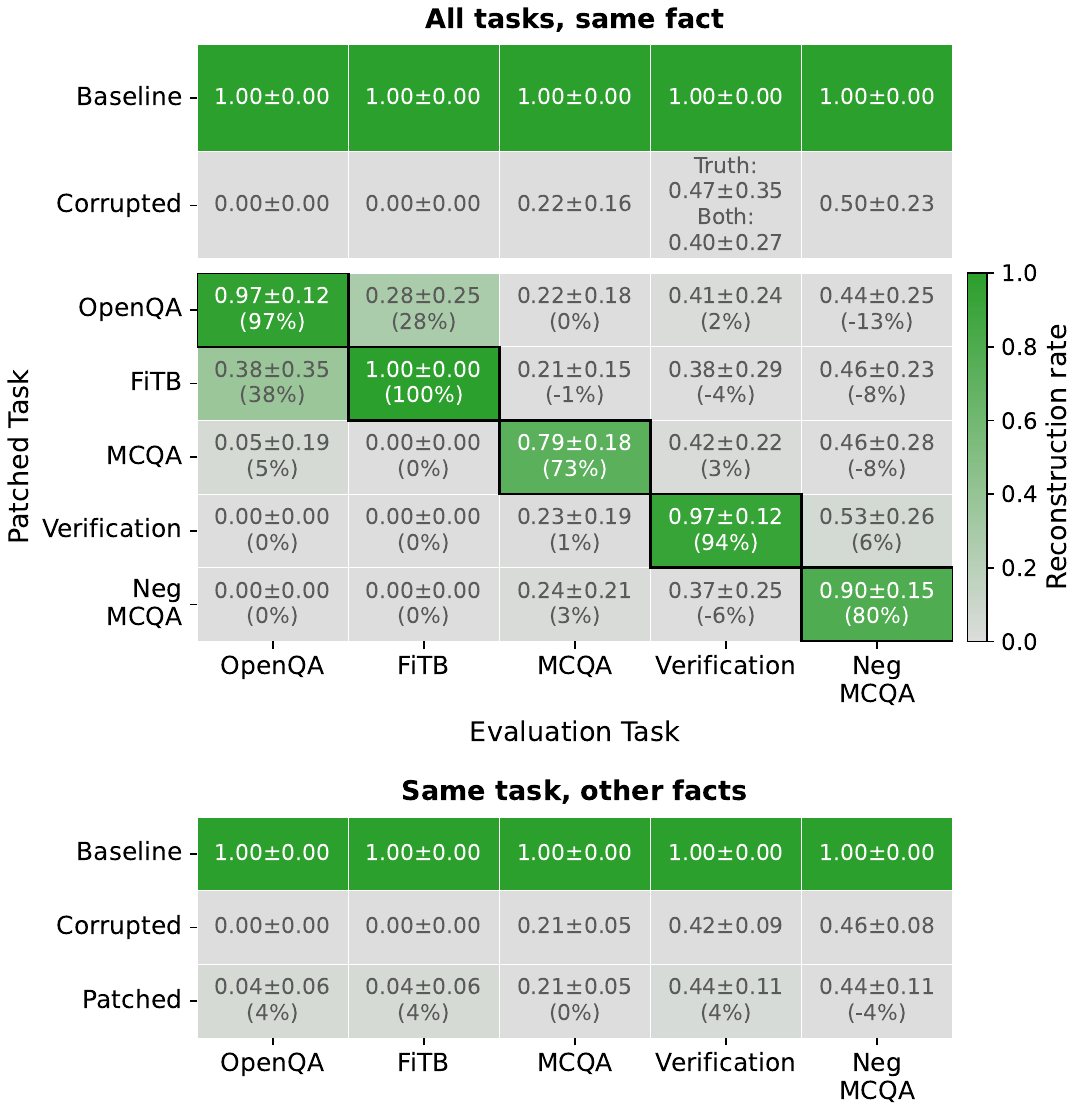}
            \caption{\texttt{(country, capital-of, city)}}
            \label{fig:sufficiency_nd_capital_gemma_9b}
        \end{subfigure}
    \end{minipage}\hfill
    \begin{minipage}[t]{0.49\linewidth}
        \centering
        \begin{subfigure}{\linewidth}
            \centering
            \includegraphics[width=0.88\linewidth,trim=0 15pt 0 0pt,clip]{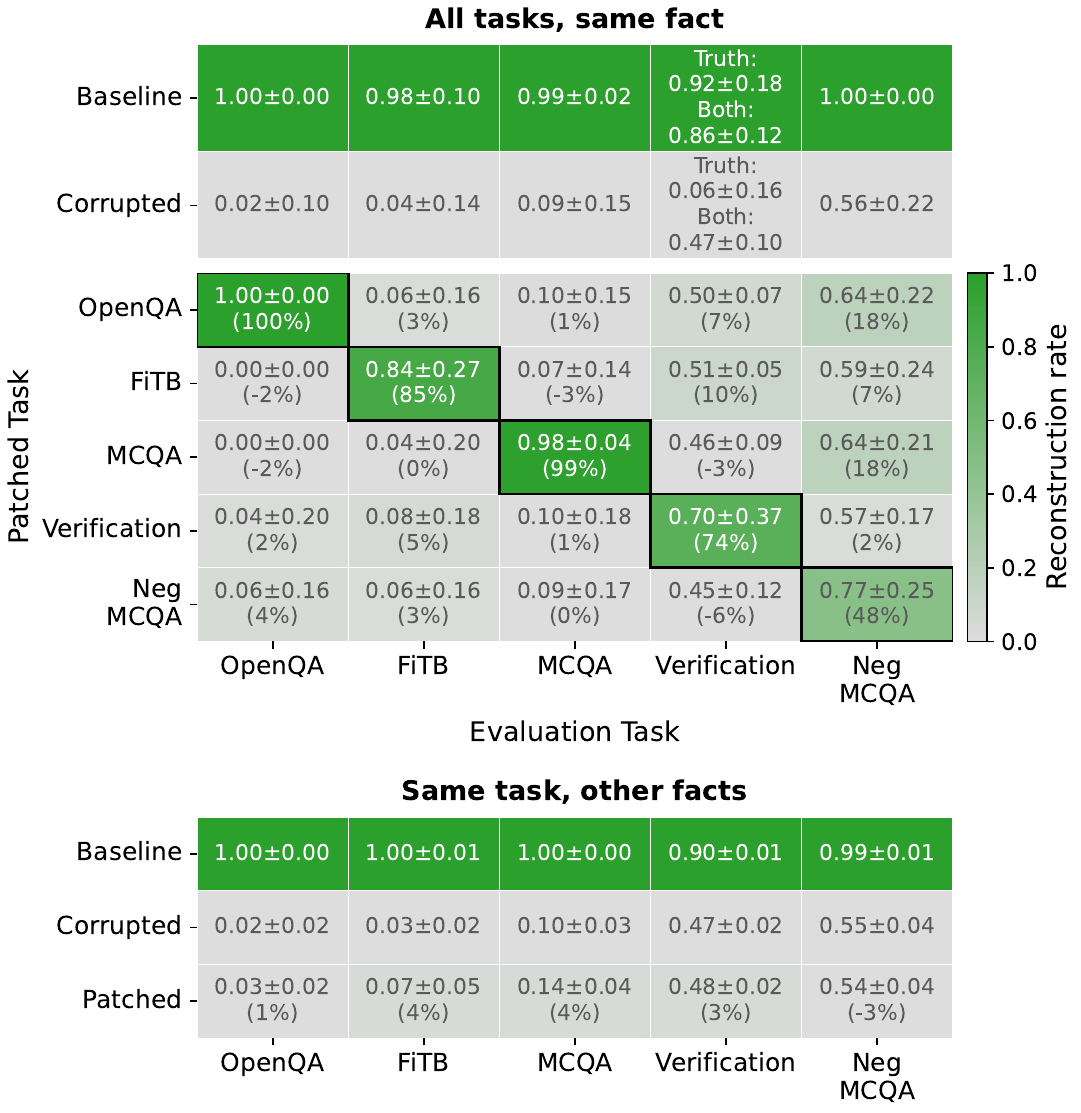}
            \caption{\texttt{(person, plays-instrument, instrument)}}
            \label{fig:sufficiency_nd_instrument_gemma_9b}
        \end{subfigure}\\
        \begin{subfigure}{\linewidth}
            \centering
            \includegraphics[width=0.88\linewidth,trim=0 15pt 0 0pt,clip]{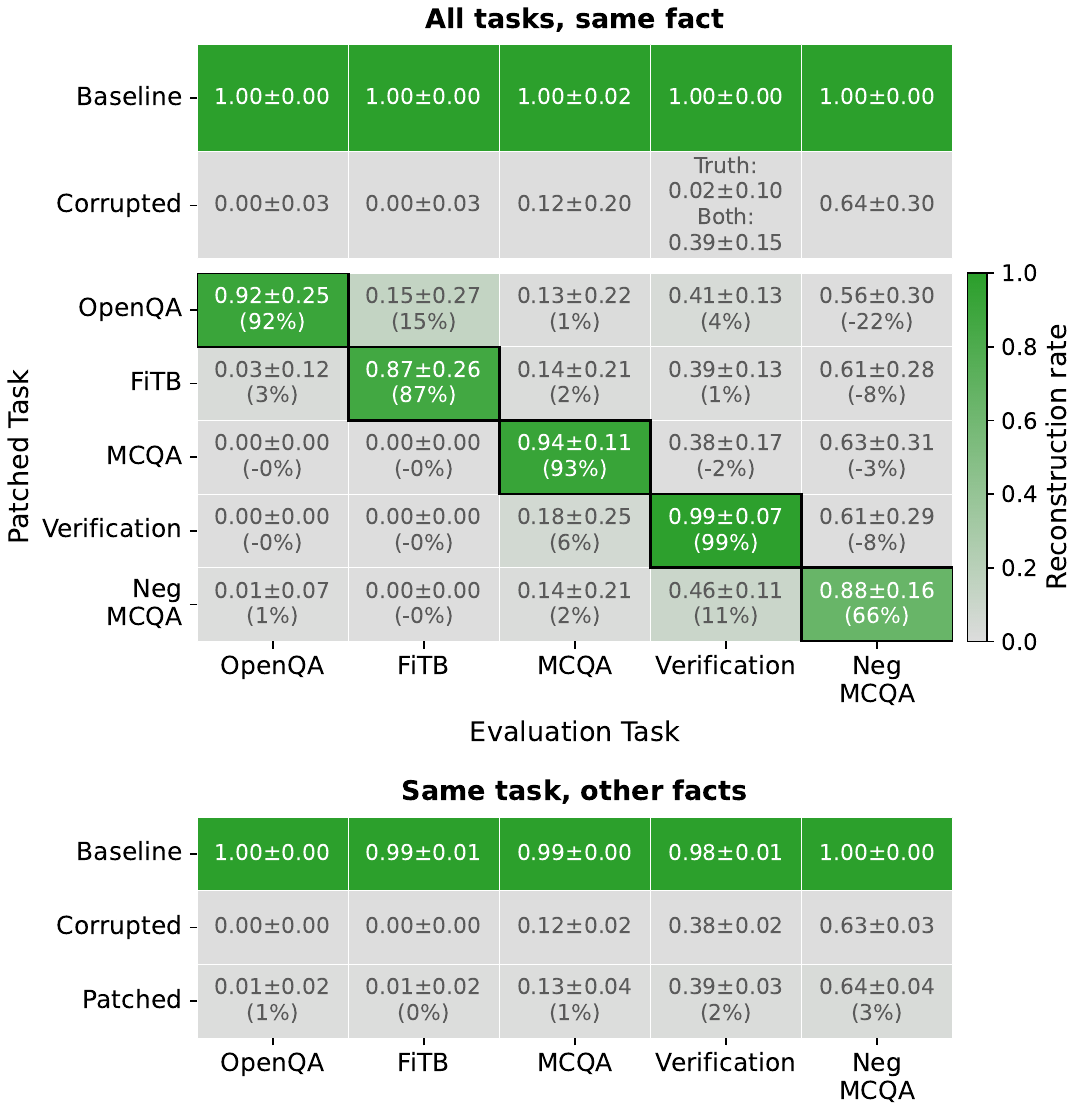}
            \caption{\texttt{(company, HQ-in-city, city)}}
            \label{fig:sufficiency_nd_company_gemma_9b}
        \end{subfigure}
    \end{minipage}
    \caption{Sufficiency results on \Gemma{} under \emph{non-disjoint} mask training. Same layout as \Cref{fig:sufficiency_gemma}.}
    \label{fig:sufficiency_nd_gemma_9b}
\end{figure*}

\begin{figure*}[t]
    \centering
    \begin{minipage}[t]{0.49\linewidth}
        \centering
        \begin{subfigure}{\linewidth}
            \centering
            \includegraphics[width=0.9\linewidth]{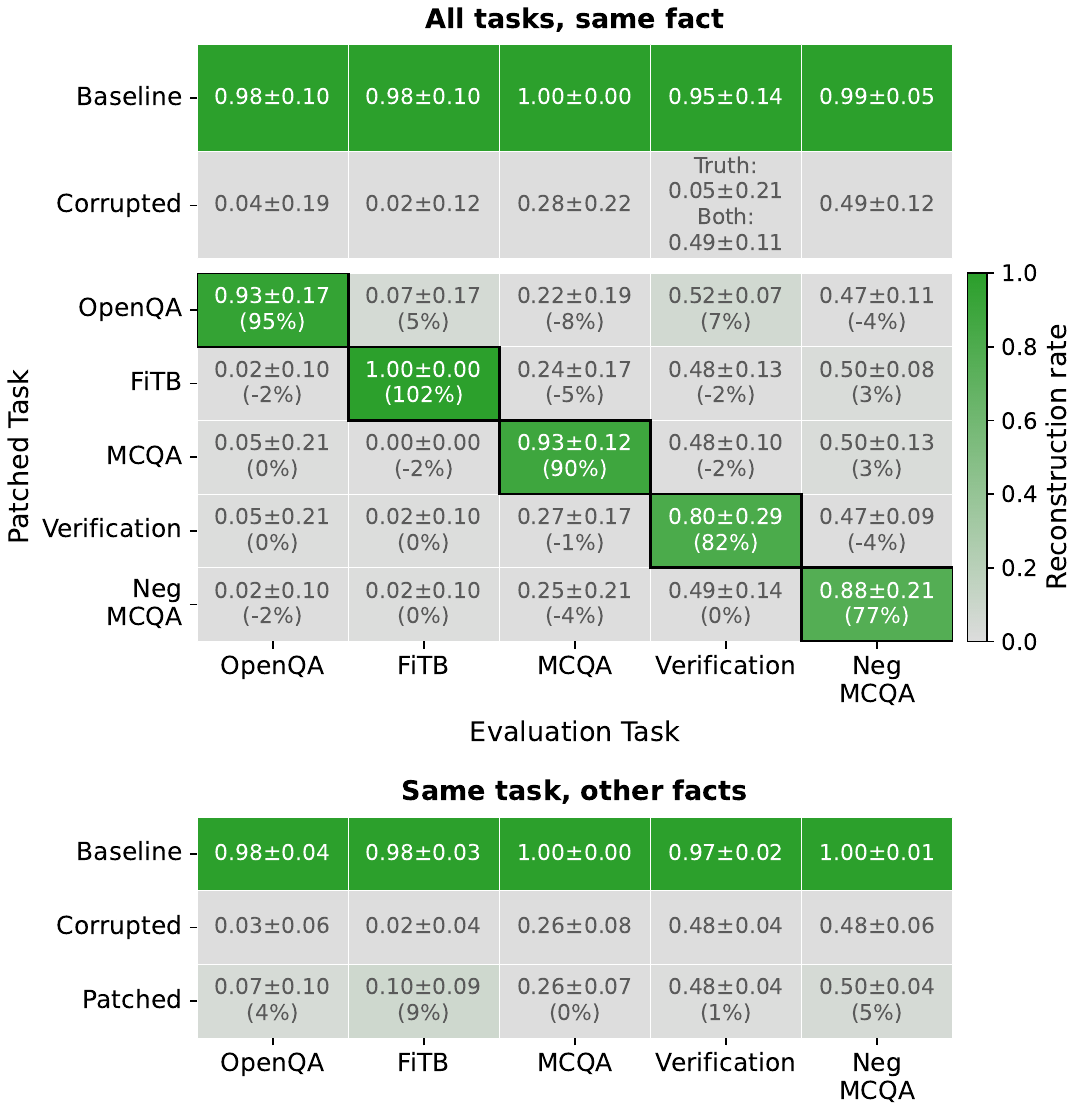}
            \caption{\texttt{(country, official language, language)}}
            \label{fig:sufficiency_nd_language_olmo_7b}
        \end{subfigure}\\
        \begin{subfigure}{\linewidth}
            \centering
            \includegraphics[width=0.9\linewidth]{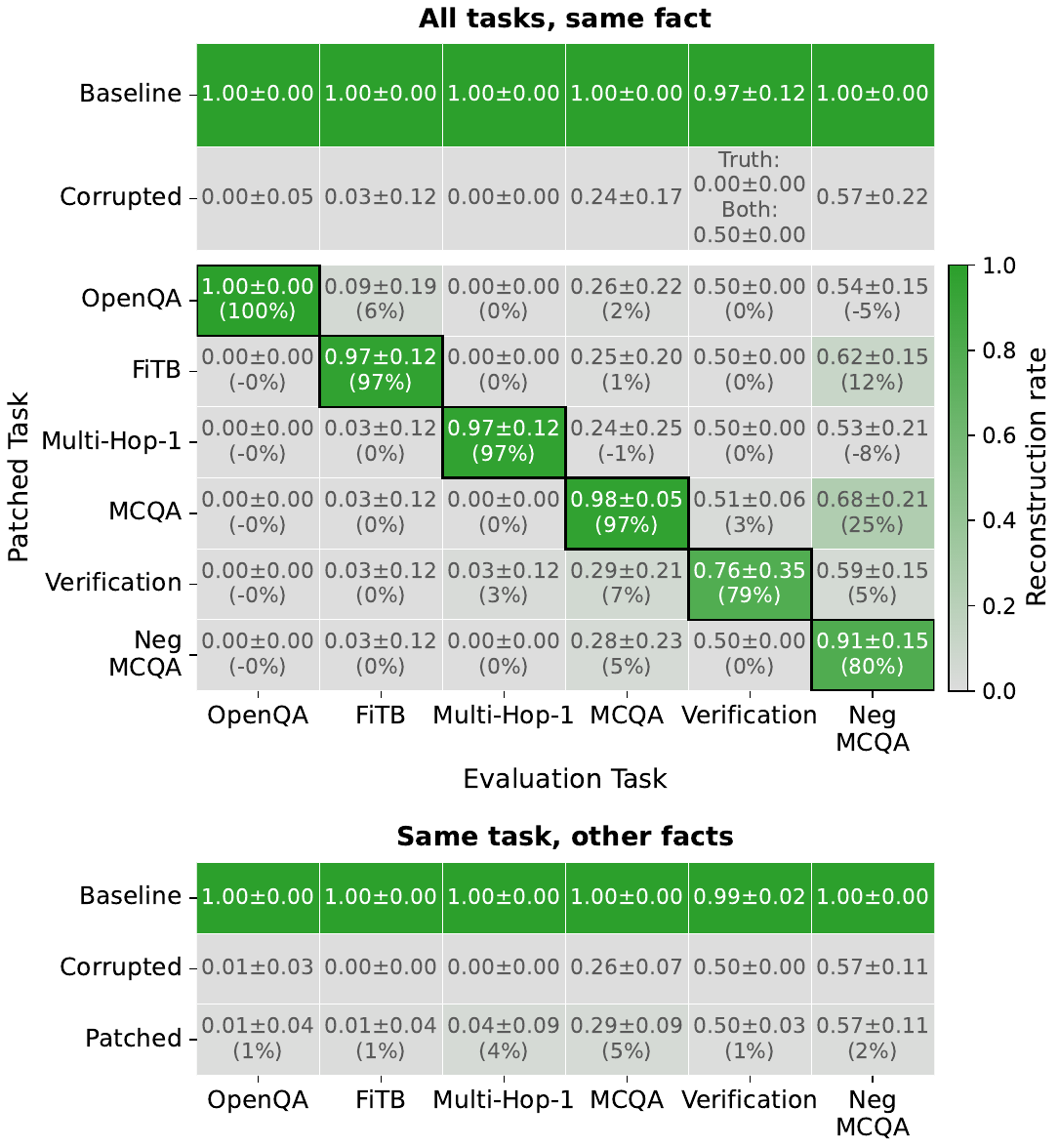}
            \caption{\texttt{(landmark, in-country, country)}}
            \label{fig:sufficiency_nd_landmark_olmo_7b}
        \end{subfigure}\\
        \begin{subfigure}{\linewidth}
            \centering
            \includegraphics[width=0.9\linewidth]{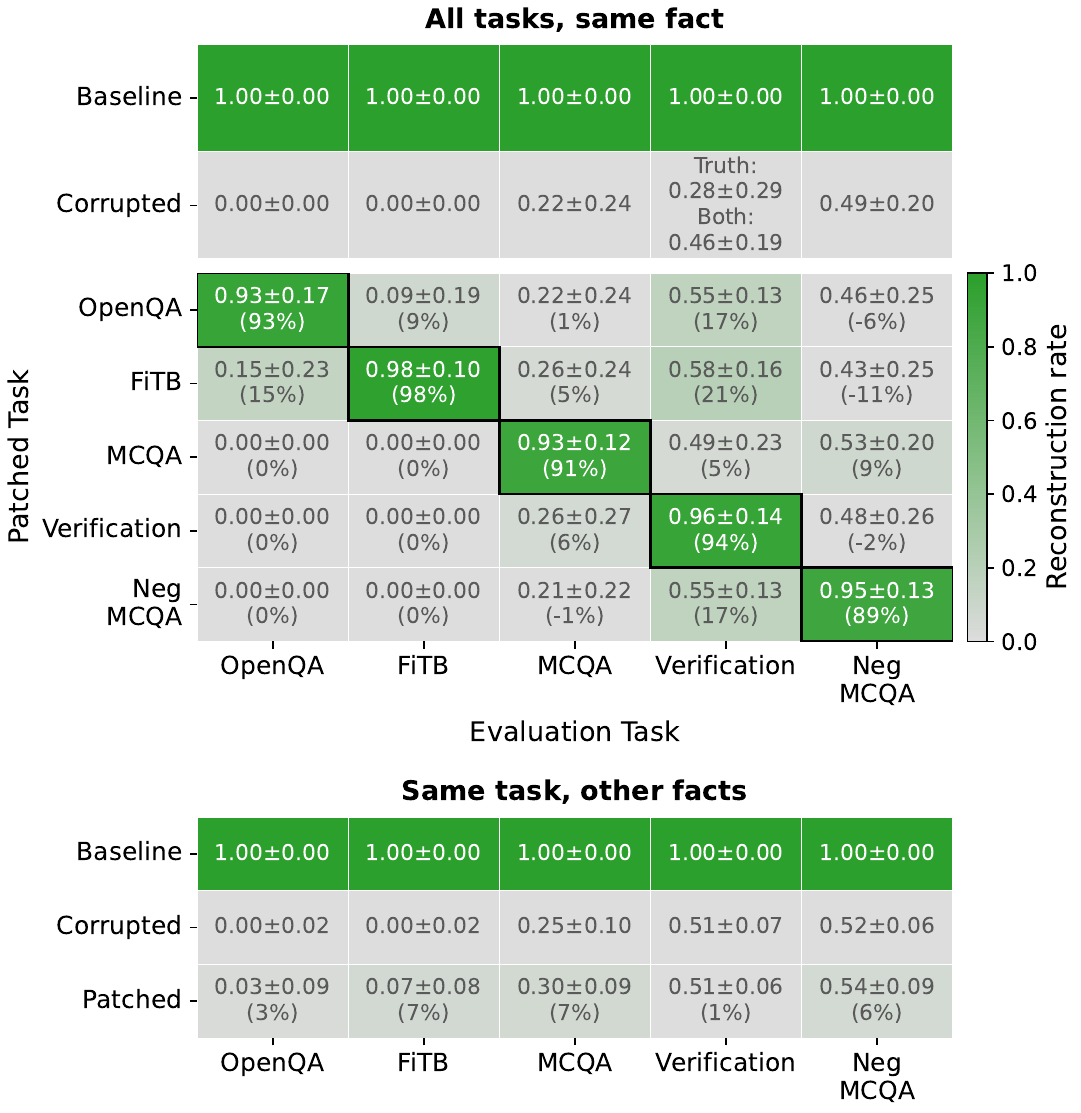}
            \caption{\texttt{(country, capital-of, city)}}
            \label{fig:sufficiency_nd_capital_olmo_7b}
        \end{subfigure}
    \end{minipage}\hfill
    \begin{minipage}[t]{0.49\linewidth}
        \centering
        \begin{subfigure}{\linewidth}
            \centering
            \includegraphics[width=0.9\linewidth]{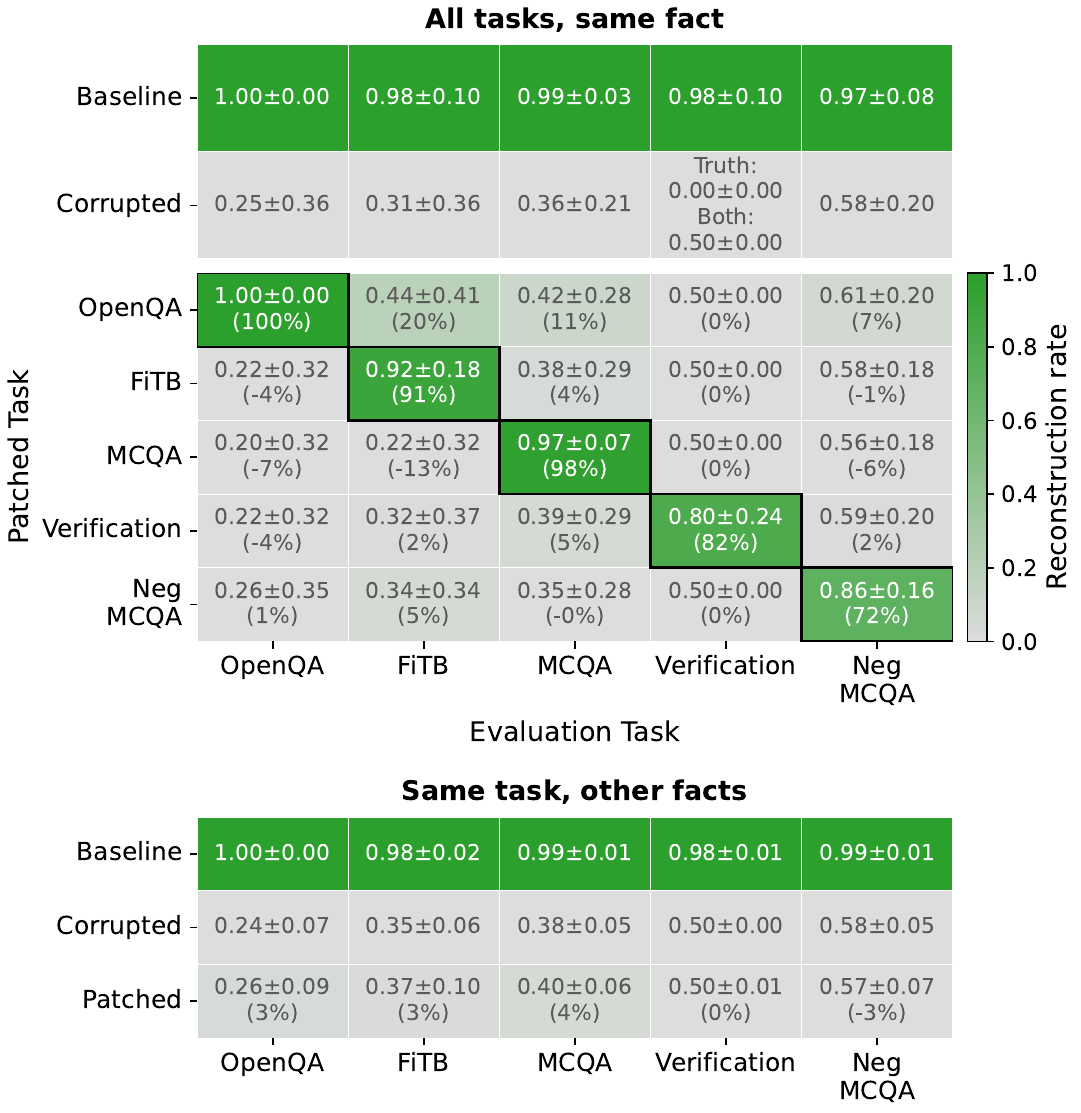}
            \caption{\texttt{(person, plays-instrument, instrument)}}
            \label{fig:sufficiency_nd_instrument_olmo_7b}
        \end{subfigure}\\
        \begin{subfigure}{\linewidth}
            \centering
            \includegraphics[width=0.9\linewidth]{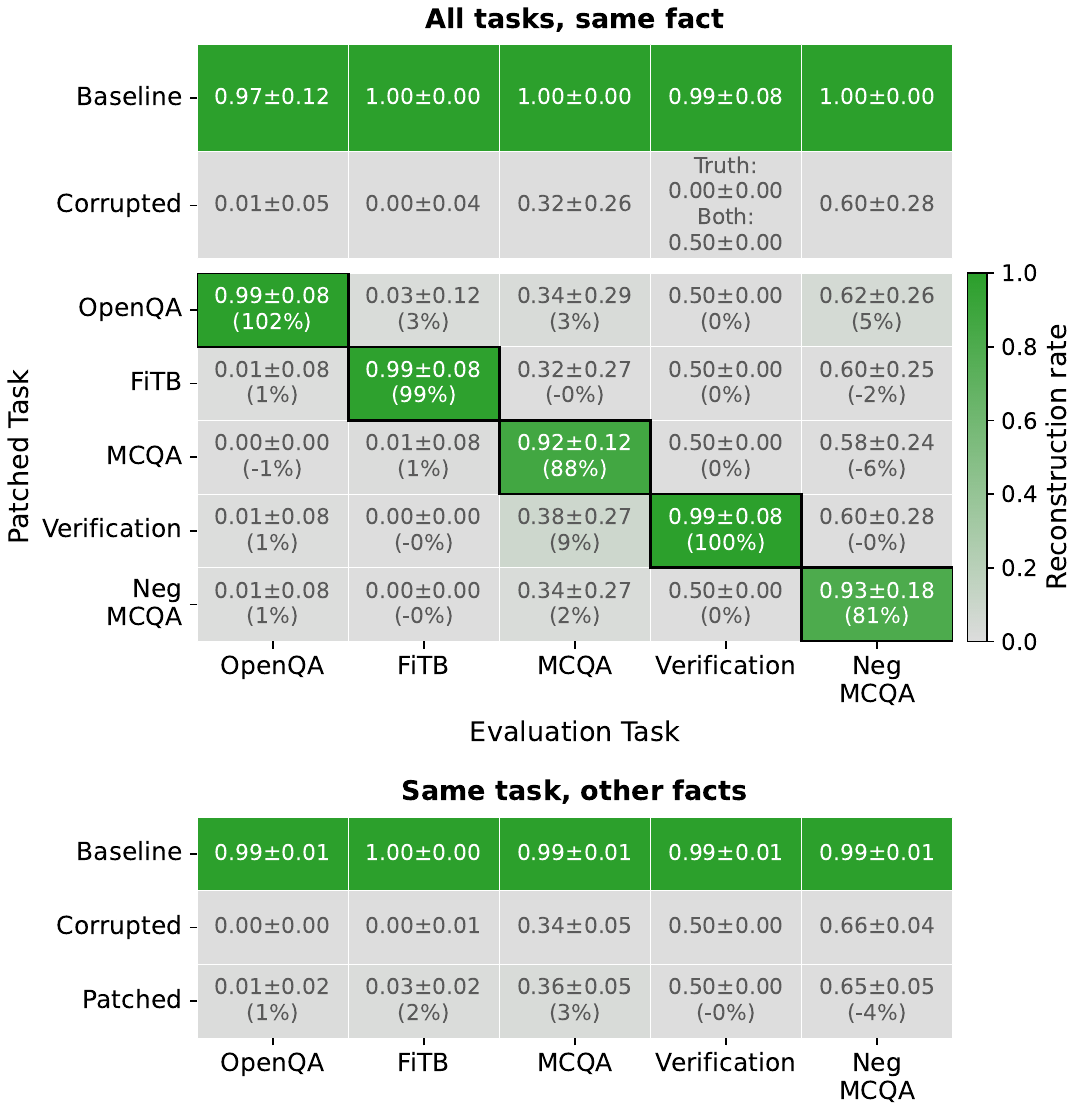}
            \caption{\texttt{(company, HQ-in-city, city)}}
            \label{fig:sufficiency_nd_company_olmo_7b}
        \end{subfigure}
    \end{minipage}
    \caption{Sufficiency results on \OlmoSevenB{} under \emph{non-disjoint} mask training. Same layout as \Cref{fig:sufficiency_gemma}.}
    \label{fig:sufficiency_nd_olmo_7b}
\end{figure*}

\begin{figure*}[t]
    \centering
    \begin{minipage}[t]{0.49\linewidth}
        \centering
        \begin{subfigure}{\linewidth}
            \centering
            \includegraphics[width=0.9\linewidth]{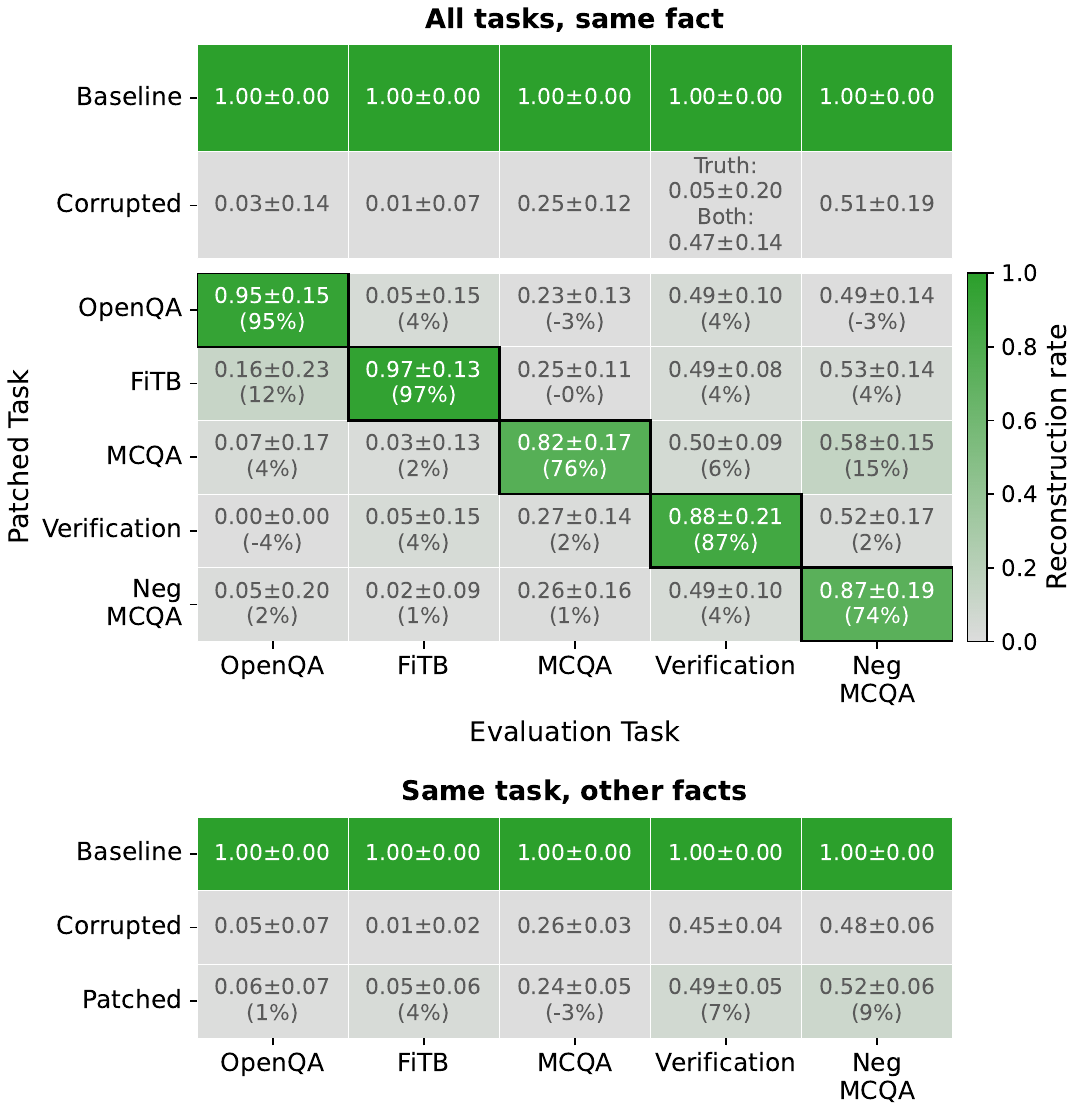}
            \caption{\texttt{(country, official language, language)}}
            \label{fig:sufficiency_nd_language_olmo_13b}
        \end{subfigure}\\
        \begin{subfigure}{\linewidth}
            \centering
            \includegraphics[width=0.9\linewidth]{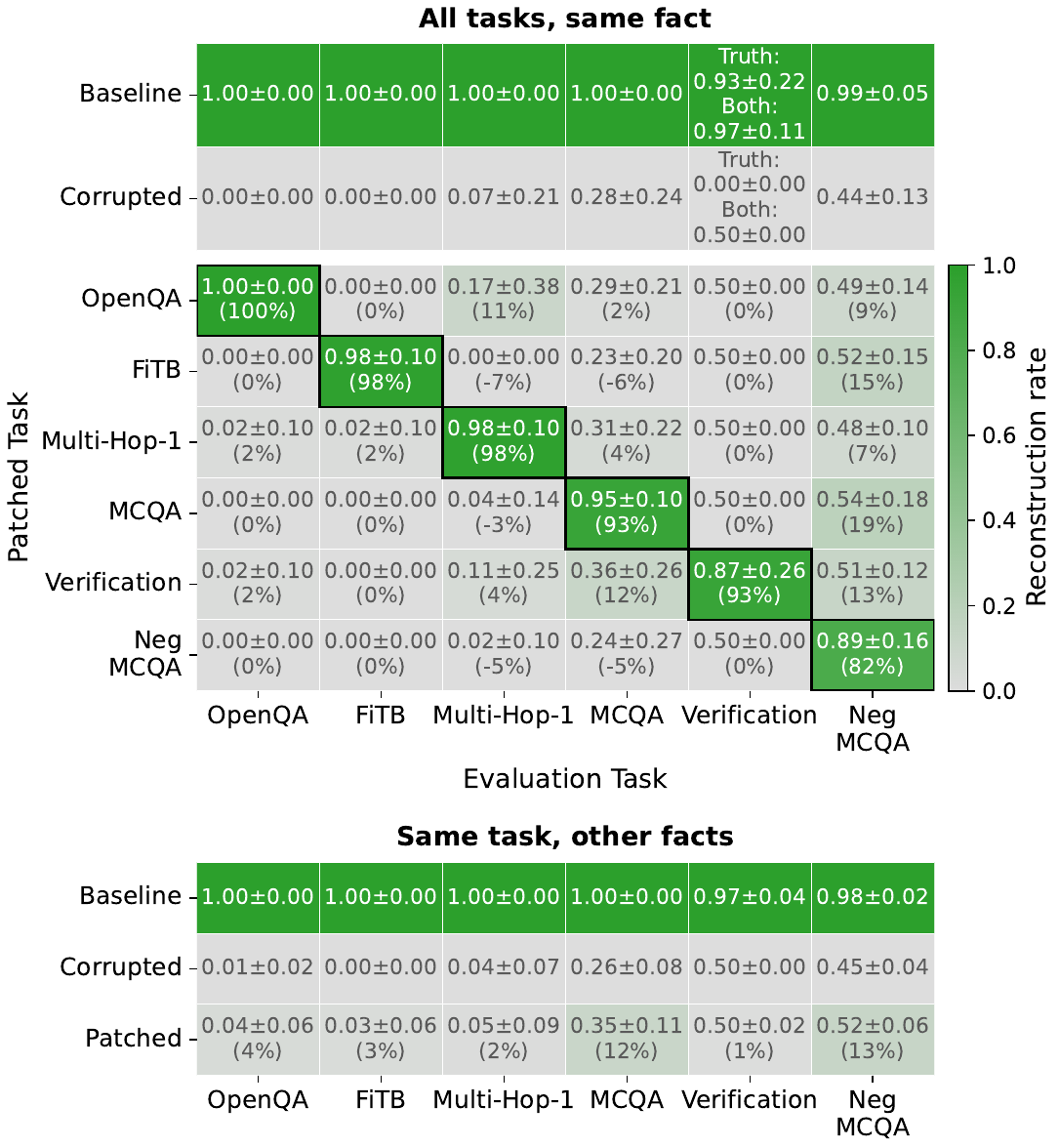}
            \caption{\texttt{(landmark, in-country, country)}}
            \label{fig:sufficiency_nd_landmark_olmo_13b}
        \end{subfigure}\\
        \begin{subfigure}{\linewidth}
            \centering
            \includegraphics[width=0.9\linewidth]{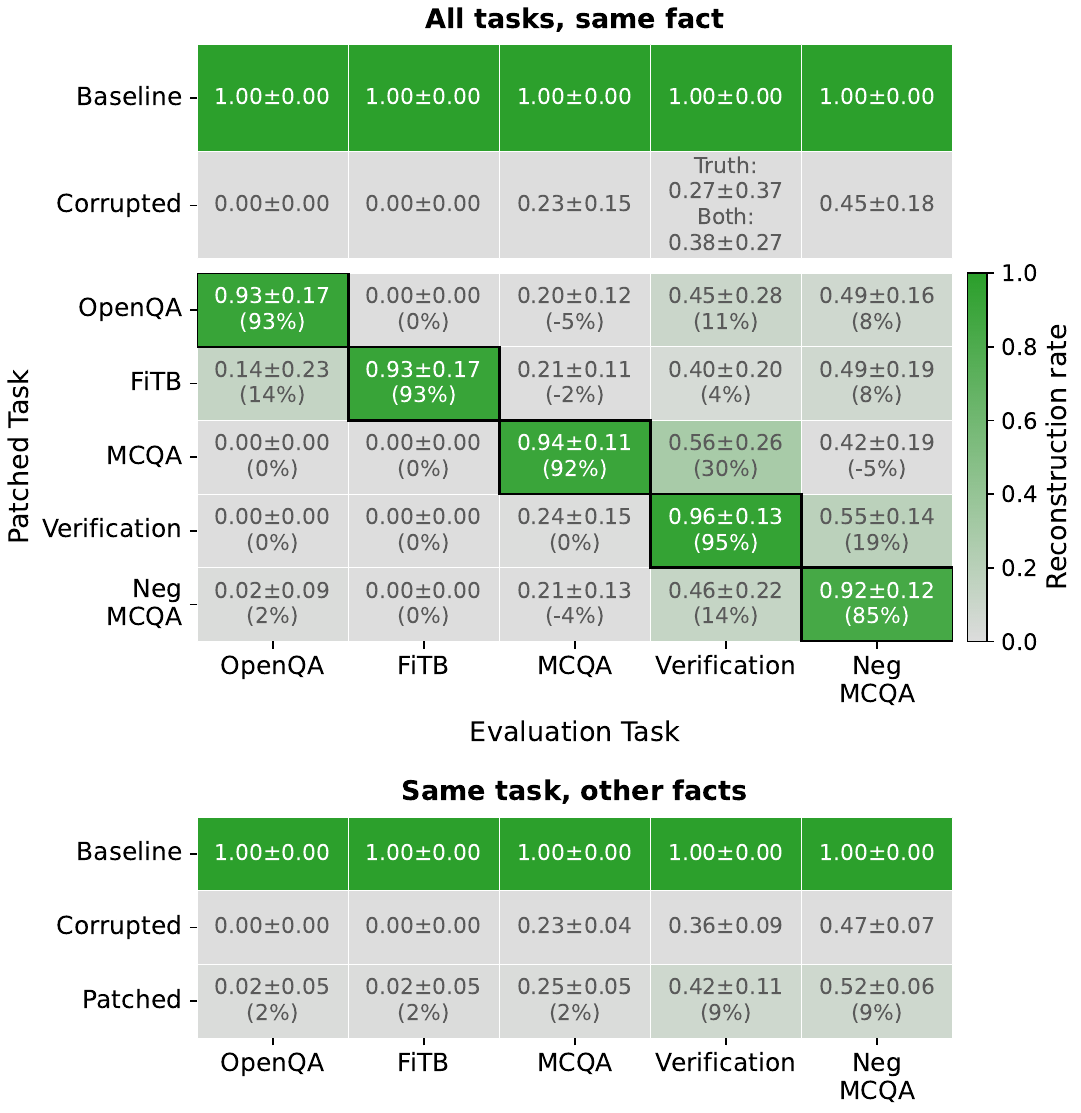}
            \caption{\texttt{(country, capital-of, city)}}
            \label{fig:sufficiency_nd_capital_olmo_13b}
        \end{subfigure}
    \end{minipage}\hfill
    \begin{minipage}[t]{0.49\linewidth}
        \centering
        \begin{subfigure}{\linewidth}
            \centering
            \includegraphics[width=0.9\linewidth]{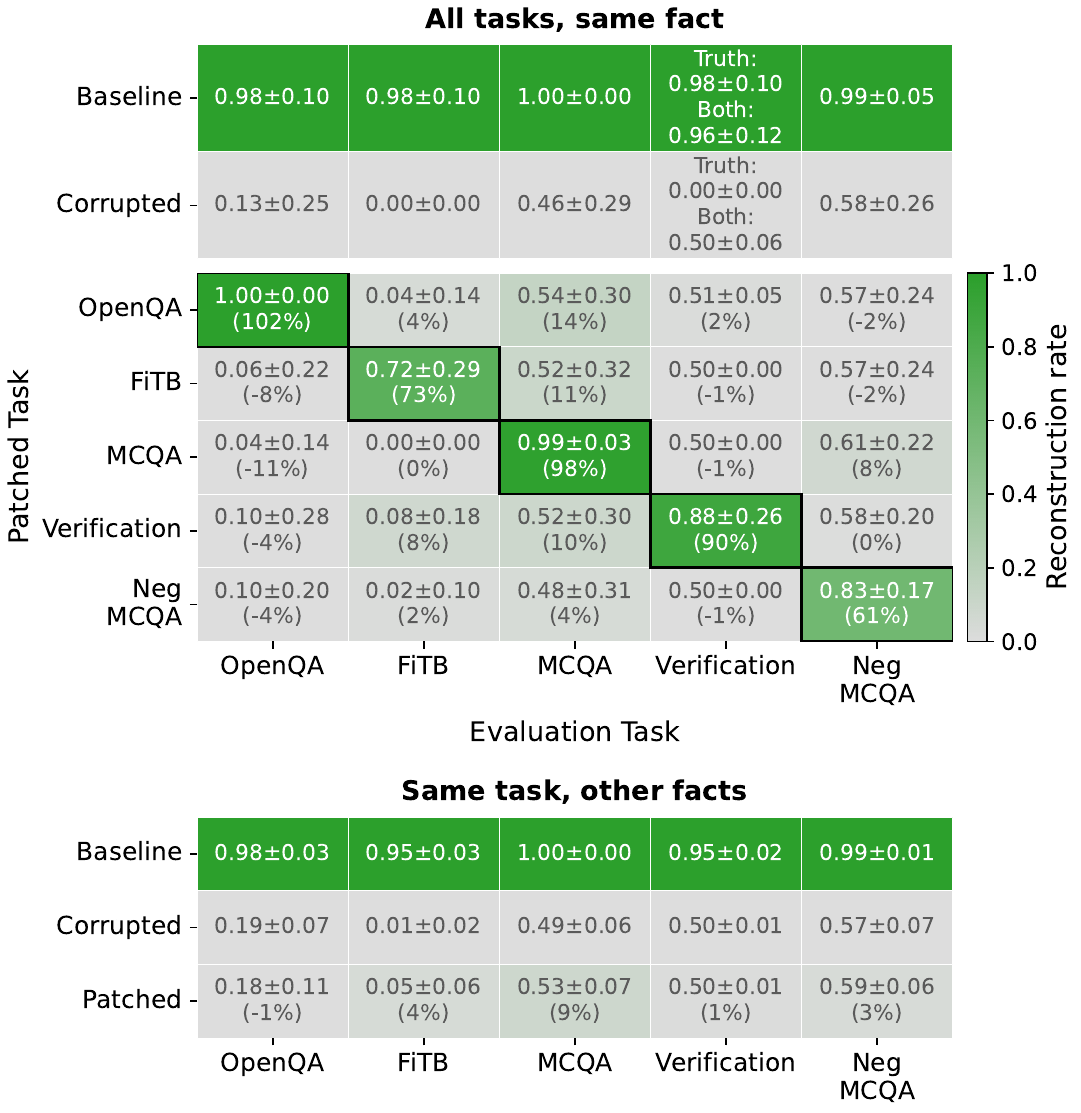}
            \caption{\texttt{(person, plays-instrument, instrument)}}
            \label{fig:sufficiency_nd_instrument_olmo_13b}
        \end{subfigure}\\
        \begin{subfigure}{\linewidth}
            \centering
            \includegraphics[width=0.9\linewidth]{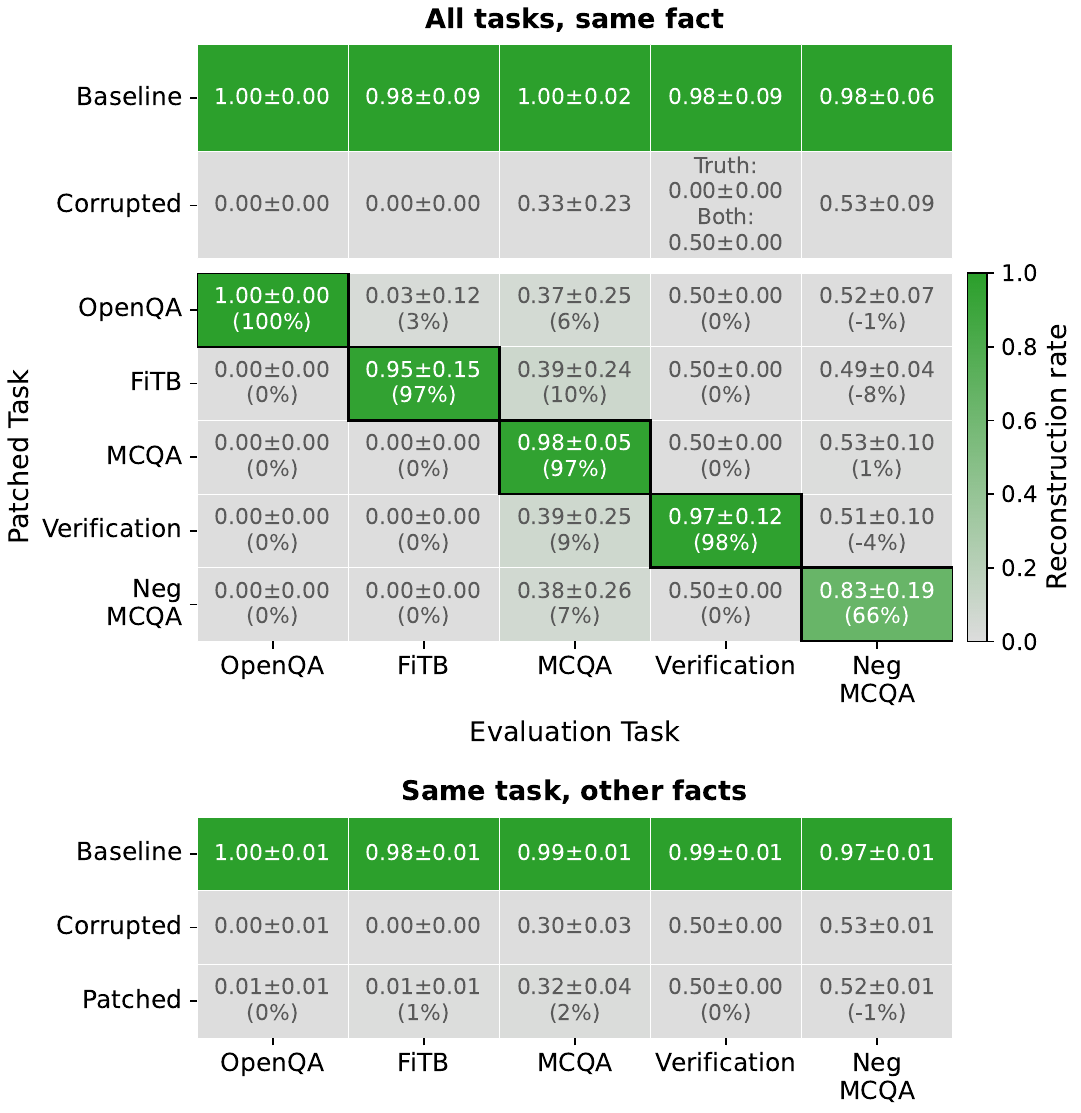}
            \caption{\texttt{(company, HQ-in-city, city)}}
            \label{fig:sufficiency_nd_company_olmo_13b}
        \end{subfigure}
    \end{minipage}
    \caption{Sufficiency results on \OlmoThirteenB{} under \emph{non-disjoint} mask training. Same layout as \Cref{fig:sufficiency_gemma}.}
    \label{fig:sufficiency_nd_olmo_13b}
\end{figure*}

\paragraph{Non-disjoint results}
Recall that for a given fact, masks are trained sequentially, with components selected by earlier masks excluded from subsequent ones. 
We imposed this disjointness constraint to test our hypothesis in its strongest form. 
We asked whether fully disjoint parameter subsets exist that are necessary, sufficient, and specific to each (fact, task) pair, and found that they do.
Nevertheless, we tested whether this constraint is what drives the disjointness by retraining the masks without it, across all models and datasets. The results are similar, with the subsets remaining necessary, sufficient, and specific to similar degrees. Ablating a (fact, task) encoding disrupts the model's performance on the targeted pair, while the same fact on other tasks and other facts on the same task stay near the baseline (\Cref{fig:necessity_nd_gemma_9b,fig:necessity_nd_olmo_7b,fig:necessity_nd_olmo_13b}). Similarly, patching the localized components' activations into a corrupted run recovers performance primarily on the targeted pair (\Cref{fig:sufficiency_nd_gemma_9b,fig:sufficiency_nd_olmo_7b,fig:sufficiency_nd_olmo_13b}). 
Importantly, the masks still end up almost disjoint. The mean Intersection-over-Union (IoU) between components localized for the same fact across task pairs is 0.05 for neurons and 0.006 for attention heads (\Cref{tab:nondisjoint_iou}). 
The IoU computed on random subsets of components sampled from the same layers is 0.004 for neurons and near zero for heads, suggesting a very small shared core but largely disjoint encodings.

\begin{table}[t]
\caption{IoU between the components localized for the same fact under different tasks, for the \emph{non-disjoint} masks. Each entry is the mean Jaccard index $|\mathcal{N}_{f,t}\cap\mathcal{N}_{f,t'}|/|\mathcal{N}_{f,t}\cup\mathcal{N}_{f,t'}|$ over facts $f$ and task pairs $t\neq t'$, where $\mathcal{N}_{f,t}$ is the neuron set localized for fact $f$ under task $t$ (attention heads: $\mathcal{H}_{f,t}$). Dashes mark combinations without localized heads.}
\label{tab:nondisjoint_iou}
\centering
\small
\begin{tabular}{lcc}
\toprule
\textbf{Dataset} & \textbf{Neurons} & \textbf{Heads} \\
\midrule
\multicolumn{3}{l}{\textbf{\Gemma}} \\
Landmark in Country & 0.07 & --- \\
Country to Capital & 0.06 & --- \\
Company to HQ & 0.05 & --- \\
Country to Off.\ Language & 0.06 & --- \\
Person Plays Instrument & 0.06 & --- \\
\midrule
\multicolumn{3}{l}{\textbf{\OlmoSevenB}} \\
Landmark in Country & 0.05 & 0.00 \\
Country to Capital & 0.05 & 0.00 \\
Company to HQ & 0.05 & 0.00 \\
Country to Off.\ Language & 0.05 & 0.00 \\
Person Plays Instrument & 0.06 & 0.00 \\
\midrule
\multicolumn{3}{l}{\textbf{\OlmoThirteenB}} \\
Landmark in Country & 0.05 & 0.01 \\
Country to Capital & 0.04 & 0.01 \\
Company to HQ & 0.04 & 0.01 \\
Country to Off.\ Language & 0.04 & 0.02 \\
Person Plays Instrument & 0.06 & 0.00 \\
\bottomrule
\end{tabular}
\end{table}

\begin{figure*}[t]
    \centering
    \begin{subfigure}{0.49\linewidth}
        \centering
        \includegraphics[width=\linewidth]{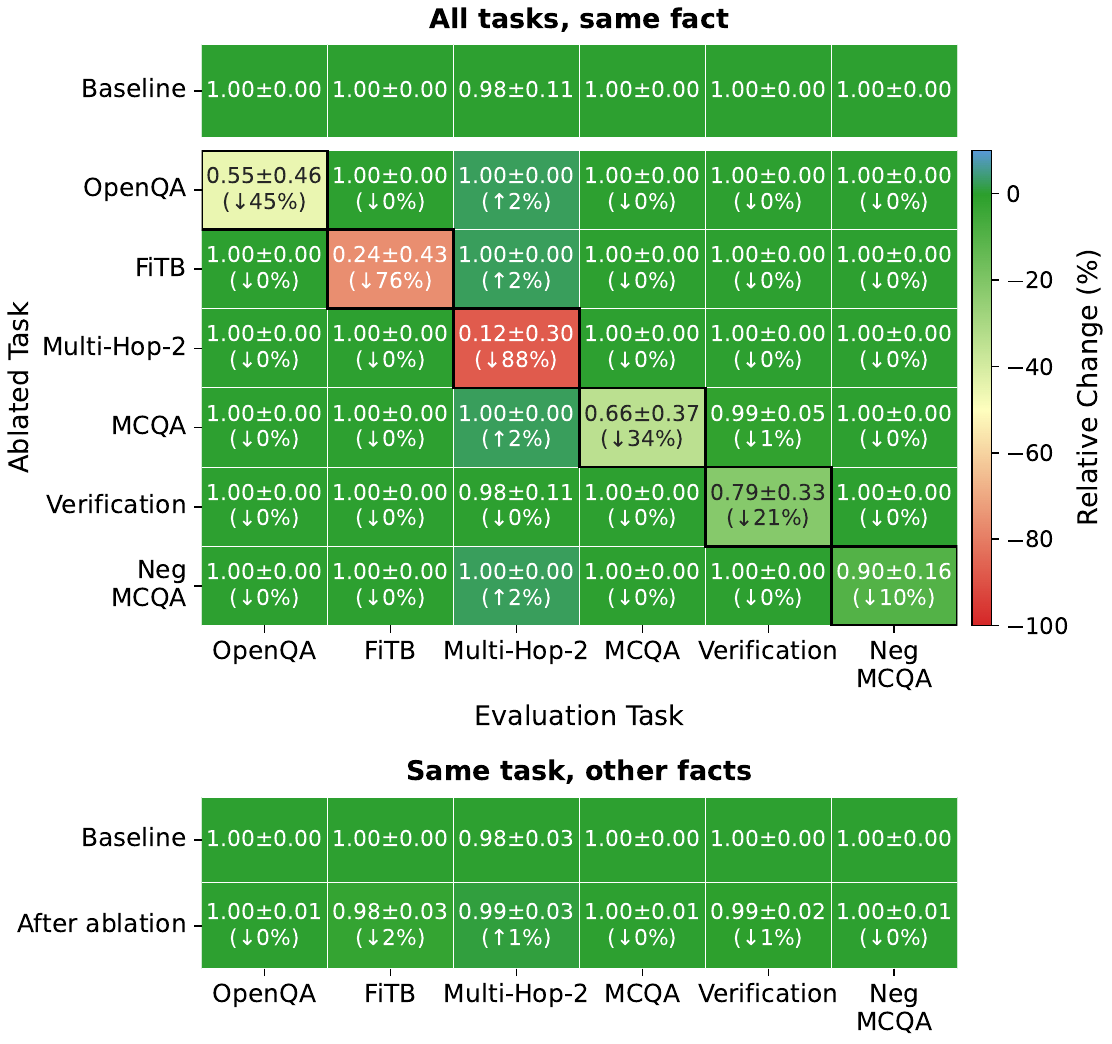}
        \caption{Necessity, \texttt{(country, capital-of, city)}}
        \label{fig:necessity_capital_gemma_27b}
    \end{subfigure}\hfill
    \begin{subfigure}{0.49\linewidth}
        \centering
        \includegraphics[width=\linewidth]{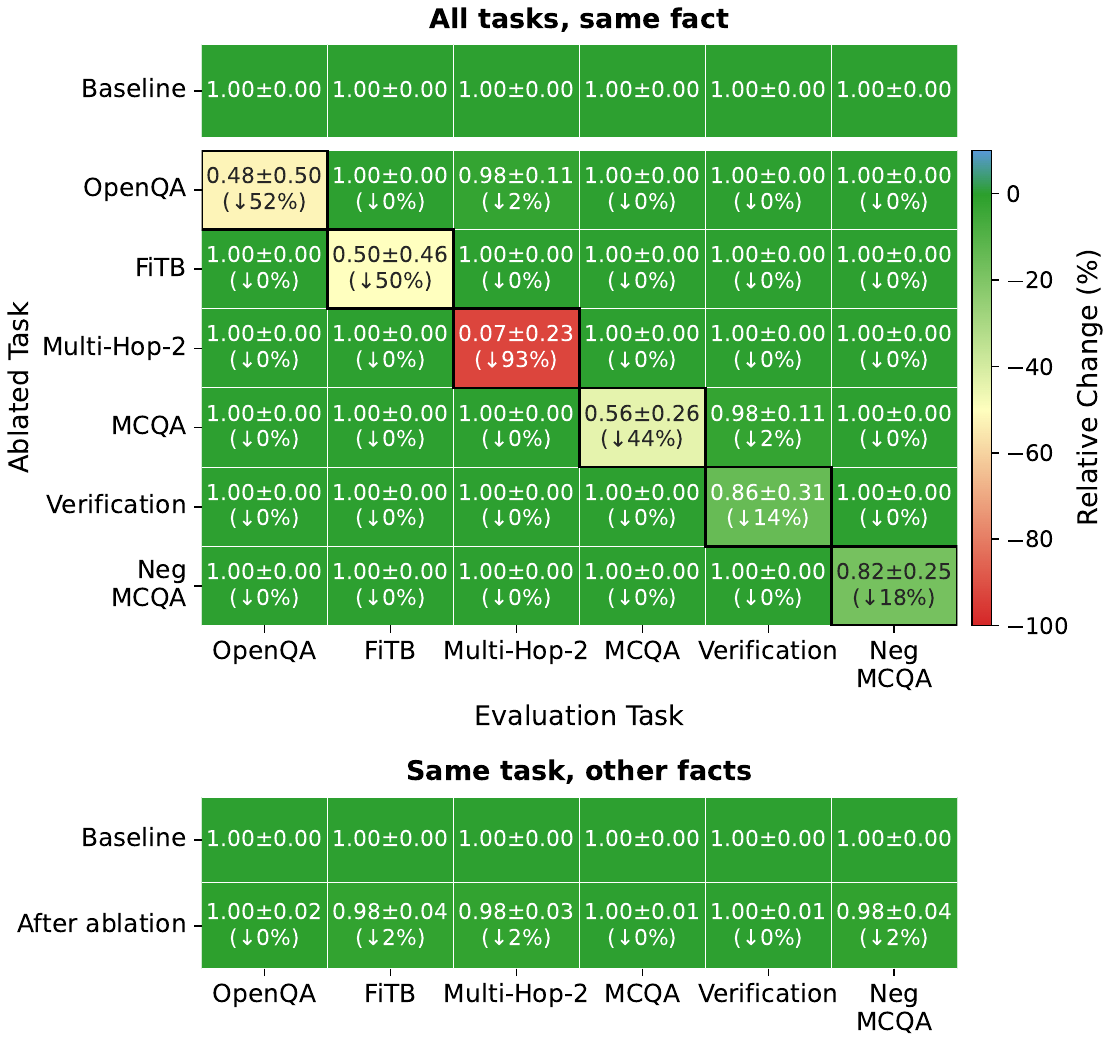}
        \caption{Necessity, \texttt{(country, official language, language)}}
        \label{fig:necessity_language_gemma_27b}
    \end{subfigure}
    \\[6pt]
    \begin{subfigure}{0.49\linewidth}
        \centering
        \includegraphics[width=\linewidth]{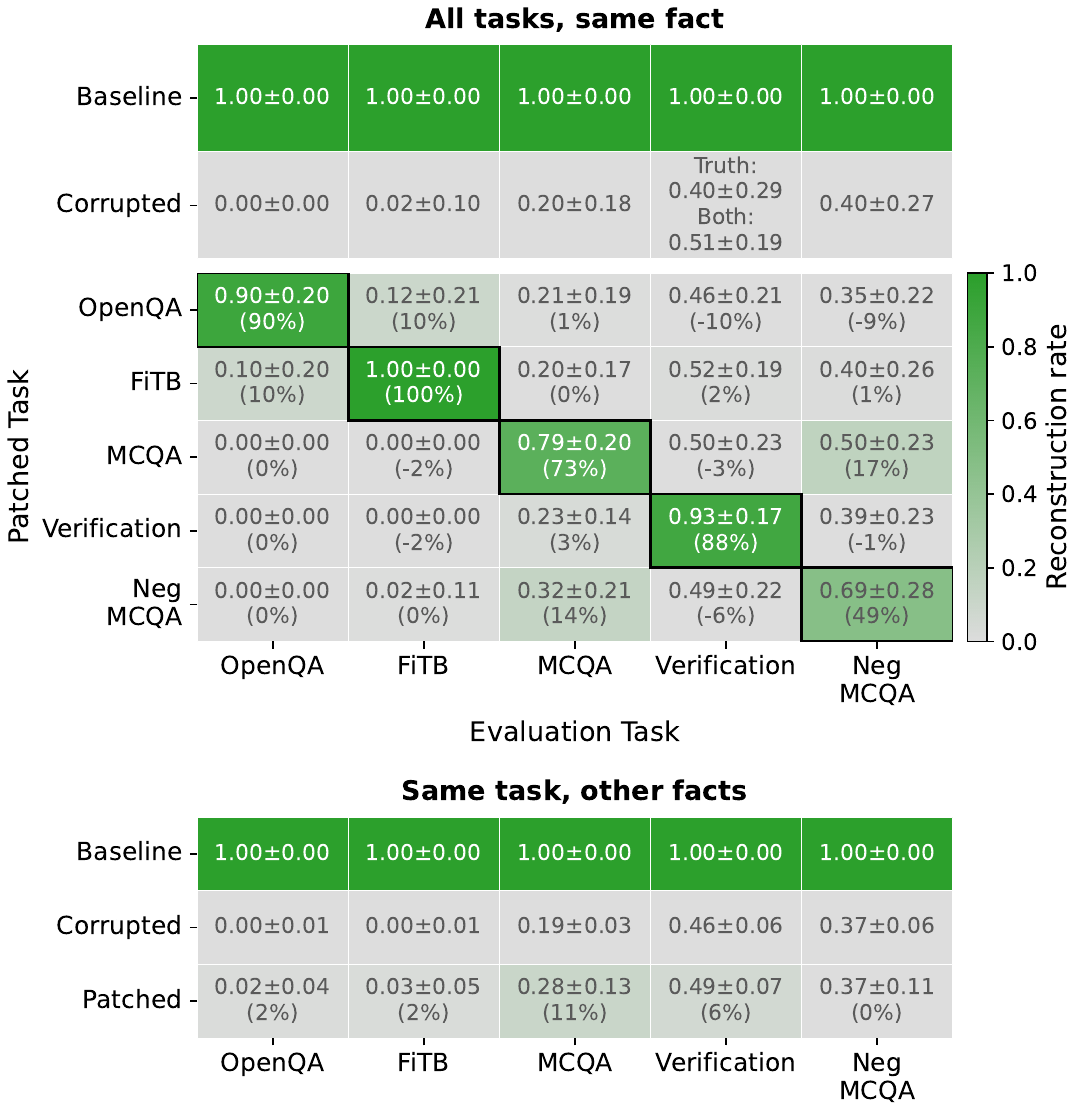}
        \caption{Sufficiency, \texttt{(country, capital-of, city)}}
        \label{fig:sufficiency_capital_gemma_27b}
    \end{subfigure}\hfill
    \begin{subfigure}{0.49\linewidth}
        \centering
        \includegraphics[width=\linewidth]{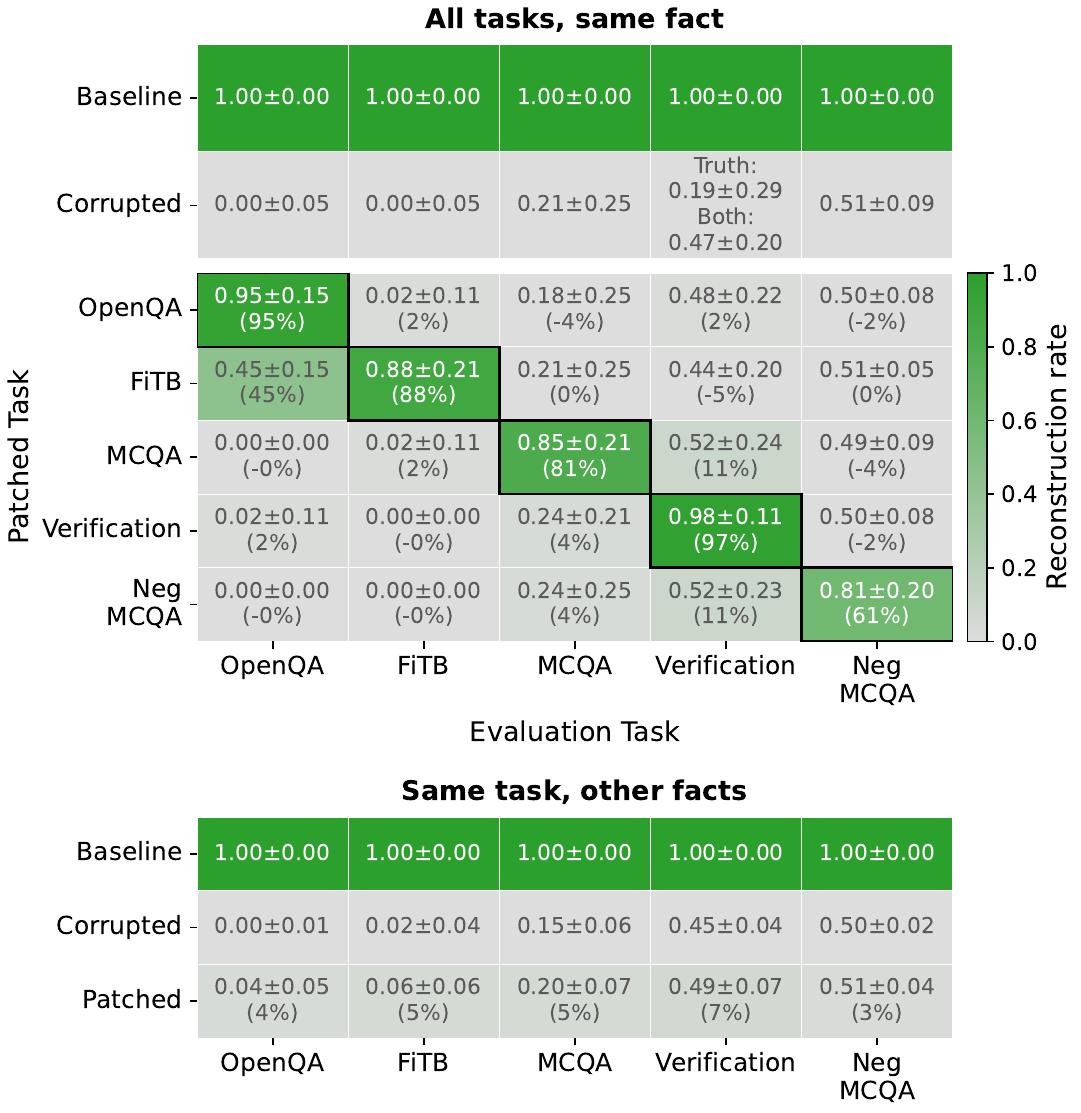}
        \caption{Sufficiency, \texttt{(country, official language, language)}}
        \label{fig:sufficiency_language_gemma_27b}
    \end{subfigure}
    \caption{Necessity (top row) and sufficiency (bottom row) results for \GemmaTwentySeven{} on the \texttt{(country, capital-of, city)} and \texttt{(country, official language, language)} datasets. Same layout as \Cref{fig:necessity} (necessity) and \Cref{fig:sufficiency_gemma} (sufficiency).}
    \label{fig:gemma27b_generalization}
\end{figure*}

\paragraph{Generalization to a larger model}
We train masks for \GemmaTwentySeven{} on the \texttt{(country, capital-of, city)} and \texttt{(country, official language, language)} datasets, 21 facts each. We use a larger sparsity weight ($\beta = 100$) than the default ($\beta = 10$), keeping all other hyperparameters unchanged (\S\ref{appendix:mask_optimization}). We observe the same task-specific pattern under both necessity and sufficiency (\Cref{fig:gemma27b_generalization}). Ablating a (fact, task) encoding drops performance on the targeted pair (diagonal drops of 10\%--93\%), while the same fact on other tasks and other facts on the same task change by at most 2\%. 
Patching the localized components' activations into a corrupted prompt recovers performance primarily on the targeted pair (49\%--100\% reconstruction rate). Non-target pairs show little recovery ($-10\%$ to $17\%$), with one exception, where patching the \textsc{FiTB} encoding recovers \textsc{OpenQA} by 45\% on \texttt{(country, official language, language)}. This shows that the necessity, sufficiency, and specificity of (fact, task) encodings generalize to a substantially larger model.

 \section{Entanglement analysis: additional details}
\label{appendix:entanglement}
 
In \S\ref{sec:entanglement} we quantified the degree of cross-task entanglement between parametric encodings of different (fact,task) pairs. Here, we provide the formal metric definitions, complete per-task entanglement scores, and task pairwise entanglement heatmaps.
 
\paragraph{Entanglement metric details}
The model performance on a (fact,task) pair is measured via variant-tolerant accuracy (\S\ref{appendix:variant_tolerance}), chance-normalized as $a' = (a - c)/(1 - c)$ with chance levels $c = 0.25$ for \textsc{MCQA}, $c = 0.5$ for \textsc{Verification} and \textsc{Neg~MCQA}, and $c = 0$ for all other tasks. 
We define two types of relative change. The \emph{target drop} $\Delta_{\text{target}}(f, t)$ measures how much ablation of the (fact,task) parametric encoding degrades the performance on the targeted pair, clamped so that performance increases receive no credit and drops beyond the baseline are capped:
\begin{equation}
\min\!\left(\max\!\left(\frac{|a'_{\text{before}}| - |a'_{\text{after}}|}{|a'_{\text{before}}|},\; 0\right),\; 1\right)
\label{eq:delta_target}
\end{equation}
The \emph{collateral change} $\Delta_{\text{coll}}(f, t)$ captures any perturbation to a non-targeted pair, capped at $1$:
\begin{equation}
\min\!\left(\frac{|a'_{\text{after}} - a'_{\text{before}}|}{|a'_{\text{before}}|},\; 1\right)
\label{eq:delta_coll}
\end{equation}
The sum over other facts on the same task in Eq.~\ref{eq:E_score} ranges over the held-out evaluation split  (see \S\ref{sec:mechanistic_analysis}). The resulting per-task scores $\operatorname{Ent}_{\text{task}}$ for all (model,dataset) pairs are reported in \Cref{tab:ent_task_full}.

\begin{table*}[t]
\caption{Per-task entanglement scores $\operatorname{Ent}_{\text{task}}$. The rightmost column shows the per-dataset mean across tasks; the bottom row shows the per-task mean across datasets. Dashes mark cells where the dataset does not contain the task.}
\label{tab:ent_task_full}
\centering
\small
\setlength{\tabcolsep}{4pt}
\begin{tabular}{llcccccc|c}
\toprule
\textbf{Model} & \textbf{Dataset} & \textsc{OpenQA} & \textsc{FiTB} & \textsc{M-Hop} & \textsc{MCQA} & \textsc{Verif.} & \textsc{Neg MCQA} & \textbf{Mean} \\
\midrule
\multirow{5}{*}{\Gemma} & (landmark, in-country)       & 0.10 & 0.17 & 0.15$^1$ & 0.14 & 0.25 & 0.23 & \textbf{0.17} \\
 & (country, capital-of)        & 0.12 & 0.10 & 0.05$^2$ & 0.15 & 0.24 & 0.27 & \textbf{0.15} \\
 & (company, HQ-in-city)             & 0.08 & 0.13 & --- & 0.15 & 0.23 & 0.27 & \textbf{0.17} \\
 & (country, language-of) & 0.16 & 0.12 & 0.06$^2$ & 0.15 & 0.18 & 0.25 & \textbf{0.15} \\
 & (person, plays-instr.)   & 0.16 & 0.24 & --- & 0.23 & 0.30 & 0.32 & \textbf{0.25} \\
\midrule
\multirow{5}{*}{\OlmoSevenB} & (landmark, in-country)       & 0.20 & 0.10 & 0.06$^1$ & 0.10 & 0.27 & 0.19 & \textbf{0.15} \\
 & (country, capital-of)        & 0.10 & 0.08 & 0.16$^2$ & 0.10 & 0.27 & 0.24 & \textbf{0.16} \\
 & (company, HQ-in-city)             & 0.03 & 0.07 & --- & 0.14 & 0.24 & 0.22 & \textbf{0.14} \\
 & (country, language-of) & 0.15 & 0.15 & 0.09$^2$ & 0.17 & 0.30 & 0.20 & \textbf{0.18} \\
 & (person, plays-instr.)   & 0.10 & 0.12 & --- & 0.16 & 0.21 & 0.23 & \textbf{0.16} \\
\midrule
\multirow{5}{*}{\OlmoThirteenB} & (landmark, in-country)       & 0.07 & 0.09 & 0.13$^1$ & 0.23 & 0.28 & 0.24 & \textbf{0.17} \\
 & (country, capital-of)        & 0.10 & 0.07 & 0.07$^2$ & 0.12 & 0.20 & 0.25 & \textbf{0.13} \\
 & (company, HQ-in-city)             & 0.06 & 0.09 & --- & 0.17 & 0.26 & 0.22 & \textbf{0.16} \\
 & (country, language-of) & 0.12 & 0.12 & 0.08$^2$ & 0.16 & 0.22 & 0.19 & \textbf{0.15} \\
 & (person, plays-instr.)   & 0.23 & 0.24 & --- & 0.15 & 0.30 & 0.22 & \textbf{0.23} \\
\midrule
\multicolumn{2}{l}{\textbf{Overall Mean}} & \textbf{0.12} & \textbf{0.13} & \textbf{0.09} & \textbf{0.15} & \textbf{0.25} & \textbf{0.24} & \textbf{0.17} \\
\bottomrule
\end{tabular}
\vspace{2pt}
{\footnotesize $^1$Multi-Hop-1; $^2$Multi-Hop-2.}
\end{table*}

\begin{figure}[t]
    \centering
    \begin{subfigure}{\linewidth}
        \centering
        \includegraphics[width=\linewidth]{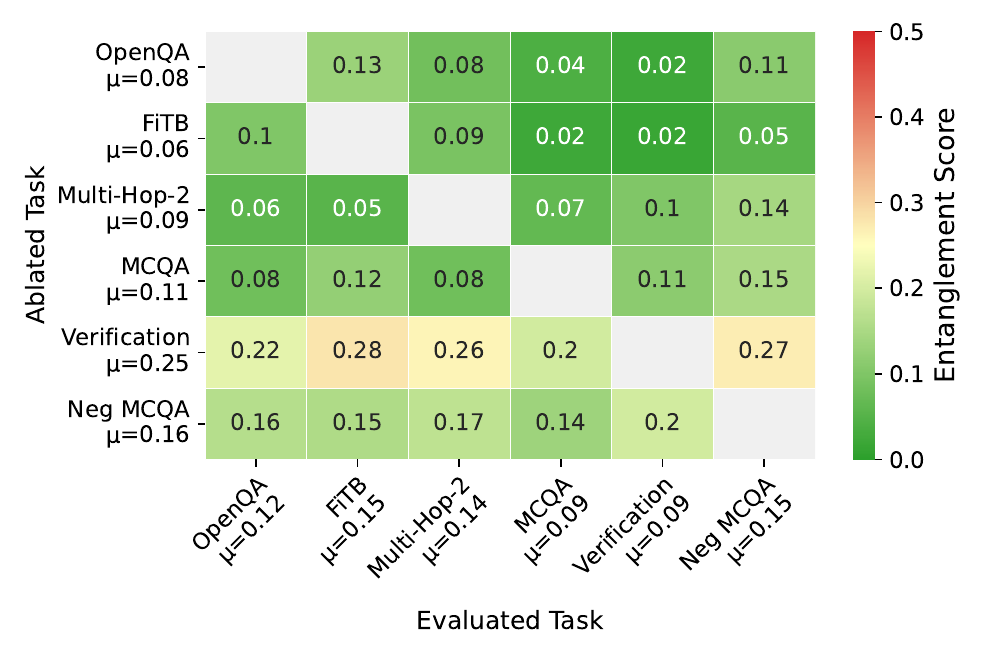}
        \caption{\texttt{(country, capital-of, city)}}
        \label{fig:pairwise_ent_capital}
    \end{subfigure}\\
    \begin{subfigure}{\linewidth}
        \centering
        \includegraphics[width=\linewidth]{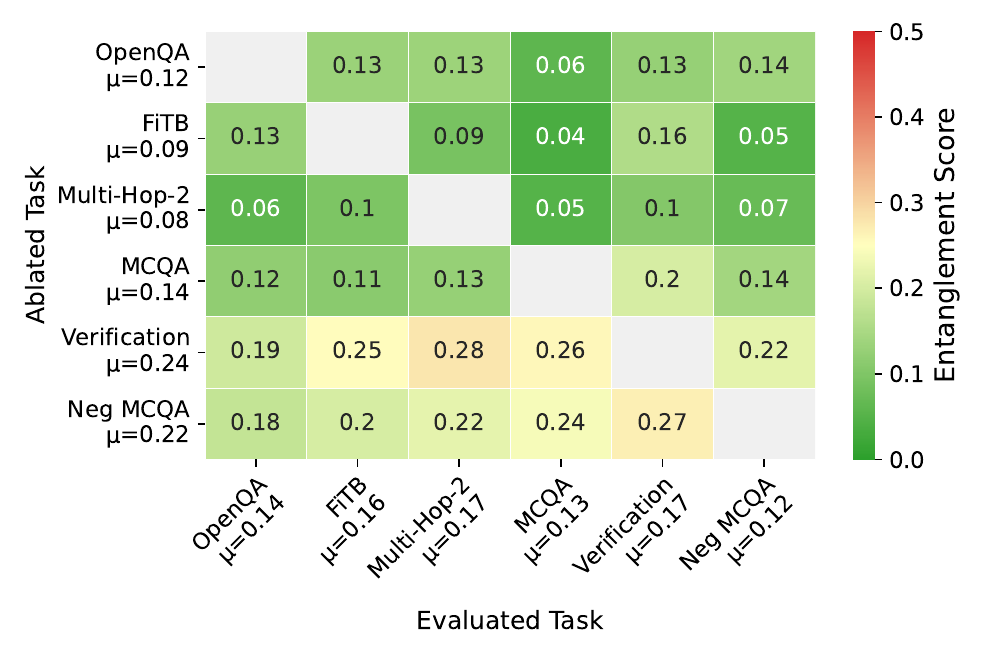}
        \caption{\texttt{(country, official language, language)}}
        \label{fig:pairwise_ent_language}
    \end{subfigure}
    \caption{Pairwise entanglement scores $\operatorname{Ent}(t_A \!\to\! t_B)$ on \OlmoSevenB{}. Rows correspond to the ablated task; columns to the evaluated task. Row and column annotations show the mean score ($\mu$). Discrimination-tasks exhibit higher entanglement with all other tasks than generation-tasks.}
    \label{fig:pairwise_ent}
\end{figure}

\paragraph{Pairwise entanglement}
To examine whether specific task pairs are more entangled than others, we train separate masks for each directed pair of tasks $(t_A, t_B)$ and compute a pairwise entanglement score $\operatorname{Ent}(t_A \!\to\! t_B)$, averaged across facts. 
This is Eq.~\ref{eq:E_score} with the cross-task sum reduced to the single term $t' = t_B$. 
Since this requires training $|\mathcal{T}|^2$ masks per fact, compared to $|\mathcal{T}|$ masks in \S\ref{sec:mechanistic_analysis}, we limit this analysis to \OlmoSevenB{} on the \texttt{(country, official language, language)} and \texttt{(country, capital-of, city)} datasets. 
\Cref{fig:pairwise_ent} presents the resulting heatmaps. Row means demonstrate another instance of the generation-discrimination split: 
ablating generation-task encodings causes modest collateral damage across evaluated tasks ($\mu = 0.06$–$0.12$), while ablating discrimination-task encodings produces broader collateral damage ($\mu = 0.11$–$0.25$).
 
\section{The role of task-specific encodings in CoT reasoning: additional details}
\label{appendix:cot_ablation}

In \S\ref{sec:cot_ablation} we tested whether chain-of-thought (CoT) reasoning engages task-specific encodings beyond those tied to the evaluation task. Here, we provide the prompt construction procedure, filtering criteria, and results for all models and datasets.

\subsection{Additional implementation details}

\paragraph{CoT prompt construction}
For the prompts in the mechanistic analysis (\S\ref{sec:mechanistic_analysis}), we use an instruction that ends with \textit{``Your response should be formatted as: `Answer: \{your answer\}'.''}
For the CoT evaluation, we replace this with \textit{``Before answering, think step by step. Your response should be formatted as: `Reasoning: \{your reasoning\}. Answer: \{your final answer\}'.''} 
After applying the model's chat template to the prompt, we augment the prompt with the string \textit{``Reasoning:''}.

\paragraph{Reasoning generation}
For each prompt we generate a reasoning trace by greedy decoding with at most 200 new tokens, truncated at the first generated \textit{``Answer:''} marker.
We place the resulting trace in the assistant turn and append the answer prefix. We then score the probability of the first token of the target answer at the end of the prefix, as in the direct-answering condition. This also allows evaluating generations that never produce an answer marker.

\paragraph{Ablations}
For each (fact, task) pair, we reuse its localized mask and zero-ablate the identified components, as described in \S\ref{sec:mechanistic_analysis}.

\paragraph{Fact filtering}
We use facts whose post-CoT accuracy meets or exceeds the per-task threshold from \S\ref{appendix:dataset_construction} on every task, namely 0.85 for \textsc{MCQA}, \textsc{Neg~MCQA}, and \textsc{Verification}, and 0.75 for \textsc{OpenQA} and \textsc{FiTB}. 
\Cref{tab:cot_facts} reports the counts.

\paragraph{Evaluation}
From each ablation we read two quantities, both reported as mean  accuracy across facts (using the formatting tolerance from \S\ref{appendix:variant_tolerance}), under direct answering and CoT: (i)~the accuracy drop on the ablated task itself (same-task effect), which is meant to test whether CoT recovers what direct answering loses; and (ii)~for each fact, the accuracy drop caused by the other task's encoding that most damages each condition (measured via accuracy drop; cross-task effect), testing whether CoT suffers more collateral damage than direct answering.
In the cross-task panels, since CoT accuracy without ablation exceeds $0.99$ for every model and dataset, a separate CoT no-ablation bar is not shown.

\begin{table*}[t]
\caption{Facts retained by the CoT filter, as $k/n$ where $n$ is the number of facts used in the mechanistic analysis (\Cref{tab:tasks_per_dataset}) and $k$ is how many facts also clear the per-task CoT threshold on every task.}
\label{tab:cot_facts}
\centering
\small
\resizebox{\textwidth}{!}{%
\begin{tabular}{lccc}
\toprule
\textbf{Dataset} & \OlmoSevenB & \OlmoThirteenB & \Gemma \\
\midrule
(country, official language, language) & 18/22 & 20/29 & 32/33 \\
(landmark, in-country, country)          & 17/17 & 19/23 & 37/38 \\
(country, capital-of, city)           & 22/23 & 23/28 & 31/32 \\
(company, HQ-in-city, city)                & 26/34 & 15/32 & 41/51 \\
(person, plays-instrument, instrument)      & 15/25 & 14/25 & 16/25 \\
\bottomrule
\end{tabular}%
}
\end{table*}

\subsection{Additional results}
\begin{figure*}[t]\centering
\begin{subfigure}{0.49\linewidth}\includegraphics[width=\linewidth]{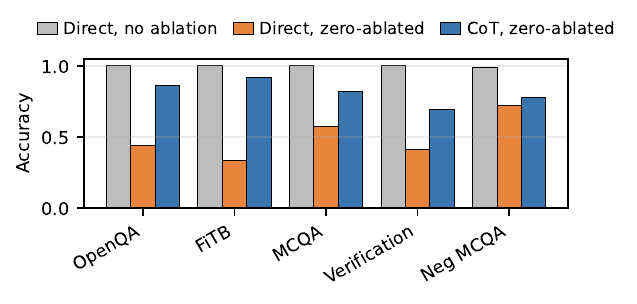}\caption{(country, official language, language), own-encoding}\end{subfigure}\hfill
\begin{subfigure}{0.49\linewidth}\includegraphics[width=\linewidth]{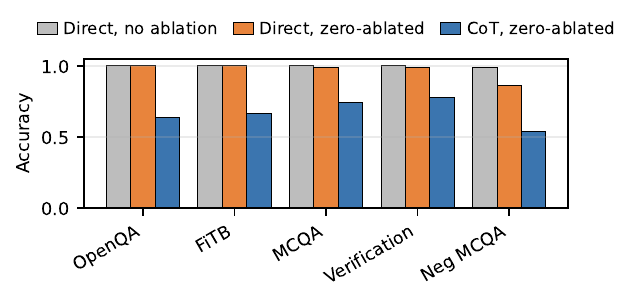}\caption{(country, official language, language), cross-task}\end{subfigure}

\begin{subfigure}{0.49\linewidth}\includegraphics[width=\linewidth]{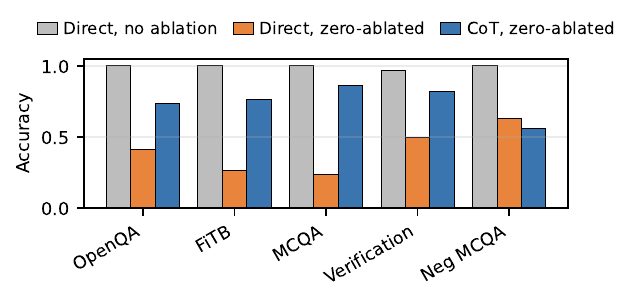}\caption{(landmark, in-country, country), own-encoding}\end{subfigure}\hfill
\begin{subfigure}{0.49\linewidth}\includegraphics[width=\linewidth]{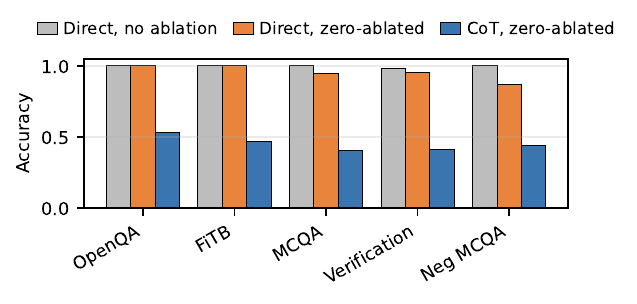}\caption{(landmark, in-country, country), cross-task}\end{subfigure}

\begin{subfigure}{0.49\linewidth}\includegraphics[width=\linewidth]{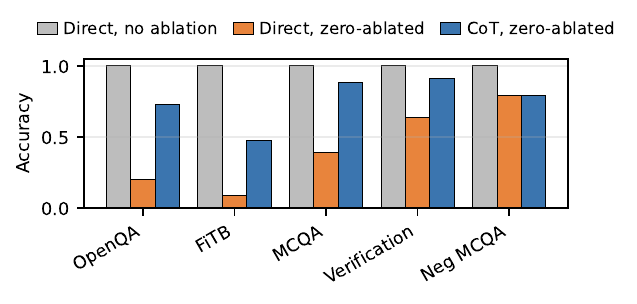}\caption{(country, capital-of, city), own-encoding}\end{subfigure}\hfill
\begin{subfigure}{0.49\linewidth}\includegraphics[width=\linewidth]{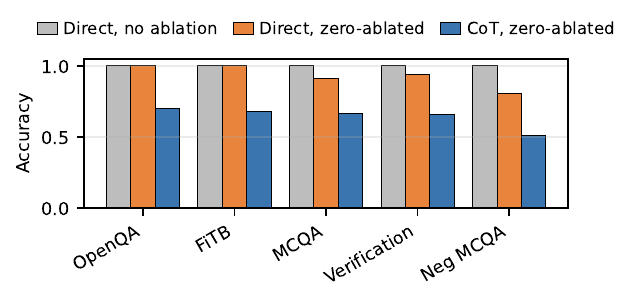}\caption{(country, capital-of, city), cross-task}\end{subfigure}

\begin{subfigure}{0.49\linewidth}\includegraphics[width=\linewidth]{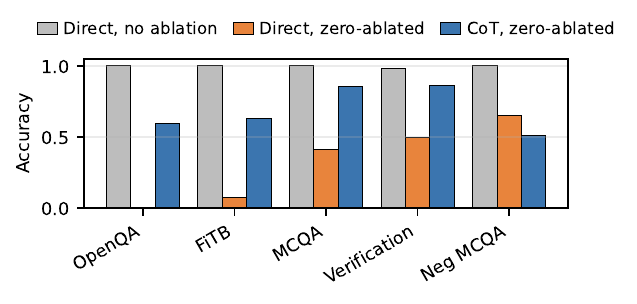}\caption{(company, HQ-in-city, city), own-encoding}\end{subfigure}\hfill
\begin{subfigure}{0.49\linewidth}\includegraphics[width=\linewidth]{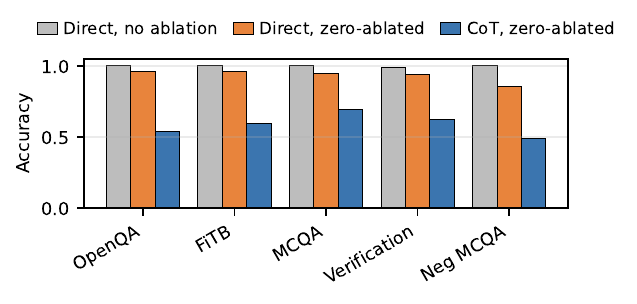}\caption{(company, HQ-in-city, city), cross-task}\end{subfigure}

\begin{subfigure}{0.49\linewidth}\includegraphics[width=\linewidth]{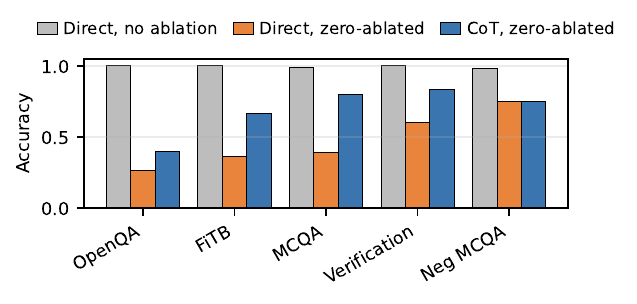}\caption{(person, plays-instrument, instrument), own-encoding}\end{subfigure}\hfill
\begin{subfigure}{0.49\linewidth}\includegraphics[width=\linewidth]{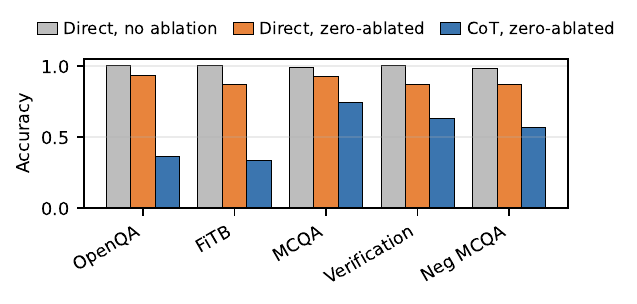}\caption{(person, plays-instrument, instrument), cross-task}\end{subfigure}
\caption{CoT vs.\ direct answering under zero-ablation, \OlmoSevenB{}.}
\label{fig:cot_recovery_olmo_7b}
\end{figure*}

\begin{figure*}[t]\centering
\begin{subfigure}{0.49\linewidth}\includegraphics[width=\linewidth]{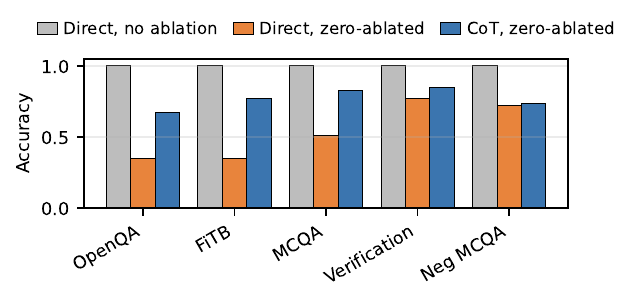}\caption{(country, official language, language), own-encoding}\end{subfigure}\hfill
\begin{subfigure}{0.49\linewidth}\includegraphics[width=\linewidth]{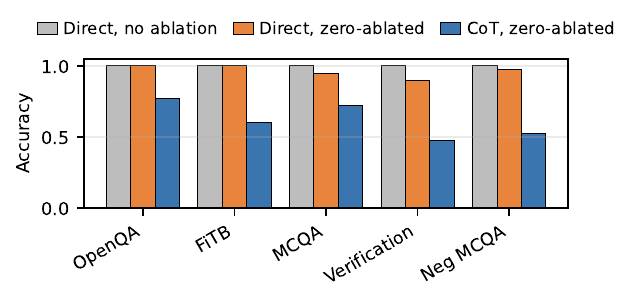}\caption{(country, official language, language), cross-task}\end{subfigure}

\begin{subfigure}{0.49\linewidth}\includegraphics[width=\linewidth]{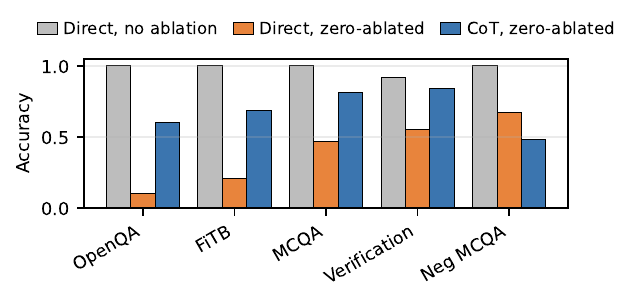}\caption{(landmark, in-country, country), own-encoding}\end{subfigure}\hfill
\begin{subfigure}{0.49\linewidth}\includegraphics[width=\linewidth]{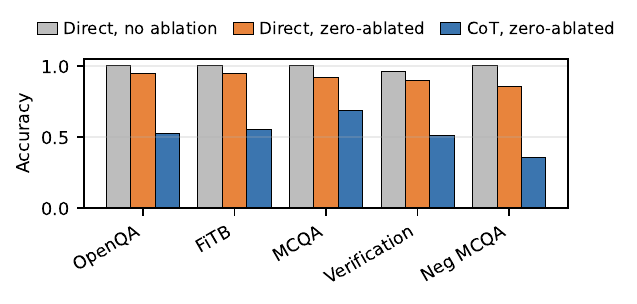}\caption{(landmark, in-country, country), cross-task}\end{subfigure}

\begin{subfigure}{0.49\linewidth}\includegraphics[width=\linewidth]{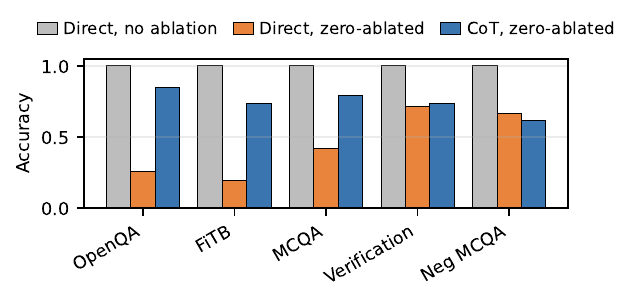}\caption{(country, capital-of, city), own-encoding}\end{subfigure}\hfill
\begin{subfigure}{0.49\linewidth}\includegraphics[width=\linewidth]{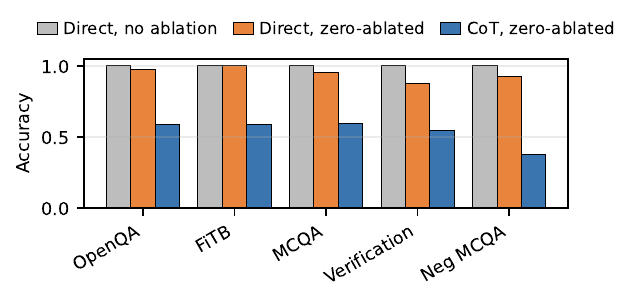}\caption{(country, capital-of, city), cross-task}\end{subfigure}

\begin{subfigure}{0.49\linewidth}\includegraphics[width=\linewidth]{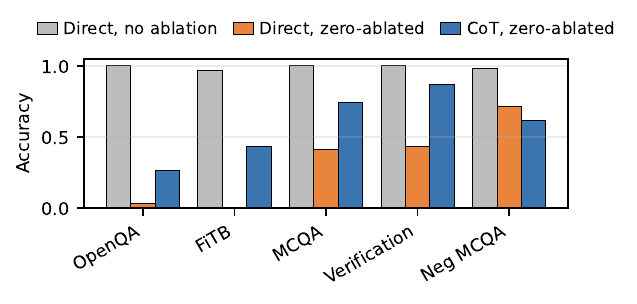}\caption{(company, HQ-in-city, city), own-encoding}\end{subfigure}\hfill
\begin{subfigure}{0.49\linewidth}\includegraphics[width=\linewidth]{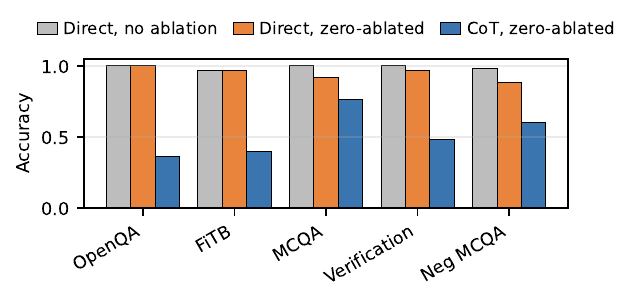}\caption{(company, HQ-in-city, city), cross-task}\end{subfigure}

\begin{subfigure}{0.49\linewidth}\includegraphics[width=\linewidth]{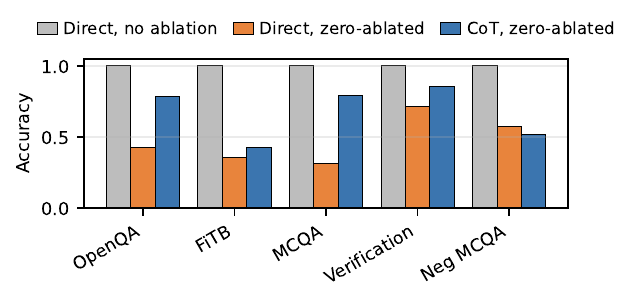}\caption{(person, plays-instrument, instrument), own-encoding}\end{subfigure}\hfill
\begin{subfigure}{0.49\linewidth}\includegraphics[width=\linewidth]{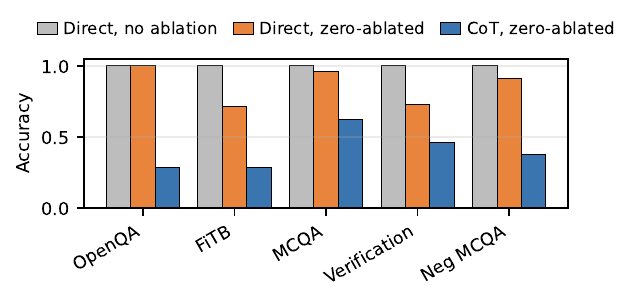}\caption{(person, plays-instrument, instrument), cross-task}\end{subfigure}
\caption{CoT vs.\ direct answering under zero-ablation, \OlmoThirteenB{}.}
\label{fig:cot_recovery_olmo_13b}
\end{figure*}

\begin{figure*}[t]\centering
\begin{subfigure}{0.49\linewidth}\includegraphics[width=\linewidth]{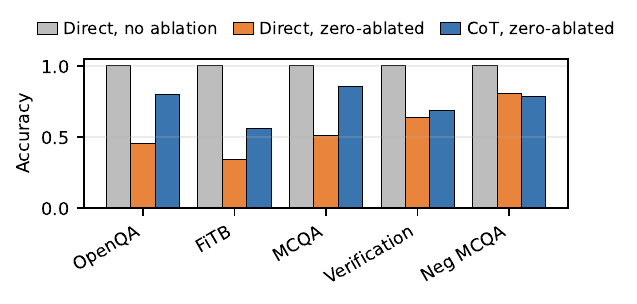}\caption{(country, official language, language), own-encoding}\end{subfigure}\hfill
\begin{subfigure}{0.49\linewidth}\includegraphics[width=\linewidth]{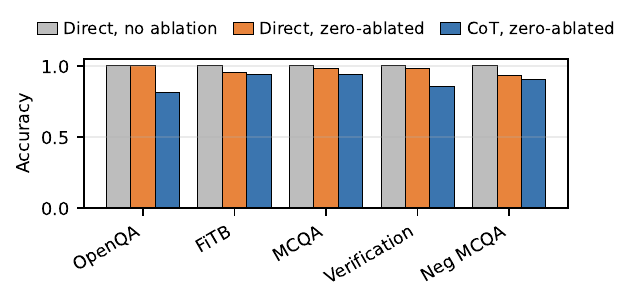}\caption{(country, official language, language), cross-task}\end{subfigure}

\begin{subfigure}{0.49\linewidth}\includegraphics[width=\linewidth]{figures/cot_recovery_own_landmark_gemma_9b_strict.pdf}\caption{(landmark, in-country, country), own-encoding}\end{subfigure}\hfill
\begin{subfigure}{0.49\linewidth}\includegraphics[width=\linewidth]{figures/cot_recovery_cross_landmark_gemma_9b_strict.pdf}\caption{(landmark, in-country, country), cross-task}\end{subfigure}

\begin{subfigure}{0.49\linewidth}\includegraphics[width=\linewidth]{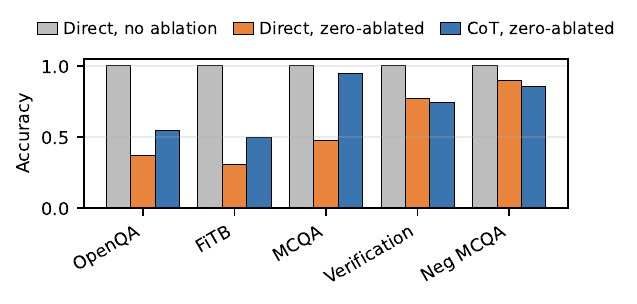}\caption{(country, capital-of, city), own-encoding}\end{subfigure}\hfill
\begin{subfigure}{0.49\linewidth}\includegraphics[width=\linewidth]{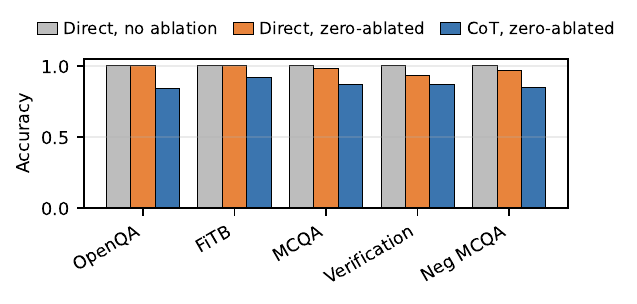}\caption{(country, capital-of, city), cross-task}\end{subfigure}

\begin{subfigure}{0.49\linewidth}\includegraphics[width=\linewidth]{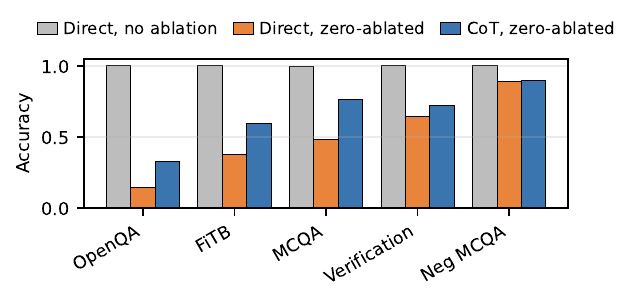}\caption{(company, HQ-in-city, city), own-encoding}\end{subfigure}\hfill
\begin{subfigure}{0.49\linewidth}\includegraphics[width=\linewidth]{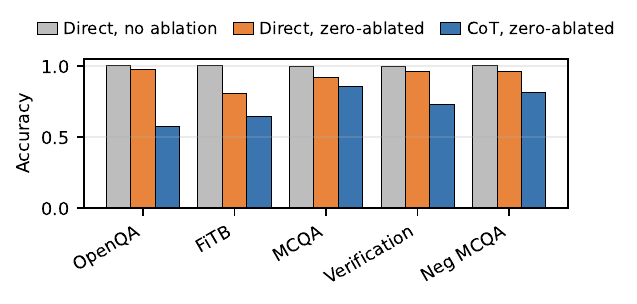}\caption{(company, HQ-in-city, city), cross-task}\end{subfigure}

\begin{subfigure}{0.49\linewidth}\includegraphics[width=\linewidth]{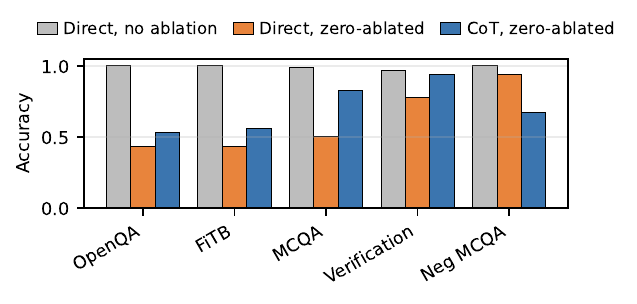}\caption{(person, plays-instrument, instrument), own-encoding}\end{subfigure}\hfill
\begin{subfigure}{0.49\linewidth}\includegraphics[width=\linewidth]{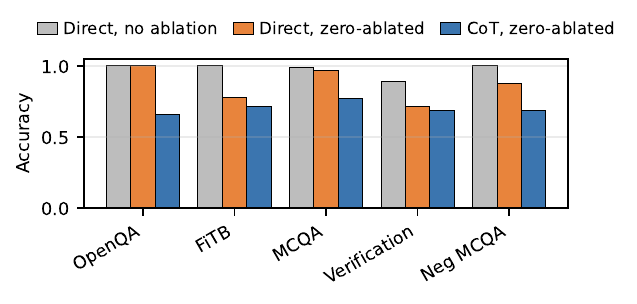}\caption{(person, plays-instrument, instrument), cross-task}\end{subfigure}
\caption{CoT vs.\ direct answering under zero-ablation, \Gemma{}.}
\label{fig:cot_recovery_gemma_9b}
\end{figure*}

\begin{figure*}[t]\centering
\begin{subfigure}{0.49\linewidth}\includegraphics[width=\linewidth]{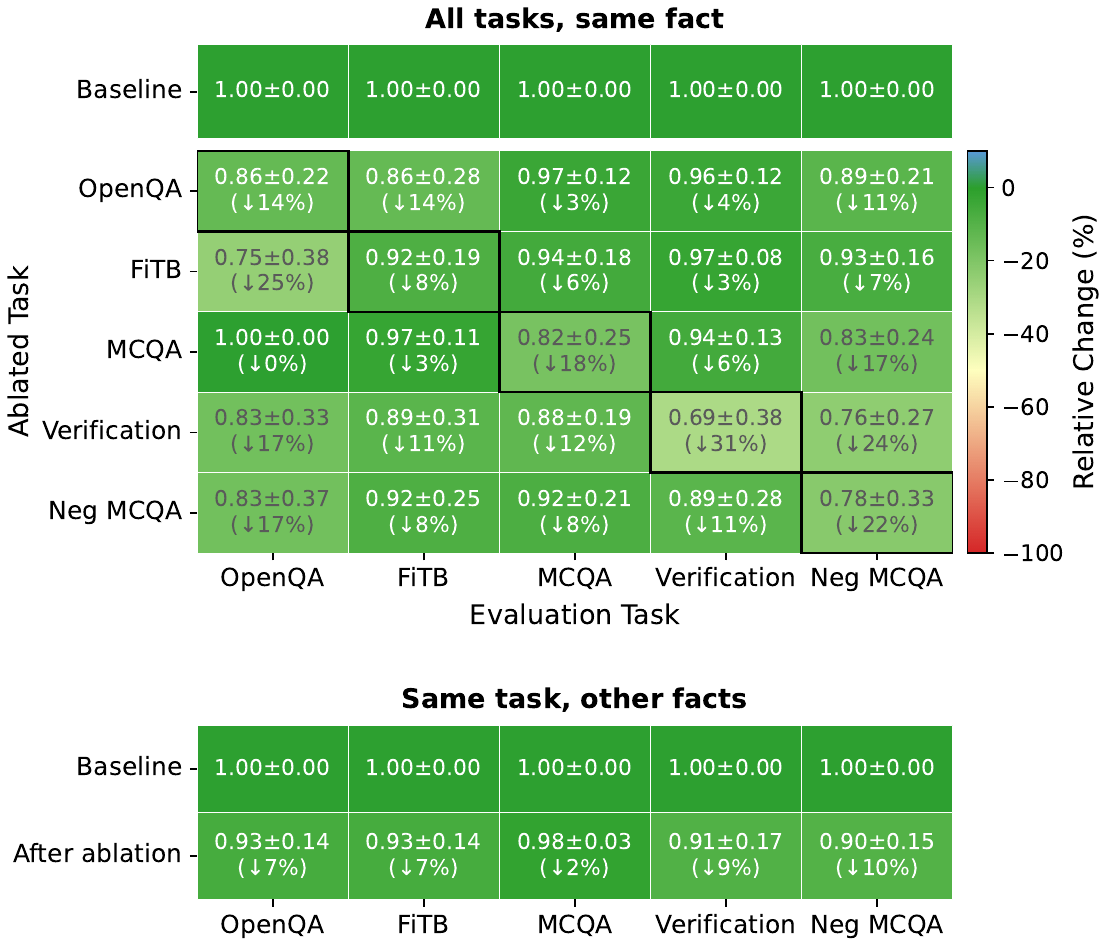}\caption{(country, official language, language)}\end{subfigure}\hfill
\begin{subfigure}{0.49\linewidth}\includegraphics[width=\linewidth]{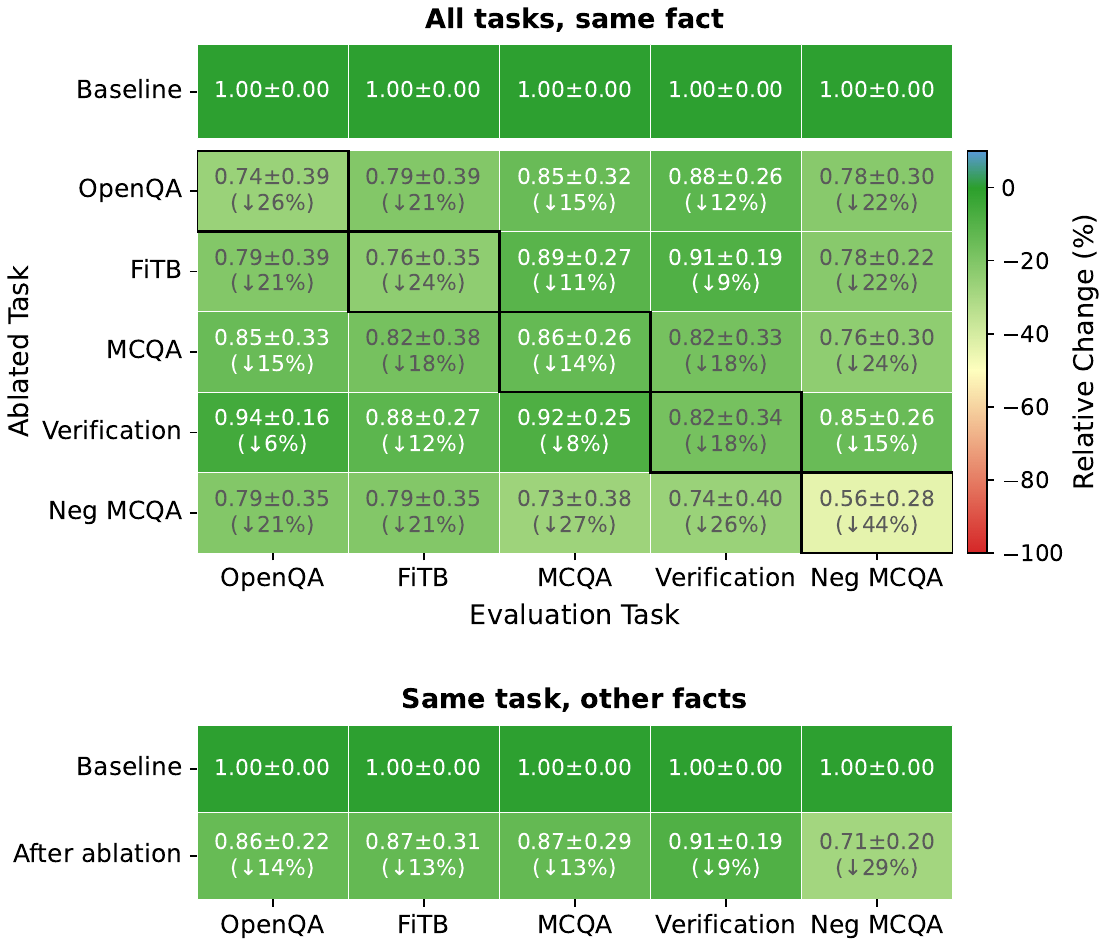}\caption{(landmark, in-country, country)}\end{subfigure}

\begin{subfigure}{0.49\linewidth}\includegraphics[width=\linewidth]{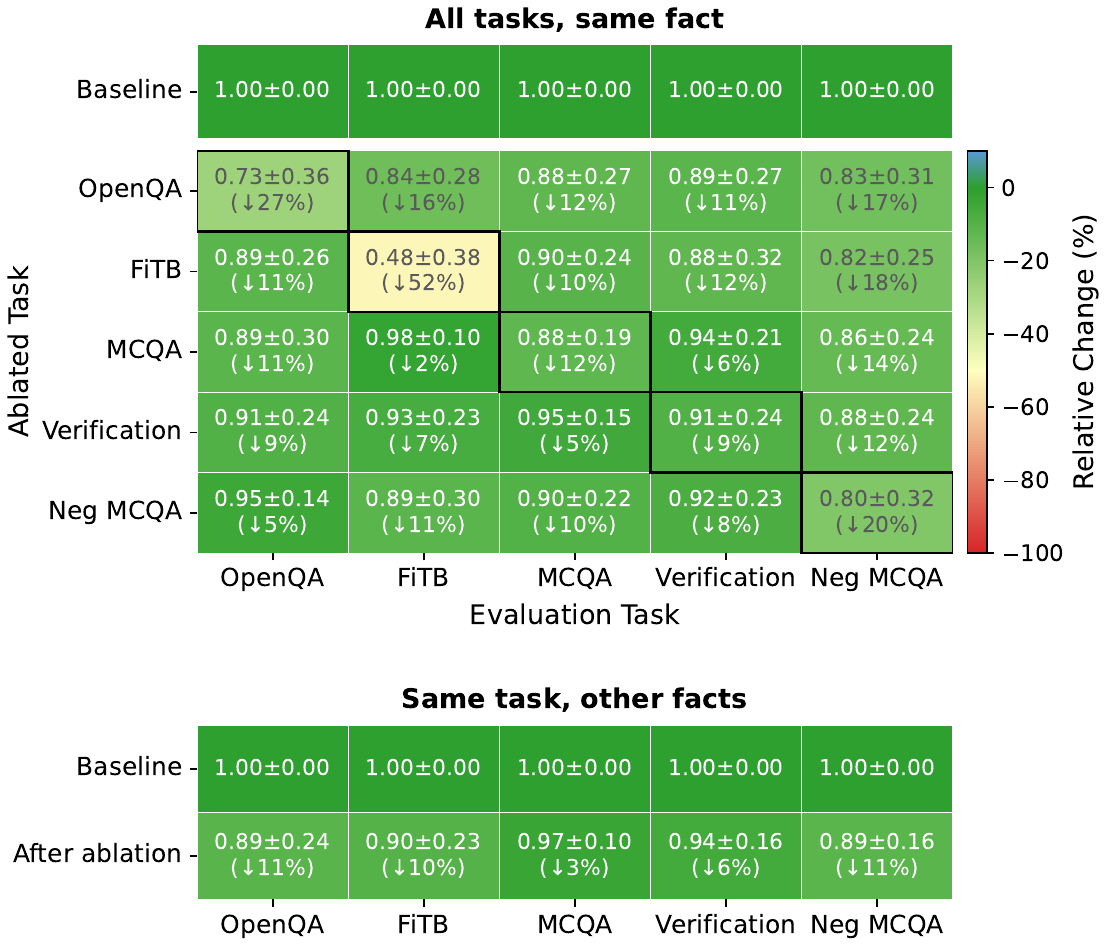}\caption{(country, capital-of, city)}\end{subfigure}\hfill
\begin{subfigure}{0.49\linewidth}\includegraphics[width=\linewidth]{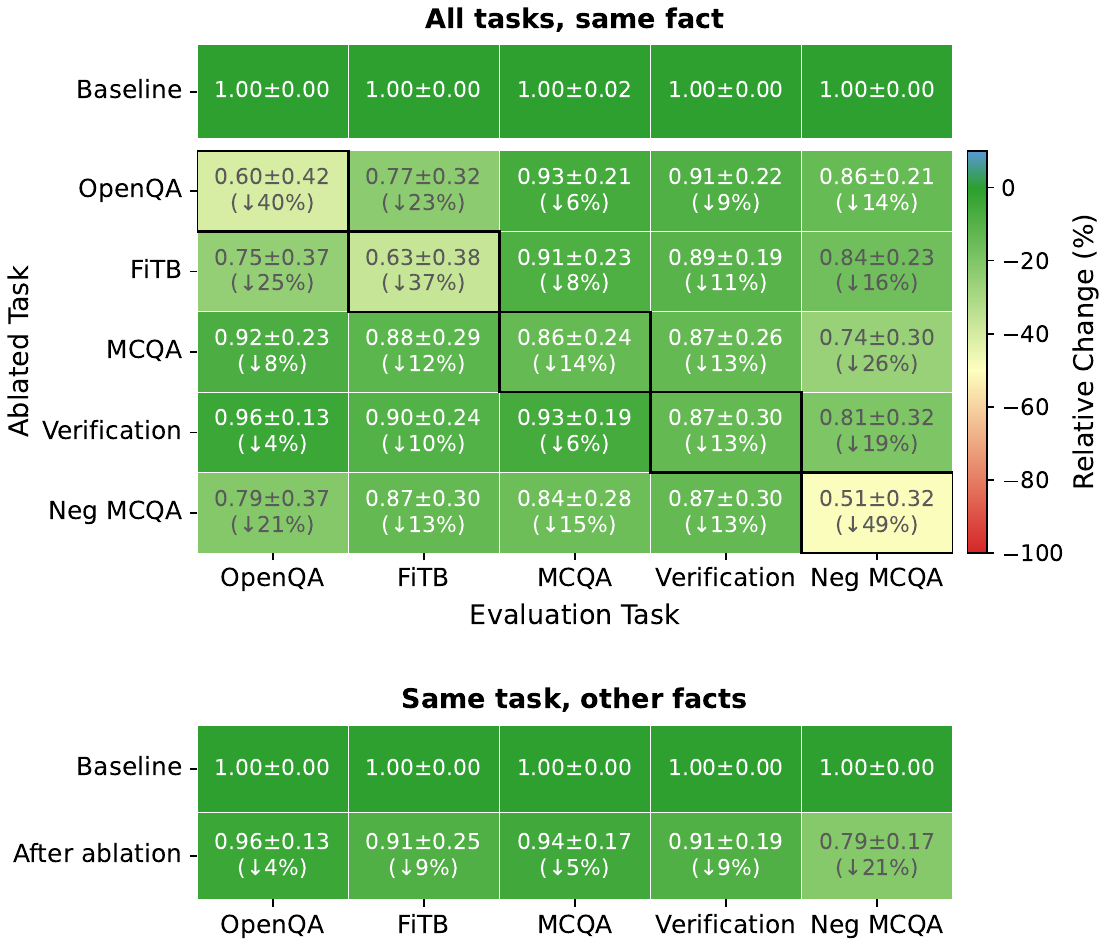}\caption{(company, HQ-in-city, city)}\end{subfigure}

\begin{subfigure}{0.49\linewidth}\includegraphics[width=\linewidth]{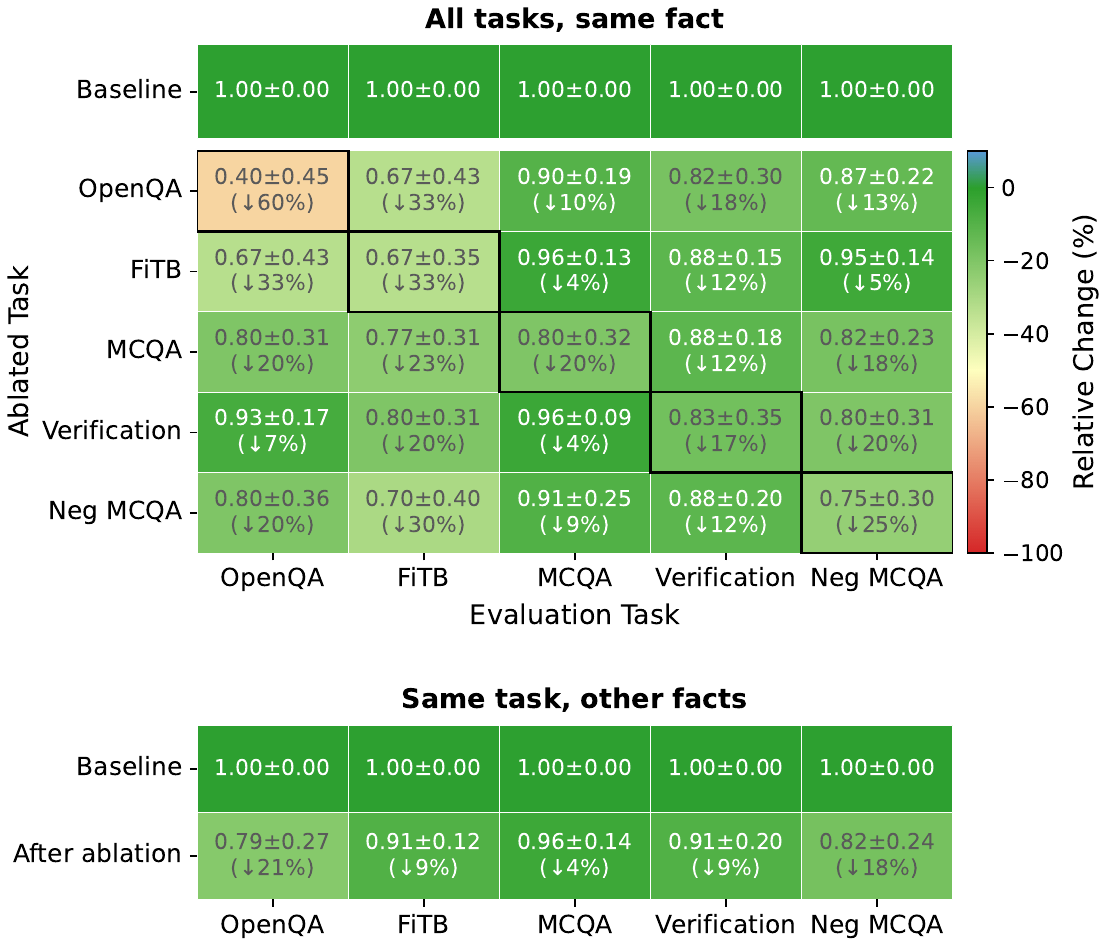}\caption{(person, plays-instrument, instrument)}\end{subfigure}
\caption{CoT-ablation heatmaps, \OlmoSevenB{}. Rows: ablated task; columns: evaluation task scored under CoT; bottom panel: same-task other-facts control.}
\label{fig:cot_heatmap_olmo_7b}
\end{figure*}

\begin{figure*}[t]\centering
\begin{subfigure}{0.49\linewidth}\includegraphics[width=\linewidth]{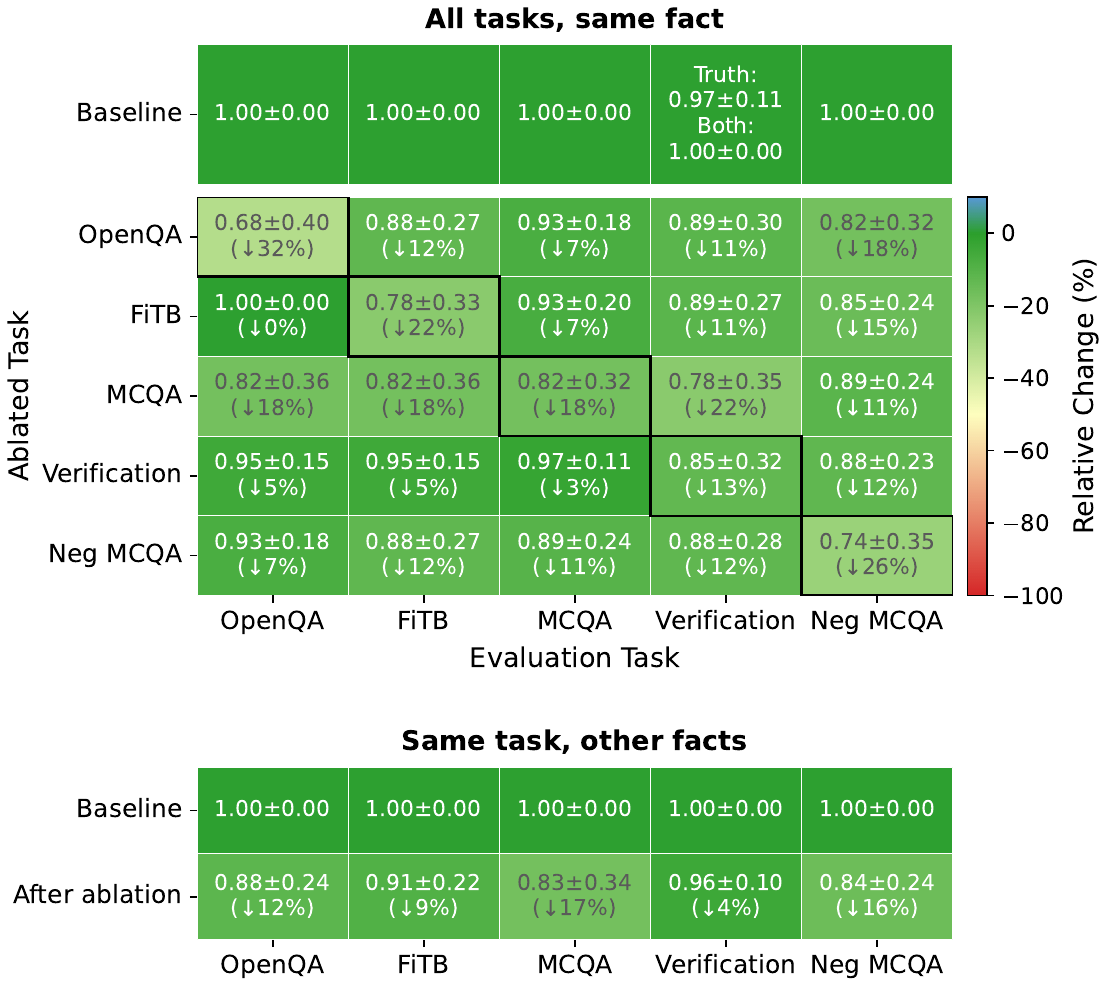}\caption{(country, official language, language)}\end{subfigure}\hfill
\begin{subfigure}{0.49\linewidth}\includegraphics[width=\linewidth]{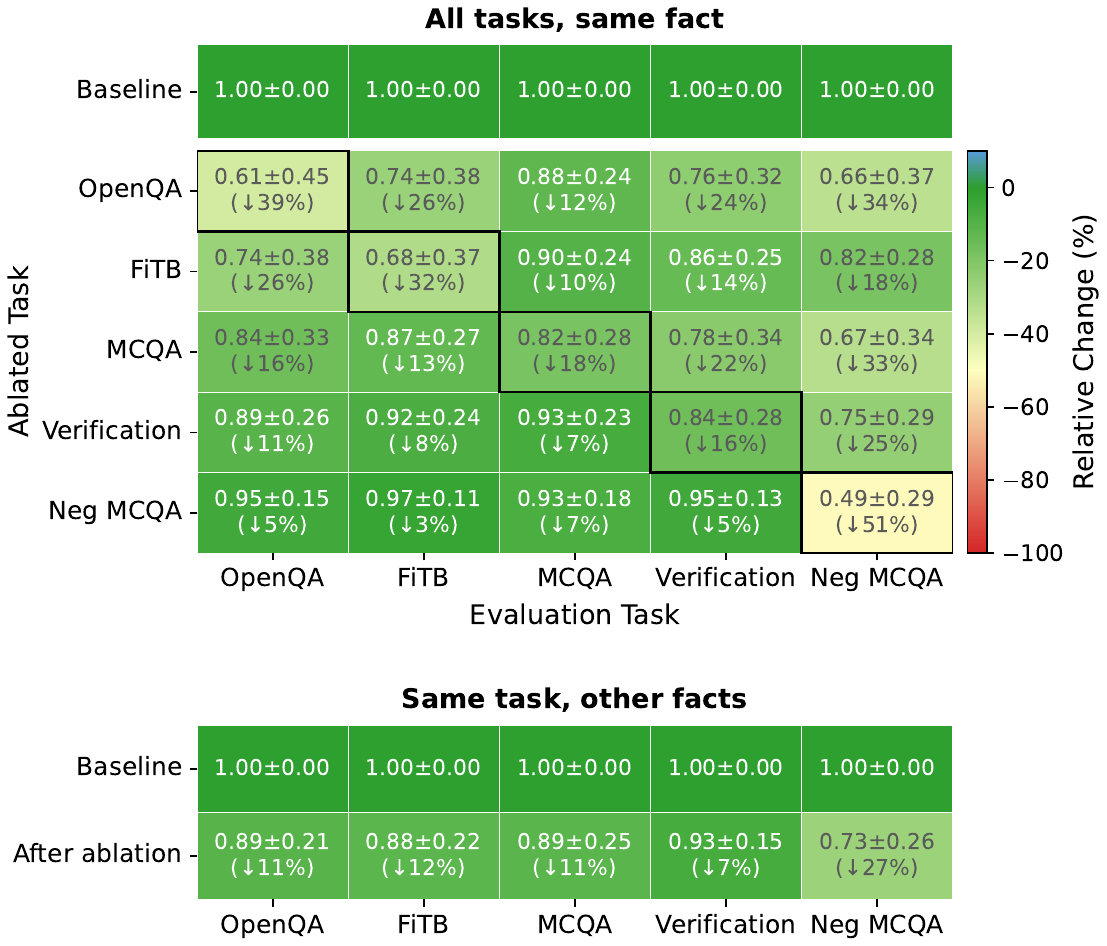}\caption{(landmark, in-country, country)}\end{subfigure}

\begin{subfigure}{0.49\linewidth}\includegraphics[width=\linewidth]{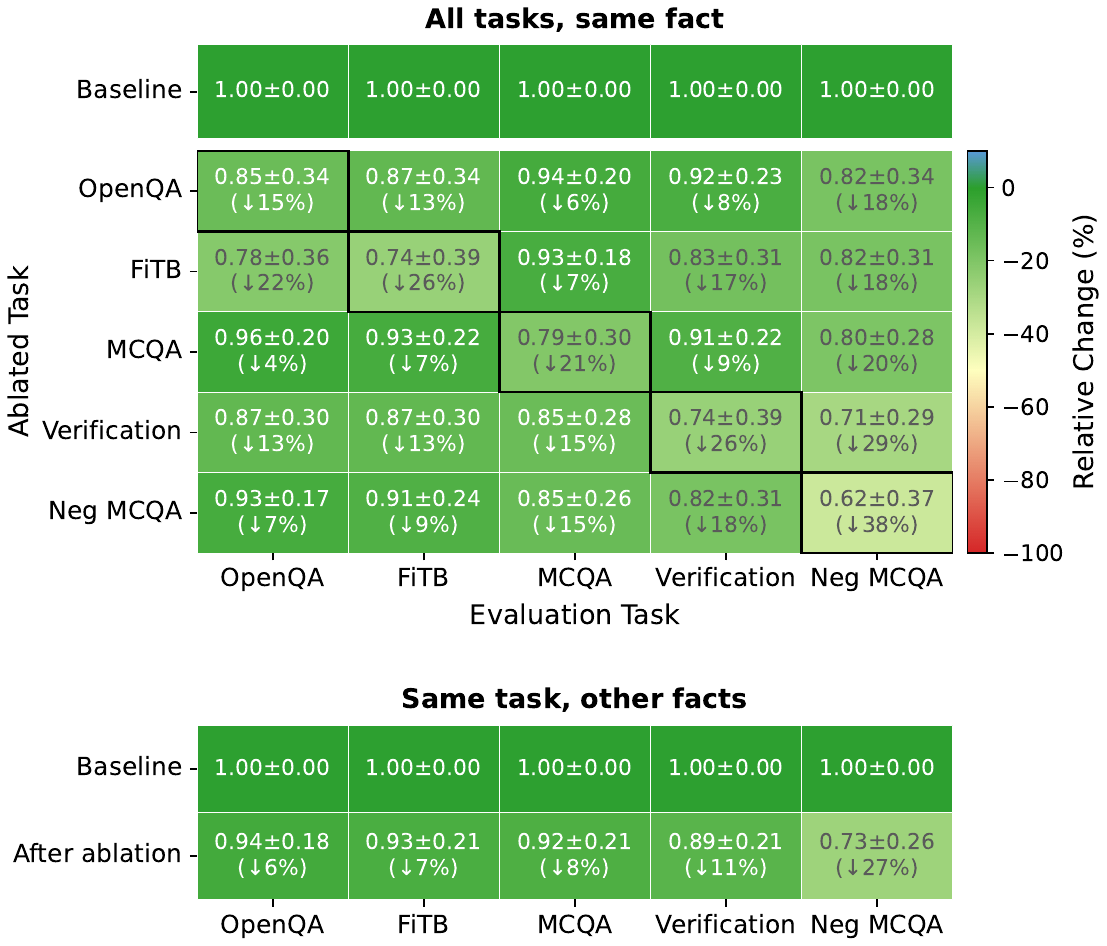}\caption{(country, capital-of, city)}\end{subfigure}\hfill
\begin{subfigure}{0.49\linewidth}\includegraphics[width=\linewidth]{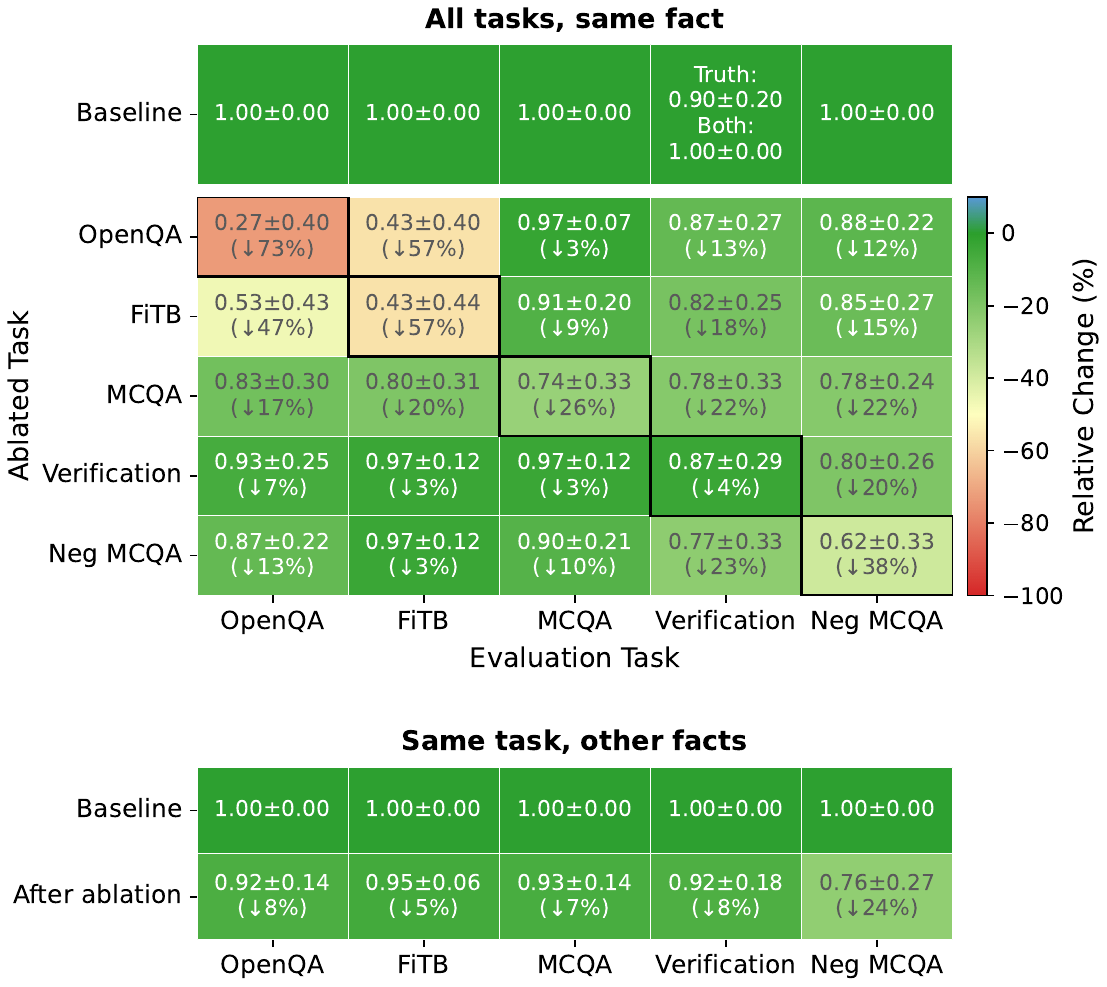}\caption{(company, HQ-in-city, city)}\end{subfigure}

\begin{subfigure}{0.49\linewidth}\includegraphics[width=\linewidth]{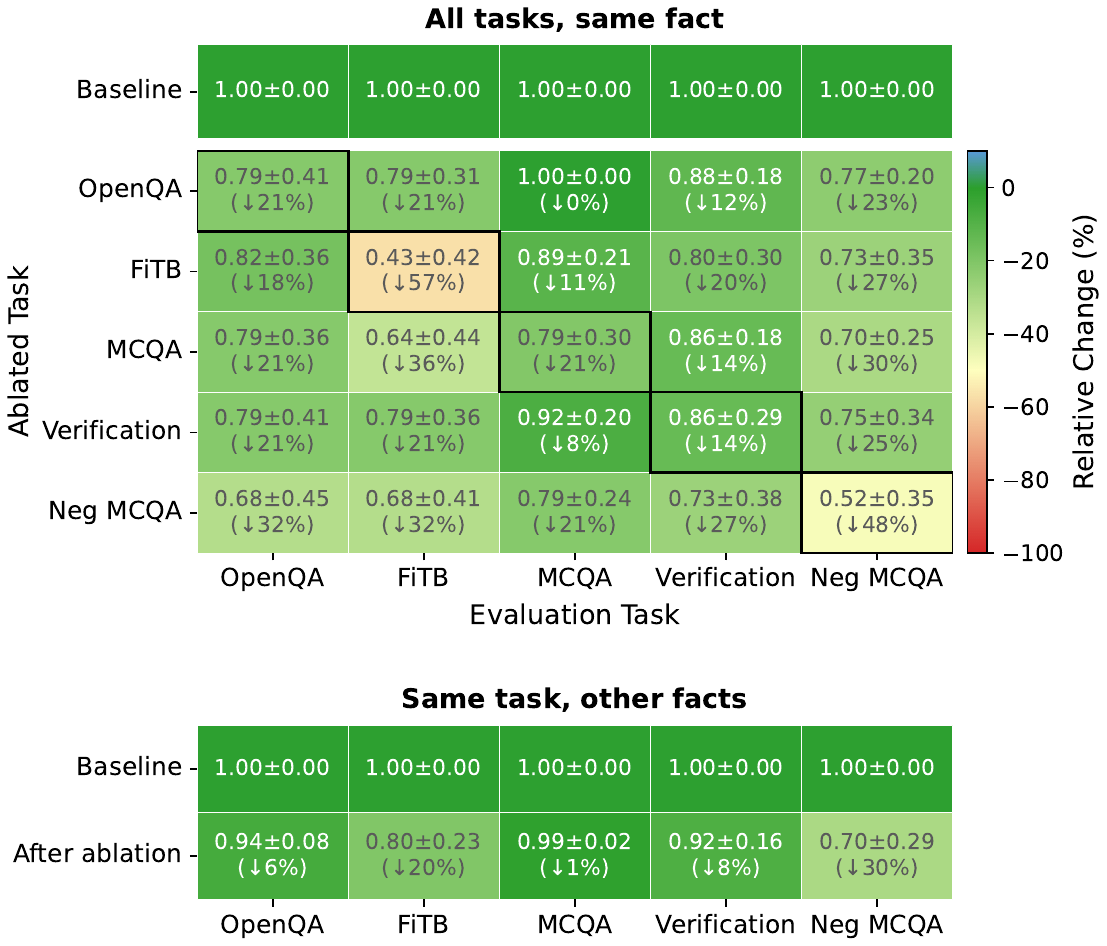}\caption{(person, plays-instrument, instrument)}\end{subfigure}
\caption{CoT-ablation heatmaps, \OlmoThirteenB{}. Same layout as \Cref{fig:cot_heatmap_olmo_7b}.}
\label{fig:cot_heatmap_olmo_13b}
\end{figure*}

\begin{figure*}[t]\centering
\begin{subfigure}{0.49\linewidth}\includegraphics[width=\linewidth]{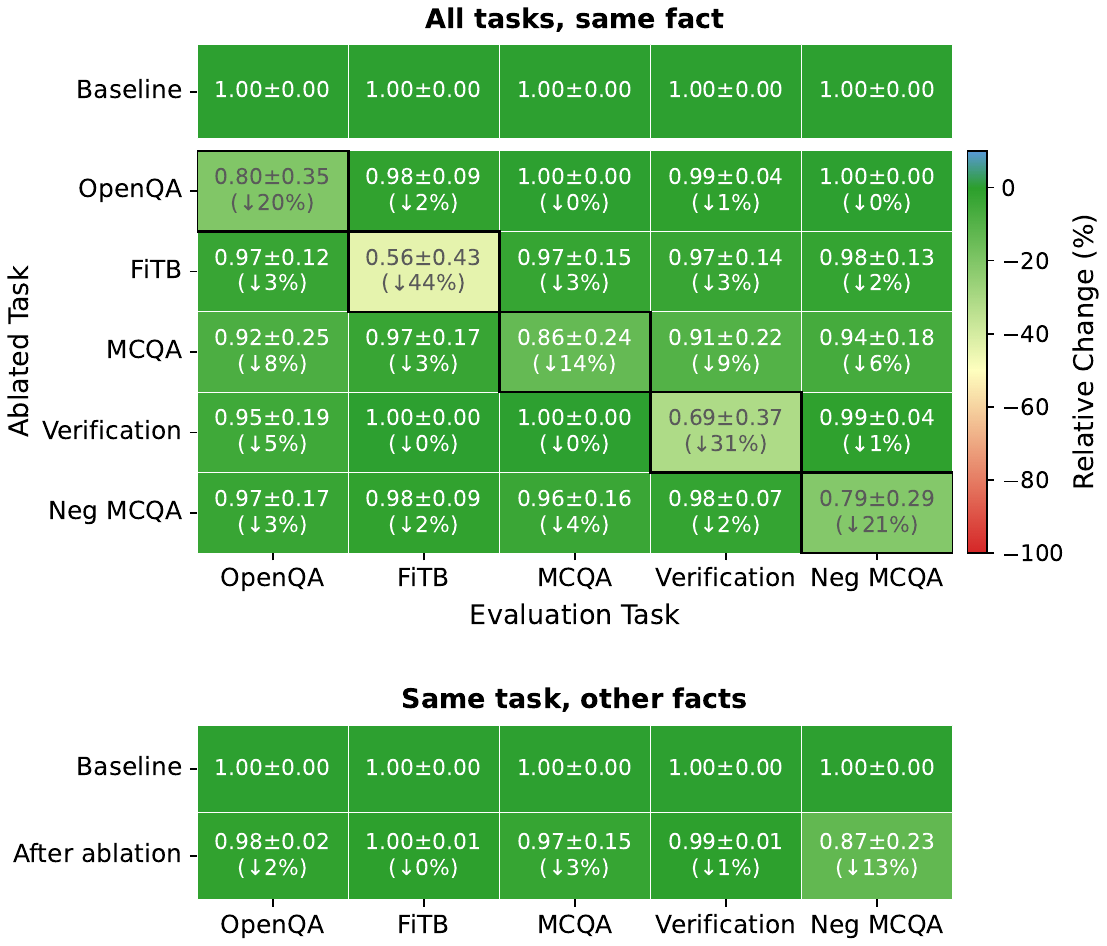}\caption{(country, official language, language)}\end{subfigure}\hfill
\begin{subfigure}{0.49\linewidth}\includegraphics[width=\linewidth]{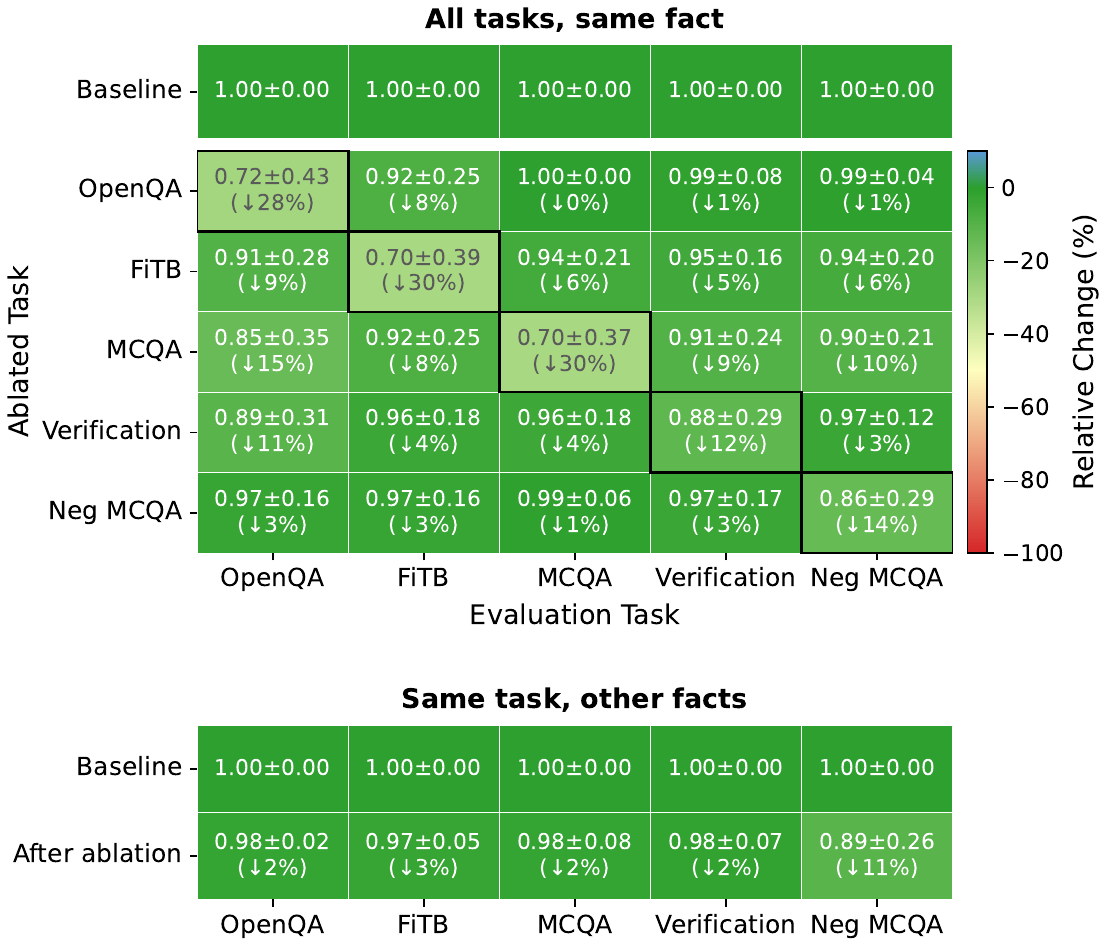}\caption{(landmark, in-country, country)}\end{subfigure}

\begin{subfigure}{0.49\linewidth}\includegraphics[width=\linewidth]{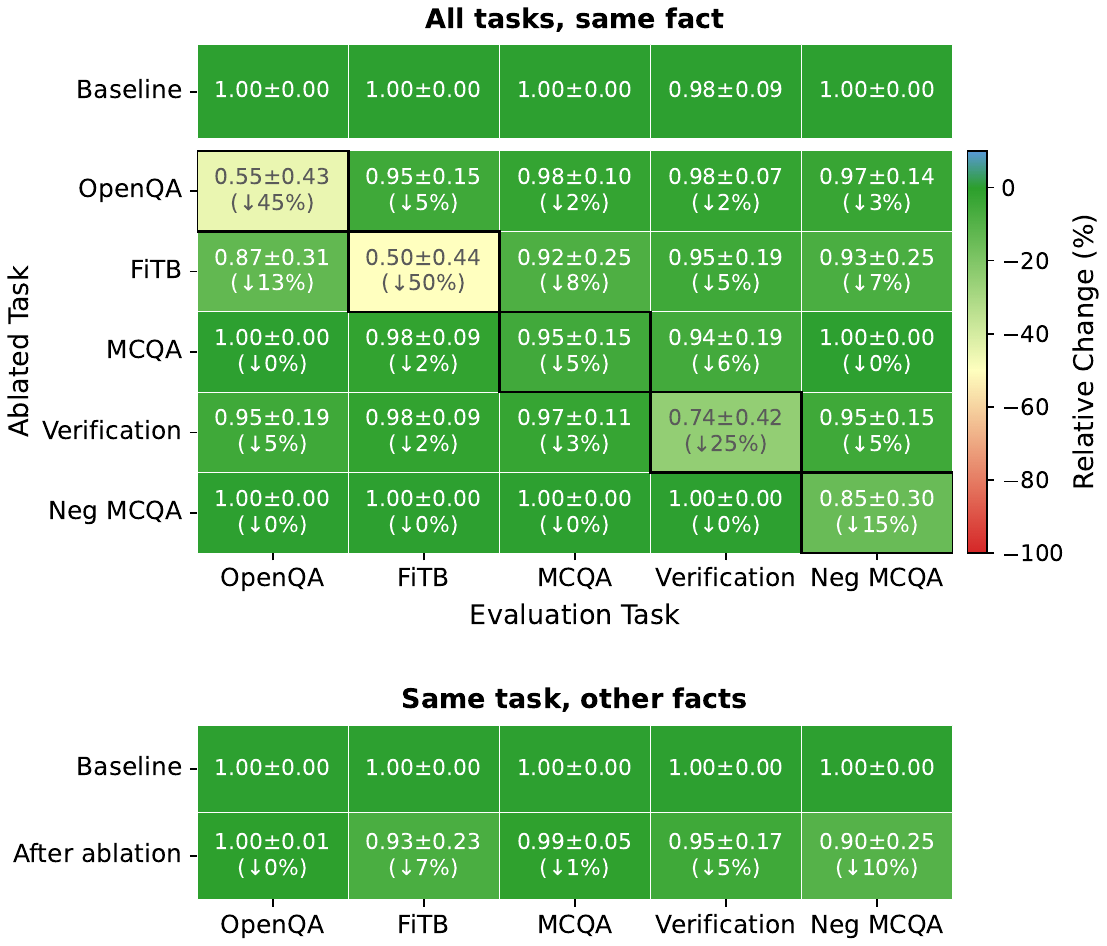}\caption{(country, capital-of, city)}\end{subfigure}\hfill
\begin{subfigure}{0.49\linewidth}\includegraphics[width=\linewidth]{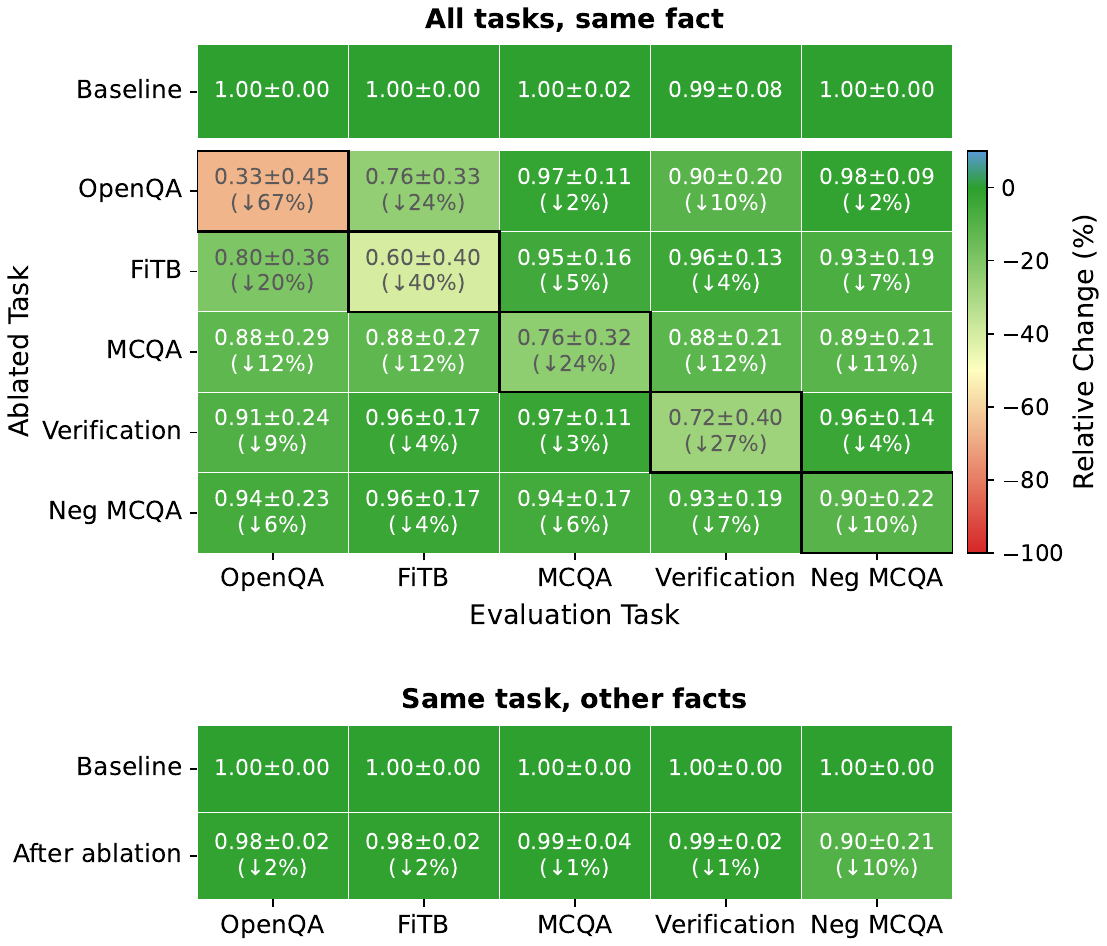}\caption{(company, HQ-in-city, city)}\end{subfigure}

\begin{subfigure}{0.49\linewidth}\includegraphics[width=\linewidth]{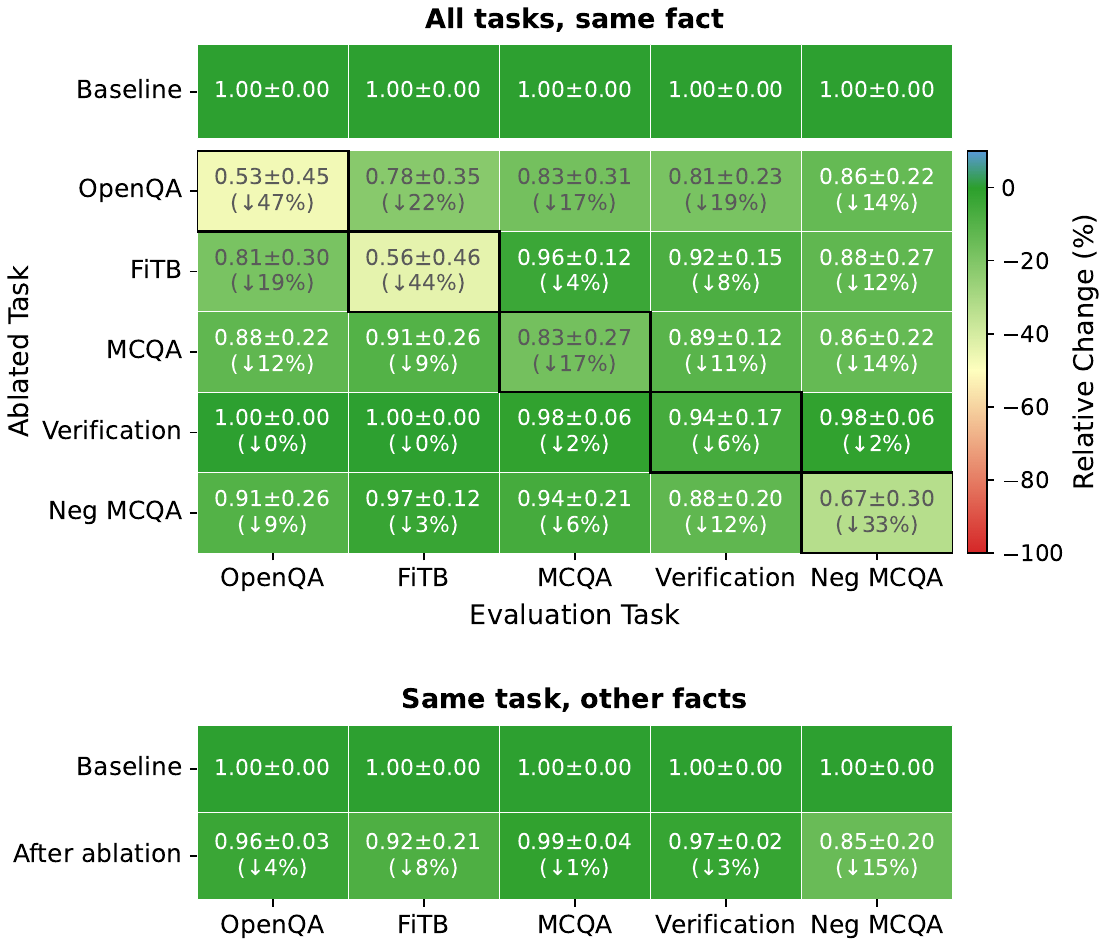}\caption{(person, plays-instrument, instrument)}\end{subfigure}
\caption{CoT-ablation heatmaps, \Gemma{}. Same layout as \Cref{fig:cot_heatmap_olmo_7b}.}
\label{fig:cot_heatmap_gemma_9b}
\end{figure*}

\Cref{fig:cot_recovery_olmo_7b,fig:cot_recovery_olmo_13b,fig:cot_recovery_gemma_9b} present the direct-versus-CoT results across all datasets for \OlmoSevenB{}, \OlmoThirteenB{}, and \Gemma{}, and \Cref{fig:cot_heatmap_olmo_7b,fig:cot_heatmap_olmo_13b,fig:cot_heatmap_gemma_9b} present the corresponding per-task ablation heatmaps. 
The pattern holds throughout. Direct answering drops significantly when the evaluation task's own parameters are ablated while CoT stays close to the unablated baseline, and direct answering is less affected than CoT by another task's ablation.

\section{Resources and packages}
\label{appendix:resources}
Our experiments use models and code from HuggingFace Transformers \citep{wolf2019transformers}. In the (fact, task) interaction analysis (\S\ref{appendix:interaction}) we used SciPy \citep{2020SciPy-NMeth} for the $F$-test.
All experiments requiring GPU were run on a single 256GB AMD MI325X GPU.
In the mechanistic experiment (\S\ref{sec:mechanistic_analysis}), we trained masks for different facts in parallel (up to two facts at a time on a single GPU). Training masks for one fact takes approximately 10 hours, yielding an effective rate of ${\approx}$5 hours per fact. Across the 437 target facts of the three main models, we estimate a total of ${\approx}$2,200 GPU hours. The remaining experiments are negligible in comparison.
\end{document}